\documentclass{article}

\usepackage[preprint]{neurips_2026}

\usepackage[utf8]{inputenc} 
\usepackage[T1]{fontenc}    
\usepackage{hyperref}       
\usepackage{url}            
\usepackage{booktabs}       
\usepackage{amsfonts}       
\usepackage{nicefrac}       
\usepackage{microtype}      
\usepackage{xcolor}         
\usepackage{amsmath}
\usepackage{amssymb}
\usepackage{amsfonts}
\usepackage{graphicx}
\usepackage{multirow}
\usepackage{xcolor}
\usepackage{tikz}
\usepackage{pgfplots}
\pgfplotsset{compat=1.18} 
\usepackage{booktabs}
\usepackage{caption}
\usepackage{pgfplots}
\usepackage{tikz}
\usepackage{subcaption}
\usepackage{wrapfig}
\usepackage{enumitem}
\definecolor{rred}{RGB}{245, 152, 153}
\definecolor{oorange}{RGB}{253, 205, 154}

\title{FlashClear: Ultra-Fast Image Content Removal via Efficient Step Distillation and Feature Caching}

\author{
  Yixin Tang$^{1}$\thanks{Equal contribution.},\enspace 
  Jiawei Guo$^{1}$\footnotemark[1],\enspace 
  Junxian Li$^{1}$,\enspace 
  Zhiteng Li$^{1}$,\enspace 
  Jixin Zhao$^{2}$,\enspace 
  Bingya Zhang$^{3}$,\enspace \\
  \textbf{Chenbo Wang$^{3}$,}\enspace 
  \textbf{Yulun Zhang}$^{1}$\thanks{Corresponding authors: Yulun Zhang, yulun100@gmail.com; Shangchen Zhou, shangchenzhou@gmail.com.}\textbf{,} \enspace
  \textbf{Shangchen Zhou$^{2\,\dagger}$}
  \\
  \textsuperscript{1}Shanghai Jiao Tong University,\enspace
  \textsuperscript{2}Nanyang Technological University,\enspace
  \textsuperscript{3}Honor Device Co., Ltd
  \vspace{-5.mm}
}

\begin{document}

\maketitle

\begin{abstract}
Recently, diffusion-based object removal models have achieved impressive results in eliminating objects and their associated visual effects. However, they indiscriminately denoise all tokens across all timesteps, ignoring that removal usually involves small foreground regions. This strategy introduces substantial computational overhead and prolonged inference times. To overcome this computational burden, we propose a latent discriminator to implement \textbf{R}egion-aware \textbf{A}dversarial \textbf{D}istillation (\textbf{RAD}), yielding a highly efficient few-step model named \textbf{FlashClear}. Furthermore, tailored to few-step diffusion models, we propose \textbf{FPAC} (\textbf{F}oreground-\textbf{P}rioritized Asymmetric \textbf{A}ttention and \textbf{C}aching), a training-free acceleration strategy. Extensive experiments demonstrate that our framework provides massive acceleration while maintaining or exceeding the performance of our base model, ObjectClear. 
Notably, on the OBER benchmark, our FlashClear achieves up to 8.26$\times$ and 122$\times$ speedup over ObjectClear and OmniPaint, respectively, while maintaining high visual quality and fidelity. 
\end{abstract}

\setlength{\abovedisplayskip}{2pt}
\setlength{\belowdisplayskip}{2pt}

\vspace{-2mm}
\section{Introduction}
\vspace{-2mm}
Object removal is a practical yet challenging form of image inpainting, aiming to completely erase a user-specified object while producing a visually plausible background. Recent diffusion-based image editing and inpainting models~\cite{meng2021sdedit, podell2023sdxl, rombach2022high, corneanu2024latentpaint, liu2024structure, yu2021wavefill, xie2023smartbrush} have greatly improved the realism and generalization ability of object removal systems. Unlike Early mask-aligned methods~\cite{ekin2024clipaway, zhuang2024task, sun2025attentive, li2025rorem}, which mainly conduct simple object erasure, thus leaving inconsistent object-induced visual effects, more recent approaches~\cite{wei2025omnieraser, winter2024objectdrop, zhu2025georemover, zhao2025objectclear} infer and remove object-induced visual effects beyond the object mask. Despite their impressive visual quality, these methods inherit the traditional multi-step diffusion paradigm~\cite{song2020denoising, ho2020denoising, rombach2022high}, which introduces massive computational overhead and latency, limiting their deployment in real-time and resource-constrained applications.

Efficiency is particularly critical for object removal. Unlike one-shot generation, it is often embedded in interactive editing loops requiring instant feedback. Consequently, diffusion sampling latency becomes a major bottleneck for instant editing, mobile deployment, and large-scale services. 
Many efforts have accelerated diffusion models. One popular direction is step distillation, which compresses models into few-step generators~\cite{sauer2024adversarial, lin2024sdxl, song2023consistency, yin2024one}. However, most distillation methods target text-to-image generation and remain under-explored for image inpainting tasks like object removal due to their strong spatial constraints and effect-aware reasoning. Another direction is training-free acceleration, particularly feature caching~\cite{ma2024deepcache, selvaraju2024fora, chen2024delta}, which exploits temporal redundancy by reusing similar features across adjacent denoising steps. However, applying feature caching to step-distilled models reveals a natural incompatibility~\cite{zou2026disca}: once the denoising trajectory is compressed, temporal redundancy is significantly reduced, making feature reuse more inaccurate and the overhead of cache selection more prominent. Consequently, directly applying existing caching to few-step models yields limited speedup and noticeable quality degradation.

\begin{figure}[t]
\centering
\vspace{-2mm}
\includegraphics[width=0.98\textwidth]{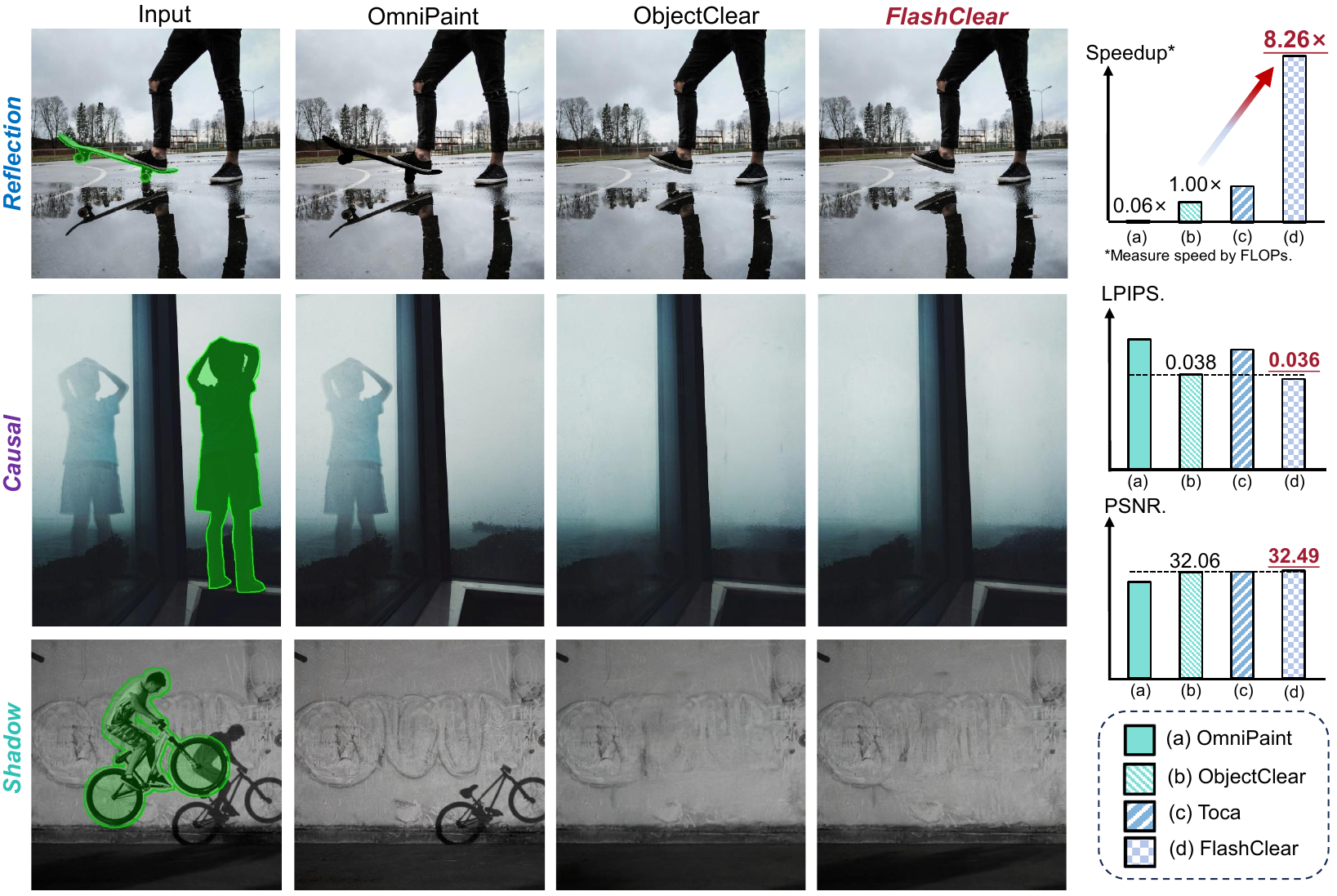}
\vspace{-1mm}
\caption{Qualitative and efficiency comparison of different object removal methods OmniPaint~\cite{yu2025omnipaint} and ObjectClear~\cite{zhao2025objectclear}, and acceleration method ToCa~\cite{zou2024accelerating} applied to ObjectClear. FlashClear effectively removes target objects and causal effects, achieving superior quality and the fastest speed.}
\label{teaser}
\vspace{-10mm}
\end{figure}

Despite the incompatibility challenge mentioned above, in this work, we show that object removal provides a unique opportunity to reconcile this conflict. Although temporal redundancy becomes weak in the few-step regime, object removal contains strong spatial redundancy. The editing process is intrinsically localized: most background regions should remain unchanged, while only the object and its associated visual effects require substantial recomputation. This property suggests that feature reuse should not rely solely on adjacent denoising steps, but should instead be guided by the spatial structure of the removal task. The key challenge is that the user-provided mask usually covers only the target object and does not explicitly indicate the full region affected by the object, such as shadows and reflections. Recent works~\cite{yan2025eedit, qin2025spotedit, zhao2025objectclear} also noticed this inconsistency between the user-provided mask and the real region that needs to be edited. To address it, ObjectClear~\cite{zhao2025objectclear} indicates that diffusion models can implicitly capture object-induced effects through internal attention responses with auxiliary guidance. These attention maps offer a natural spatial prior for distinguishing tokens that require recomputation from invariant background tokens that can be safely cached.

Motivated by this observation, we propose FlashClear, an efficient object removal framework that jointly exploits few-step distillation and attention-guided feature caching. \textbf{Firstly}, instead of treating step reduction and feature reuse as two independent or even contradicting acceleration techniques, FlashClear first constructs a low-cost, few-step removal model and then adapts feature caching specifically to this compressed sampling regime. \textbf{Secondly} and concretely, we introduce region-aware adversarial distillation (RAD), which uses the U-Net of the base removal model as a latent discriminator to distill the multi-step removal process into a few-step generator. RAD substantially reduces the sampling cost while preserving the model's ability to perceive object-induced effects through internal attention maps. \textbf{Thirdly}, based on the distilled model, we further propose FPAC, an attention-guided feature caching mechanism that selectively reuses features of invariant background tokens and recomputes features for foreground and effect-related regions. By shifting the basis of caching from temporal redundancy to task-specific spatial redundancy, FPAC remains effective even when only a few denoising steps are available. Our contributions are summarized as follows:
\begin{itemize}[nosep, leftmargin=*]
    \item We propose FlashClear, an efficient object removal framework that integrates few-step distillation with feature caching, addressing the efficiency bottleneck of diffusion-based object removal.
    \vspace{0.5mm}
    \item We propose RAD, a region-aware adversarial distillation strategy tailored for object removal to reduce inference steps while maintaining effect-aware capabilities.
    \vspace{0.5mm}
    \item We design FPAC, an attention-guided caching mechanism that exploits background spatial redundancy, enabling substantially lower computational cost without sacrificing removal quality.
\end{itemize}
\vspace{-6mm}
\section{Related Work}
\vspace{-2mm}
\textbf{Object Removal and Image Editing.} Object removal is a challenging inpainting task that requires removing not only the target object but also its associated visual effects (\textit{e.g.}, shadows and reflections) while synthesizing a coherent background. Diffusion-based methods have recently achieved strong results in this task. Early approaches rely on synthetic training data~\cite{suvorov2022resolution,lugmayr2022repaint,jiang2025smarteraser,liu2025erase}, which limits their ability to model real-world object-scene interactions. Alongside these, text-guided diffusion models have revolutionized general image editing and inpainting~\cite{nichol2021glide,saharia2022palette,avrahami2022blended,brooks2023instructpix2pix,kawar2023imagic}. Furthermore, to improve spatial controllability and regional consistency, various structural guidance and plug-and-play modules have been proposed~\cite{zhang2023adding,ju2024brushnet,chen2024anydoor,manukyan2023hd}. Recent methods specifically target physical effect removal by constructing higher-quality data from real videos, simulations, or fixed-camera captures~\cite{zhao2025objectclear,zhu2025georemover,wei2025omnieraser}. However, almost all the models mentioned use multi-step diffusion inference, which is computationally expensive, restricting practical deployment.

\vspace{-0.5mm}
\textbf{Step Distillation and Fast Sampling.} Accelerating diffusion models traditionally involves compressing the sampling trajectory into fewer denoising steps. Orthogonal to distillation, advanced training-free ODE solvers~\cite{lu2022dpm,zhao2023unipc,liu2022pseudo} significantly reduce sampling steps by analytically modeling the numerical trajectory. Within the realm of distillation, representative methods include progressive distillation~\cite{salimans2022progressive}, consistency-based models~\cite{song2023consistency,luo2023latent,lu2024simplifying}, distribution matching distillation~\cite{yin2024one}, and rectified-flow distillation~\cite{liu2023instaflow}. In the extremely low-step regime, adversarial and score-based distillation methods such as ADD~\cite{sauer2024adversarial}, SDXL-Lightning~\cite{lin2024sdxl}, DMD2~\cite{yin2024improved}, UFOGen~\cite{xu2024ufogen}, and SwiftBrush~\cite{nguyen2024swiftbrush} have demonstrated strong visual quality in one or few steps. Inspired by this line of work, we propose Region-aware Adversarial Distillation (RAD), which adapts adversarial distillation to object removal by emphasizing both global realism and local removal fidelity.

\vspace{-0.5mm}
\textbf{Architectural Efficiency and Feature Caching.} Feature caching provides a complementary training-free acceleration strategy by reusing intermediate computations during diffusion inference. With the recent paradigm shift towards transformer-based diffusion models (\textit{e.g.}, DiT~\cite{peebles2023scalable} and U-ViT~\cite{bao2023all}), token-level redundancy has been heavily exploited. Existing methods either directly cache features~\cite{ma2024deepcache}, dynamically select cache steps or tokens~\cite{zou2024accelerating,zhang2025training,selvaraju2024fora}, or extrapolate cached representations~\cite{feng2025hicache,wu2025quantcache,chen2024delta}. Besides caching, techniques like token merging (ToMe)~\cite{bolya2023token}, attention-map reuse (T-Gate)~\cite{zhang2024cross}, and architectural compression~\cite{li2023snapfusion,kim2023architectural} have been developed to fundamentally reduce computational overhead. Despite their effectiveness, these methods are mainly designed for general multi-step generation and often rely on sufficient denoising steps or additional priors~\cite{feng2025hicache,liu2025reusing}. This makes them less suitable for our 4-step object removal setting. We therefore introduce FPAC, a foreground-prioritized caching strategy tailored precisely to low-step object removal dynamics.

\vspace{-4mm}
\section{Methodology}
\vspace{-2mm}
\subsection{Preliminary}
\vspace{-2mm}
Image inpainting tasks aim to edit specified image regions conditioned on inputs such as reference images, masks, and text prompts. 
As a subcase of the image inpainting task, object removal can be formulated as a conditional generation problem operated within a latent diffusion framework~\cite{rombach2022high}.\par
Following SDXL-inpainting~\cite{lin2024sdxl}, the input reference image $I_{ref}$ $\in \mathbb{R}^{H \times W \times 3}$ is first encoded into a latent $z_0 \in \mathbb{R}^{4 \times h \times w}$ by a frozen VAE~\cite{kingma2013auto} encoder, where $h$ and $w$ denote the downsampled spatial dimensions. The forward diffusion process continually adds Gaussian noise $\epsilon \sim \mathcal{N}(0, \textbf{I})$ to the initial latent $z_0$ over $\textbf{T}$ steps. The noisy latent $z_t$ with $t \in [1, T]$ can be obtained as: 
\begin{equation}
    a(z_t|z_0)=\mathcal{N}(z_t;\sqrt{\bar{\alpha_t}z_0, (1-\bar{\alpha_t}\textbf{I})},
\end{equation}
where $\bar{\alpha_t}$ is the pre-defined noise schedule parameter.\par
Let $M\in \{0,1\}^{1\times h \times w}$ represent the binary mask of input resized to the latent space, where $1$ indicates the foreground region to be removed, and $0$ indicates the unmasked background. 
The joint input is formulated by concatenating the noisy latent $z_t$, the binary mask $M$, and the reference latent $z_{ref}=z_0$ along the channel dimension as follows:
\begin{equation}
    z_{in} = \text{Concat}(z_t, M, z_{ref}) \in \mathbb{R}^{9 \times h \times w}.
\end{equation}\par
During inference, DDIM~\cite{song2020denoising} scheduler is used to get the timesteps $\mathcal{T} = \{t_N, t_{N-1}, ...,t_1\}$. Starting from a Gaussian noise $z_{\tau_4} \sim \mathcal{N}(0, \mathbf{I})$, the model iteratively computes the next latent $z_{\tau_{i-1}}$ using the deterministic DDIM sampling mechanism:

\begin{equation}
    z_{\tau_{i-1}} = \sqrt{\bar{\alpha}_{\tau_{i-1}}} \left( \frac{z_{\tau_i} - \sqrt{1 - \bar{\alpha}_{\tau_i}} \hat{\epsilon}_\theta^{(i)}}{\sqrt{\bar{\alpha}_{\tau_i}}} \right) + \sqrt{1 - \bar{\alpha}_{\tau_{i-1}}} \hat{\epsilon}_\theta^{(i)},
\end{equation}
where $\hat{\epsilon}_\theta^{(i)} = \mathcal{G}_\theta(\text{Concat}(z_{\tau_i}, M, z_{ref}), \tau_i, c)$ represents the predicted noise at step $\tau_i$ and $\mathcal{G}_\theta$ is the denoising U-Net.
In our specific object removal context, we adopt the post-fused Clip text-image embeddings $c_{fusion}$ following the strategy proposed in ObjectClear as the condition $c$.

\begin{figure}[t]
\centering
\includegraphics[width=1\linewidth]{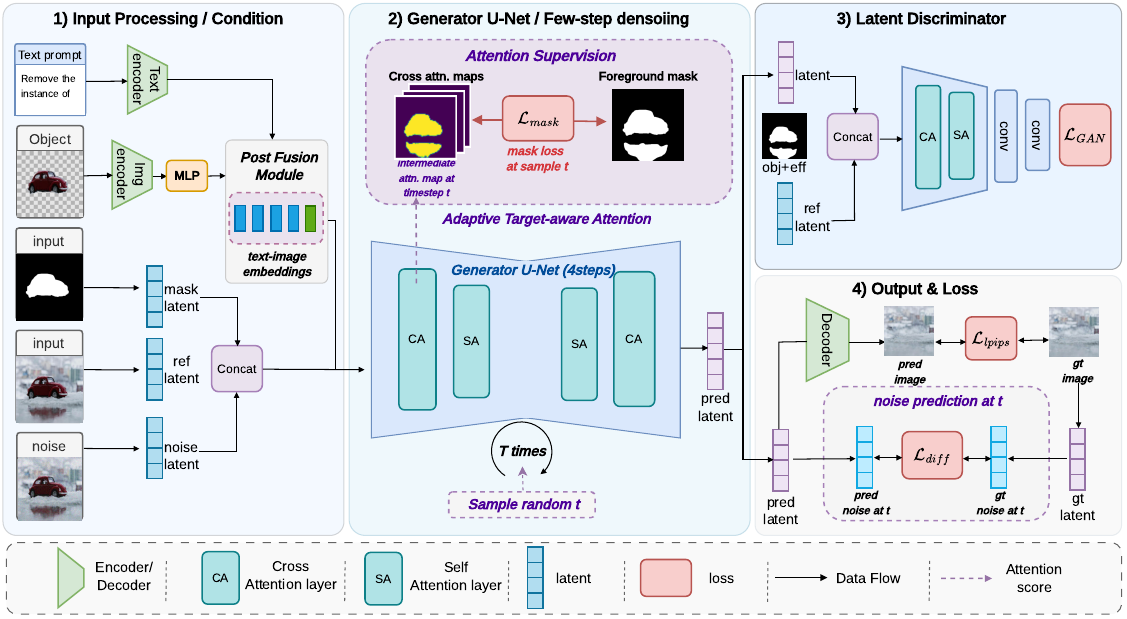}
\vspace{-3mm}
\caption{Illustration of our Region-aware Adversarial Distillation. To accelerate object removal without introducing inconsistent visual effects, we propose an asymmetric masking strategy. The generator is conditioned strictly on the target object mask (obj mask) to predict the denoised latent. In contrast, the latent discriminator evaluates the generation quality using an expanded mask (obj+eff mask) that includes both the object and its causal effects. Additionally, we utilize a post-fusion module to transform the object image token into the text embedding and apply attention-guided object localization loss to preserve spatial priors.}
\label{fig:rad_pipeline}
\vspace{-5mm}
\end{figure}

\vspace{-2mm}
\subsection{Region-aware Adversarial Distillation (RAD)}
\vspace{-2mm}

As an effective way to mitigate the computational cost during inference, few-step distillation enables our base model to remove the masked object with a much lower inference cost. Our goal is to minimize the inference steps as much as possible, while avoiding unnatural artifacts caused by step reduction. Furthermore, since we will subsequently conduct attention-guided feature caching for the distilled model (as detailed in Sec.~\ref{section-fpac}), preserving the model's ability to perceive the object with its associated physical effects, such as shadows and reflections, is paramount.

Although progressive distillation~\cite{salimans2022progressive} and distribution matching methods~\cite{yin2024improved, yin2024one} demonstrate great potential in diffusion-based image generation acceleration, accelerating highly constrained tasks like object removal remains under-explored. RORem~\cite{li2025rorem} adopts the latent consistency model (LCM~\cite{luo2023latent}) to distill a four-step model for object removal. However, consistency-based trajectories often introduce visual blurring and mode-averaging into the inpainting regions. What's more, existing distillation methods merely account for the specific characteristics of object removal tasks, namely, the need to remove both the object and its associated physical effects, such as reflections and shadows. 

To circumvent this, we employ adversarial distillation~\cite{sauer2024adversarial} with tailored settings to obtain a high-fidelity four-step model. Following the paradigm in sdxl-lightning~\cite{yin2024improved}, we utilize the pre-trained U-Net encoder attached with trainable convolutional heads as our latent discriminator, which is initialized from our base ObjectClear~\cite{zhao2025objectclear} model. \par
\textbf{GAN Architecture.} As shown in Figure~\ref{fig:rad_pipeline}, since our U-Net backbone is built upon the sdxl-inpainting model~\cite{podell2023sdxl}, our distillation can be efficiently conducted in the latent space with much lower cost. Specifically, the 9-channel input configuration of the inpainting U-Net allows us to design different strategies for the generator and discriminator, so as to achieve tailored region-aware adversarial supervision. The input tuple can be formulated as $\langle \mathbf{z}_t, \mathbf{z}_{ref}, \mathbf{M} \rangle$, which formals our model's input as 
\begin{equation}
\mathbf{z}_{in} = \text{Concat}(\mathbf{z}_t, \mathbf{M}, \mathbf{z}_{ref}) \in \mathbb{R}^{9 \times H \times W},
\end{equation}
where $z_t$ denotes the noise map, $z_{ref}$ represents the reference image latent, and $M$ is the corresponding mask, which is the object-only mask $M_{{obj}}$ for generator and object effect mask $M_{\text{obj+effect}}$ for discriminator, respectively. 

\begin{figure}[t] 
\centering
\includegraphics[width=1\linewidth]{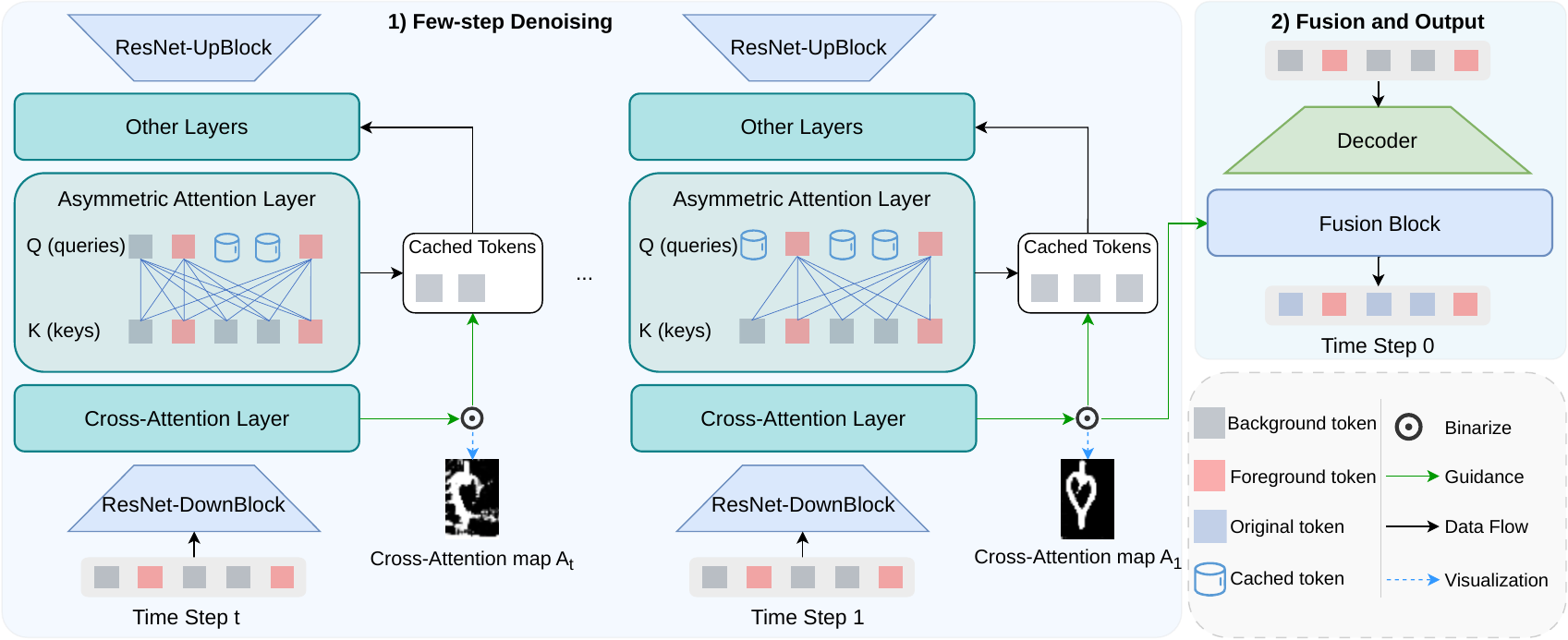} 
\vspace{-3.5mm}
\caption{Overview of FPAC. Foreground and background tokens are dynamically separated via cross-attention maps. In self-attention layers, cached tokens contribute only Key and Value (K, V) pairs without generating Queries (Q) or undergoing updates, while bypassing computation entirely in other modules (\textit{e.g.}, MLPs). The progressive refinement of cross-attention maps throughout the denoising process enables fine-grained token caching. Finally, the accurate attention map at the last timestep guides the seamless fusion of generated foreground tokens with original background tokens.}
\label{fig:FPAC}
\vspace{-6mm}
\end{figure}

\textbf{Region-aware Adversarial Distillation.} During distillation, the generator is conditioned on the same tight object mask $M_{{obj}}$ as during inference. On the other hand, we feed the discriminator with an expanded mask $M_{obj+eff}$, which encompasses both the primary object and its associated physical effects. This design makes $M_{obj+eff}$ act as a spatial prior to explicitly guide the discriminator to penalize any unnatural remnants (\textit{e.g.}, leftover shadows or reflections) within the effect regions. 

In summary, the additional five channels of the SDXL-inpainting U-Net enable both generator and discriminator to comprehensively understand the physical scene rather than merely blending the tight bounding box during adversarial distillation, ensuring seamless and artifact-free object removal. The final discriminator loss is defined as:
\begin{equation}
    \mathcal{L}_{adv}^{(D)} = \mathbb{E}_{\mathbf{z}_{real}, t}[\log \mathcal{D}(\mathbf{z}, \mathbf{z}_{ref}, \mathbf{M}_{\mathrm{obj+eff}}, t)] + \mathbb{E}_{\mathbf{z}_{pred}, t}[\log(1 - \mathcal{D}(\mathbf{z}_{pred}, \mathbf{z}_{ref}, \mathbf{M}_{\mathrm{obj+eff}}, t))],
\end{equation}
where $\mathcal{D}$ denotes the discriminator, $z_{pred}=\mathcal{G}(z_t, c, M_{obj}, t)$ is the generated image latent, and $\mathcal{G}$,  $z$, $z_t$, $z_{pred}$, $z_{ref}$ represent fine-tuned generator, ground truth latent, noised latent, reference latent, respectively. 

In terms of generator loss, we further introduce the commonly used perceptual loss LPIPS~\cite{zhang2018unreasonable} for semantic supervision and the standard diffusion $l_2$ loss for better convergence. Generator loss can be written as follows:
\begin{equation}
    L_{adv}^{(G)}=\lambda_1 \cdot \mathcal{L}_{lpips} + \lambda_2 \cdot ||z_{pred}-z||^2 + \lambda_3 \cdot \mathbb{E}_{z_t,c,t}[-\mathcal{D}(z_{pred}, z_{ref}, M_{obj}, t)],
\end{equation}
where $M_{obj}$ denotes tight object-only mask provided for inference. Additionally, to facilitate generator to focus on the region associated with object and the corresponding physical effect, we reintroduce the mask loss $L_{mask}$ proposed in ObjectClear~\cite{zhao2025objectclear}, where we extract the cross-attention maps $\textbf{A}$ corresponding to Clip encoder's visual embedding to get supervised with the foreground object-effect masks $M_{fg}$ in training dataset. The objective can be formalized as:
\begin{equation}
    \mathcal{L}_{mask}=\text{mean}(A[1-M_{fg}]) - \text{mean}(A[M_{fg}]).
\end{equation}\par
Through the above implementations, we succeed in guiding our distilled model to completely remove the object and effect to edit within a four-step inference process, while preserving the comparable ability to perceive and obtain region-aware attention maps for our further foreground-prioritized asymmetric attention \& caching.

\begin{table*}[t]
\centering
\caption{Quantitative comparison on the OBER-Test (512$\times$512) and RORD-Val (960$\times$540) datasets. The best and second-best performances are marked in \colorbox{rred}{red} and \colorbox{oorange}{orange}, respectively.}
\vspace{-2mm}
\label{tab:removal}
\renewcommand{\arraystretch}{1.35} 
\resizebox{\textwidth}{!}{%
\setlength{\tabcolsep}{3pt}
\begin{tabular}{l rcccc rcccc}
\toprule
\multirow{2}{*}{Method} & \multicolumn{5}{c}{OBER-Test (512$\times$512)} & \multicolumn{5}{c}{{RORD-Val (960$\times$540)}} \\
\cmidrule(lr){2-6} \cmidrule(lr){7-11}
& FLOPs (T) $\downarrow$ & LPIPS $\downarrow$ & LPIPS-Local $\downarrow$ & PSNR $\uparrow$ & PSNR-mask $\uparrow$ 
& FLOPs (T) $\downarrow$ & LPIPS $\downarrow$ & LPIPS-Local $\downarrow$ & PSNR $\uparrow$ & PSNR-mask $\uparrow$ \\
\midrule
SDXL-INP~\cite{podell2023sdxl} & 256.9 & 0.1310 & 0.4409 & 22.43 & 12.26 & 256.9 & 0.1808 & 0.3995 & 20.23 & 12.26 \\
PowerPaint~\cite{zhuang2024task} & 122.4 & 0.1583 & 0.3489 & 23.02 & 15.35 & 122.4 & 0.1903 & 0.2983 & 21.76 & 16.78 \\
GeoRemover~\cite{zhu2025georemover} & 3,897.4 & 0.1454 & 0.1859 & 24.67 & 21.65 & 3,897.4 & 0.1200 & \colorbox{oorange}{0.2094} & 24.49 & 19.41 \\
DesignEdit~\cite{jia2024designedit} & 1,727.2 & 0.1302 & 0.2548 & 26.39 & 20.60 & 1,727.2 & 0.1937 & 0.3101 & 23.27 & \colorbox{oorange}{19.68} \\
CLIPAway~\cite{ekin2024clipaway} & \colorbox{oorange}{80.3} & 0.1328 & 0.3614 & 22.22 & 14.43 & \colorbox{oorange}{80.3} & 0.1620 & 0.3015 & 21.10 & 15.63 \\
Omnieraser~\cite{wei2025omnieraser} & 2,097.3 & 0.2102 & 0.2630 & 24.35 & 21.29 & 2,097.3 & 0.2289 & 0.3030 & 22.11 & 18.63 \\
Attentive Eraser~\cite{sun2025attentive}& 549.2 & 0.0809 & 0.2436 & 27.17 & 20.85 & 549.2 & 0.1399 & 0.3014 & 24.10 & 17.84 \\
RORem~\cite{li2025rorem} & 331.3 & 0.0979 & 0.2391 & 26.21 & 19.14 & 331.3 & \colorbox{oorange}{0.0979} & 0.2390 & \colorbox{oorange}{26.21} & 19.14 \\
Omnipaint~\cite{yu2025omnipaint} & 1,057.5 & \colorbox{oorange}{0.0521} & \colorbox{rred}{0.1299} & \colorbox{oorange}{29.05} & \colorbox{oorange}{23.56} & 2,015.5 & 0.1178 & 0.2380 & 22.74 & 17.62 \\
FlashClear (ours) & \colorbox{rred}{8.6} & \colorbox{rred}{0.0351} & \colorbox{oorange}{0.1396} & \colorbox{rred}{33.05} & \colorbox{rred}{24.12} & \colorbox{rred}{15.6} & \colorbox{rred}{0.0698} & \colorbox{rred}{0.1996} & \colorbox{rred}{27.66} & \colorbox{rred}{20.02} \\
\bottomrule
\end{tabular}
}
\vspace{-8mm}
\end{table*}

\vspace{-2mm}
\subsection{Foreground-Prioritized Asymmetric Attention and Caching (FPAC)} \label{section-fpac}
\vspace{-2mm}

While RAD successfully compresses the denoising trajectory into minimal steps, substantial spatial redundancy still exists during the generation process. Most background regions remain invariant, whereas the target object and its complex physical effects require intensive recomputation. However, exploiting this spatial redundancy is non-trivial, as user-provided masks cannot explicitly cover these object-induced effects. Furthermore, existing text-to-image caching methods are poorly suited for this scenario. They often incur substantial computational overhead (\textit{e.g.}, SiTo~\cite{zhang2025training}), lack token-level sparsity in self-attention layers (\textit{e.g.}, ToCa~\cite{zou2024accelerating}), or rely on temporal extrapolations that fail in extreme few-step regimes (\textit{e.g.}, HiCache~\cite{feng2025hicache}).

To overcome these limitations and leverage the spatial prior captured by our distilled model, we propose Foreground-Prioritized Asymmetric Attention and Caching (FPAC), with two core components:

\begin{wrapfigure}{r}{0.5\textwidth}
    \centering
    \vspace{-4mm}

    \makebox[\linewidth][c]{%
    \begin{subfigure}[b]{0.19\linewidth}
        \centering
        \includegraphics[width=\linewidth]{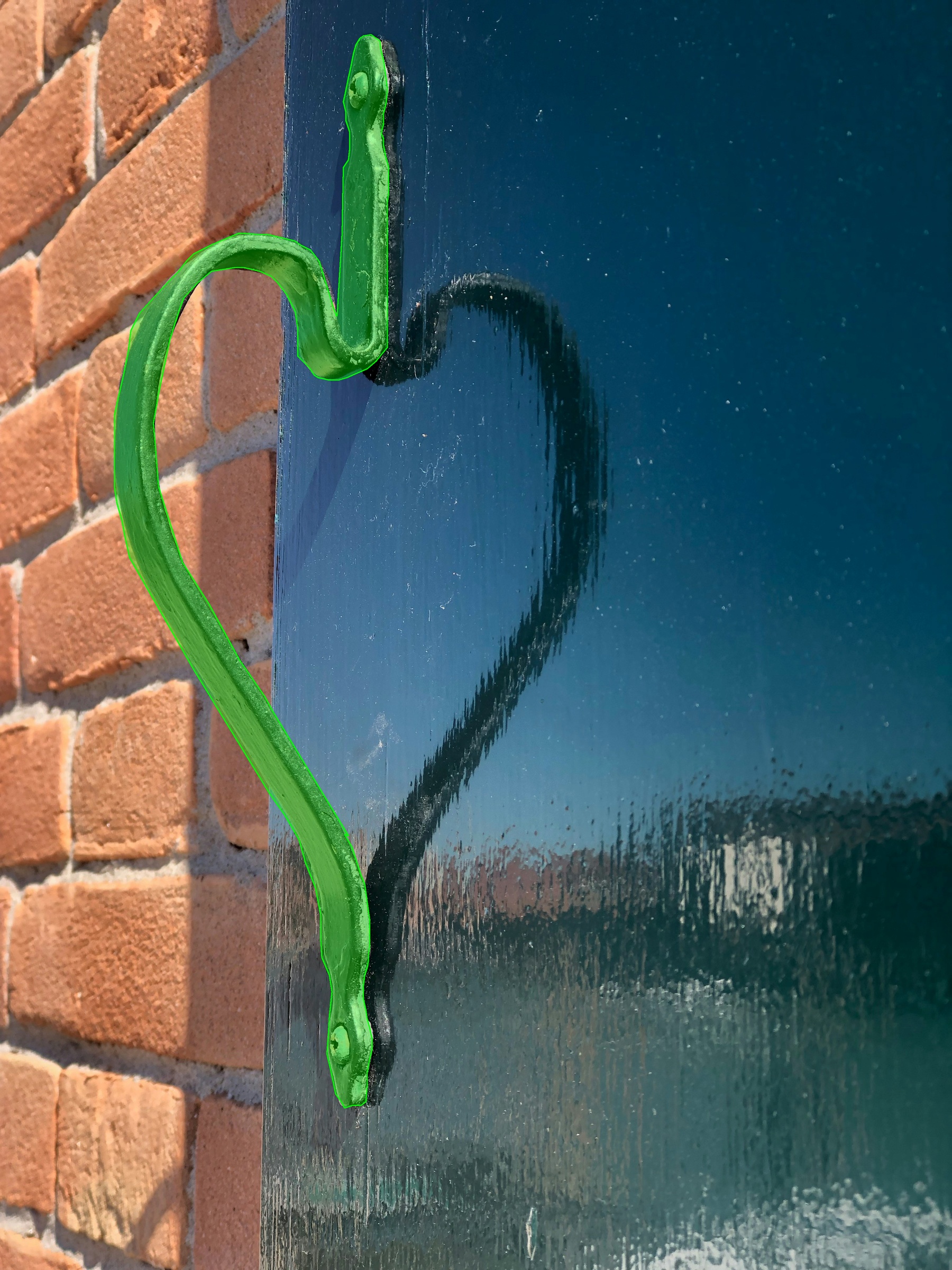}
        \caption*{\footnotesize \\\vspace{-15pt} Input}
    \end{subfigure}%
    \hfill%
    \begin{subfigure}[b]{0.19\linewidth}
        \centering
        \includegraphics[width=\linewidth]{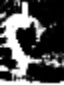}
        \caption*{\footnotesize \\\vspace{-15pt} $\mathcal{M}_1$}
    \end{subfigure}%
    \hfill%
    \begin{subfigure}[b]{0.19\linewidth}
        \centering
        \includegraphics[width=\linewidth]{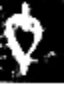}
        \caption*{\footnotesize \\\vspace{-15pt} $\mathcal{M}_2$}
    \end{subfigure}%
    \hfill%
    \begin{subfigure}[b]{0.19\linewidth}
        \centering
        \includegraphics[width=\linewidth]{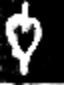}
        \caption*{\footnotesize \\\vspace{-15pt} $\mathcal{M}_3$}
    \end{subfigure}%
    \hfill%
    \begin{subfigure}[b]{0.19\linewidth}
        \centering
        \includegraphics[width=\linewidth]{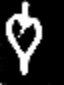}
        \caption*{\footnotesize \\\vspace{-15pt} $\mathcal{M}_4$}
    \end{subfigure}%
    }

    \vspace{-2mm}
    \caption{\small The mask is colored in \textcolor{green}{green}. The binary spatial map $\mathcal{M}_i$ is refined gradually.}
    \label{fig:fpac_vis}
    \vspace{-5mm}
\end{wrapfigure}

\textbf{Foreground-Prioritized Caching.} As illustrated in Figure~\ref{fig:FPAC}, we leverage cross-attention maps to establish a dynamic threshold, yielding a binary spatial mask $\mathcal{M} \in \{0, 1\}^N$, where $\mathcal{M}_i = 0$ identifies background tokens and $\mathcal{M}_i = 1$ denotes the foreground tokens. As shown in Figure~\ref{fig:fpac_vis}, $\mathcal{M}_i$ is refined gradually in the process. By utilizing $\mathcal{M}$, we elegantly cache tokens that belong to the background regions. By incorporating an attention-guided fusion mechanism at the final stage---formulated as $\mathbf{F}_{fused} = \alpha \odot \mathbf{F}_{pred} + (1 - \alpha) \odot \mathbf{F}_{orig}$, where $\alpha$ represents the attention-derived fusion weight and $\odot$ denotes element-wise multiplication---we theoretically achieve lossless caching of background tokens.

\textbf{Asymmetric Attention.} To mitigate the computational bottleneck of the attention mechanism and prevent error accumulation caused by local zero-padding, we introduce Asymmetric Attention. Let $\mathbf{Q}, \mathbf{K}, \mathbf{V} \in \mathbb{R}^{N \times d}$ denote the standard Query, Key, and Value matrices. As illustrated in Figure~\ref{fig:FPAC}, we define a pruned query matrix $\mathbf{Q}'$ such that for the $i$-th token, $\mathbf{q}'_i = \mathbf{q}_i$ if $\mathcal{M}_i = 1$ (foreground), and $\mathbf{q}'_i = \mathbf{0}$ otherwise. Equivalently, this can be expressed as $\mathbf{Q}' = \text{diag}(\mathcal{M})\mathbf{Q}$. The asymmetric attention is then computed as $\mathbf{O} = \text{Softmax}\left((\mathbf{Q}'\mathbf{K}^T)/\sqrt{d}\right) \mathbf{V}$.

Through this formulation, the foreground tokens actively query the continuously updated $\mathbf{K}$ and $\mathbf{V}$ from the entire image space. For the background tokens ($\mathcal{M}_i = 0$), the redundant attention computation is bypassed, and their features are directly populated from the layer cache $\mathbf{H}^{cache}$. Specifically, the final token representations $\mathbf{H}^{out}$ are obtained via:
\begin{equation}
\mathbf{H}^{out}_i = \begin{cases}
\mathbf{O}_i, & \text{if } \mathcal{M}_i = 1 \\
\mathbf{H}^{cache}_i, & \text{if } \mathcal{M}_i = 0.
\end{cases}
\end{equation}

This structural asymmetry ensures that the foreground regions to be synthesized can still attend to the most up-to-date background context, while the background itself naturally bypasses redundant computational updates and remains strictly unchanged before the final fusion step.

Built on FlashClear, our method (denoted as FlashClear-C) sets a new SOTA for object removal. It achieves training-free, lossless complexity reduction, effectively complementing few-step distillation.

\vspace{-2mm}
\section{Experiments}
\vspace{-3mm}
\subsection{Experiment Settings}
\vspace{-2mm}
\textbf{Implementation Details.} Our proposed FlashClear is built upon SDXL~\cite{podell2023sdxl} architecture. The model is fine-tuned using Low-Rank Adaptation (LoRA ~\cite{hu2022lora}) with $\text{rank}$$=$$256$ and $\alpha$=$256$. We train the network for 10,000 steps using the AdamW optimizer ($\beta_1=0.9$, $\beta_2=0.999$) with a learning rate of $10^{-5}$ for both the generator and discriminator. Training is conducted on 2 NVIDIA A800 GPUs with a total batch size of 16 in bfloat16 mixed precision. Inference is conducted on a single NVIDIA A6000 GPU. The model is configured with a fixed four-timestep scheduler, and CFG is disabled. 

\textbf{Evaluation Data.} We evaluate on two benchmarks: the \textit{OBER-Test}~\cite{zhao2025objectclear} dataset (163 samples) and the \textit{RORD-Val}~\cite{sagong2022rord} dataset (343 samples selected in~\cite{zhao2025objectclear}). \textit{OBER-Test} evaluates the general object removal capability at 512$\times$512 resolution, whereas \textit{RORD-Val} contains higher-resolution images (960$\times$540) with complex scenes, providing a challenging testbed for practical applications.

\textbf{Evaluation Metrics.} For computational efficiency, measuring raw latency can be easily confounded by hardware disparities, system states, and varying low-level kernel implementations (\textit{e.g.}, FlashAttention or xformers). To ensure a fair and objective comparison, we adopt the theoretical denoising FLOPs as the primary criterion for acceleration in our main text. For visual quality assessment, we employ the widely used perceptual metric LPIPS~\cite{zhang2018unreasonable} and the pixel-level similarity metric PSNR. Furthermore, object removal heavily relies on the coherence between the manipulated area and its surroundings. To rigorously investigate the local restoration quality and foreground-background consistency, we introduce PSNR-mask and LPIPS-Local. These localized metrics specifically evaluate the fidelity of the masked regions and assess the naturalness of the transition boundaries.

\begin{figure}[t] 
    \centering
    
    \setlength{\tabcolsep}{0.5pt}      
    \renewcommand{\arraystretch}{1.2}

    \begin{tabular}{@{}ccccccc@{}}

        \includegraphics[width=0.14\linewidth]{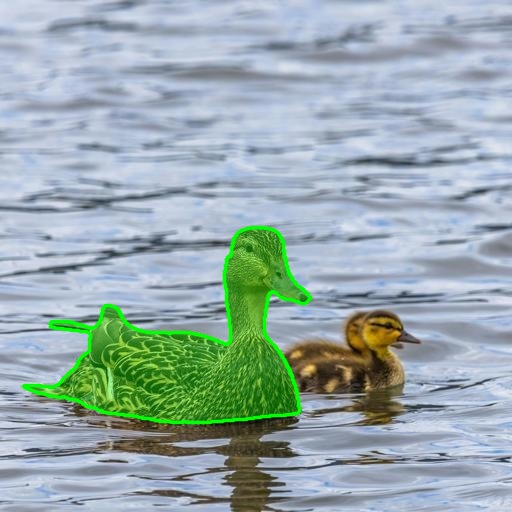} &
        \includegraphics[width=0.14\linewidth]{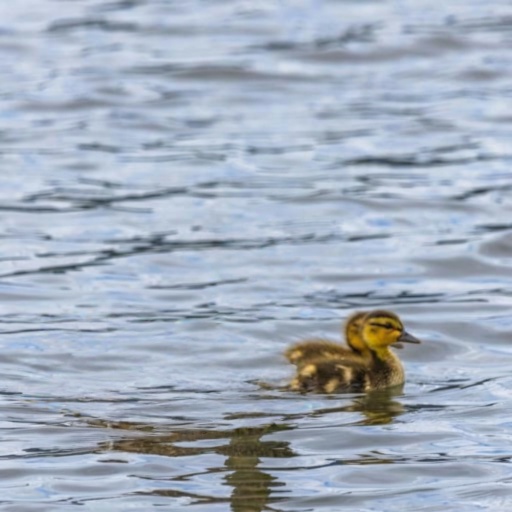} &
        \includegraphics[width=0.14\linewidth]{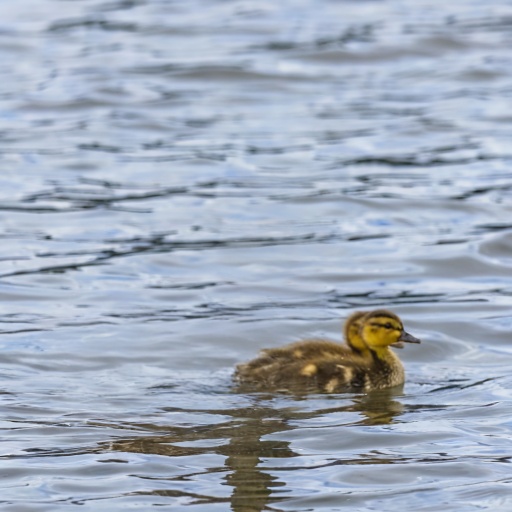} &
        \includegraphics[width=0.14\linewidth]{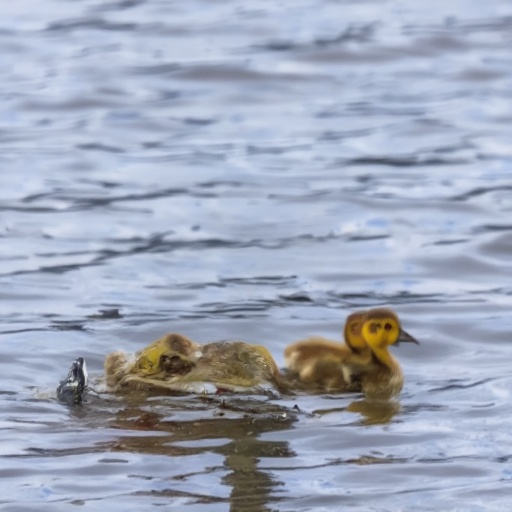} &
        \includegraphics[width=0.14\linewidth]{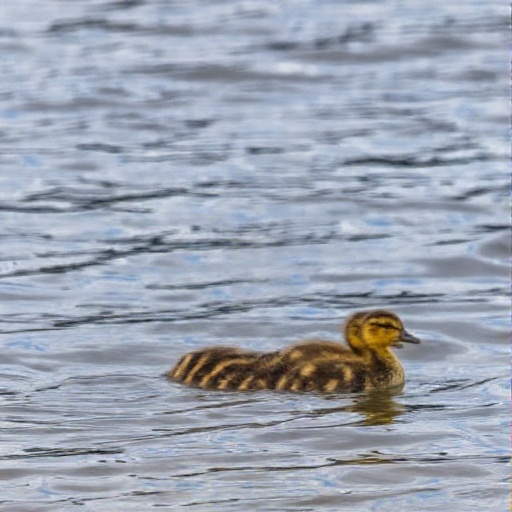} &
        \includegraphics[width=0.14\linewidth]{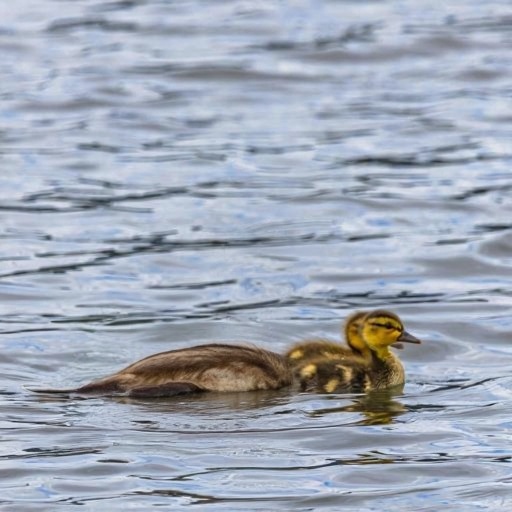} &
        \includegraphics[width=0.14\linewidth]{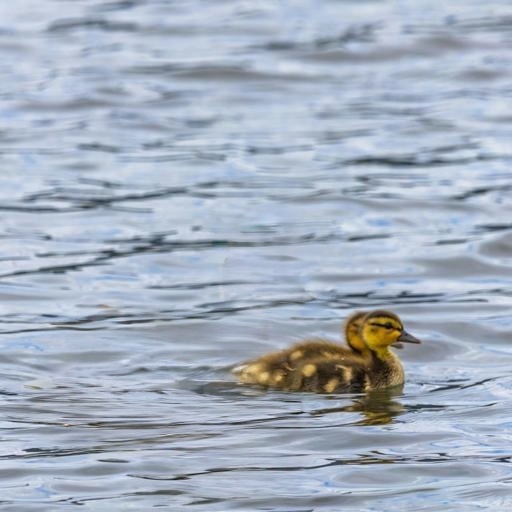} \\[-2pt]

        \includegraphics[width=0.14\linewidth]{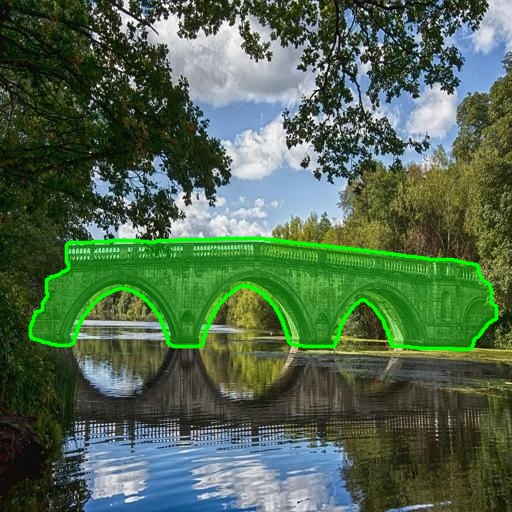} &
        \includegraphics[width=0.14\linewidth]{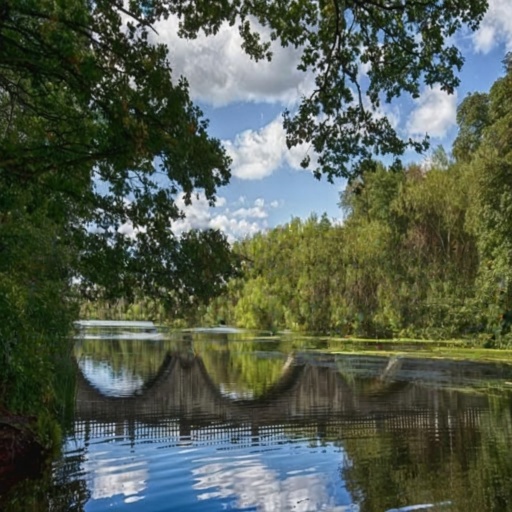} &
        \includegraphics[width=0.14\linewidth]{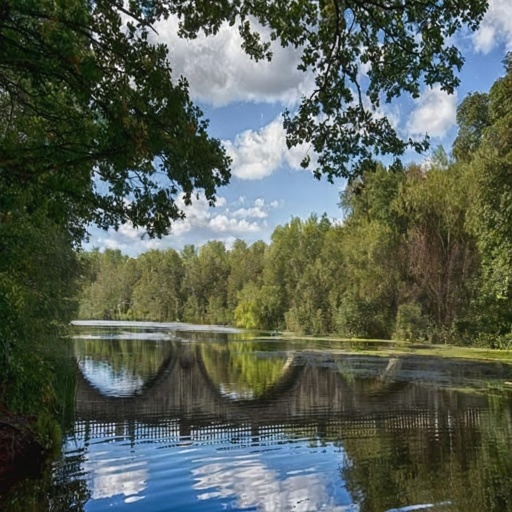} &
        \includegraphics[width=0.14\linewidth]{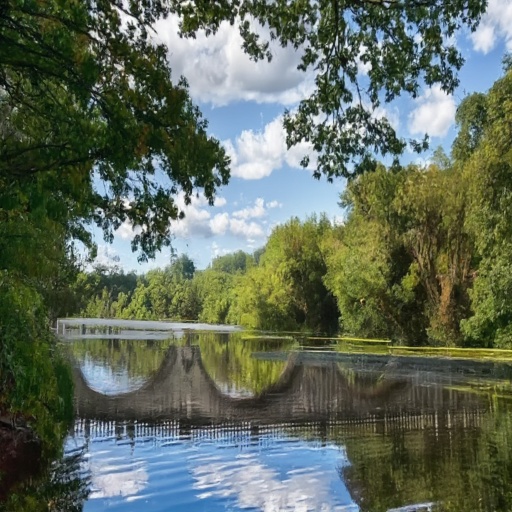} &
        \includegraphics[width=0.14\linewidth]{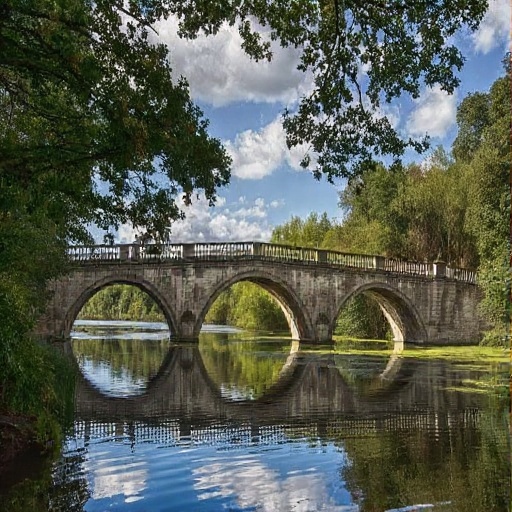} &
        \includegraphics[width=0.14\linewidth]{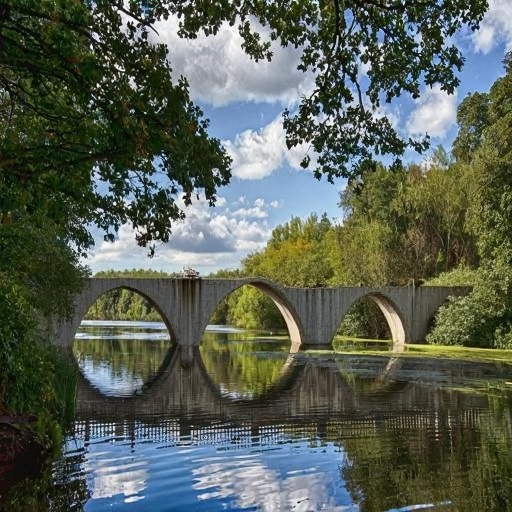} &
        \includegraphics[width=0.14\linewidth]{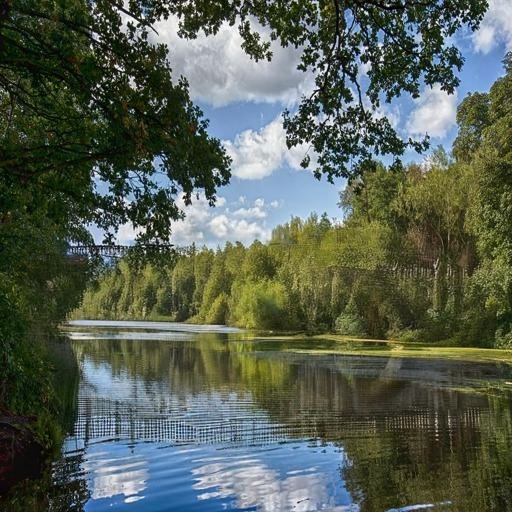} \\[-2pt]

       \includegraphics[width=0.14\linewidth]{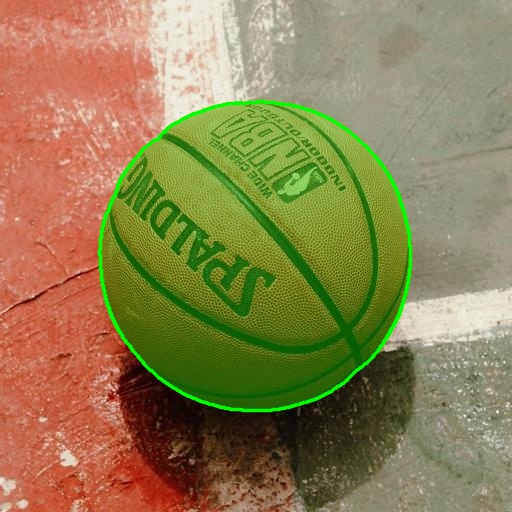} &
        \includegraphics[width=0.14\linewidth]{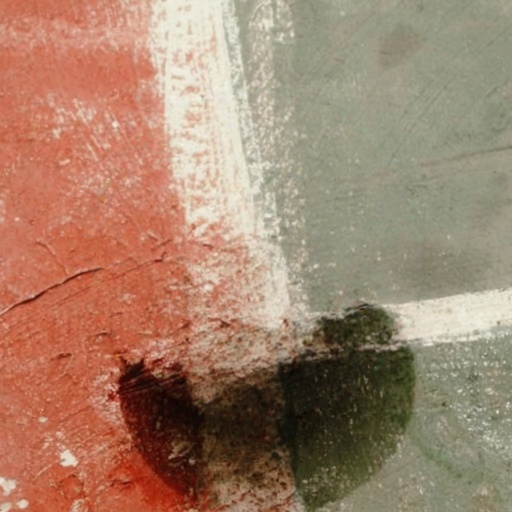} &
        \includegraphics[width=0.14\linewidth]{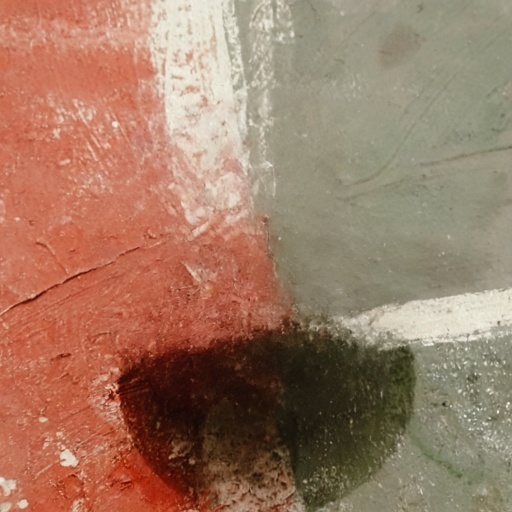} &
        \includegraphics[width=0.14\linewidth]{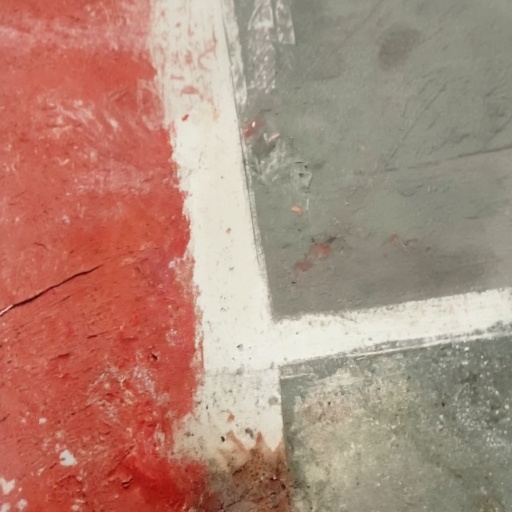} &
        \includegraphics[width=0.14\linewidth]{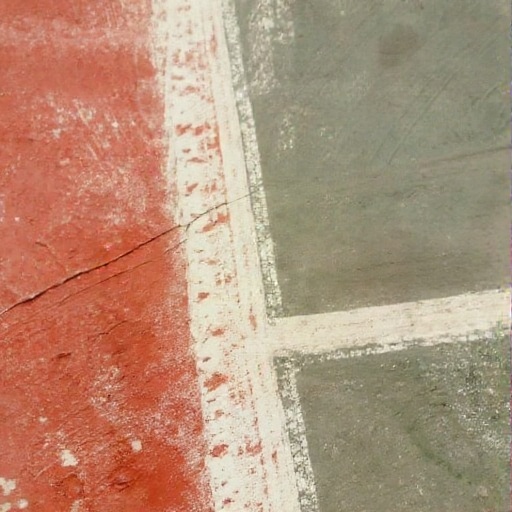} &
        \includegraphics[width=0.14\linewidth]{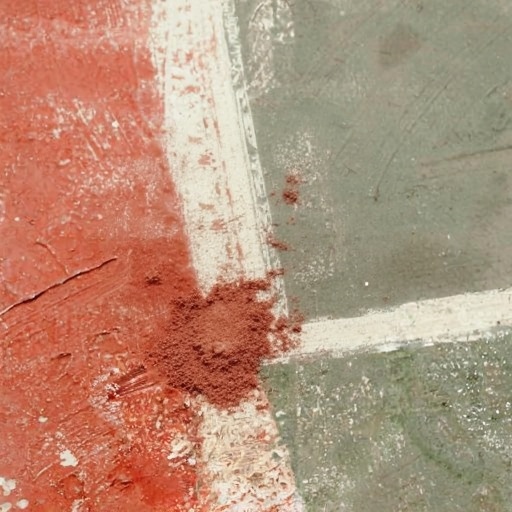} &
        \includegraphics[width=0.14\linewidth]{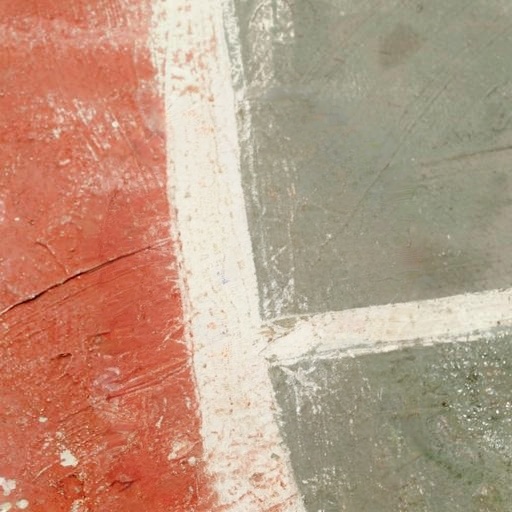} \\[-2pt]

        \includegraphics[width=0.14\linewidth]{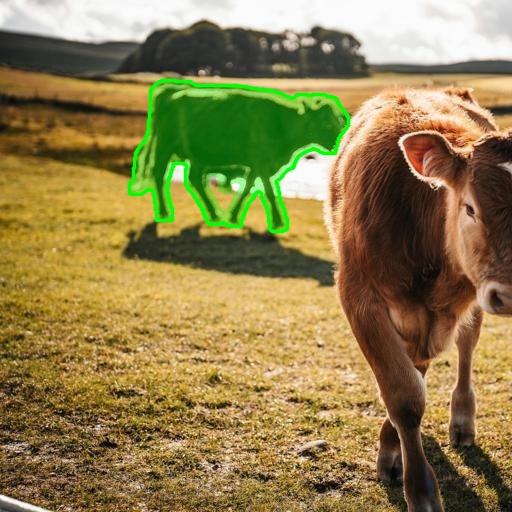} &
        \includegraphics[width=0.14\linewidth]{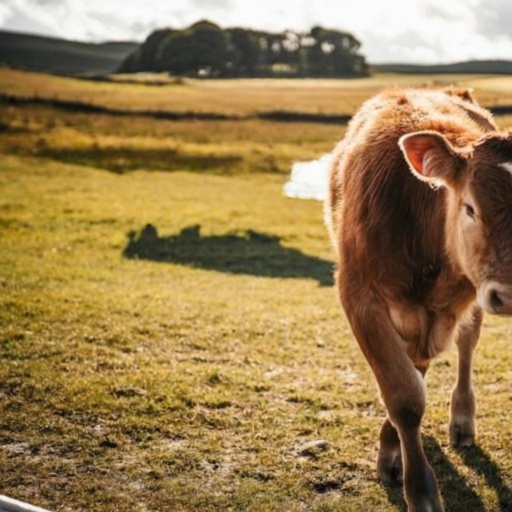} &
        \includegraphics[width=0.14\linewidth]{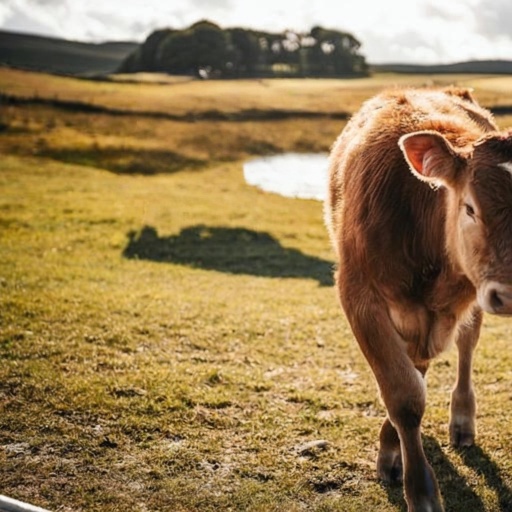} &
        \includegraphics[width=0.14\linewidth]{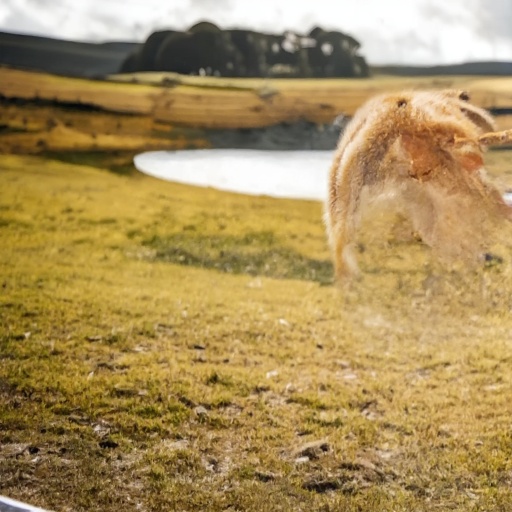} &
        \includegraphics[width=0.14\linewidth]{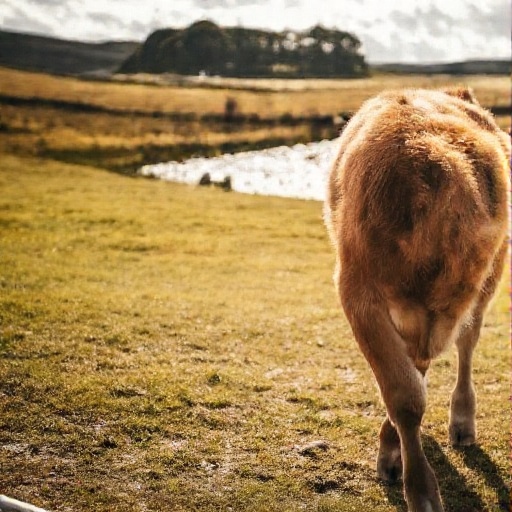} &
        \includegraphics[width=0.14\linewidth]{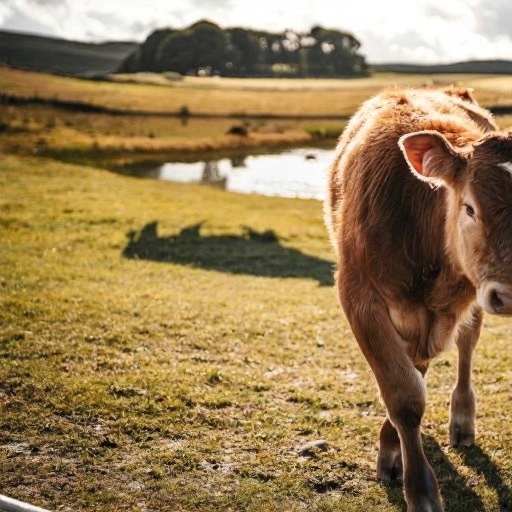} &
        \includegraphics[width=0.14\linewidth]{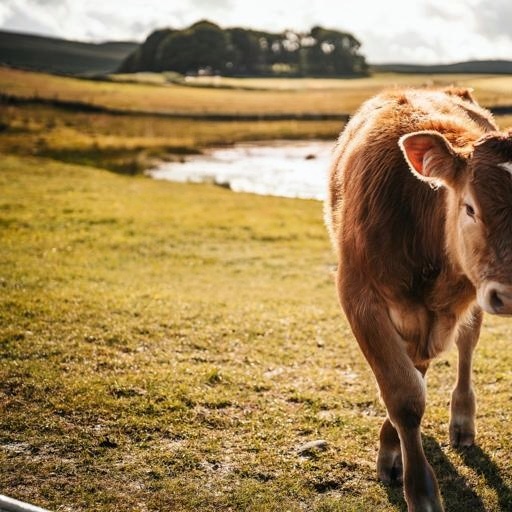} \\[-2pt]

        \includegraphics[width=0.14\linewidth]{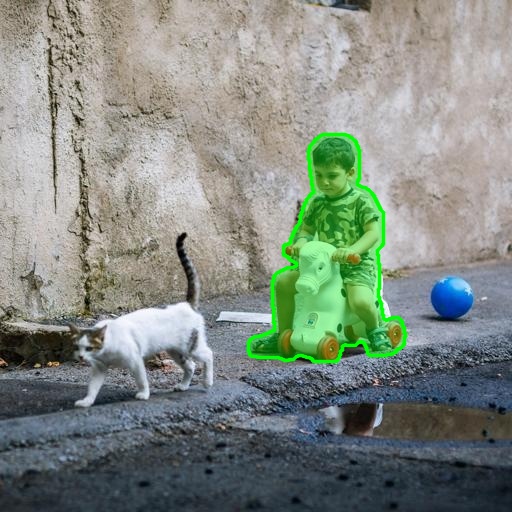} &
        \includegraphics[width=0.14\linewidth]{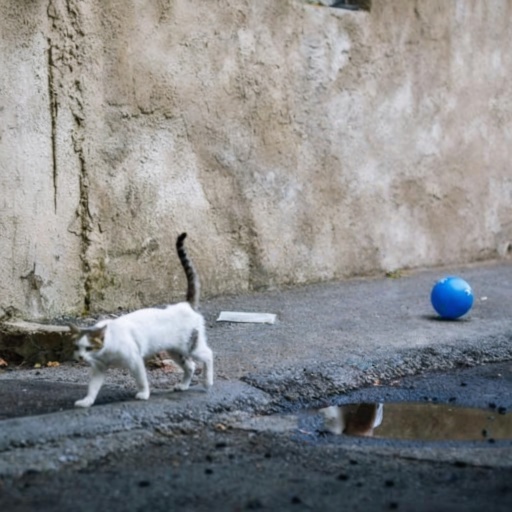} &
        \includegraphics[width=0.14\linewidth]{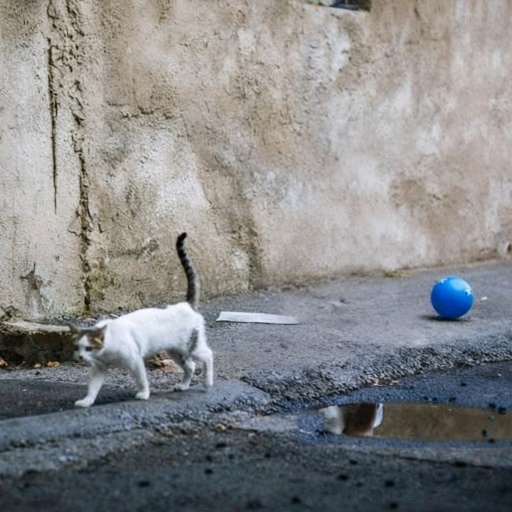} &
        \includegraphics[width=0.14\linewidth]{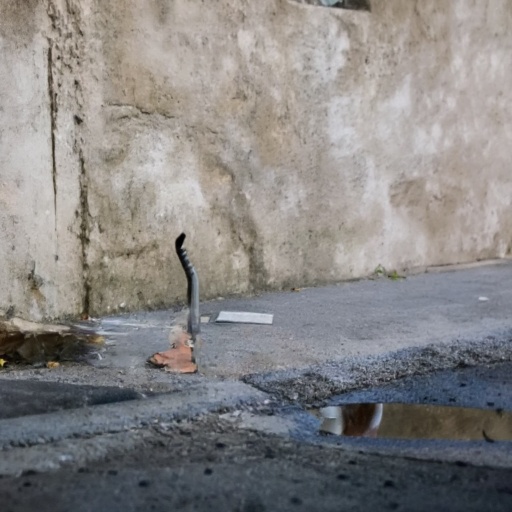} &
        \includegraphics[width=0.14\linewidth]{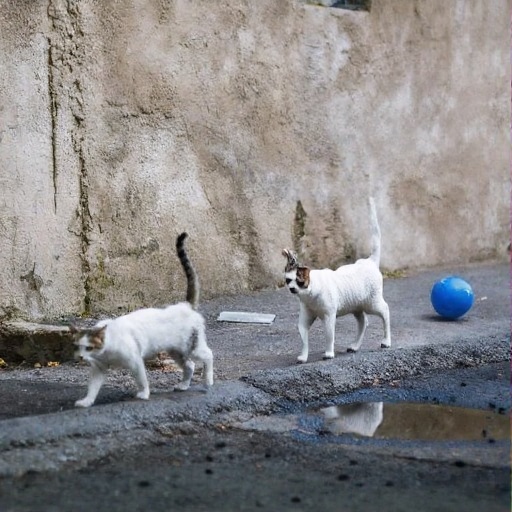} &
        \includegraphics[width=0.14\linewidth]{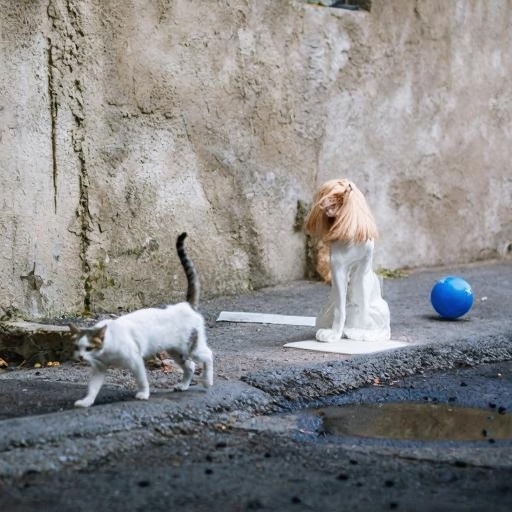} &
        \includegraphics[width=0.14\linewidth]{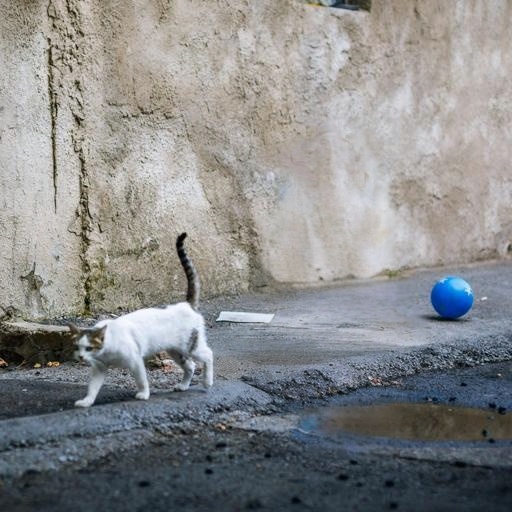} 
        \\[-2pt]

        \footnotesize Input w/ Mask & 
        \footnotesize Atten.Eraser~\cite{sun2025attentive} & 
        \footnotesize RORem~\cite{li2025rorem} & 
        \footnotesize OmniEraser~\cite{wei2025omnieraser} & 
        \footnotesize GeoRemover\cite{zhu2025georemover} & 
        \footnotesize OmniPaint~\cite{yu2025omnipaint} & 
        \footnotesize FlashClear (ours) \\
        
    \end{tabular}
    
\vspace{-2mm} 
\caption{Qualitative comparison of object removal methods. Our method shows stronger object removal ability and can generate more visually reasonable results compared to other methods.}
\label{fig:visual comparison}
\vspace{-5mm}
\end{figure}

\newcommand{\acccasedir}{case/Acc_latex_case}
\newlength{\accfigw}
\setlength{\accfigw}{0.121\textwidth}
\newcommand{\accimg}[1]{\makebox[\accfigw][c]{\includegraphics[width=\accfigw]{\acccasedir/#1}}}
\newcommand{\acclabel}[1]{\makebox[\accfigw][c]{\scriptsize #1}}

\begin{figure*}[t]
    \centering
    \setlength{\tabcolsep}{0.5pt}
    \renewcommand{\arraystretch}{0.4}
    \begin{tabular}{@{}cccccccc@{}}
        \accimg{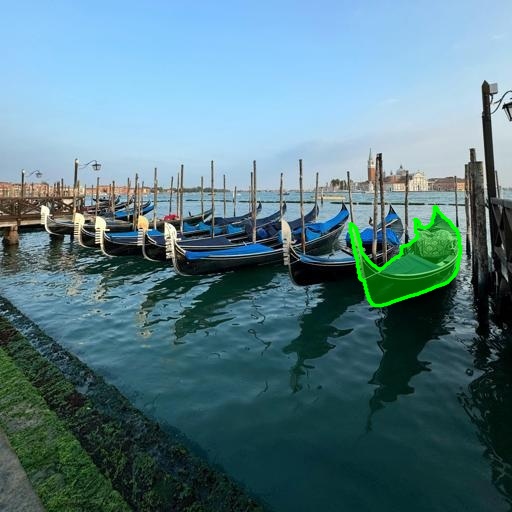} &
        \accimg{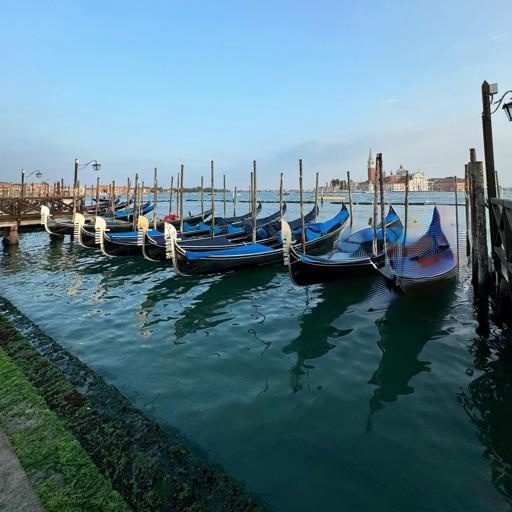} &
        \accimg{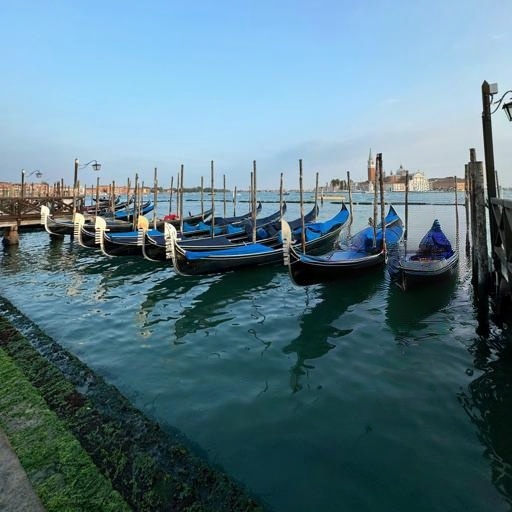} &
        \accimg{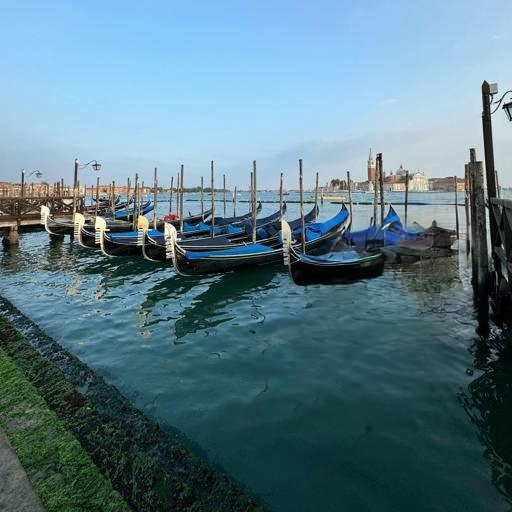} &
        \accimg{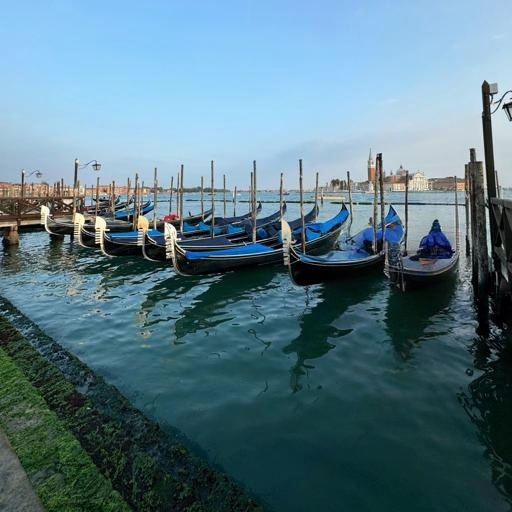} &
        \accimg{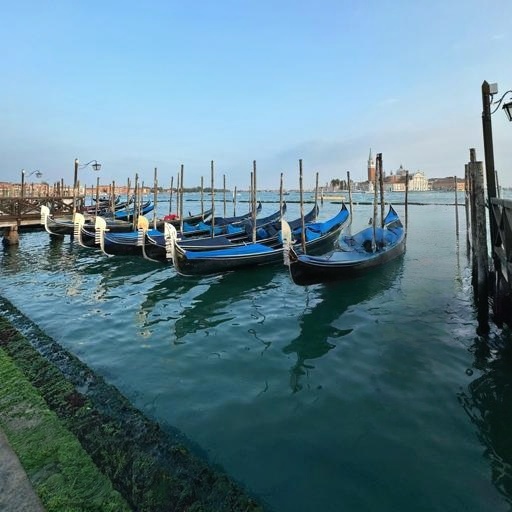} &
        \accimg{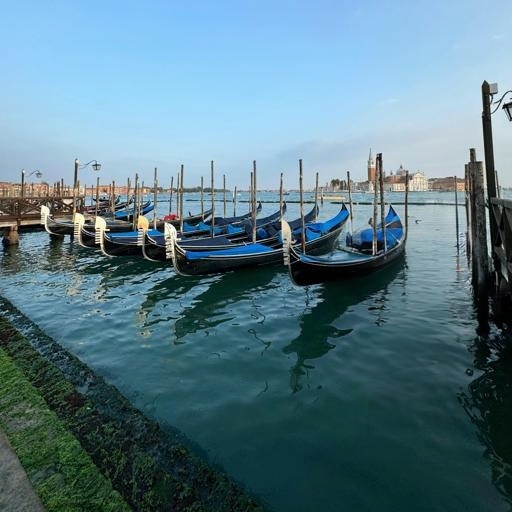} &
        \accimg{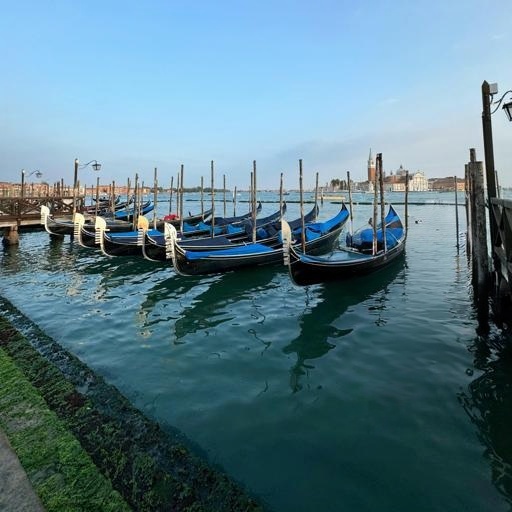} \\[0pt]
        \accimg{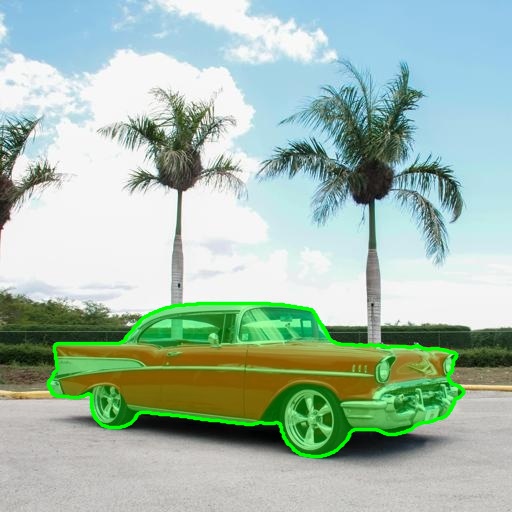} &
        \accimg{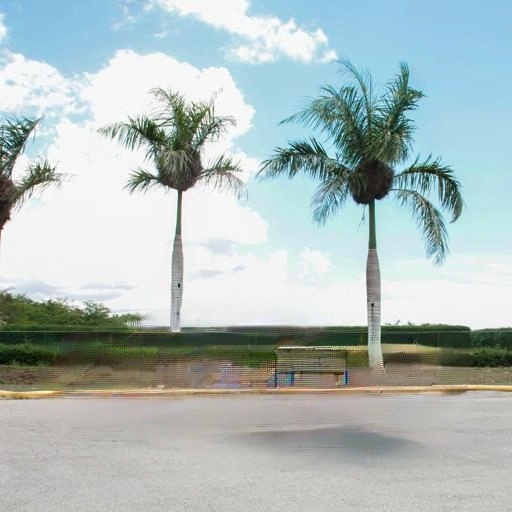} &
        \accimg{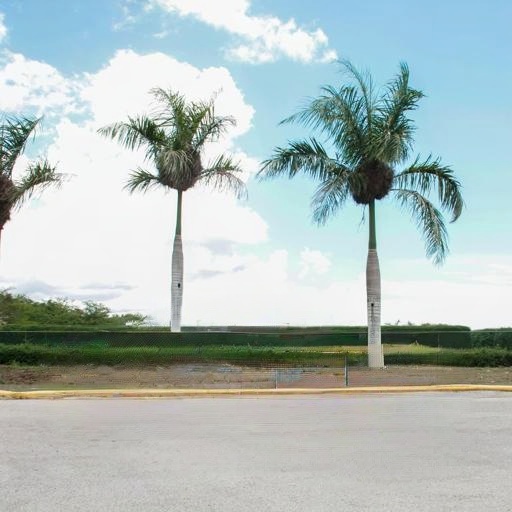} &
        \accimg{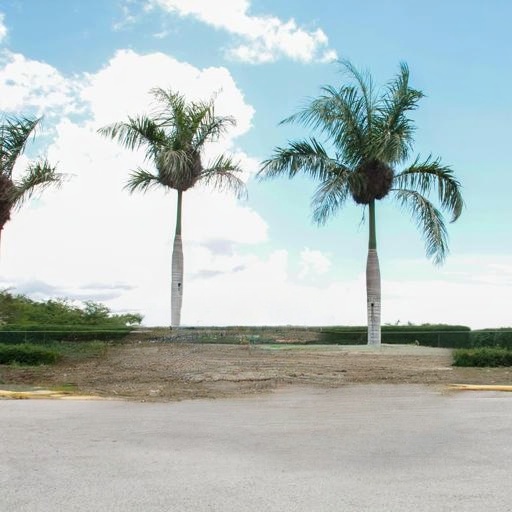} &
        \accimg{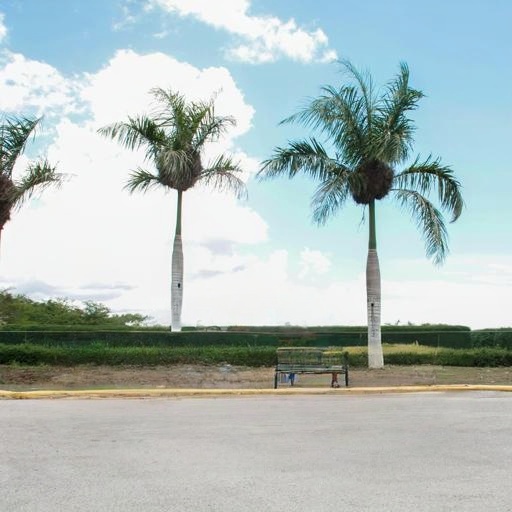} &
        \accimg{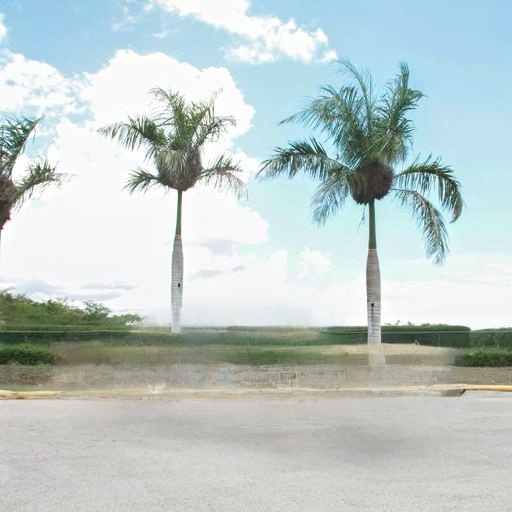} &
        \accimg{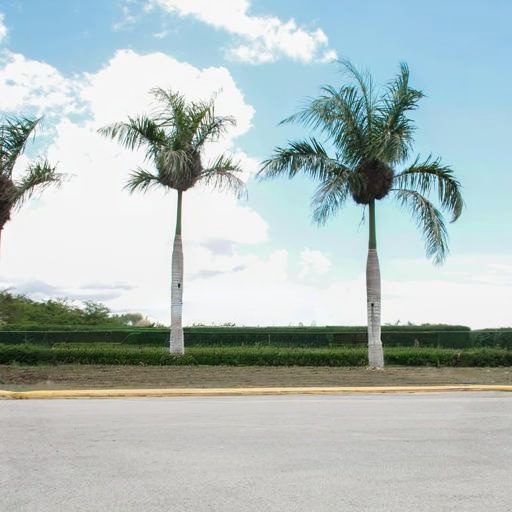} &
        \accimg{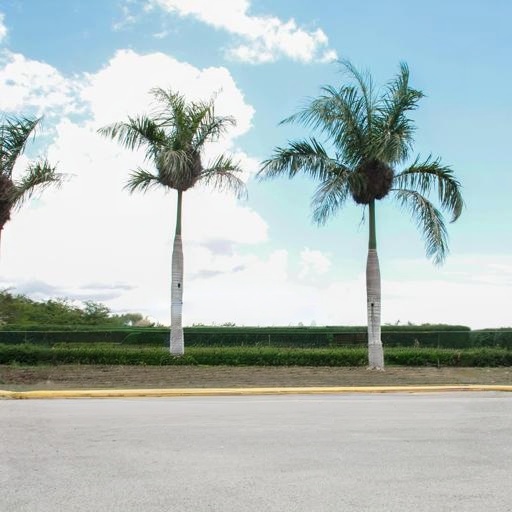} \\[0pt]
        \accimg{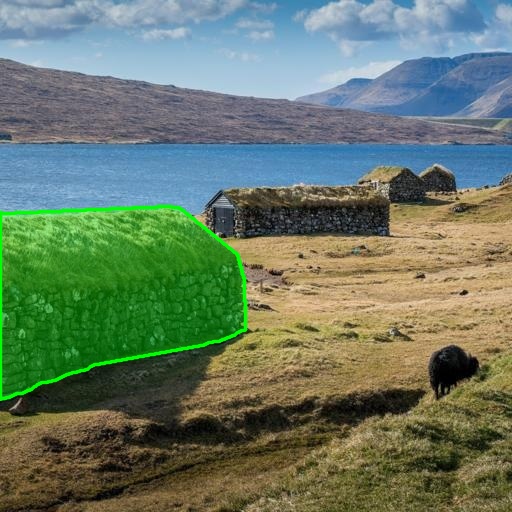} &
        \accimg{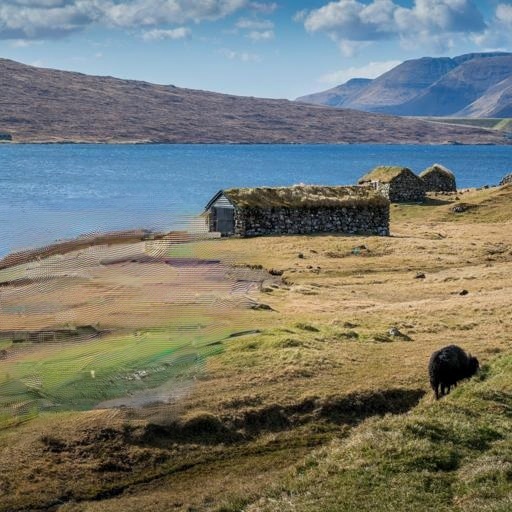} &
        \accimg{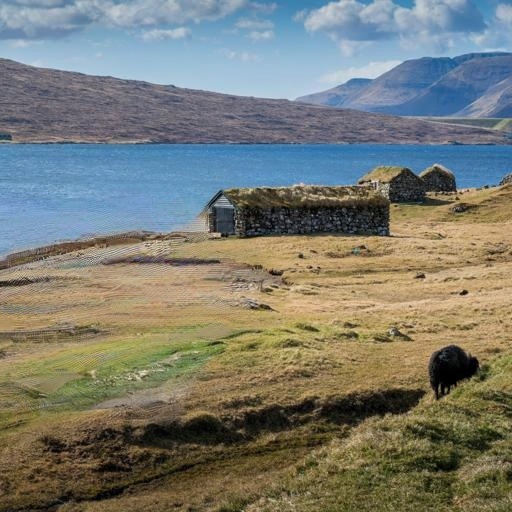} &
        \accimg{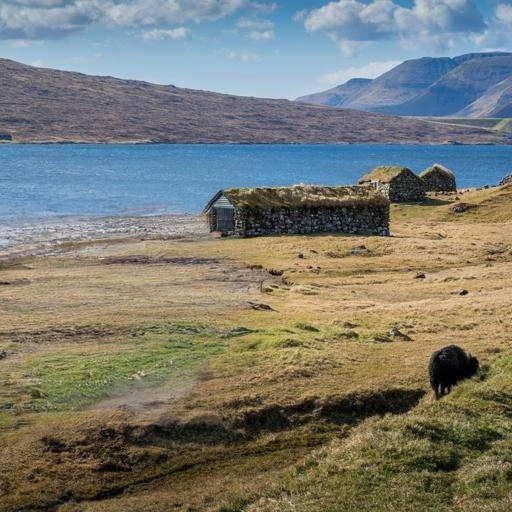} &
        \accimg{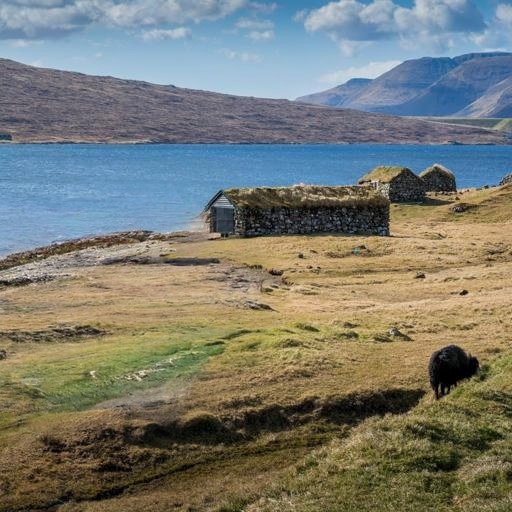} &
        \accimg{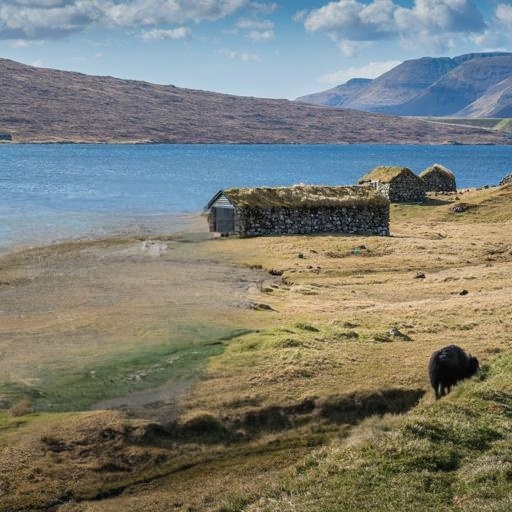} &
        \accimg{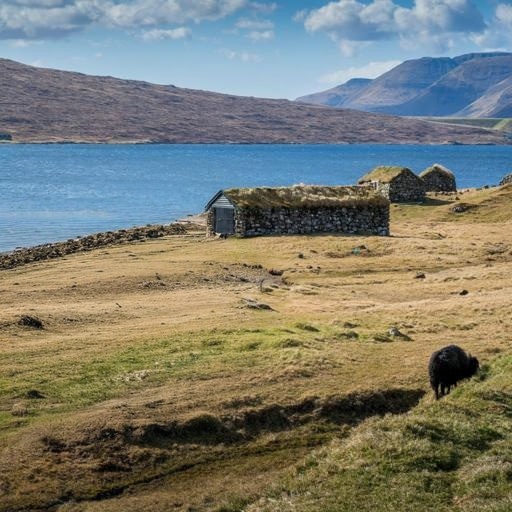} &
        \accimg{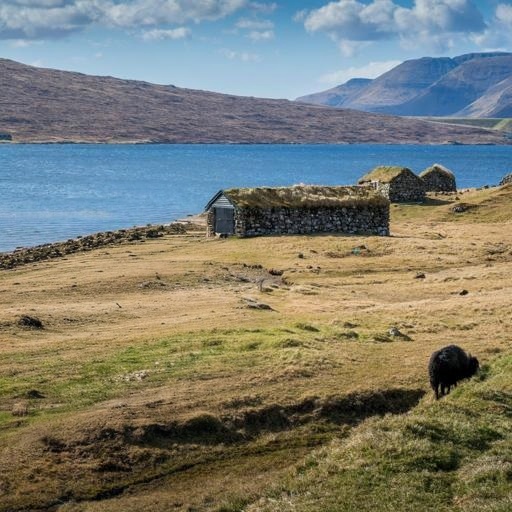} \\[0pt]
        \accimg{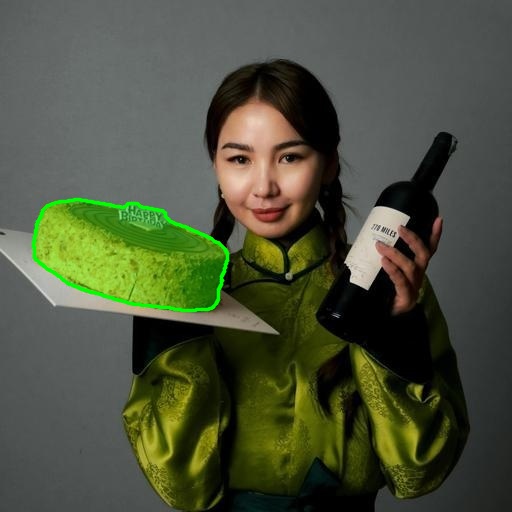} &
        \accimg{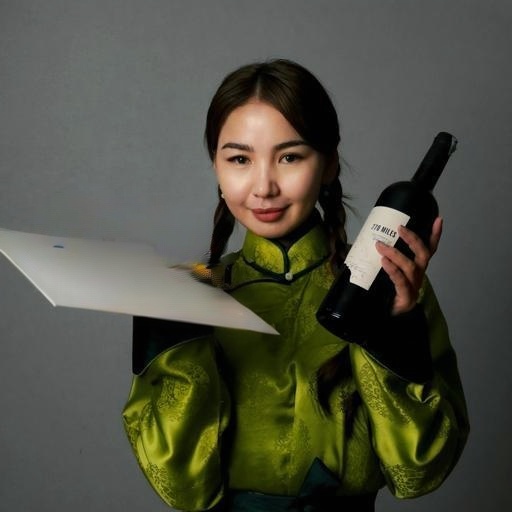} &
        \accimg{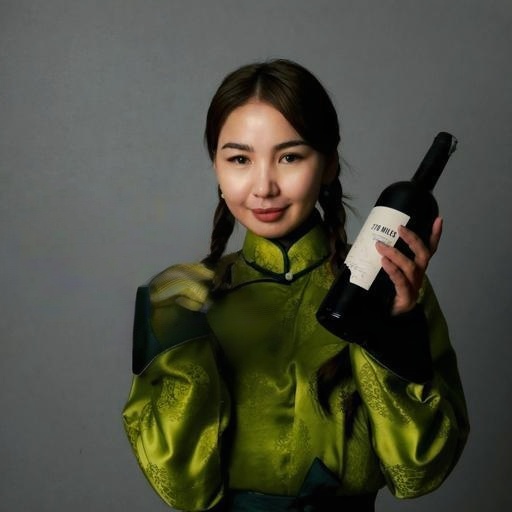} &
        \accimg{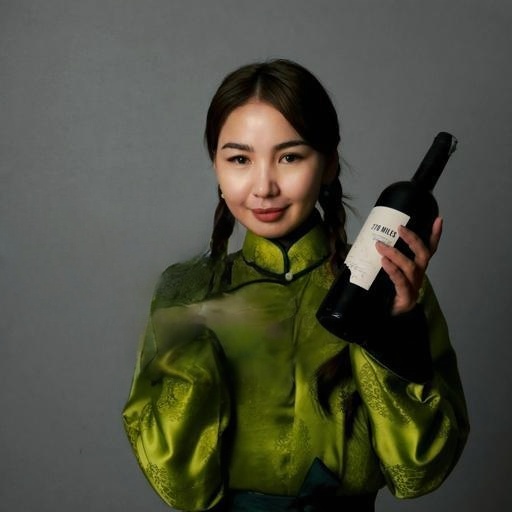} &
        \accimg{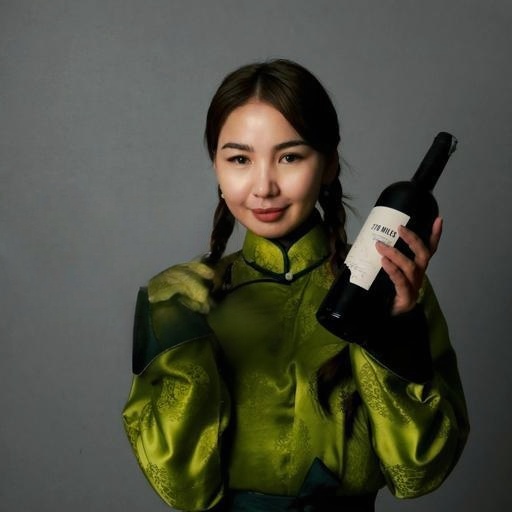} &
        \accimg{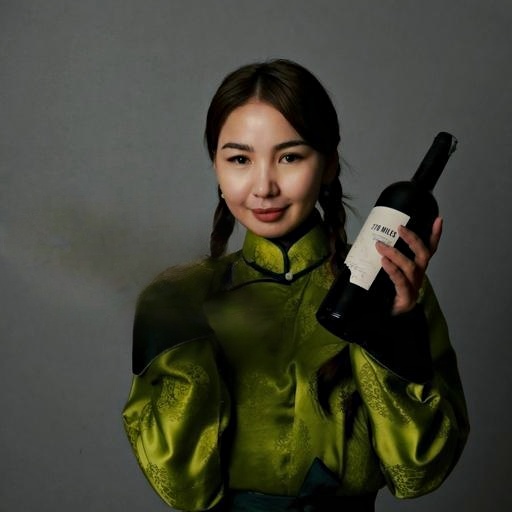} &
        \accimg{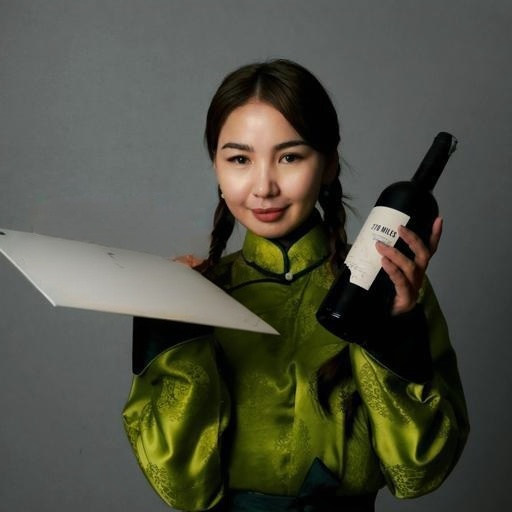} &
        \accimg{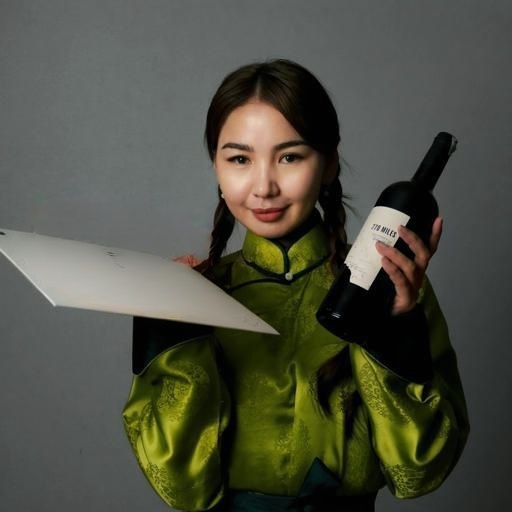} \\[4pt]
        \acclabel{Image w/ mask} &
        \acclabel{20\% step} &
        \acclabel{ToCa~\cite{zou2024accelerating}} &
        \acclabel{SiTo~\cite{zhang2025training}} &
        \acclabel{HiCache~\cite{feng2025hicache}} &
        \acclabel{QuantCache~\cite{wu2025quantcache}} &
        \acclabel{FlashClear-C} &
        \acclabel{FlashClear}
    \end{tabular}
    \vspace{-1mm}
    \caption{
        Qualitative comparison on acceleration methods based on ObjectClear~\cite{zhao2025objectclear}.
        We show the image with \textcolor{green}{green} mask, and compare 20\% step sampling, ToCa~\cite{zou2024accelerating}, SiTo~\cite{zhang2025training}, HiCache~\cite{feng2025hicache}, QuantCache~\cite{wu2025quantcache}, FlashClear-C (ours), and FlashClear (ours).
    }
\label{fig:acc_qualitative_comparison}
\vspace{-2.5mm}
\end{figure*}

\begin{table*}[t]
\centering
\small
\caption{Ablation study and quantitative comparison of different accelerating strategies on OBER-Test (512$\times$512). The best and second performances are marked in \colorbox{rred}{red} and \colorbox{oorange}{orange}.}
\label{tab:ablation_and_acc}
\vspace{-2mm}
\begin{subtable}[t]{0.58\textwidth}
    \centering
    \scriptsize
    \renewcommand{\arraystretch}{1.15}
    \resizebox{\linewidth}{!}{
    \setlength{\tabcolsep}{1pt}
    \begin{tabular}{l cc cc cc}
    \toprule
    Method 
    & FLOPs (T) $\downarrow$
    & Accel $\uparrow$ 
    & LPIPS $\downarrow$ 
    & LPIPS-L. $\downarrow$ 
    & PSNR $\uparrow$ 
    & PSNR-mask $\uparrow$ \\
    \midrule
    ObjectClear~\cite{zhao2025objectclear} 
    & 63.6 & 1.00$\times$ & 0.03803 & 0.1540 & 32.06 & 23.34 \\
    \midrule
    20\% steps 
    & 12.7 & 5.00$\times$ & 0.05007 & 0.2186 & 32.48 & 23.24 \\
    ToCa~\cite{zou2024accelerating} 
    & 35.3 & 1.79$\times$ & 0.04801 & 0.1919 & 32.18 & \colorbox{oorange}{23.53} \\
    SiTo~\cite{zhang2025training} 
    & 56.2 & 1.13$\times$ & 0.05024 & 0.2228 & 30.71 & 21.74 \\
    HiCache~\cite{feng2025hicache} 
    & 39.9 & 1.59$\times$ & 0.03813 & 0.1574 & 32.36 & 23.43 \\
    QuantCache~\cite{wu2025quantcache} 
    & 25.4 & 2.49$\times$ & 0.05205 & 0.2006 & 28.10 & 21.51 \\
    FlashClear-C (ours)
    & \colorbox{rred}{7.7} 
    & \colorbox{rred}{8.26$\times$} 
    & \colorbox{oorange}{0.03623} 
    & \colorbox{oorange}{0.1438} 
    & \colorbox{oorange}{32.49} 
    & 23.51 \\    
    FlashClear (ours) 
    & \colorbox{oorange}{8.6} 
    & \colorbox{oorange}{7.35$\times$} 
    & \colorbox{rred}{0.03506} 
    & \colorbox{rred}{0.1396} 
    & \colorbox{rred}{33.05} 
    & \colorbox{rred}{24.12} \\

    \bottomrule
    \end{tabular}
    }
    \caption{Acceleration comparison.}
    \label{tab:acc_strategy}
\end{subtable}
\hfill
\begin{subtable}[t]{0.39\textwidth}
    \centering
    \scriptsize
    \renewcommand{\arraystretch}{1.34}
    \renewcommand{\tabcolsep}{2mm}
    \resizebox{\linewidth}{!}{
      \begin{tabular}{lcccc}
      \toprule
      \multicolumn{5}{c}{Loss Components} \\
      \midrule
      Loss Setting
      & I 
      & II 
      & III 
      & IV (ours) \\

      \midrule
      Diffusion 
      & $\checkmark$ 
      & $\checkmark$ 
      & $\checkmark$ 
      & $\checkmark$ \\
      LPIPS 
      & 
      & $\checkmark$ 
      & $\checkmark$ 
      & $\checkmark$ \\
      GAN 
      & 
      & 
      & $\checkmark$ 
      &  \\
      RAD 
      & 
      & 
      & 
      & $\checkmark$ \\
      \midrule
      LPIPS $\downarrow$ 
      & 0.0575 
      & 0.0416 
      & 0.0360 
      & \textbf{0.0351} \\
      LPIPS-L. $\downarrow$ 
      & 0.2649 
      & 0.1762 
      & 0.1423 
      & \textbf{0.1396} \\
      \bottomrule
      \end{tabular}
    }
    \caption{Ablation study on distillation settings.}
    \label{tab:ablation_loss}
\end{subtable}
\vspace{-7mm}
\end{table*}

\vspace{-3mm}
\subsection{Removal Performance}
\label{sec:removal}
\vspace{-2mm}
We compare FlashClear with recent methods across three tasks: image inpainting, object removal, and image editing. The selected baselines include inpainting models (PowerPaint~\cite{zhuang2024task}, SDXL-INP~\cite{podell2023sdxl}), object removal methods (GeoRemover~\cite{zhu2025georemover}, Omnieraser~\cite{wei2025omnieraser}, Attentive Eraser~\cite{sun2025attentive}, CLIPAway~\cite{ekin2024clipaway}, and RORem~\cite{li2025rorem}), and image editing frameworks (DesignEdit~\cite{jia2024designedit}, Omnipaint~\cite{yu2025omnipaint}). 

\textbf{Quantitative Evaluation.} As reported in Table~\ref{tab:removal}, our evaluations span datasets with varying image resolutions. The results demonstrate that FlashClear consistently surpasses previous SOTA methods across almost all visual metrics. Crucially, it achieves this superior restoration quality with a drastically reduced computational burden. As evidenced by the significantly lower FLOPs, FlashClear operates at a fraction of the computational cost of existing models, demonstrating that our method can deliver higher-quality object removal with vastly improved efficiency.   

\textbf{Qualitative Evaluation.}
Figure~\ref{fig:visual comparison} shows qualitative comparisons on challenging object removal cases. For reflection-dominated scenes, such as the bridge, pavilion, and duck examples, existing methods often leave object-related traces or introduce inconsistent structures, whereas our method removes both the target object and its reflection while preserving coherent water textures. For scenes involving shadows or local appearance changes, such as the basketball and cow examples, our method produces fewer residual artifacts and more natural background transitions. The cluttered last-row example further shows that our method can remove the masked child while maintaining nearby objects and wall textures. Overall, these results demonstrate that our method removes not only objects but also associated visual effects, achieving realistic restoration within only four steps.

\vspace{-2mm}
\subsection{Acceleration Performance}
\label{sec:Acc}
\vspace{-2mm}
Building upon ObjectClear~\cite{zhao2025objectclear}, we compare our proposed distillation framework (RAD) and training-free acceleration method denoted as FlashClear-C (FPAC) against naive step reduction and state-of-the-art open-source caching and token pruning techniques, including ToCa~\cite{zou2024accelerating}, HiCache~\cite{feng2025hicache}, QuantCache~\cite{wu2025quantcache}, and SiTo~\cite{zhang2025training}. As shown in Figure~\ref{fig:acc_qualitative_comparison}, our approach significantly outperforms existing acceleration methods. Notably, it achieves performance comparable to the uncompressed original model while requiring substantially less computational overhead according to Table~\ref{tab:acc_strategy}.

\begin{figure*}[t]
  \centering

  \begin{minipage}{0.47\textwidth}
    \centering
    \setlength{\tabcolsep}{1pt}
    \renewcommand{\arraystretch}{0.5}

    \begin{tabular}{@{}ccc@{}}
        \includegraphics[width=0.32\linewidth]{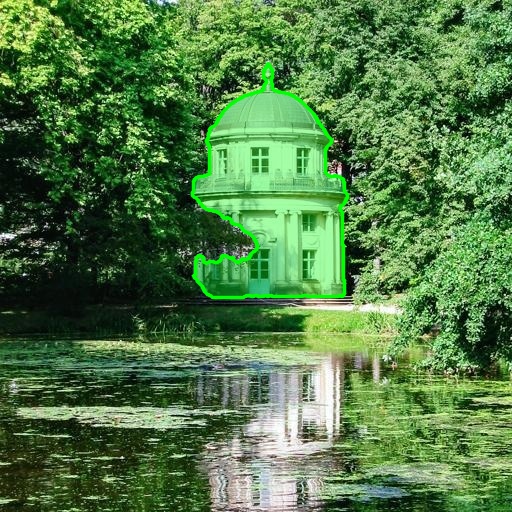} &
        \includegraphics[width=0.32\linewidth]{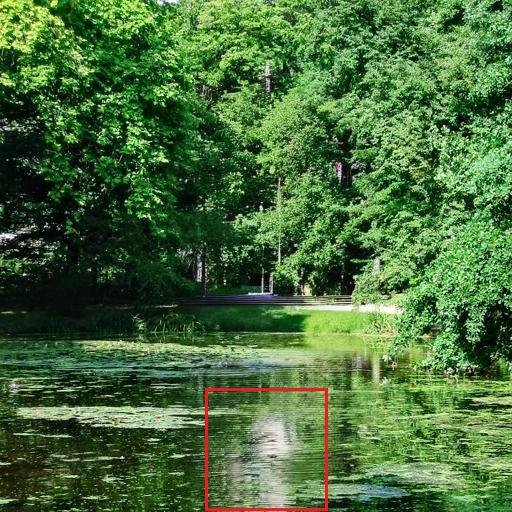} &
        \includegraphics[width=0.32\linewidth]{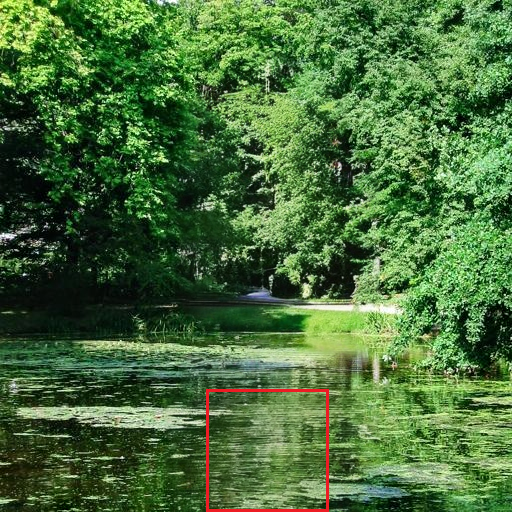} \\

        \includegraphics[width=0.32\linewidth]{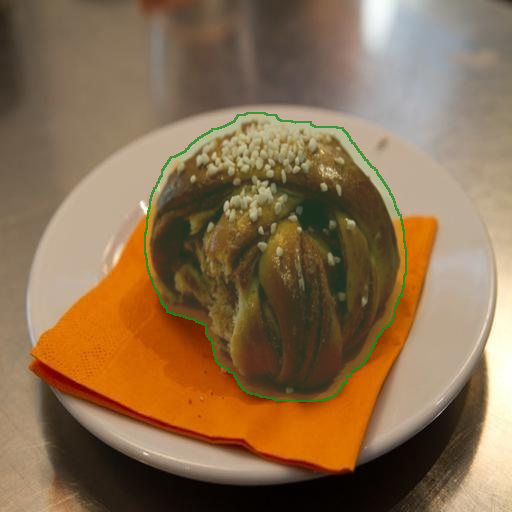} &
        \includegraphics[width=0.32\linewidth]{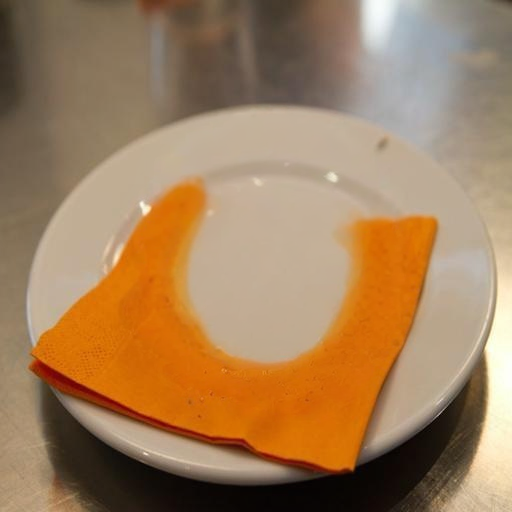} &
        \includegraphics[width=0.32\linewidth]{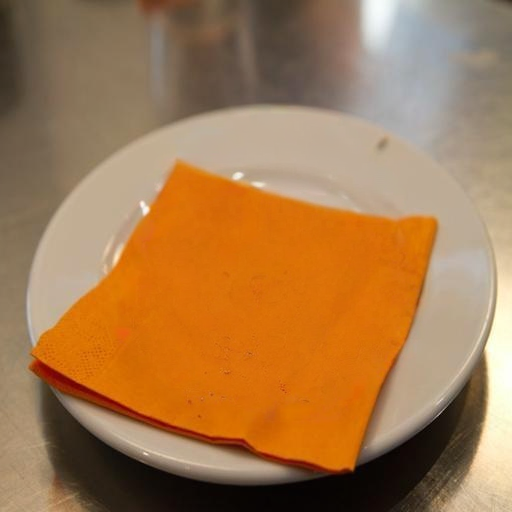} \\

        \scriptsize Input with mask &
        \scriptsize GAN (Clip) &
        \scriptsize RAD \\
    \end{tabular}

    \vspace{-1mm}
    \captionof{figure}{Visual comparison of different distillation losses. The mask is colored by \textcolor{green}{green}.}
    \label{fig:rad_ablation_visual}
  \end{minipage}
  \hfill
  \begin{minipage}{0.49\textwidth}
    \centering
    \begin{tikzpicture}
      \begin{axis}[
          width=\linewidth,
          height=0.75\linewidth,
          xlabel={FLOPs (T) $\downarrow$},
          ylabel={LPIPS $\downarrow$},
          xmin=5.0, xmax=9.5,
          ymin=0.0350, ymax=0.0405,
          grid=both,
          major grid style={line width=.2pt,draw=gray!50},
          minor grid style={line width=.1pt,draw=gray!20},
          tick label style={font=\footnotesize},
          label style={font=\footnotesize}, 
          axis on top=false
      ]
      \fill[red!10] (axis cs:5.0, 0.0364) rectangle (axis cs:9.5, 0.0405);

      \addplot [domain=5.0:9.5, samples=2, color=red, dashed, thick] {0.0364};

      \node[red!80!black, anchor=north west, font=\tiny\bfseries]
        at (axis cs: 4.97, 0.037) {Region with visible artifacts};

      \addplot[color=blue!80!black, mark=*, mark size=1.8pt, line width=1.2pt] coordinates {
          (6.485, 0.03898)
          (6.606, 0.03950)
          (7.700, 0.03623)
          (8.647, 0.03506)
      };

      \node[anchor=east, font=\tiny, xshift=-2pt]
        at (axis cs:6.445, 0.03898) {Ours (3 steps)};
      \node[anchor=south, font=\tiny, yshift=2pt]
        at (axis cs:6.606, 0.03935) {ToCa~\cite{zou2024accelerating}};
      \node[anchor=south west, font=\tiny, xshift=2pt, yshift=2pt]
        at (axis cs:7.600, 0.03623) {FPAC};
      \node[anchor=south west, font=\tiny, xshift=2pt, yshift=2pt]
        at (axis cs:8.547, 0.03508) {Ours};

      \end{axis}
    \end{tikzpicture}
\vspace{-2.8mm}
\captionof{figure}{LPIPS and FLOPs (T) trade-off. LPIPS spikes for generally designed acceleration methods like ToCa~\cite{zou2024accelerating}.}
    \label{fig:flops_vs_lpips}
  \end{minipage}
\vspace{-5mm}
\end{figure*}

\noindent\textbf{Effectiveness of RAD.}
We evaluate our four-step model's performance with different distillation losses. As shown in Table~\ref{tab:ablation_loss}, Diffusion denotes diffusion loss and GAN means using a standard semantic discriminator (Clip) in pixel space, while RAD denotes our proposed region-aware adversarial distillation in latent space. Our method outperforms other ablation settings in both global and local perceptual metrics. What's more, as illustrated in Fig.~\ref{fig:visual comparison}, the standard generator with semantic discriminator fails to maintain the background's integrity, while our region-aware adversarial distillation shows better object and effect removal performance and a more robust ability to maintain background consistency after removal. These results indicate the effectiveness of our proposed distillation method in preserving model's capability to remove objects and their causal physical effects.

\begin{wrapfigure}{r}{0.5\textwidth}
    \centering
    \vspace{-12pt} 
    \begin{subfigure}[b]{0.24\linewidth}
        \centering
        \includegraphics[width=\linewidth]{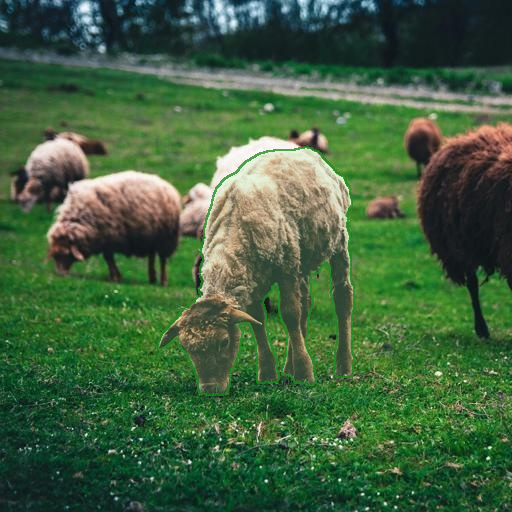}
        \caption*{\tiny \\\vspace{-12pt} Input with mask} 
    \end{subfigure}%
    \hfill
    \begin{subfigure}[b]{0.24\linewidth}
        \centering
        \includegraphics[width=\linewidth]{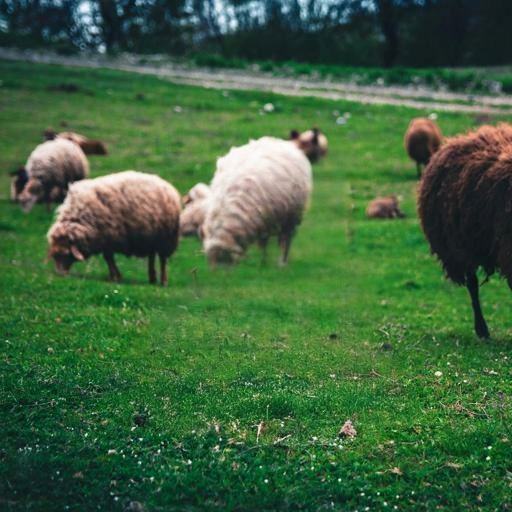}
        \caption*{\tiny \\\vspace{-12pt} 3 steps}
    \end{subfigure}%
    \hfill
    \begin{subfigure}[b]{0.24\linewidth}
        \centering
        \includegraphics[width=\linewidth]{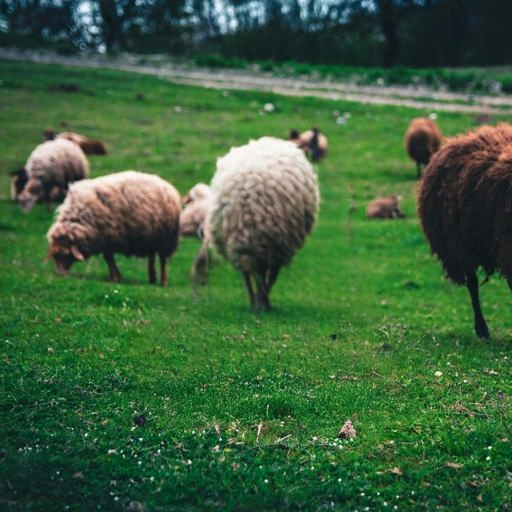}
        \caption*{\tiny \\\vspace{-12pt} ToCa~\cite{zou2024accelerating}}
    \end{subfigure}%
    \hfill
    \begin{subfigure}[b]{0.24\linewidth}
        \centering
        \includegraphics[width=\linewidth]{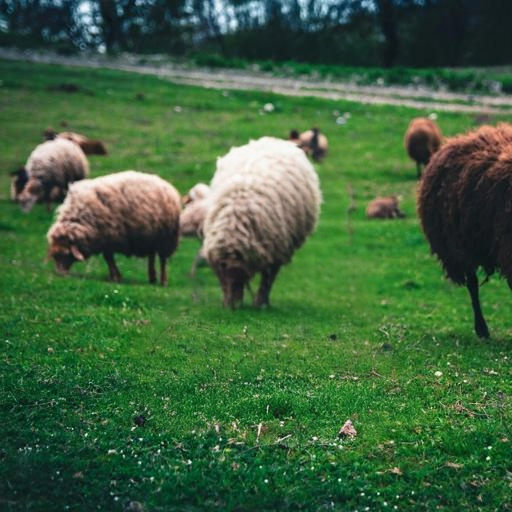}
        \caption*{\tiny \\\vspace{-12pt} FPAC}
    \end{subfigure}

    \vspace{-4pt}
    \caption{\small The mask is colored by \textcolor{green}{green}. Compared with the 3-steps and ToCa~\cite{zou2024accelerating} strategies, FPAC maintains removal quality without introducing visual artifacts.}
    \label{fig:fpac_comparison}
    \vspace{-10pt}
\end{wrapfigure}
\textbf{Effectiveness of FPAC.} To further reduce computational overhead while preserving high visual fidelity, we introduce FPAC, an acceleration plug-in specifically tailored for object removal tasks. Most existing training-free methods either require a substantial amount of prior information~\cite{feng2025hicache} or only become effective during the final denoising steps~\cite{wu2025quantcache}, rendering them incompatible with our extreme 4-step setting. Even when compared to applicable general-purpose acceleration methods like ToCa~\cite{zou2024accelerating}, our approach achieves comparable FLOPs reduction without suffering from severe degradation in generation quality. As shown in Figure~\ref{fig:flops_vs_lpips}, FPAC strikes an optimal balance between image quality and computational efficiency. Notably, as shown in Figure~\ref{fig:fpac_comparison}, it successfully avoids visible artifacts that typically accompany aggressive acceleration, maintaining robust performance at a significantly lower computational cost.

\textbf{Complementary Nature of RAD and FPAC.} This advantage comes from the spatial redundancy of object removal: most background regions should remain unchanged, while only foreground-related regions require intensive recomputation. Benefiting from RAD's targeted training and loss design, the few-step Cross-Attention maps in FlashClear can accurately cover both the target object and its object-induced effects, making cache reuse feasible even in the extremely few-step regime. By prioritizing removal-relevant tokens and safely reusing stable background information, FPAC avoids the quality drop caused by indiscriminate token pruning or caching, providing additional acceleration on top of RAD while maintaining visually consistent removal results.

\vspace{-3mm}
\section{Conclusion}
\vspace{-3mm}
In this paper, we propose FlashClear, an ultra-fast and highly efficient diffusion-based object removal model. To overcome the severe latency of multi-step inference, we propose Region-aware Adversarial Distillation (RAD), which successfully compresses the denoising process into minimal steps while strictly preserving the model's capability to perceive and eliminate complex object-induced visual effects. Furthermore, we designed FPAC, an attention-guided caching mechanism that resolves the fundamental conflict between few-step distillation and feature reuse by exploiting the inherent spatial redundancy of background regions. Extensive experiments demonstrate that FlashClear achieves up to an 8.26$\times$ speedup alongside state-of-the-art visual fidelity, effectively bridging the gap between high-quality content removal and real-time interactive deployment.

\clearpage
\bibliographystyle{plain}
\bibliography{main}

\newpage
\appendix

\section{Overview}
This supplementary material provides additional details and analyses to complement the main paper. We first describe the implementation details of our proposed method, including the training configuration and loss components. We then present our model's performance and analyses with 2-step distillation, including more quantitative comparisons and qualitative examples on diverse object removal scenarios, to further prove the effectiveness of our region-aware adversarial distillation. We also provide additional details and comparisons of our proposed FPAC and concurrent SOTA cache acceleration method Hicache. To further evaluate whether FPAC still preserves human-preferred object removal results while maintaining substantially higher inference efficiency, we conduct a user study to compare our accelerated variants with the baseline model. Finally, we discuss limitations to give a more comprehensive understanding of the proposed framework.

Specifically, Section~\ref{sec:implementation_details} presents the implementation details of FlashClear. Section~\ref{sec:2steps} presents the quantitative and qualitative results of 2-step FlashClear. Section~\ref{sec:more_results_hicache} reports additional explanation on the incapability of HiCache combined with our distilled model to prove the validity of our proposed FPAC. Section~\ref{sec:user study} provides a user study on perceptual removal quality. Section~\ref{sec:more results} provides more qualitative comparisons with existing object removal and acceleration methods. Section~\ref{sec:limitations} discusses limitations and potential future directions.

\section{Implementation Details}
\label{sec:implementation_details}
As listed in Table~\ref{tab:implementation_details}, all training experiments are conducted on NVIDIA A800 GPUs with 80GB memory. The training of FlashClear takes approximately 24 hours for 10K iterations. The inference and evaluation are conducted on a single A6000 GPU with 48GB memory, with the shorter side of test images resized to 512.
We report FLOPs to provide a hardware-independent measure of computational cost, and wall-clock latency to provide speed information for reference only.
\paragraph{RAD Implementation Details}We implement FlashClear based on the SDXL-based ObjectClear model. 
The model is initialized from the pretrained ObjectClear checkpoint and trained with LoRA adaptation. 
Unless otherwise specified, we use the same training configuration for all experiments. 
The model is trained on the OBER dataset for 10K iterations with a per-device batch size of 8. 
We use the AdamW optimizer with a learning rate of $1\times10^{-5}$ for both the generator and the discriminator. 
Mixed-precision training with bfloat16 is adopted to reduce memory consumption. 
The distillation process compresses the original multi-step object removal model into a 4-step generator. 
The training objective consists of diffusion loss, perceptual loss, adversarial loss, and object localization supervision (mask loss), where $\lambda_{\mathrm{diff}}=1.0$, $\lambda_{\mathrm{LPIPS}}=5.0$, $\lambda_{\mathrm{GAN}}=0.5$, and $\lambda_{\mathrm{mask}}=0.01$.

\paragraph{FPAC Implementation Details.}
We implement FPAC on top of the distilled few-step ObjectClear model without introducing additional trainable parameters. During inference, FPAC reuses features only at the last denoising step, where the attention maps become more stable and provide a more reliable indication of the regions that require modification. Specifically, we apply caching to the attention modules with the following feedforward MLP layers in \texttt{down\_blocks.1.attentions}, \texttt{down\_blocks.2.attentions}, \texttt{mid\_block.attentions}, \texttt{up\_blocks.0.attentions}, and \texttt{up\_blocks.1.attentions}. These layers are selected because they contain rich spatial-semantic representations while accounting for a large portion of the U-Net computation. We use the same cache configuration for all experiments.
\begin{table*}[h]
\centering
\small
\renewcommand{\arraystretch}{1.05}
\caption{Implementation details of FlashClear. Left: cache settings of FPAC. Right: key training hyperparameters of RAD.}
\label{tab:implementation_details}

\begin{subtable}[t]{0.35\textwidth}
\centering
\caption{FPAC cache settings.}
\label{tab:fpac_cache_settings}
\renewcommand{\arraystretch}{1.1}
\renewcommand{\tabcolsep}{3.0mm}
\begin{tabular}{l c}
\toprule
\textbf{Cached Module} & \textbf{Cached Step} \\
\midrule
\texttt{down\_blocks.1.attentions} & 4 \\
\texttt{down\_blocks.2.attentions} & 4 \\
\texttt{mid\_block.attentions} & 4 \\
\texttt{up\_blocks.0.attentions} & 4 \\
\texttt{up\_blocks.1.attentions} & 4 \\
\bottomrule
\end{tabular}
\end{subtable}
\hfill
\begin{subtable}[t]{0.5\textwidth}
\centering
\caption{Key training hyperparameters.}
\label{tab:training_hyperparams}
\resizebox{\linewidth}{!}{%
\begin{tabular}{l c l c}
\toprule
\textbf{Config.} & \textbf{Value} 
& \textbf{Config.} & \textbf{Value} \\
\midrule
Base model & ObjectClear 
& Mixed precision & bfloat16 \\
Initialization & ObjectClear 
& LoRA rank & 256 \\
Dataset & OBER 
& Distillation steps & 4 \\
Iterations & 10K 
& $\lambda_{\mathrm{diff}}$ & 1.0 \\
Batch size / GPU & 8 
& $\lambda_{\mathrm{LPIPS}}$ & 5.0 \\
Optimizer & AdamW 
& $\lambda_{\mathrm{GAN}}$ & 0.5 \\
Learning rate & $1\times10^{-5}$ 
& $\lambda_{\mathrm{mask}}$ & 0.01 \\
\bottomrule
\end{tabular}
}
\end{subtable}

\end{table*}

\section{Additional Results of the Two-Step Model}\label{sec:2steps}
In this section, we further provide the qualitative and quantitative results of our distilled two-step model with our proposed region-aware adversarial distillation. Similar to the metrics used in the four-step model test in the main text, we test the two-step model across various quantitative metrics like LPIPS, LPIPS-mask, PSNR, and PSNR-mask, demonstrating the effectiveness and robustness of our distillation method in a more extreme step schedule.

\subsection{Quantitative Results}

To further evaluate the robustness of our acceleration framework under more aggressive sampling schedules, we additionally train a two-step variant of our model. Compared with the four-step model used in the main paper, the two-step model further reduces the number of denoising steps and therefore provides a more challenging setting for object removal. As shown in Table~\ref{tab:two_step_quant}, the two-step model achieves substantially lower computational cost while maintaining competitive perceptual and local restoration quality. Although a slight degradation can be observed compared with the four-step model, the performance drop is relatively moderate considering the significant reduction in sampling steps. This result indicates that the proposed distillation strategy remains effective even under extremely low-step inference.
\begin{table*}[ht]
\centering
\small
\renewcommand{\arraystretch}{1.15}
\renewcommand{\tabcolsep}{3.3mm}
\caption{Quantitative comparison of the two-step model and the four-step model. The best and second performances are marked in \colorbox{rred}{red} and \colorbox{oorange}{orange}.}
\label{tab:two_step_quant}
\resizebox{\linewidth}{!}{
\begin{tabular}{l cc cc cc}
\toprule
Method 
& FLOPs(T) $\downarrow$
& Latency $\downarrow$
& LPIPS $\downarrow$ 
& LPIPS-mask $\downarrow$
& PSNR $\uparrow$ 
& PSNR-mask $\uparrow$ \\
\midrule

ObjectClear 
& 63.6331 
& 2.2901 
& 0.03803 
& 0.1540 
& 32.06
& \colorbox{oorange}{23.34} \\

4steps (ours) 
& 8.6477 
& 0.9476 
& \colorbox{rred}{0.03506} 
& \colorbox{rred}{0.1396} 
& \colorbox{rred}{33.05} 
& \colorbox{rred}{24.12} \\

2steps (ours) 
& \colorbox{oorange}{4.3238} 
& \colorbox{rred}{0.6210} 
& \colorbox{oorange}{0.03700} 
& \colorbox{oorange}{0.1438} 
& \colorbox{oorange}{32.49} 
& 23.27 \\

2steps+FPAC (ours)
& \colorbox{rred}{3.4442} 
& \colorbox{oorange}{0.8039} 
& 0.04149
& 0.1640
& 32.37
& 23.11 \\

\bottomrule
\end{tabular}
}
\vspace{-5mm}
\end{table*}
\subsection{Qualitative Results}

\begin{figure*}[t]
\centering
\setlength{\tabcolsep}{2pt}
\renewcommand{\arraystretch}{1.0}

\newlength{\visimgw}
\setlength{\visimgw}{0.19\textwidth}

\begin{tabular}{ccccc}
\includegraphics[width=\visimgw]{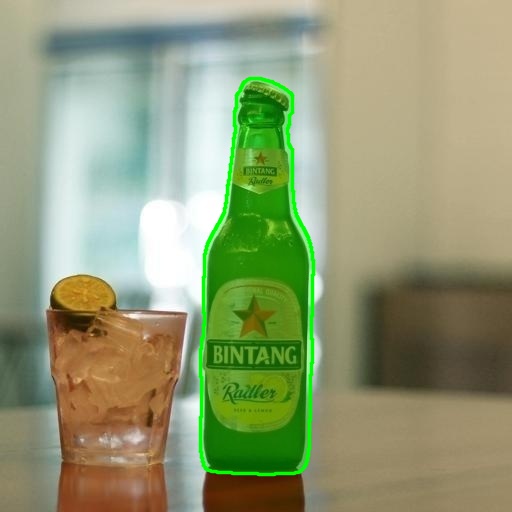} &
\includegraphics[width=\visimgw]{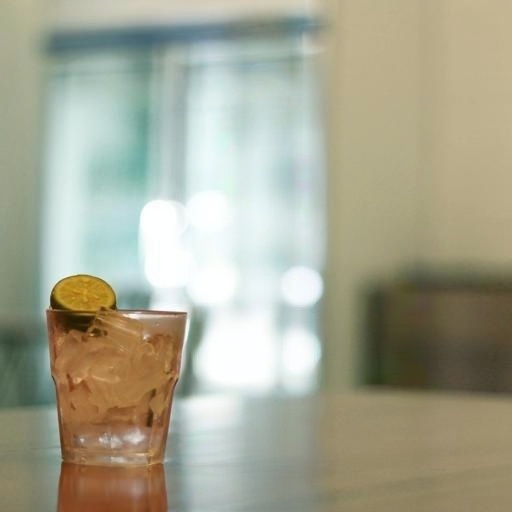} &
\includegraphics[width=\visimgw]{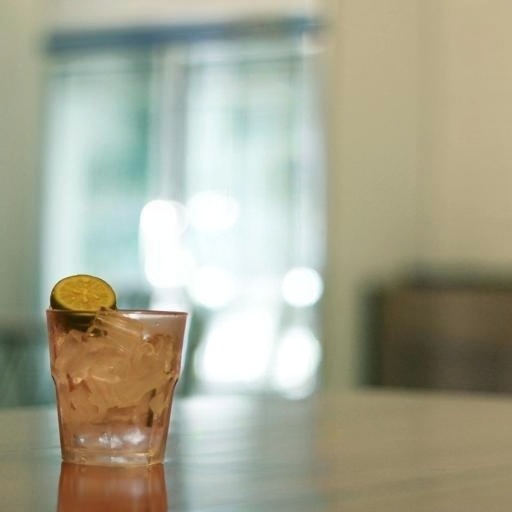} &
\includegraphics[width=\visimgw]{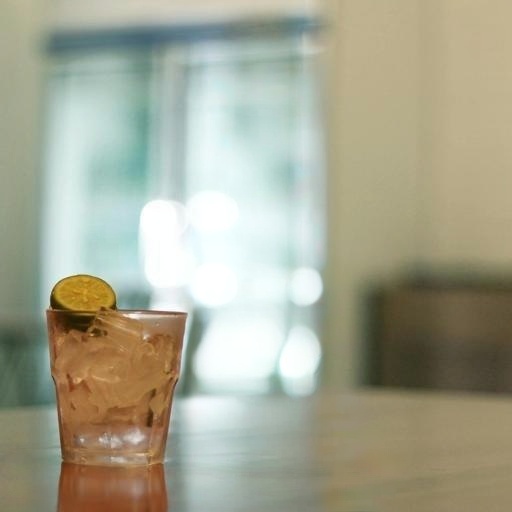} &
\includegraphics[width=\visimgw]{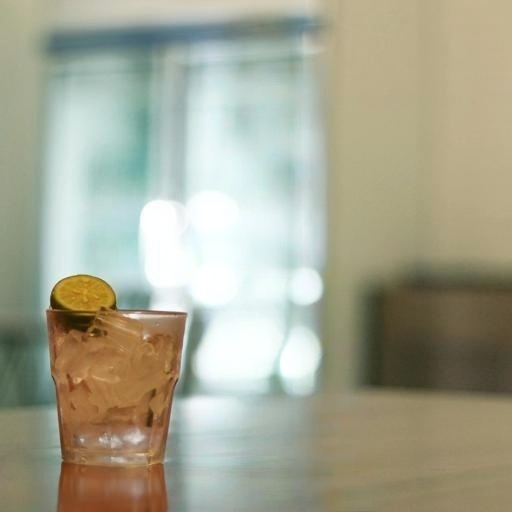} \\[4pt]

\includegraphics[width=\visimgw]{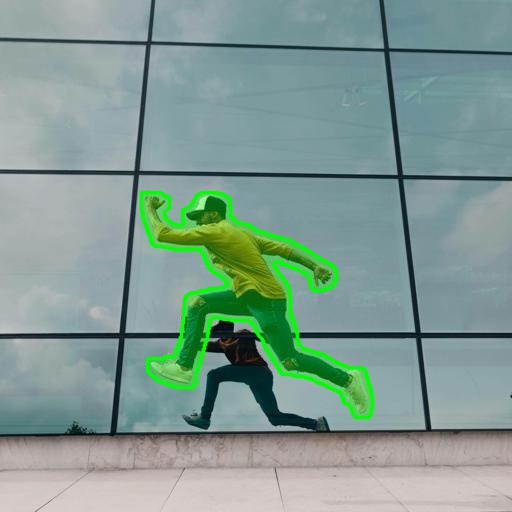} &
\includegraphics[width=\visimgw]{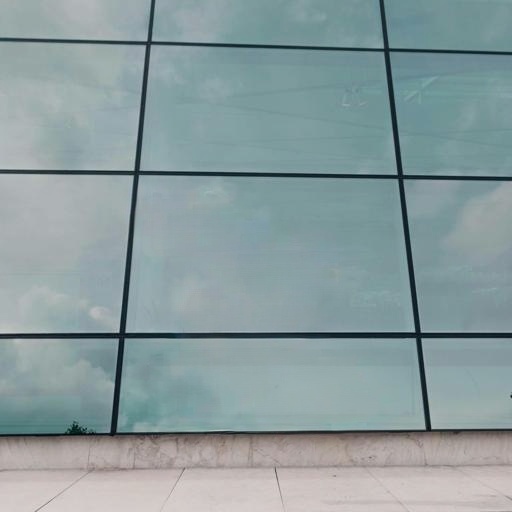} &
\includegraphics[width=\visimgw]{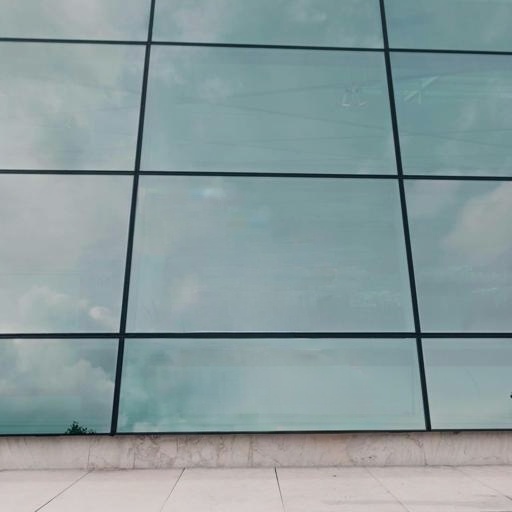} &
\includegraphics[width=\visimgw]{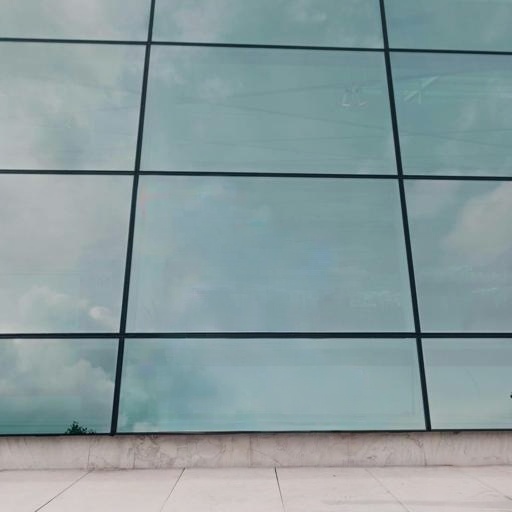} &
\includegraphics[width=\visimgw]{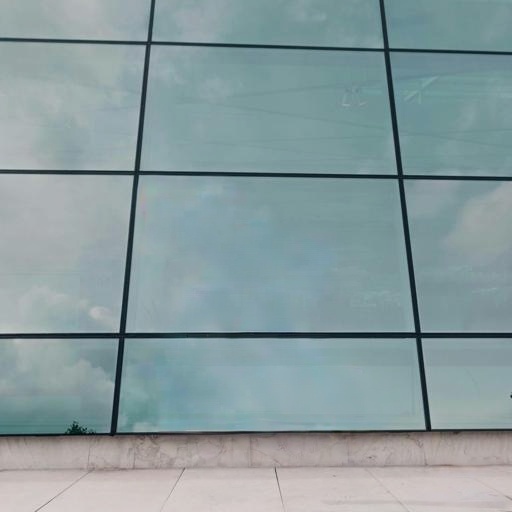} \\[4pt]

\includegraphics[width=\visimgw]{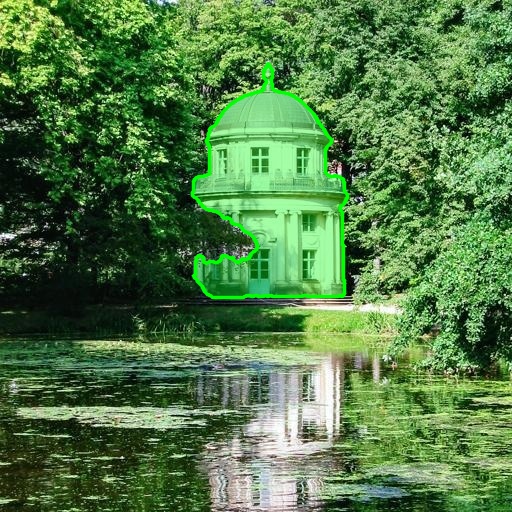} &
\includegraphics[width=\visimgw]{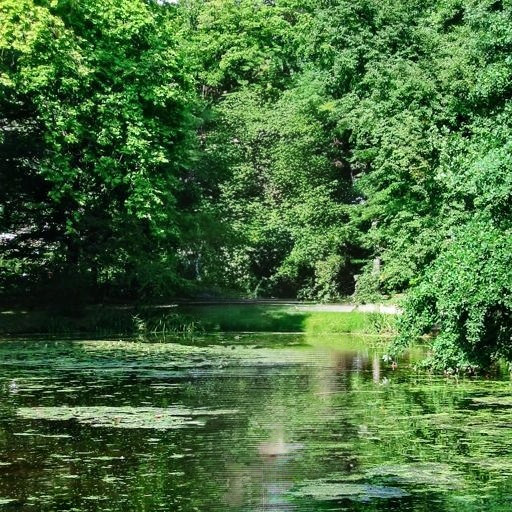} &
\includegraphics[width=\visimgw]{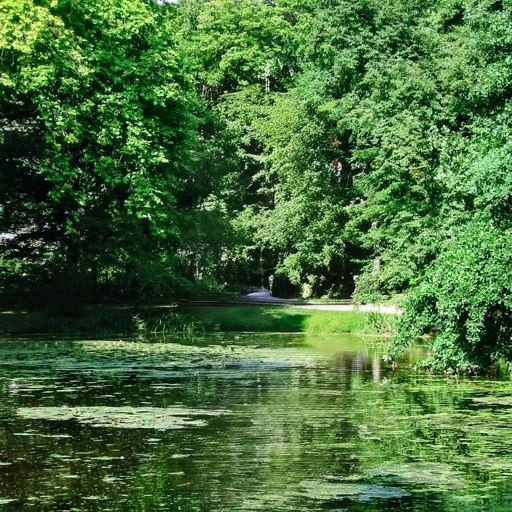} &
\includegraphics[width=\visimgw]{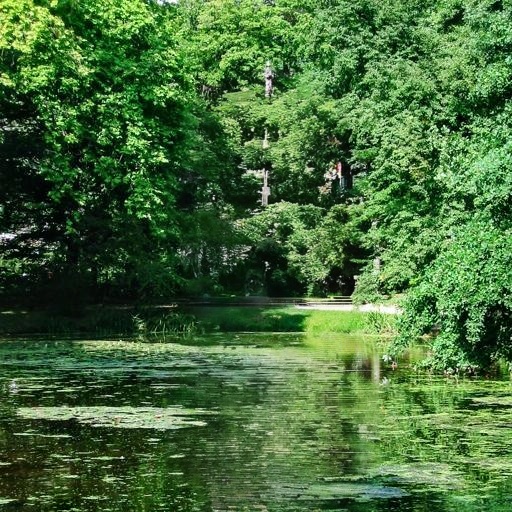} &
\includegraphics[width=\visimgw]{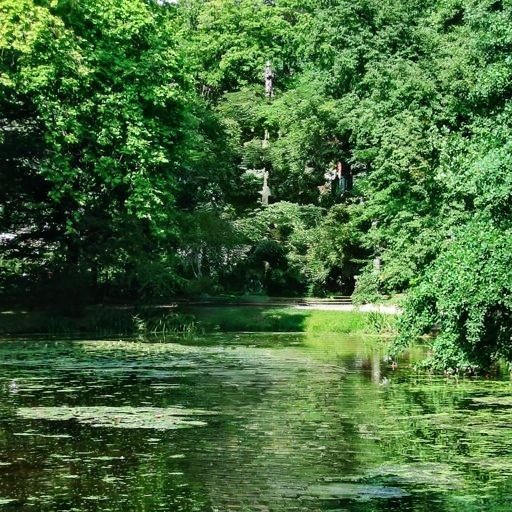} \\[4pt]

\includegraphics[width=\visimgw]{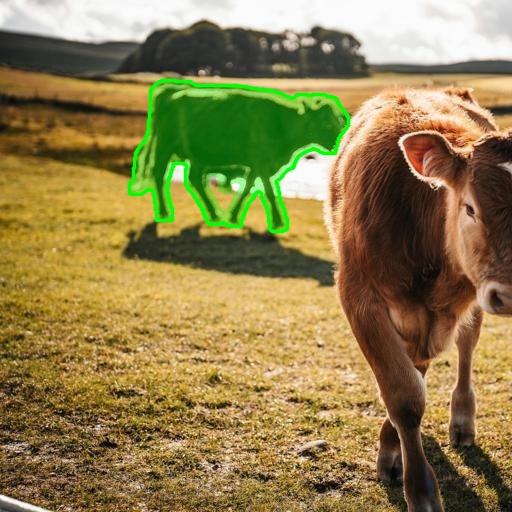} &
\includegraphics[width=\visimgw]{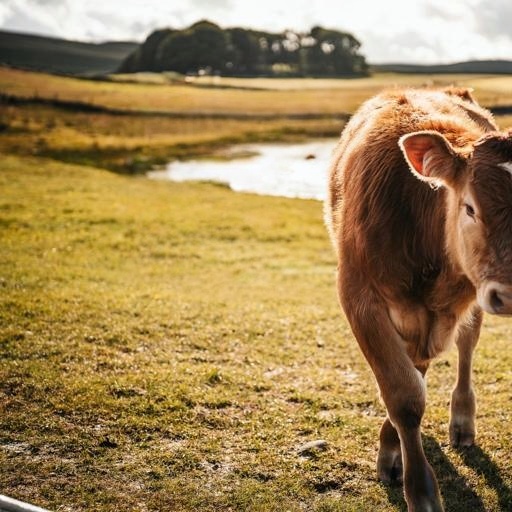} &
\includegraphics[width=\visimgw]{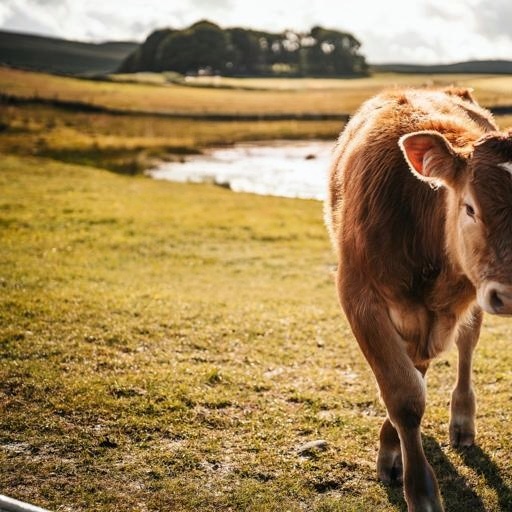} &
\includegraphics[width=\visimgw]{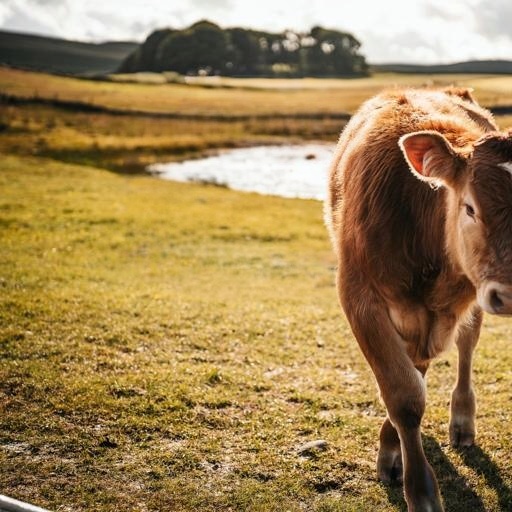} &
\includegraphics[width=\visimgw]{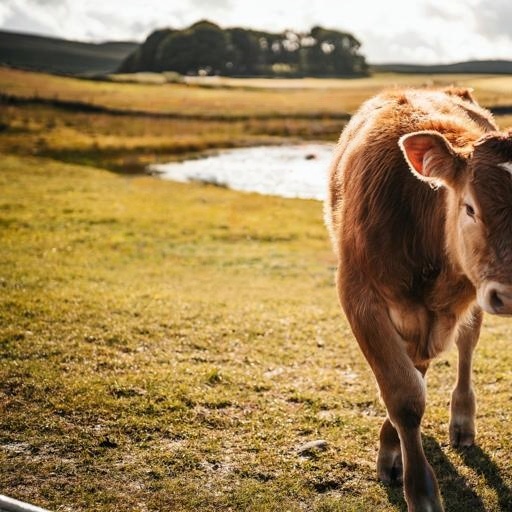} \\

{Input w/ Mask} &
{ObjectClear} &
{ours-4steps} &
{ours-2steps} &
{ours-2steps+FPAC} \\[3pt]
\end{tabular}
\vspace{-2mm}
\caption{
Qualitative comparison of different sampling and acceleration settings. 
ObjectClear denotes the original 20-step baseline, while the 4-step and 2-step variants correspond to our accelerated models with or without cache. 
The proposed low-step models preserve visually coherent object removal results, and FPAC further reduces computation while maintaining plausible visual quality.
}
\label{fig:appendix_two_step_visual}
\vspace{-5mm}
\end{figure*}

Figure~\ref{fig:appendix_two_step_visual} presents qualitative results of our two-step model on challenging object removal cases. The model can still remove objects from complex scenes and generate visually plausible backgrounds despite using only two denoising steps. In reflection-dominated scenes, such as objects located near water surfaces, the two-step model removes not only the foreground object but also its correlated reflection, leading to coherent background textures. In shadow-related examples, the model suppresses object-induced dark traces and reconstructs the surrounding surface or grassland with natural appearance. These results suggest that the two-step model preserves the essential removal capability learned from the multi-step teacher, including the ability to eliminate object-associated visual effects rather than merely filling the masked region.

\section{More Results on Ours + HiCache}\label{sec:more_results_hicache}

In this section, we provide supplementary data and detailed analysis to explain why HiCache is not utilized to accelerate our four-step model in the main text. Specifically, the mathematical formulation of HiCache inherently requires a minimum of four complete steps to accurately calculate the coefficients for its fitting formula. Forcibly applying HiCache to a four-step generation process violates this prerequisite, which leads to a sudden and drastic degradation in overall performance. 

To empirically demonstrate this limitation, we conduct a comparison between our model accelerated by HiCache and our proposed FPAC method. As detailed in Table~\ref{tab:hicache_vs_fpac}, the forced integration of HiCache results in significantly inferior quantitative metrics across FLOPs, LPIPS, LPIPS-Local, PSNR, and PSNR-mask when compared to FPAC.

\begin{figure*}[t]
  \centering
  \begin{tabular}{@{}c@{\hspace{1mm}}c@{\hspace{1mm}}c@{}}
    \includegraphics[width=0.32\linewidth]{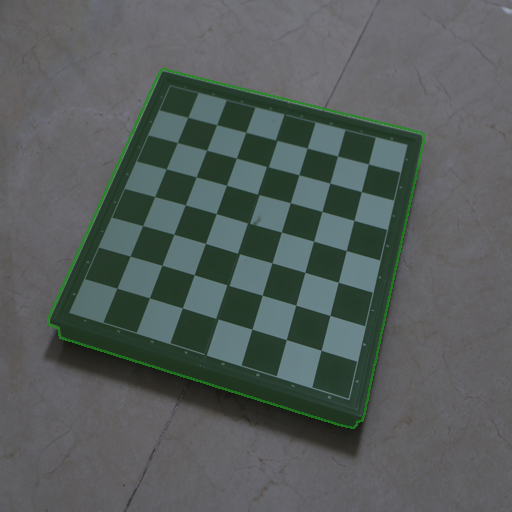} &
    \includegraphics[width=0.32\linewidth]{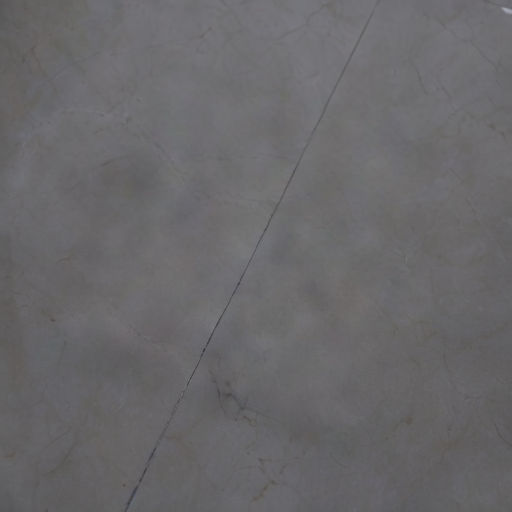} &
    \includegraphics[width=0.32\linewidth]{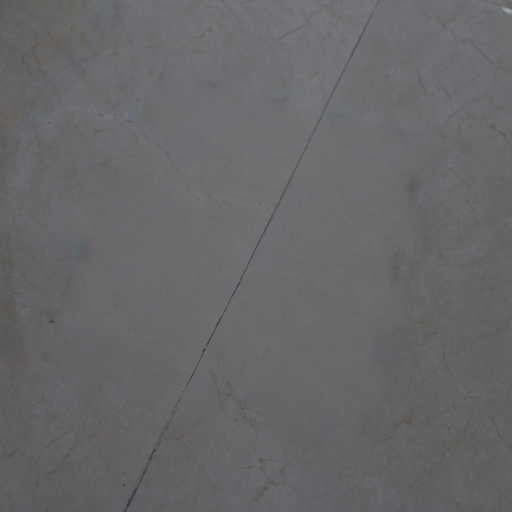} \\
    \includegraphics[width=0.32\linewidth]{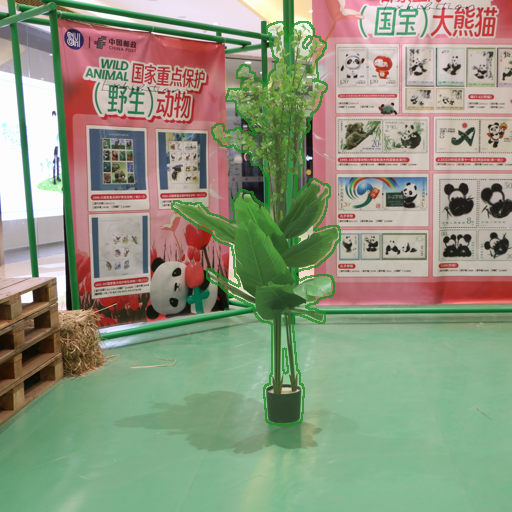} &
    \includegraphics[width=0.32\linewidth]{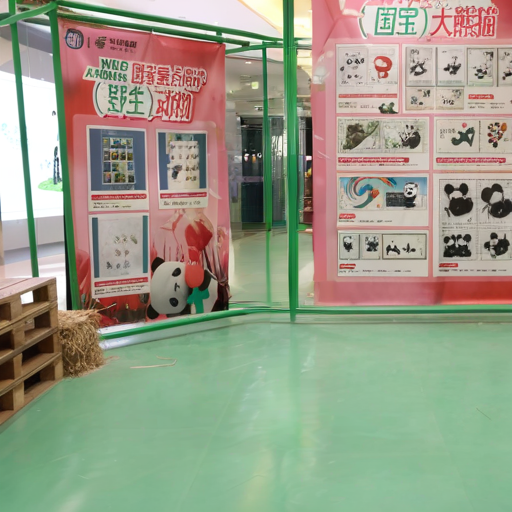} &
    \includegraphics[width=0.32\linewidth]{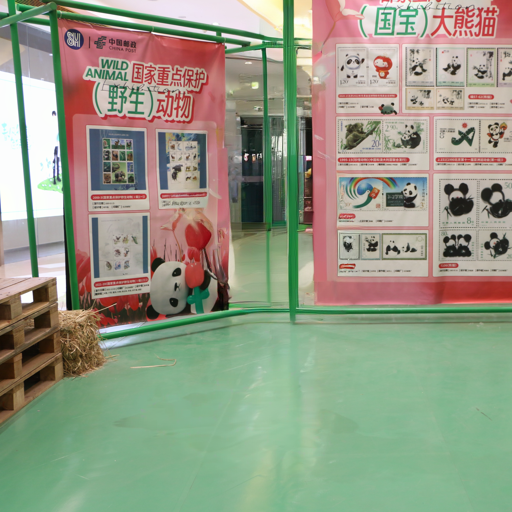} \\
    (a) Input with Mask & (b) HiCache & (c) FPAC \\
  \end{tabular}
  \caption{Visual comparison of HiCache and FPAC on different object removal cases.}
  \label{fig:hicache_vs_fpac_comparison}
\end{figure*}

Furthermore, we provide qualitative comparisons in Figure~\ref{fig:hicache_vs_fpac_comparison} to corroborate these findings. The visual results, showcasing two different cases of object removal, clearly illustrate that forcing the use of HiCache introduces severe foreground object retention and a massive amount of visual artifacts in the unmasked regions. These results firmly validate that HiCache itself is fundamentally incapable of accelerating models that operate in four steps or fewer.
\vspace{-3.5mm}
\begin{table}[htbp]
  \centering
  \small
  \caption{Quantitative comparison between Ours + HiCache and Ours + FPAC. Forcing HiCache on a four-step model leads to noticeable performance drops across all metrics.}
  \vspace{1mm}
  \begin{tabular}{lccccc}
    \toprule
    Method & FLOPs(T) $\downarrow$ & LPIPS $\downarrow$ & LPIPS-Local $\downarrow$ & PSNR $\uparrow$ & PSNR-mask $\uparrow$ \\
    \midrule
    Ours + HiCache & 7.4126 & 0.08250 & 0.1575 & 27.49 & 23.41 \\
    Ours + FPAC   & 7.7000 & 0.03623 & 0.1438 & 32.48 & 23.51 \\
    \bottomrule
  \end{tabular}
  \label{tab:hicache_vs_fpac}
  \vspace{-2mm}
\end{table}

\section{User Study on Perceptual Removal Quality}\label{sec:user study}
\label{app:user_study}
\vspace{-2mm}
To further evaluate whether the proposed acceleration strategy affects the perceptual quality of object removal, we conduct a user study comparing our accelerated variants with the baseline ObjectClear model. Unlike pixel-level metrics, this study focuses on whether human observers can perceive noticeable degradation in the final removal results.

\begin{figure}[t]
    \centering
    \includegraphics[width=1\linewidth]{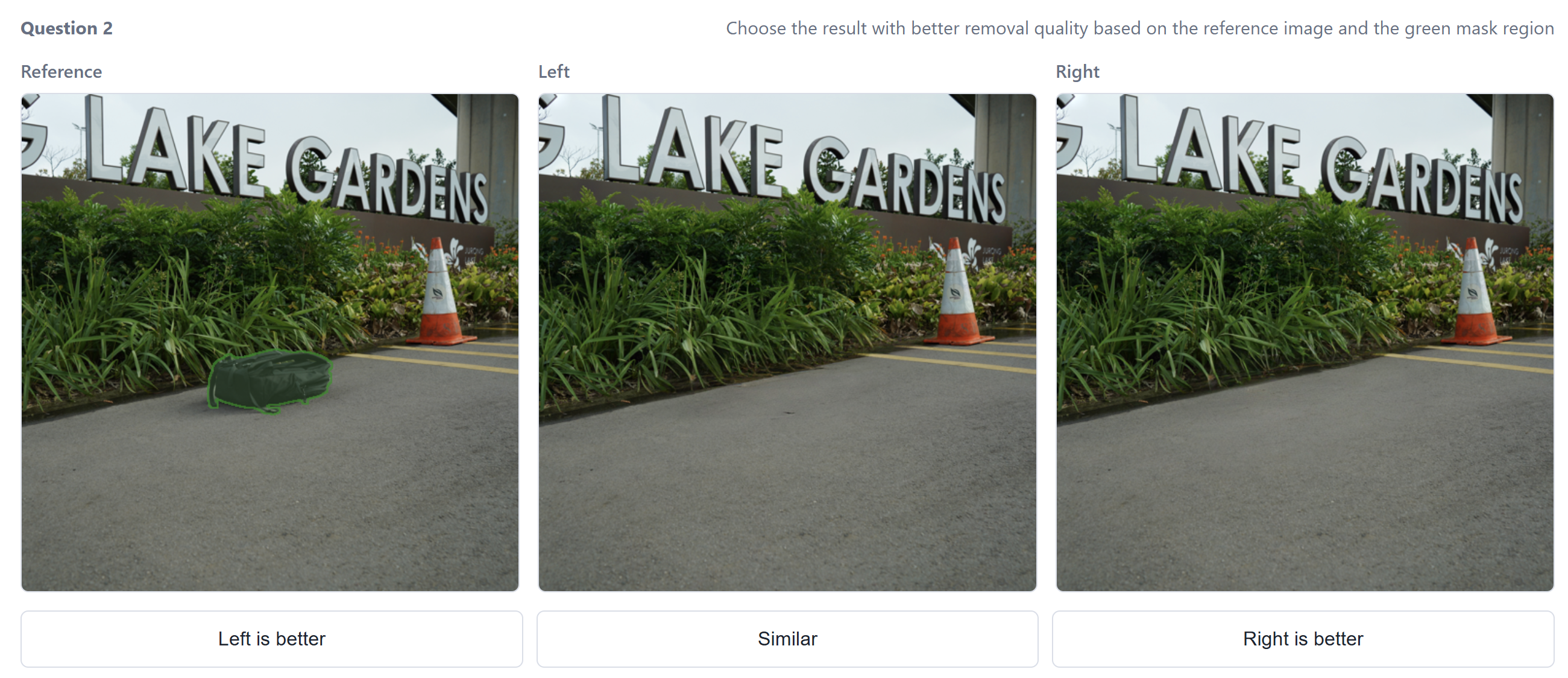}
    \caption{
    Example interface of the user study. Each question shows the reference image with the target removal region highlighted in green, along with two anonymized removal results. Participants are asked to choose the result with better perceptual removal quality or select ``similar'' when the two results are visually comparable.
    }
    \label{fig:user_study_interface}
    \vspace{-3mm}
\end{figure}

\textbf{Study Design.} We design a two-alternative forced-choice user study with an additional tie option. As shown in Figure~\ref{fig:user_study_interface}, each question presents one reference image in which the target removal region is highlighted by a green mask, together with two anonymized removal results displayed side by side. Participants are asked to judge which result has better removal quality according to the reference image and the masked target region, or to select ``similar'' if the two results are visually comparable.

The compared methods are anonymized and randomly assigned to the left or right side for each question. Each questionnaire contains 20 questions randomly sampled from a pool of 163 test images. Among them, 10 questions compare FlashClear with ObjectClear, while the other 10 compare FlashClear+FPAC (denoted as FlashClear-C) with ObjectClear. ObjectClear is included in every comparison as the baseline. The sampled images, question order, and left-right positions are independently randomized for each questionnaire to reduce ordering and presentation bias.

\textbf{Evaluation Protocol.} For each comparison, we report the win rate, tie rate, and loss rate of the accelerated method against ObjectClear. A win means that the participant prefers the accelerated method; a tie means that the participant considers the two results visually similar; and a loss means that ObjectClear is preferred. Formally, for a method $M \in \{\text{FlashClear}, \text{FlashClear+FPAC}\}$, we compute
\begin{equation}
    \mathrm{WinRate}(M) =
    \frac{N_{\mathrm{win}}(M)}
    {N_{\mathrm{win}}(M) + N_{\mathrm{tie}}(M) + N_{\mathrm{loss}}(M)},
\end{equation}
\begin{equation}
    \mathrm{TieRate}(M) =
    \frac{N_{\mathrm{tie}}(M)}
    {N_{\mathrm{win}}(M) + N_{\mathrm{tie}}(M) + N_{\mathrm{loss}}(M)},
\end{equation}
\begin{equation}
    \mathrm{LoseRate}(M) =
    \frac{N_{\mathrm{loss}}(M)}
    {N_{\mathrm{win}}(M) + N_{\mathrm{tie}}(M) + N_{\mathrm{loss}}(M)}.
\end{equation}

A high tie rate, together with a balanced win/loss distribution, indicates that the accelerated method preserves perceptual removal quality relative to ObjectClear.

\begin{table}[t]
\vspace{-5mm}
    \centering
    \caption{
    User study results comparing the perceptual removal quality of accelerated methods against ObjectClear. Each questionnaire contains 10 ObjectClear-vs-FlashClear comparisons and 10 ObjectClear-vs-FlashClear+FPAC comparisons. Win, tie, and lose rates are computed from the perspective of the accelerated method.
    }
    \vspace{1mm}
    \label{tab:user_study}
    \begin{tabular}{lcccc}
        \toprule
        Method & \#Questions & Win Rate & Tie Rate & Lose Rate \\
        \midrule
        FlashClear vs. ObjectClear
        & 200 & 22.5\% & 55.5\% & 22.0\% \\
        FlashClear-C vs. ObjectClear
        & 200 & 22.0\% & 53.5\% & 24.5\% \\
        \bottomrule
    \end{tabular}
\vspace{-3mm}    
\end{table}

\textbf{Results and Discussion.} The user study results are summarized in Table~\ref{tab:user_study}. FlashClear obtains a win rate of 22.5\%, a tie rate of 55.5\%, and a lose rate of 22.0\% against ObjectClear. The win and lose rates are nearly balanced, while more than half of the responses fall into the tie category. This suggests that participants usually perceive the removal results of FlashClear and ObjectClear as visually comparable, indicating that the proposed acceleration strategy does not introduce noticeable perceptual degradation.

For the more aggressive accelerated variant, FlashClear+FPAC, the preference statistics remain close to those of ObjectClear, with a win rate of 22.0\%, a tie rate of 53.5\%, and a lose rate of 24.5\%. Although FlashClear+FPAC applies a stronger acceleration strategy, its loss rate is only slightly higher than its win rate, and the majority of responses are still ties. These results indicate that FlashClear+FPAC largely preserves the perceptual quality of object removal while further improving inference efficiency.

Overall, the user study confirms that our acceleration pipeline maintains removal quality at the perceptual level. Combined with the quantitative acceleration results reported in the main paper, these findings demonstrate that FlashClear achieves a favorable trade-off between inference efficiency and visual quality.
\vspace{-2mm}
\section{More Visual Results}
\vspace{-2mm}
\label{sec:more results}

\begin{figure*}[t]
    \centering
    \setlength{\tabcolsep}{0.5pt}
    \renewcommand{\arraystretch}{0.4}
    \begin{tabular}{@{}cccccccc@{}}
        \accimg{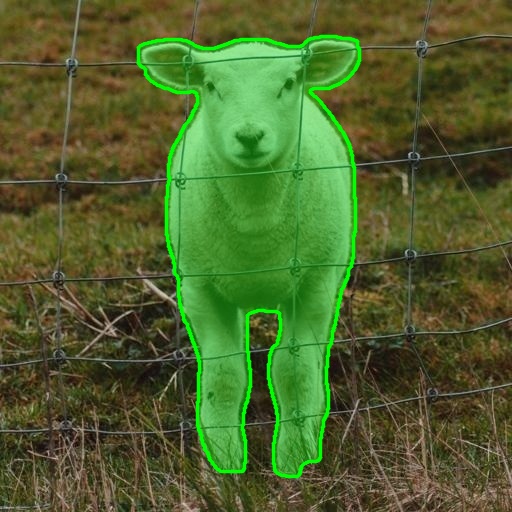} &
        \accimg{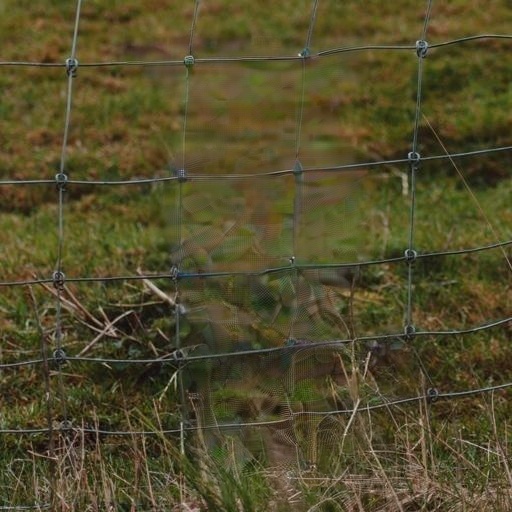} &
        \accimg{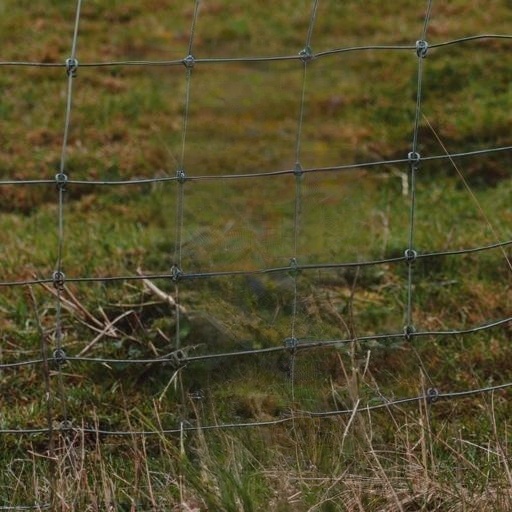} &
        \accimg{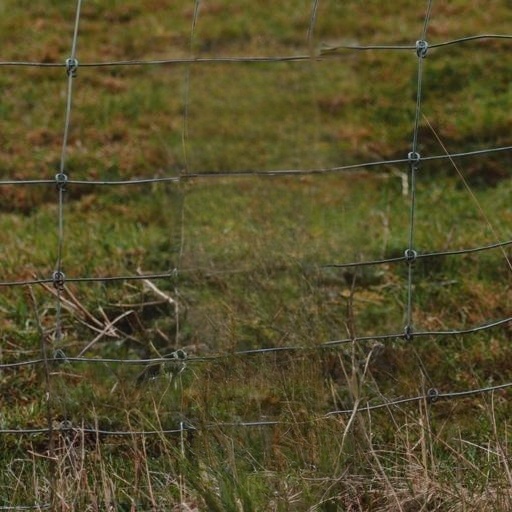} &
        \accimg{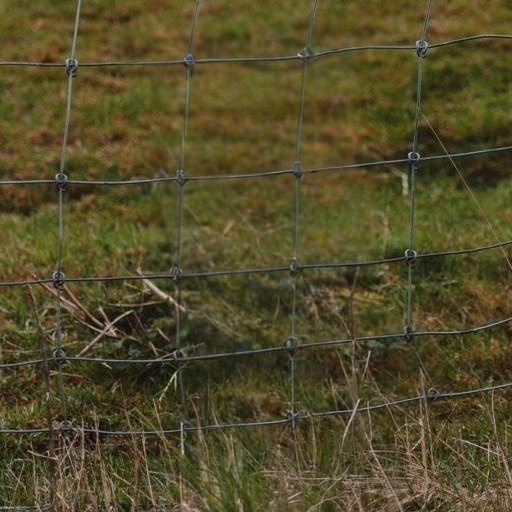} &
        \accimg{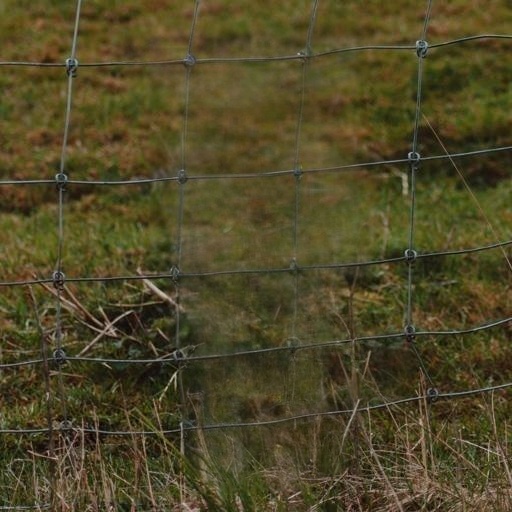} &
        \accimg{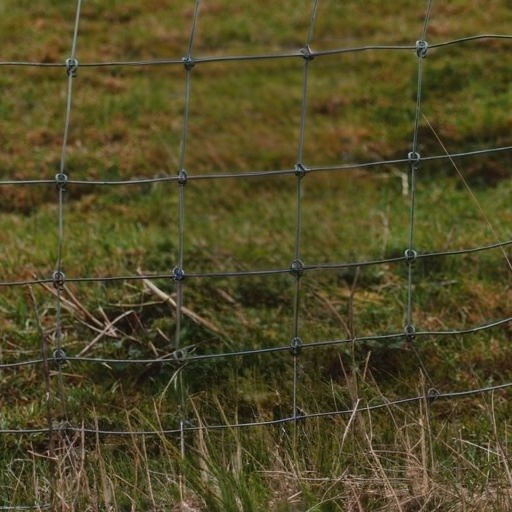} &
        \accimg{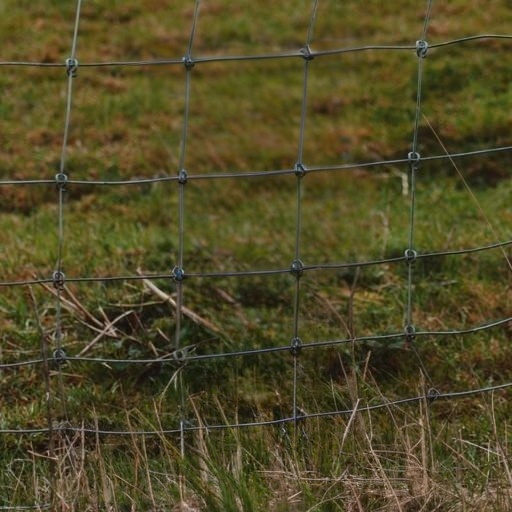} \\[0pt]
        \accimg{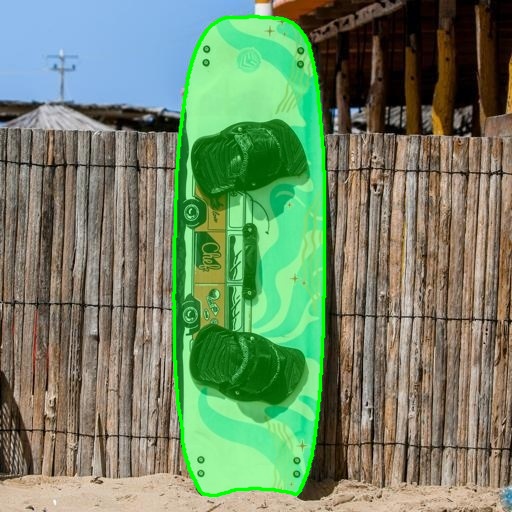} &
        \accimg{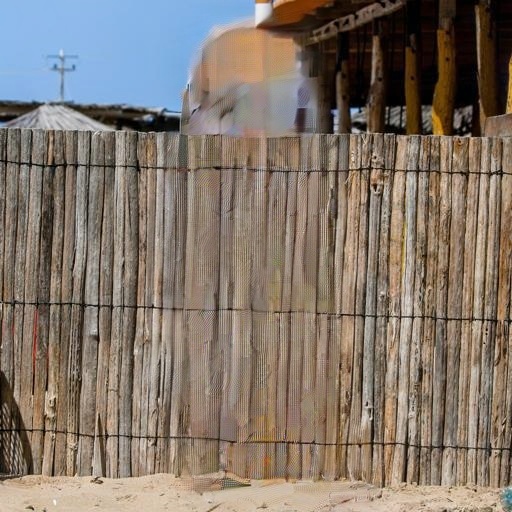} &
        \accimg{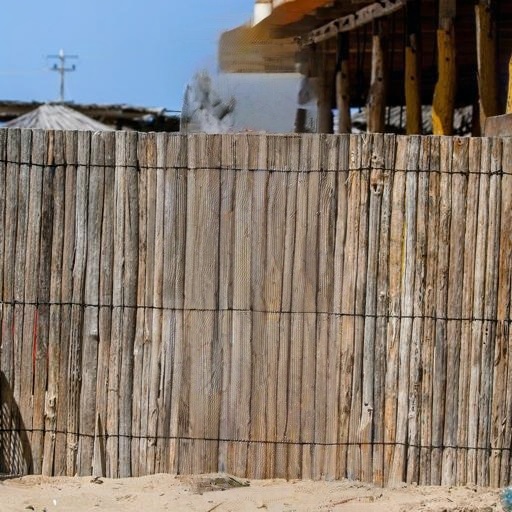} &
        \accimg{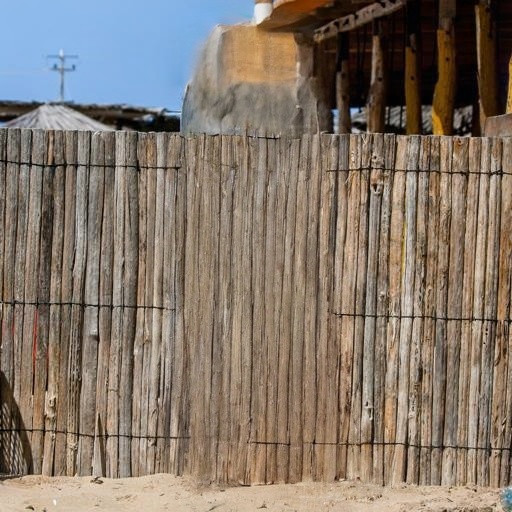} &
        \accimg{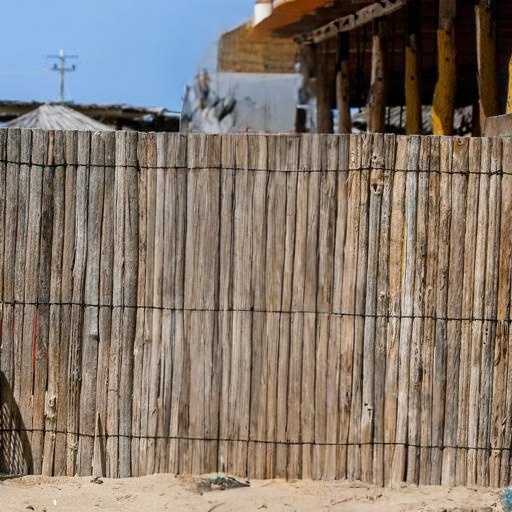} &
        \accimg{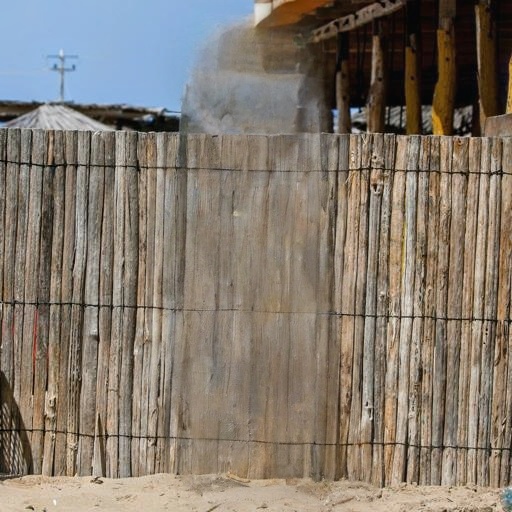} &
        \accimg{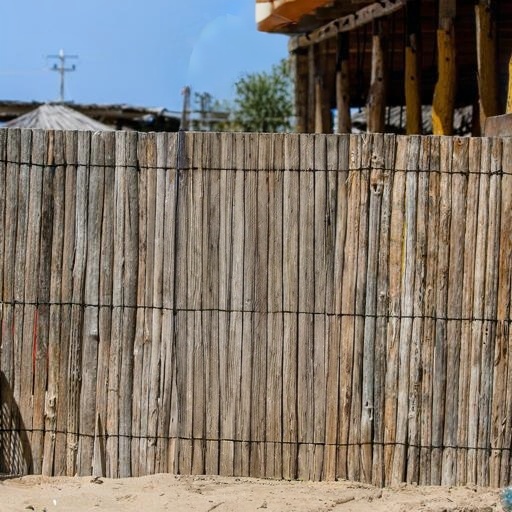} &
        \accimg{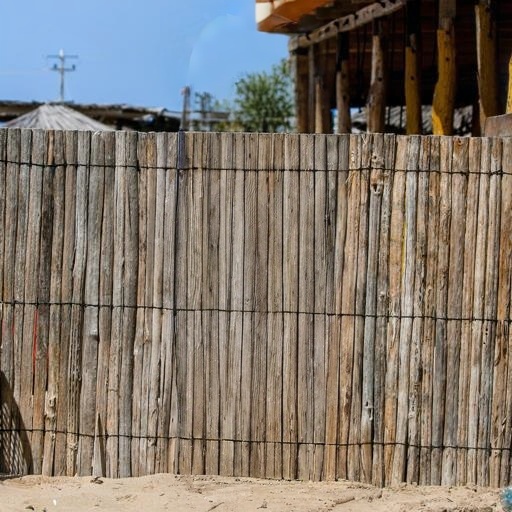} \\[0pt]
        \accimg{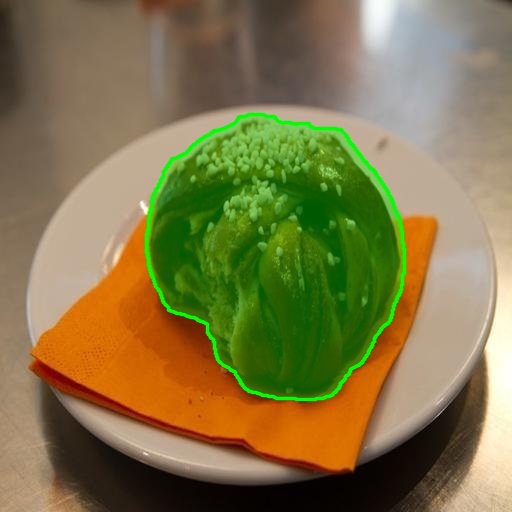} &
        \accimg{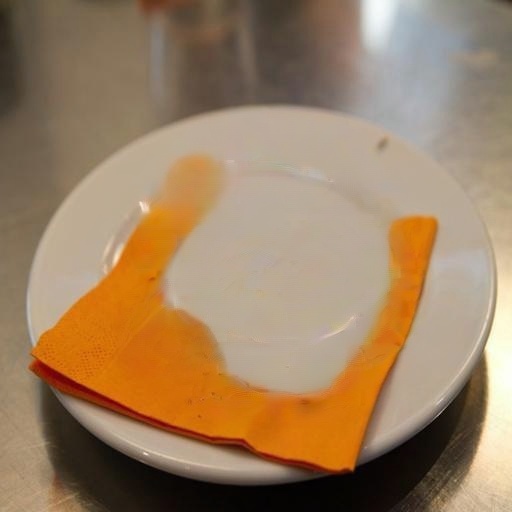} &
        \accimg{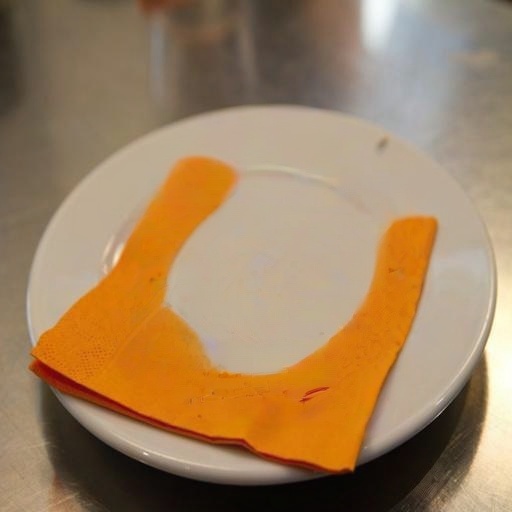} &
        \accimg{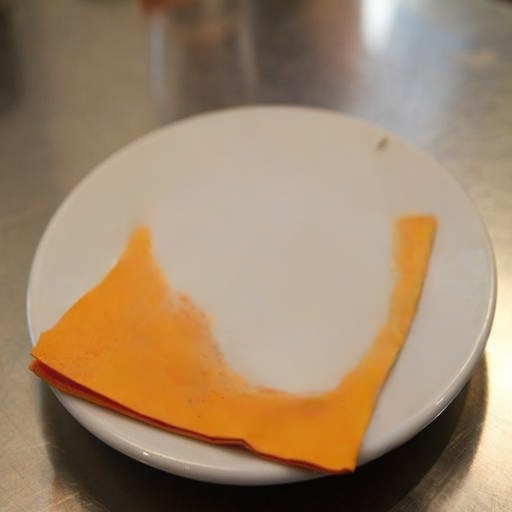} &
        \accimg{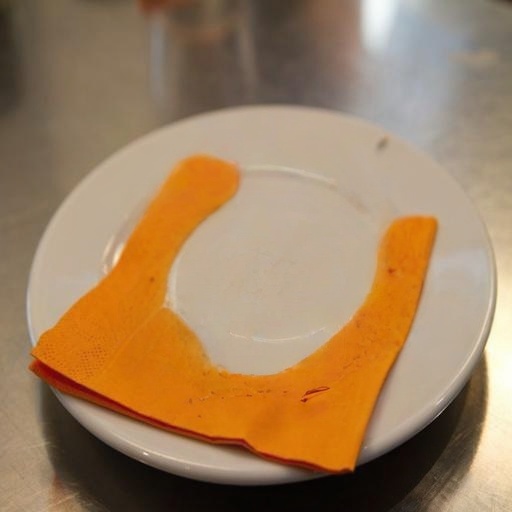} &
        \accimg{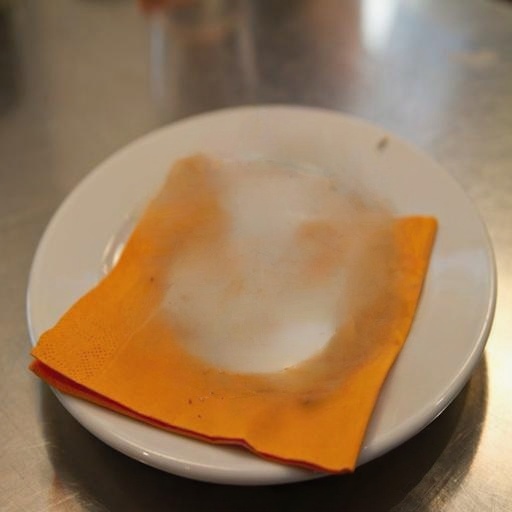} &
        \accimg{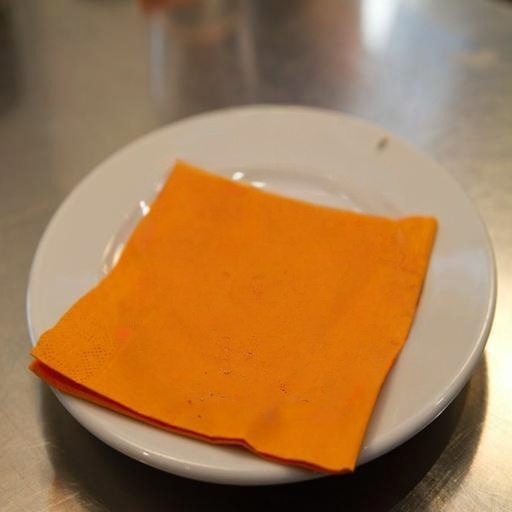} &
        \accimg{00000426_pred_FlashClear.jpg} \\[0pt]
        \accimg{00000366.jpg} &
        \accimg{00366_4step.jpg} &
        \accimg{00366_ToCa.jpg} &
        \accimg{00366_sito.jpg} &
        \accimg{00366_HiCache.jpg} &
        \accimg{00366_quantcache.jpg} &
        \accimg{00366_ours+F-PAC.jpg} &
        \accimg{00366_ours.jpg} \\[0pt]
        \accimg{00000388.jpg} &
        \accimg{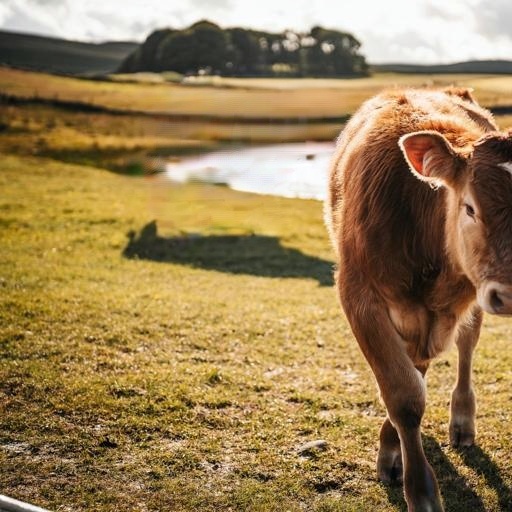} &
        \accimg{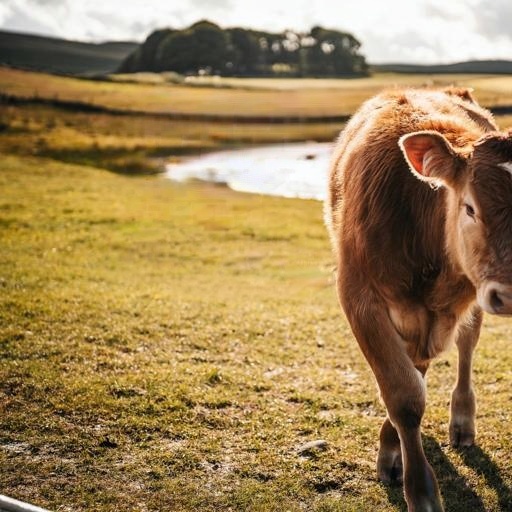} &
        \accimg{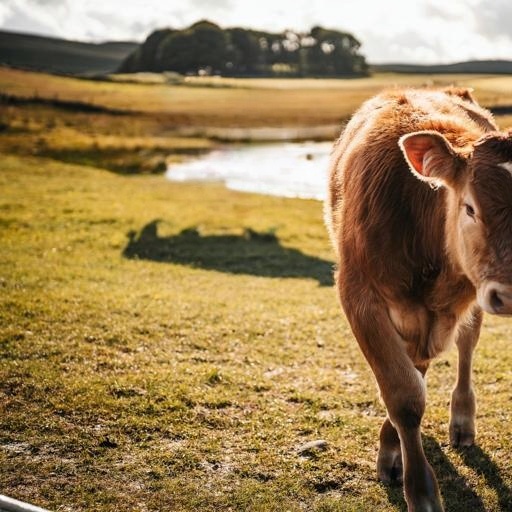} &
        \accimg{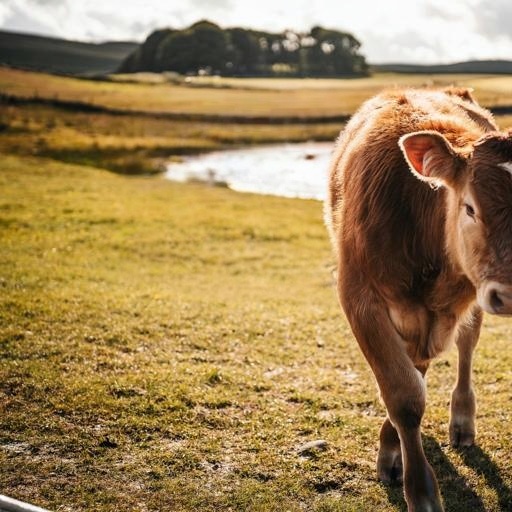} &
        \accimg{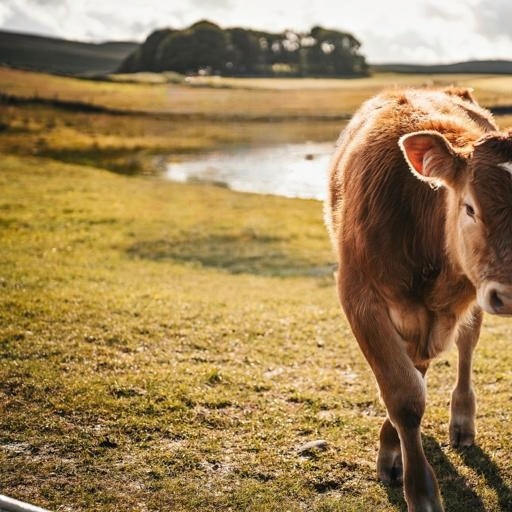} &
        \accimg{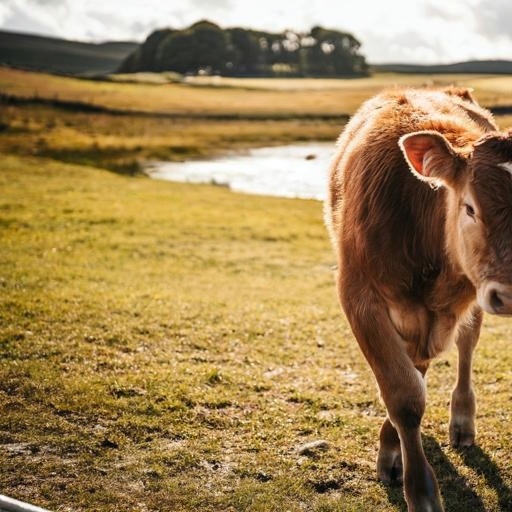} &
        \accimg{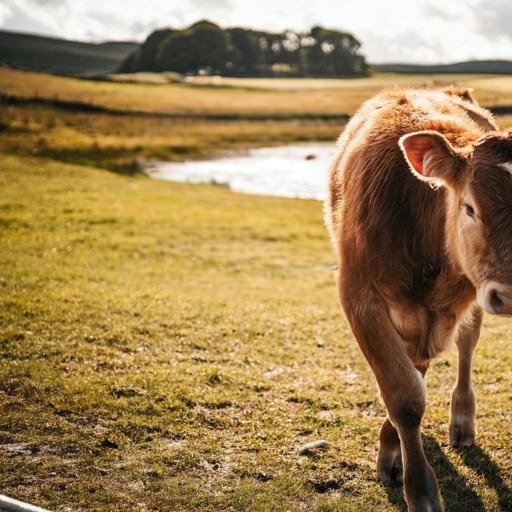} \\[0pt]
        \accimg{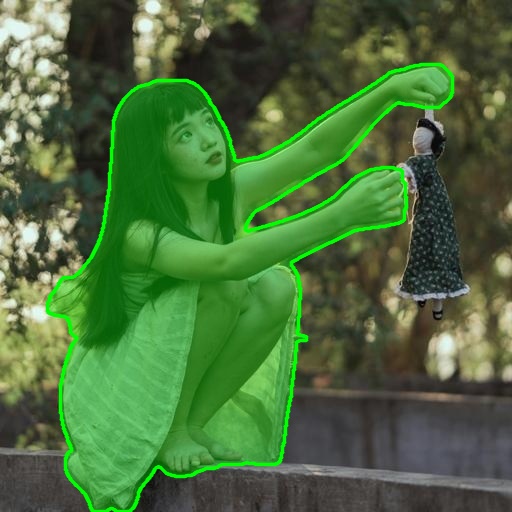} &
        \accimg{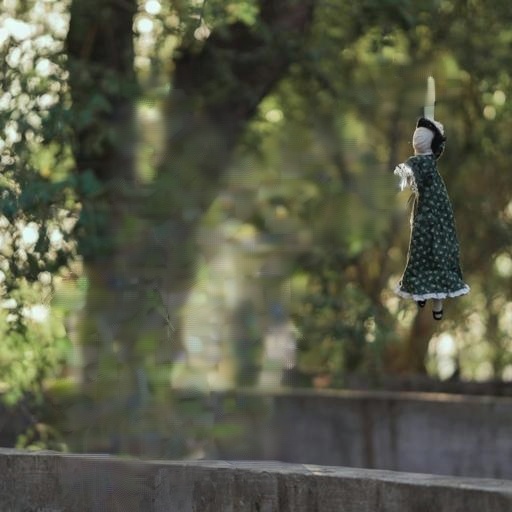} &
        \accimg{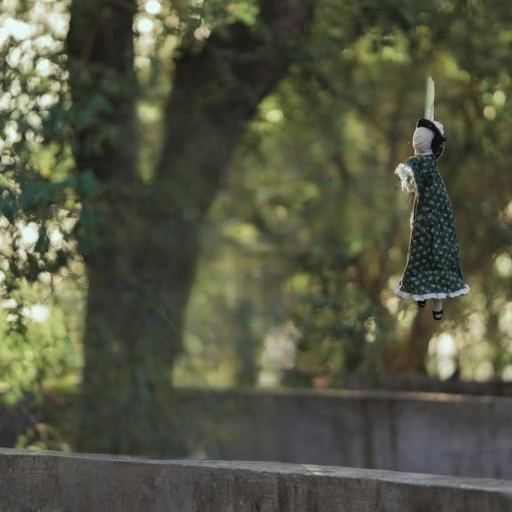} &
        \accimg{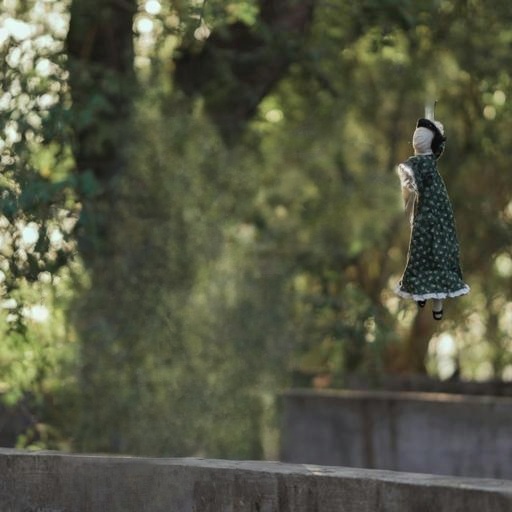} &
        \accimg{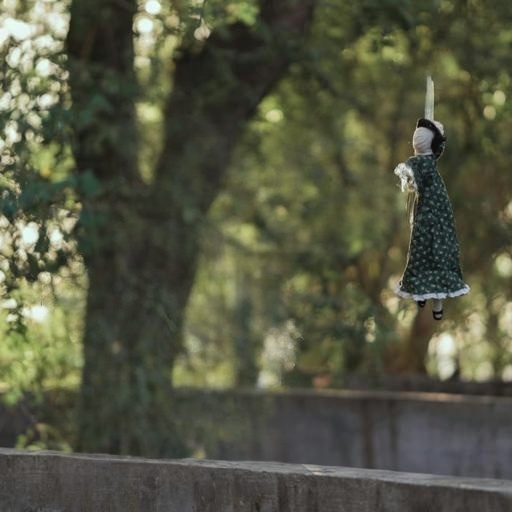} &
        \accimg{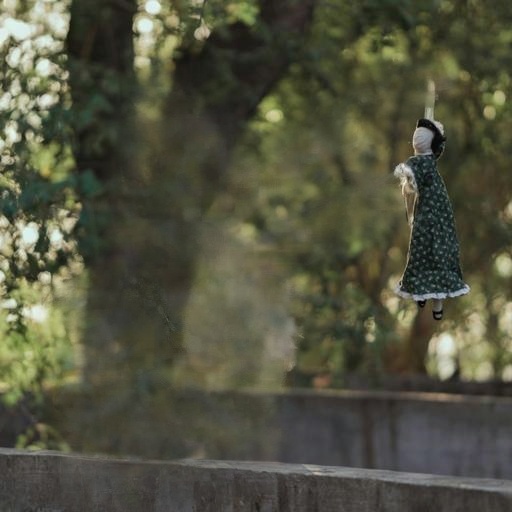} &
        \accimg{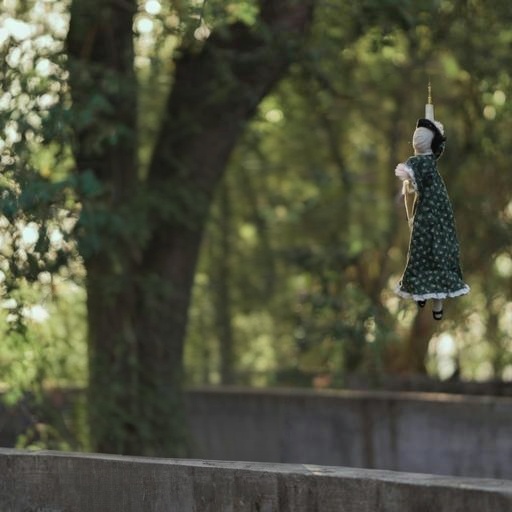} &
        \accimg{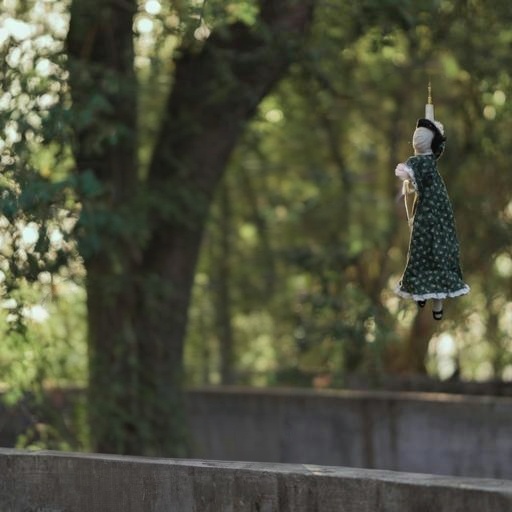} \\[0pt]
        \accimg{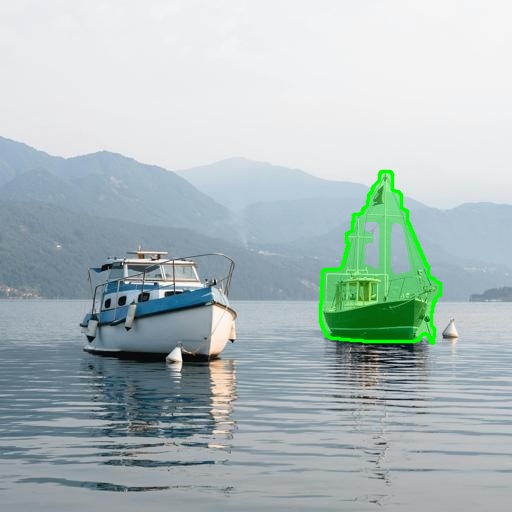} &
        \accimg{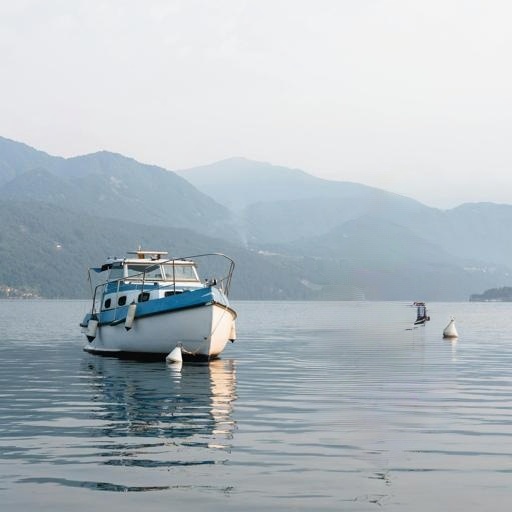} &
        \accimg{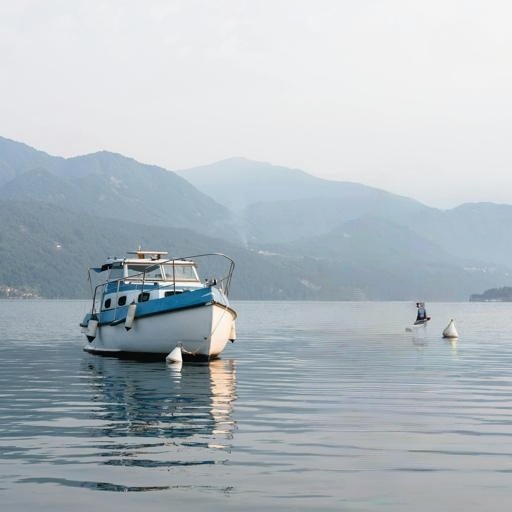} &
        \accimg{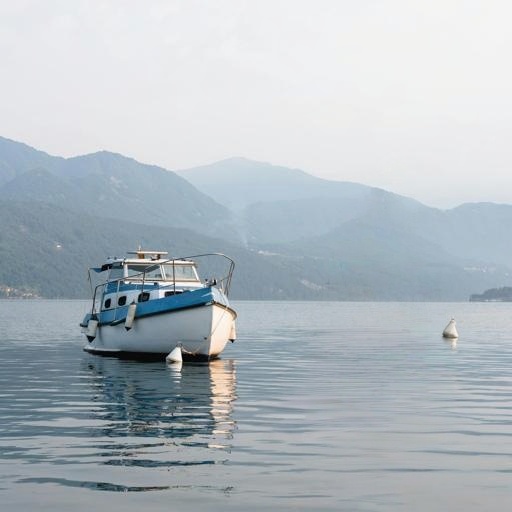} &
        \accimg{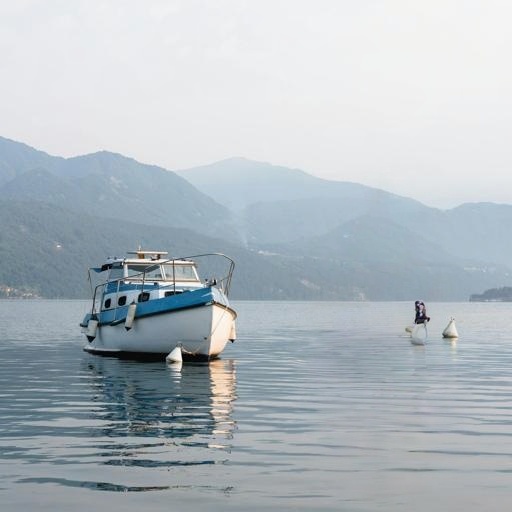} &
        \accimg{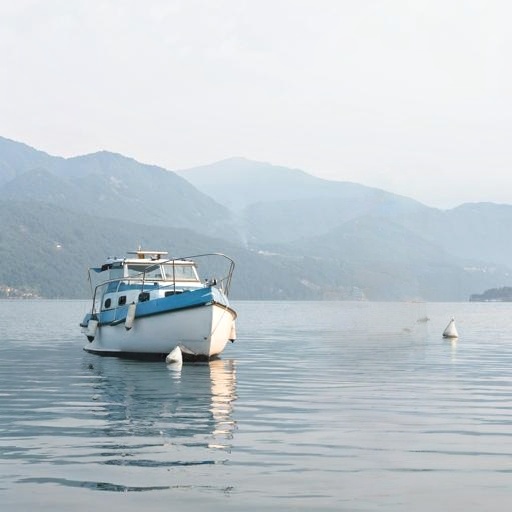} &
        \accimg{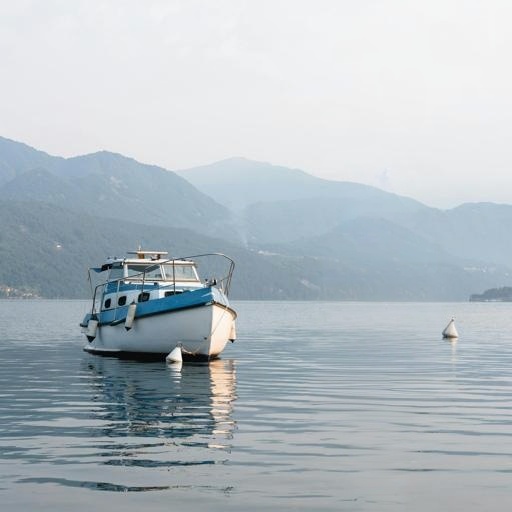} &
        \accimg{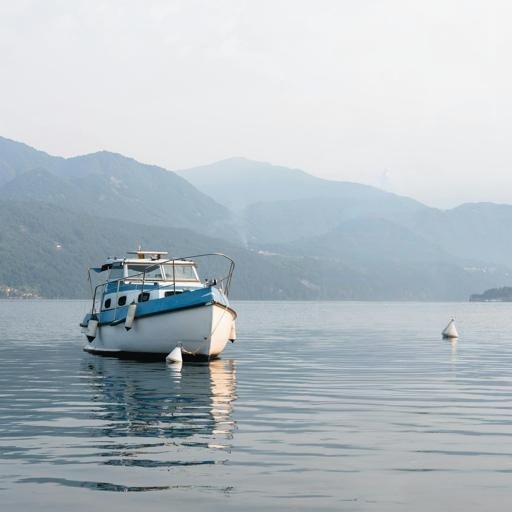} \\[0pt]
        \accimg{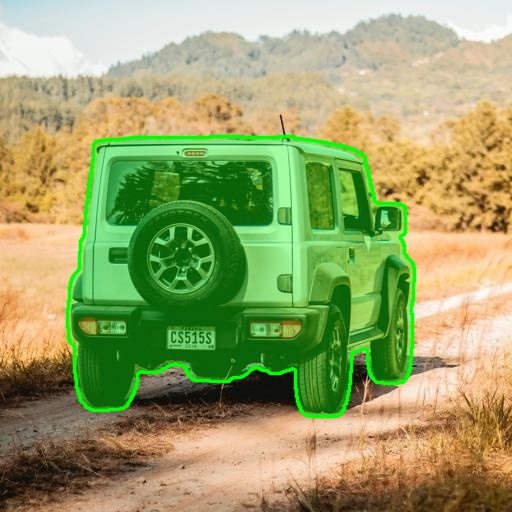} &
        \accimg{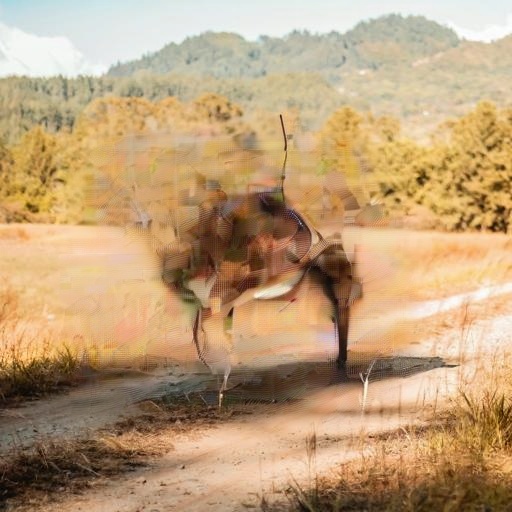} &
        \accimg{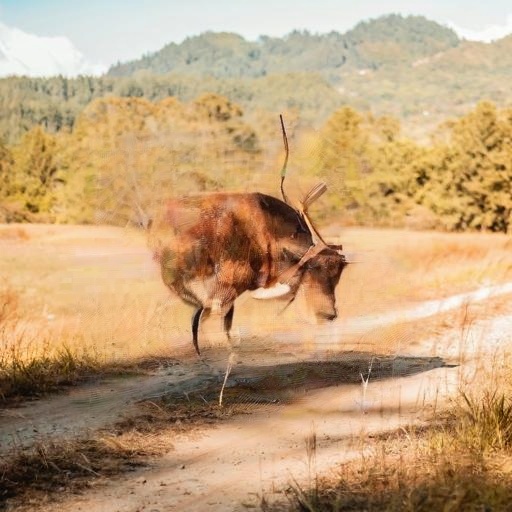} &
        \accimg{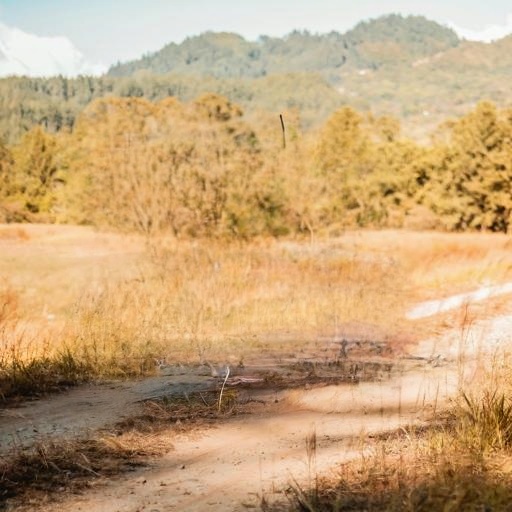} &
        \accimg{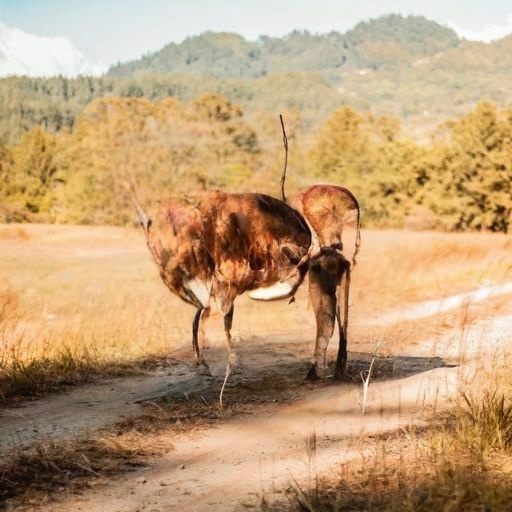} &
        \accimg{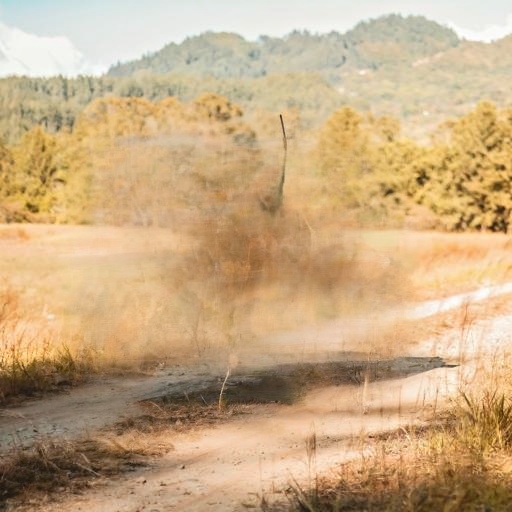} &
        \accimg{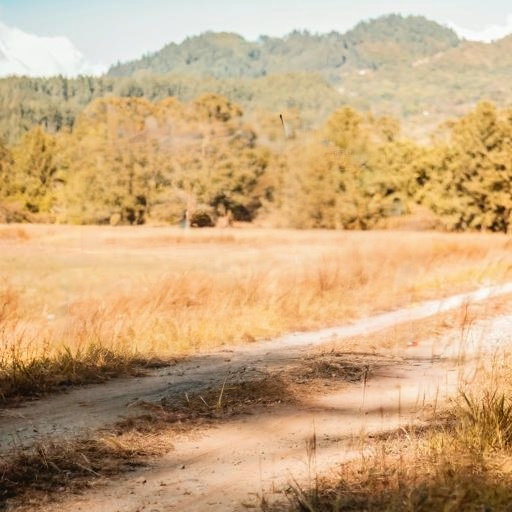} &
        \accimg{00000269_pred.jpg} \\[0pt]
        \accimg{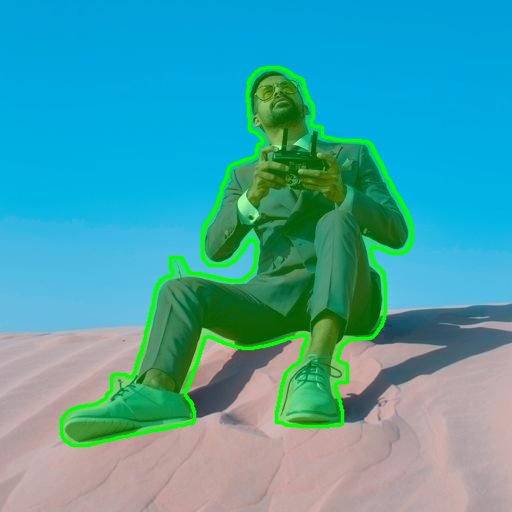} &
        \accimg{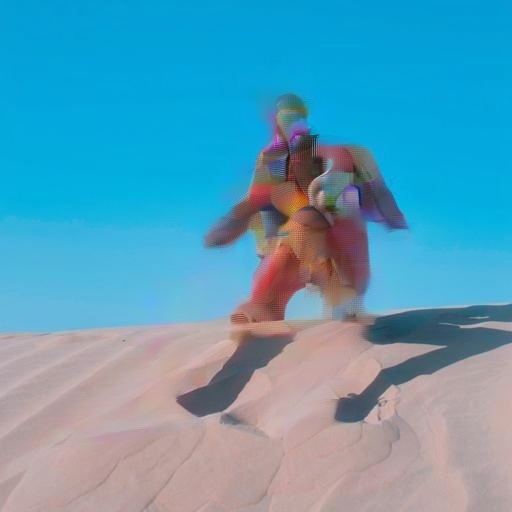} &
        \accimg{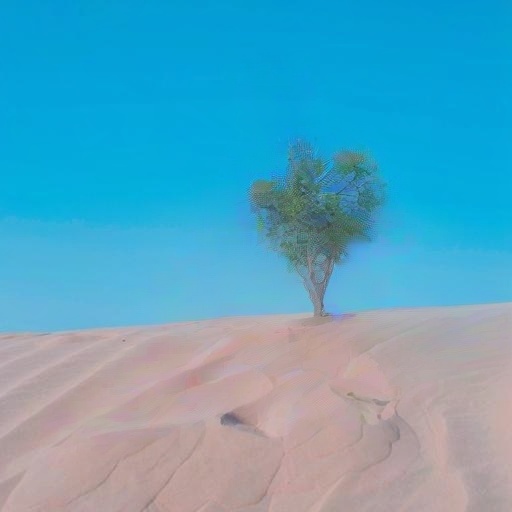} &
        \accimg{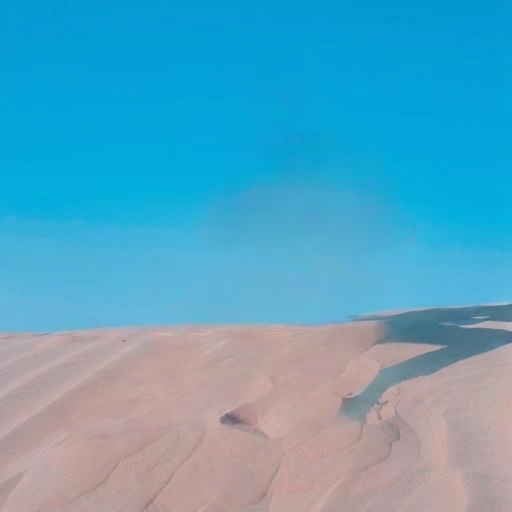} &
        \accimg{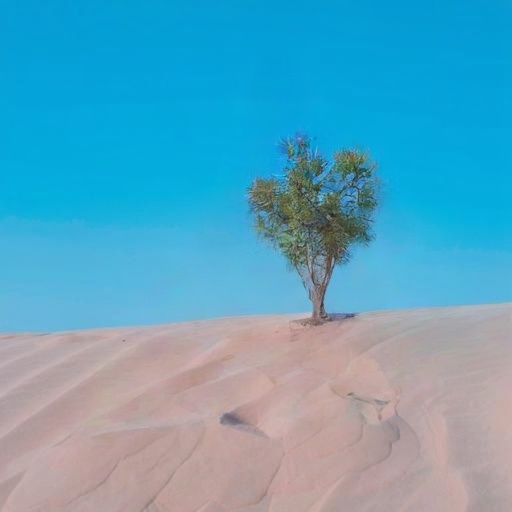} &
        \accimg{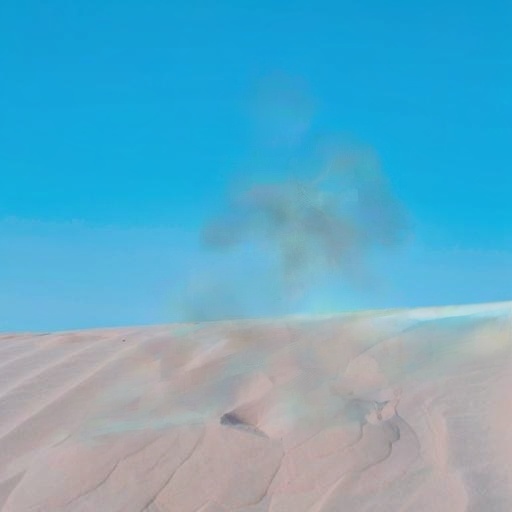} &
        \accimg{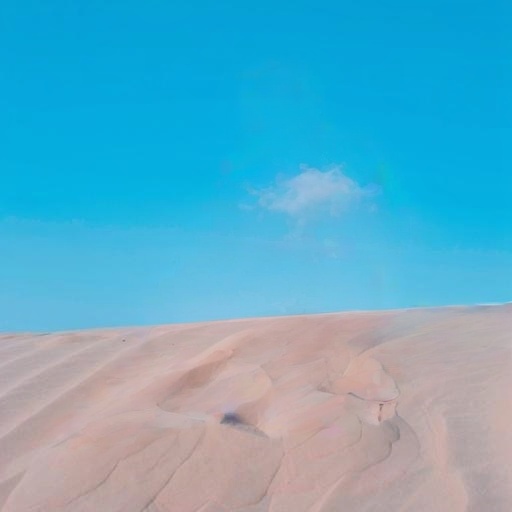} &
        \accimg{00000302_pred.jpg} \\[0pt]
        \accimg{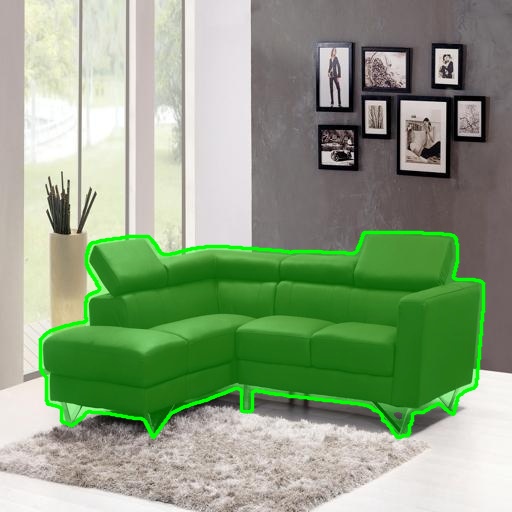} &
        \accimg{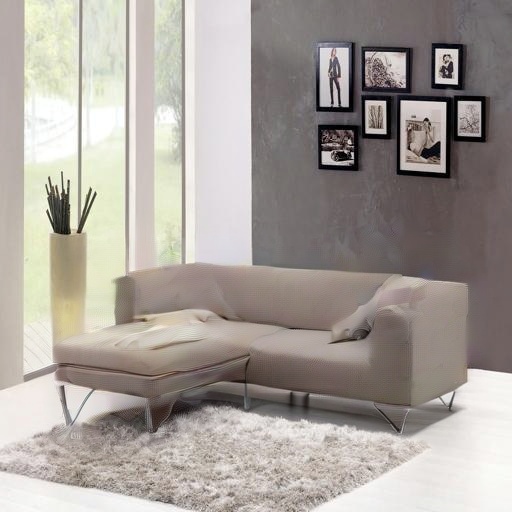} &
        \accimg{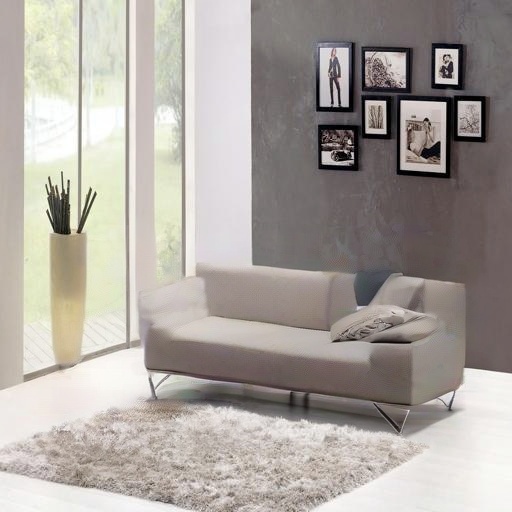} &
        \accimg{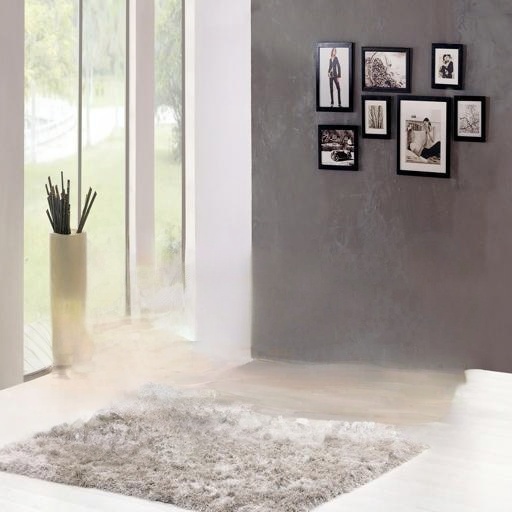} &
        \accimg{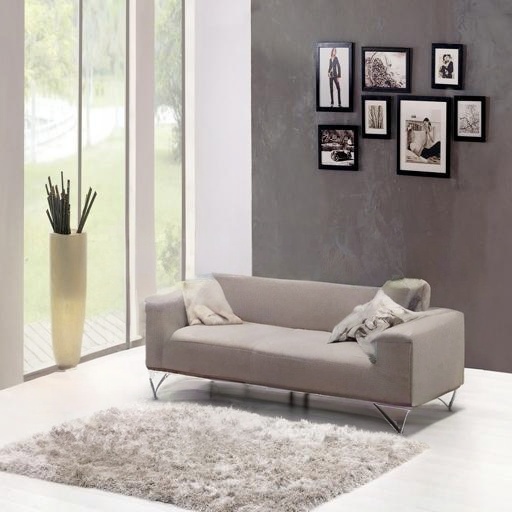} &
        \accimg{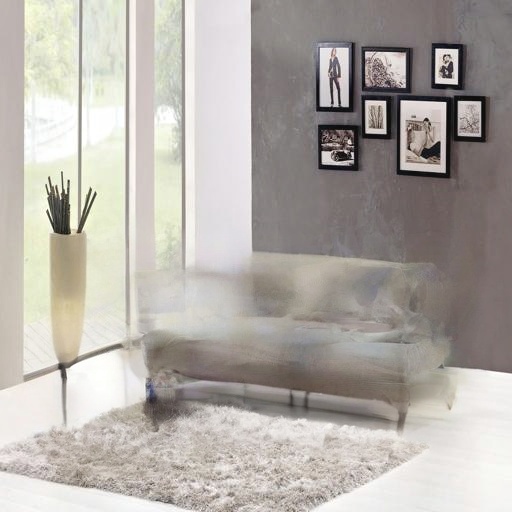} &
        \accimg{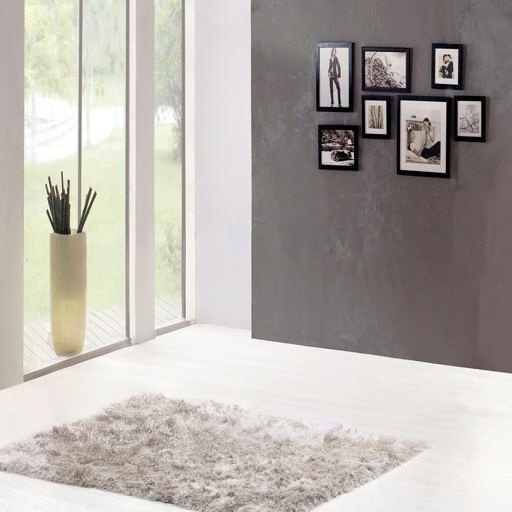} &
        \accimg{00000319_pred.jpg} \\[0pt]
        
        \\[4pt]
        \acclabel{Image w/ mask} &
        \acclabel{20\% step} &
        \acclabel{ToCa~\cite{zou2024accelerating}} &
        \acclabel{SiTo~\cite{zhang2025training}} &
        \acclabel{HiCache~\cite{feng2025hicache}} &
        \acclabel{QuantCache~\cite{wu2025quantcache}} &
        \acclabel{FlashClear-C} &
        \acclabel{FlashClear}
    \end{tabular}
    \vspace{-1mm}
    \caption{More qualitative comparison on acceleration methods based on ObjectClear~\cite{zhao2025objectclear}.}
\label{fig:acc_qualitative_comparison_more}
\vspace{-2.5mm}
\end{figure*}

\begin{figure}[h] 
    \centering
    \setlength{\tabcolsep}{0.5pt}      
    \renewcommand{\arraystretch}{1.2} 
    
    \begin{tabular}{@{}cccccc@{}}
        
        \includegraphics[width=0.16\linewidth]{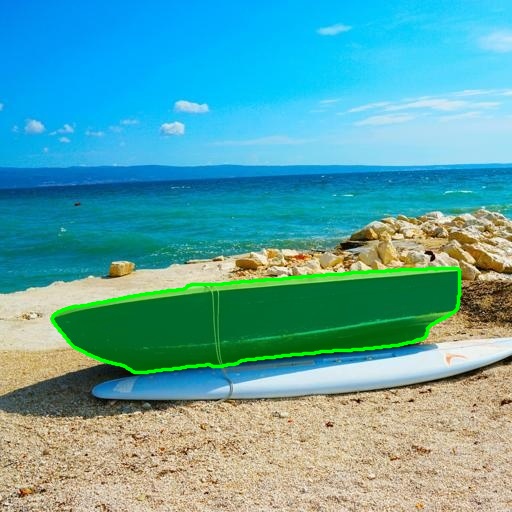} &
        \includegraphics[width=0.16\linewidth]{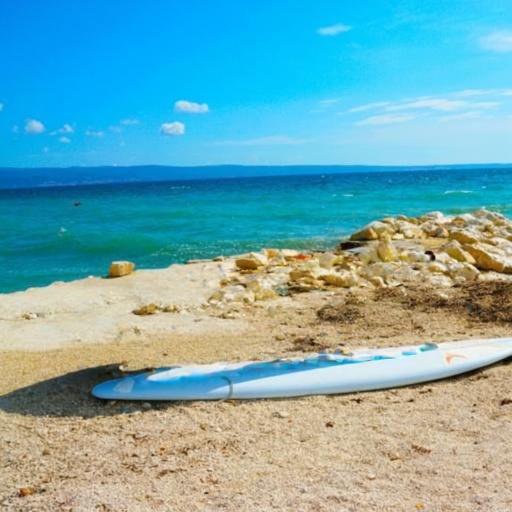} &
        \includegraphics[width=0.16\linewidth]{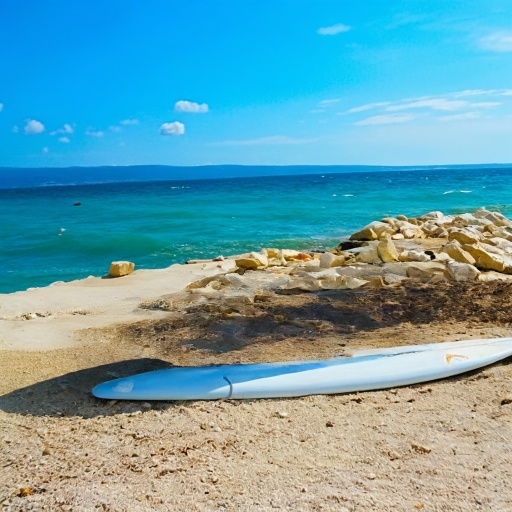} &
        \includegraphics[width=0.16\linewidth]{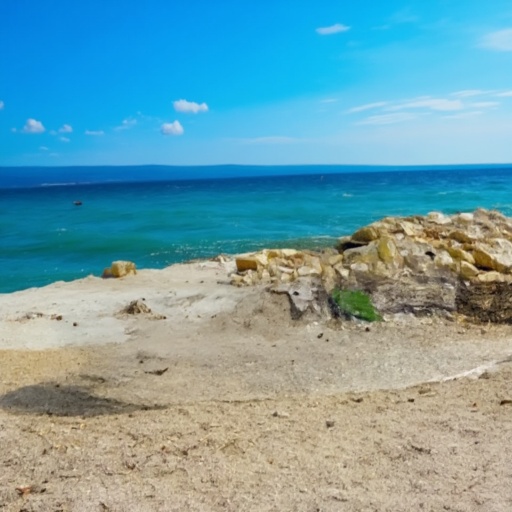} &
        \includegraphics[width=0.16\linewidth]{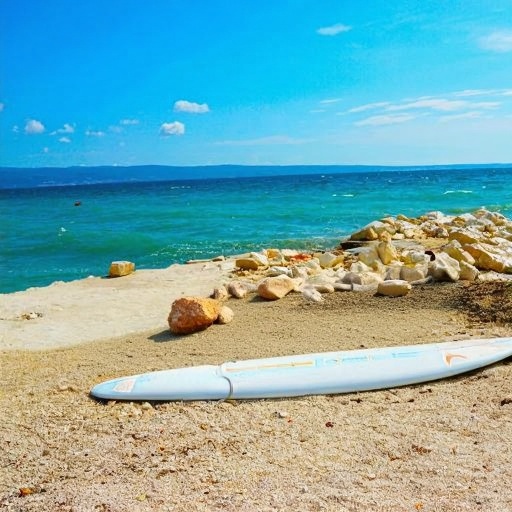} &
        \includegraphics[width=0.16\linewidth]{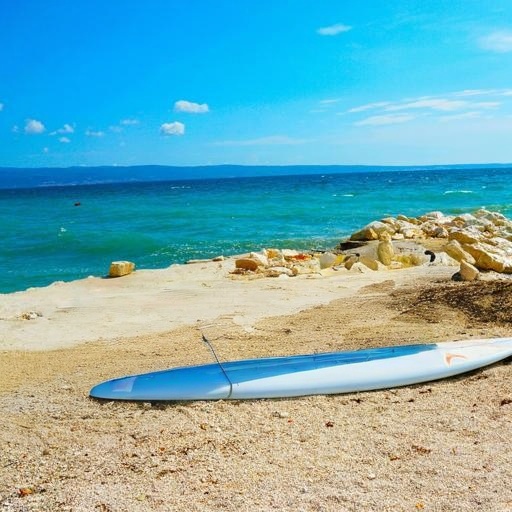} \\[-2pt]

        \includegraphics[width=0.16\linewidth]{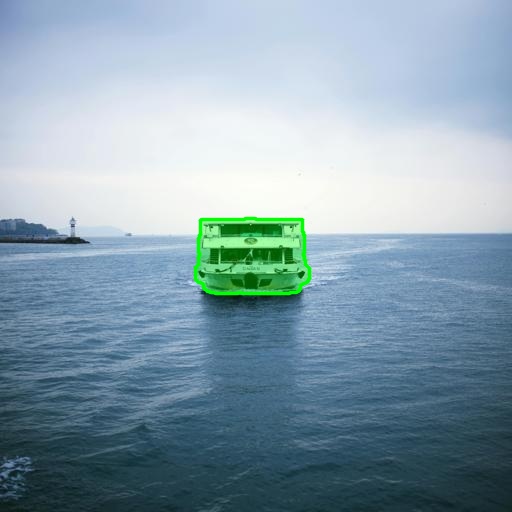} &
        \includegraphics[width=0.16\linewidth]{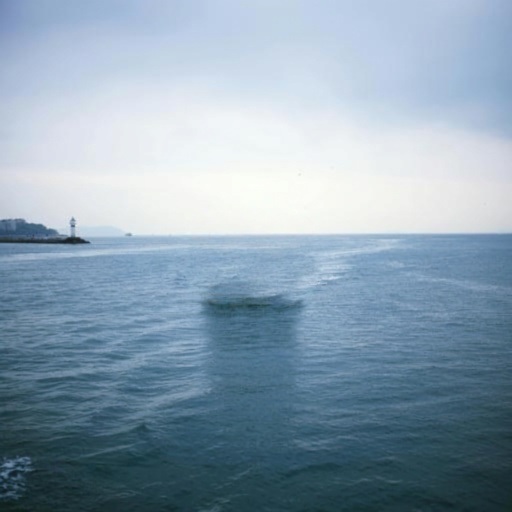} &
        \includegraphics[width=0.16\linewidth]{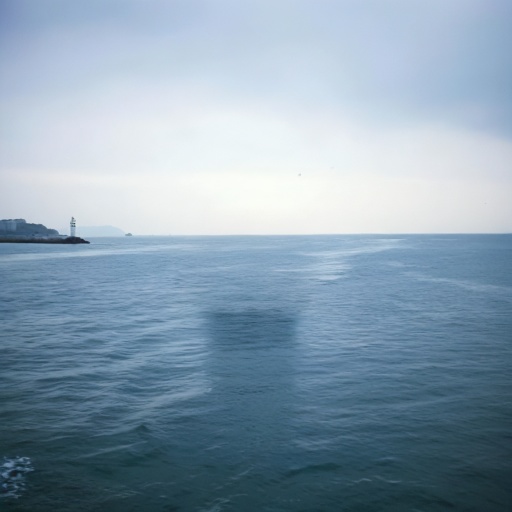} &
        \includegraphics[width=0.16\linewidth]{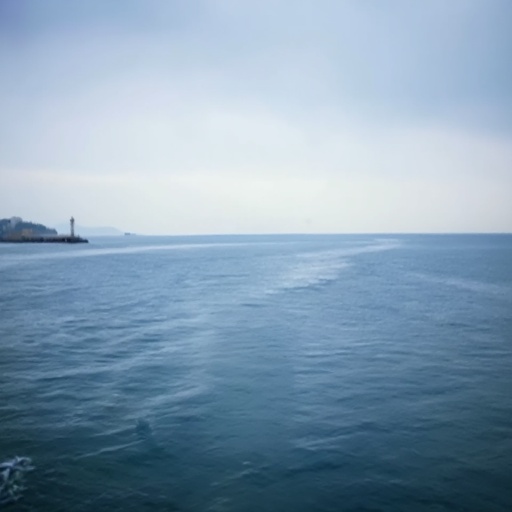} &
        \includegraphics[width=0.16\linewidth]{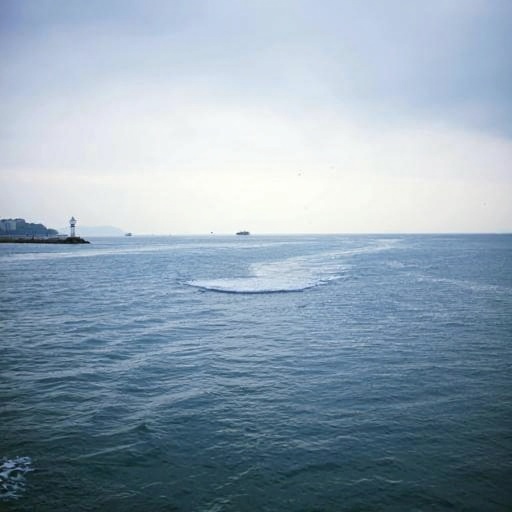} &
        \includegraphics[width=0.16\linewidth]{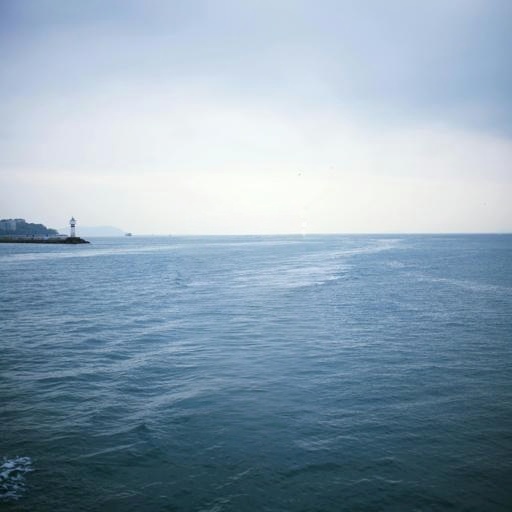} \\[-2pt]

        \includegraphics[width=0.16\linewidth]{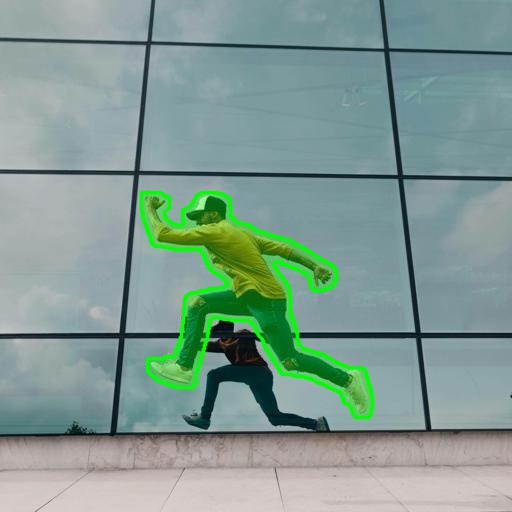} &
        \includegraphics[width=0.16\linewidth]{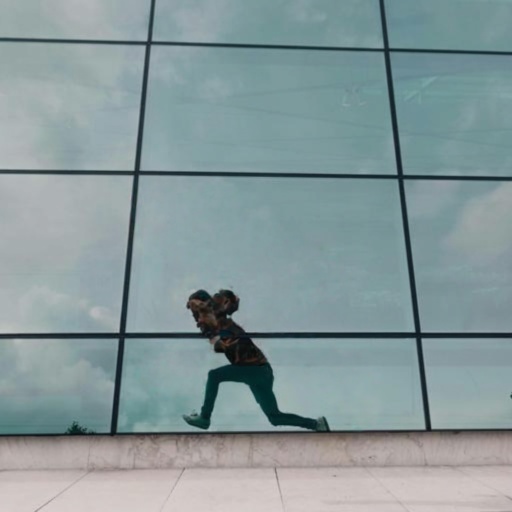} &
        \includegraphics[width=0.16\linewidth]{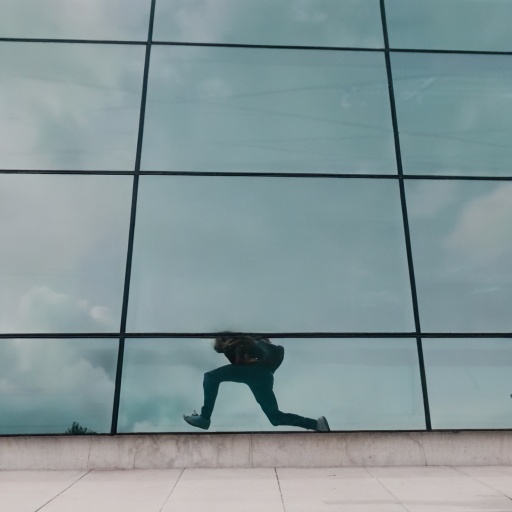} &
        \includegraphics[width=0.16\linewidth]{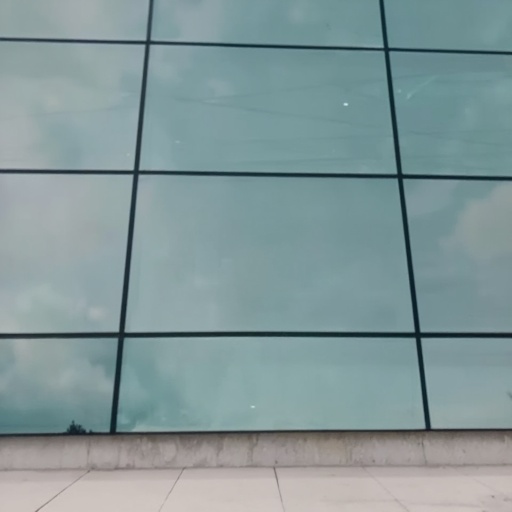} &
        \includegraphics[width=0.16\linewidth]{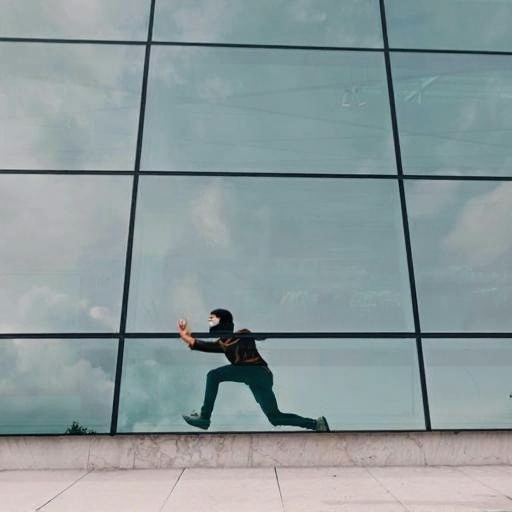} &
        \includegraphics[width=0.16\linewidth]{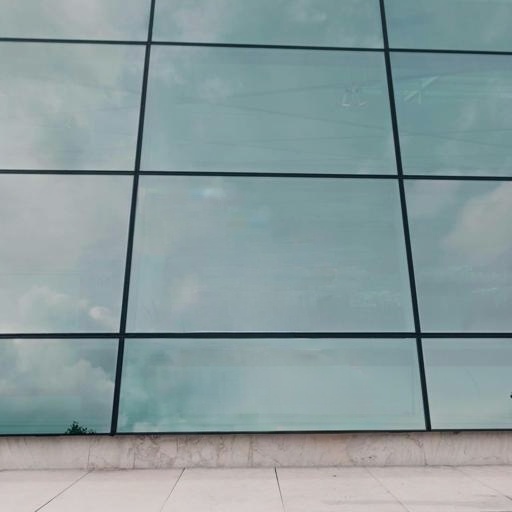} \\[-2pt]

        \includegraphics[width=0.16\linewidth]{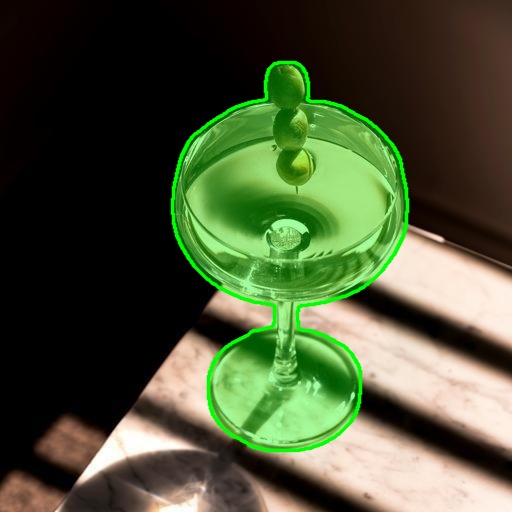} &
        \includegraphics[width=0.16\linewidth]{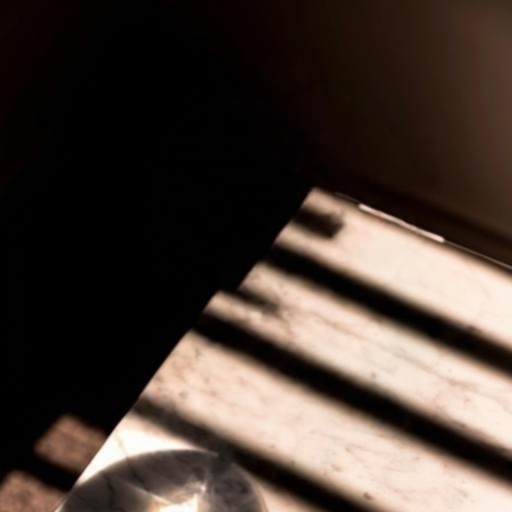} &
        \includegraphics[width=0.16\linewidth]{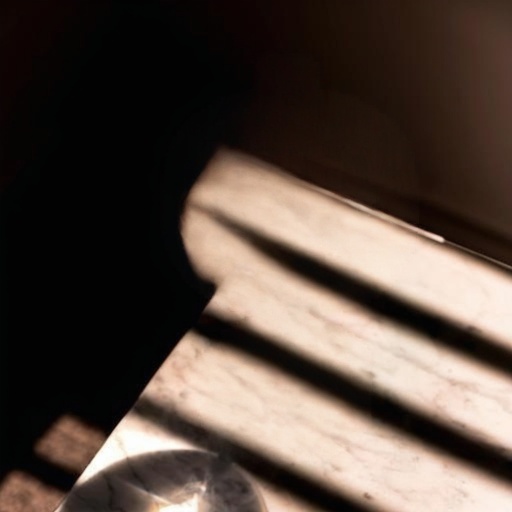} &
        \includegraphics[width=0.16\linewidth]{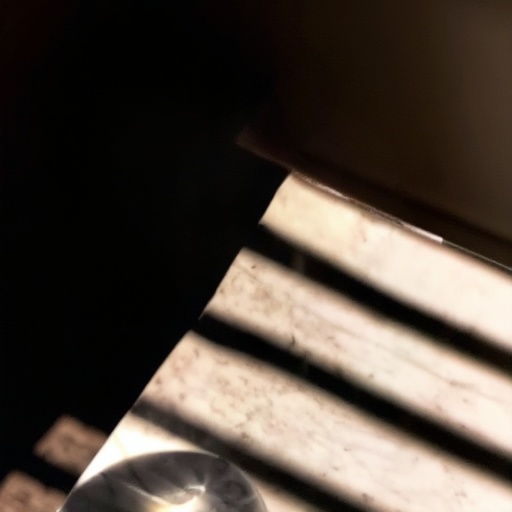} &
        \includegraphics[width=0.16\linewidth]{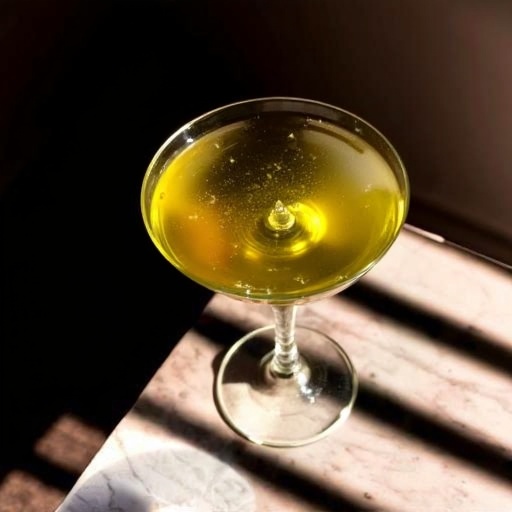} &
        \includegraphics[width=0.16\linewidth]{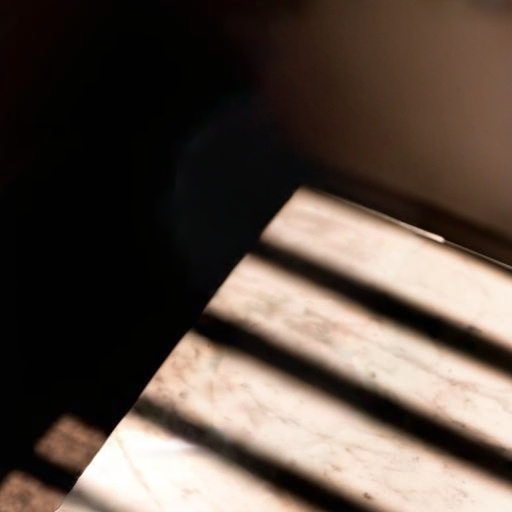} \\[-2pt]

        \includegraphics[width=0.16\linewidth]{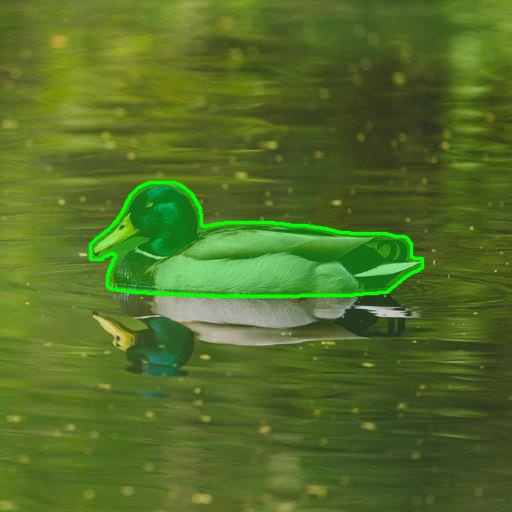} &
        \includegraphics[width=0.16\linewidth]{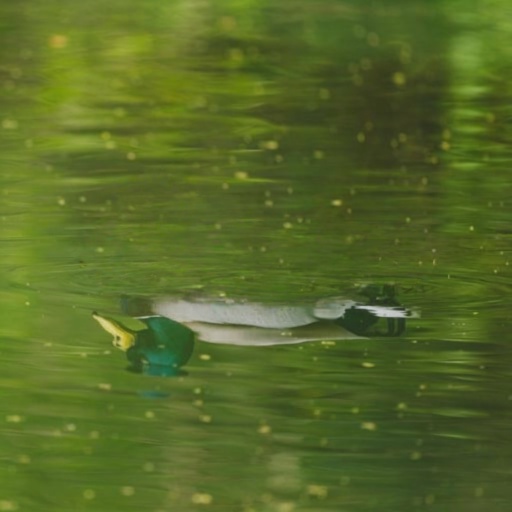} &
        \includegraphics[width=0.16\linewidth]{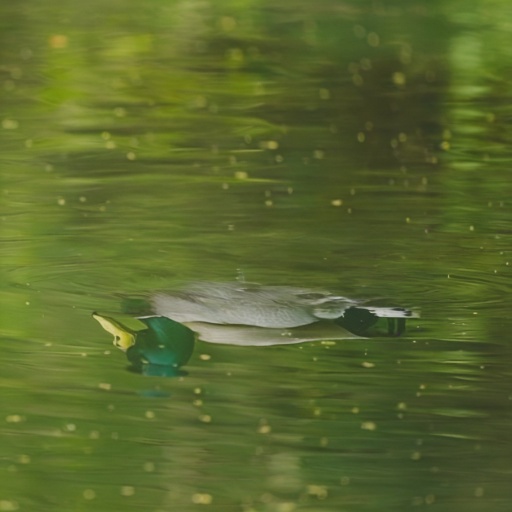} &
        \includegraphics[width=0.16\linewidth]{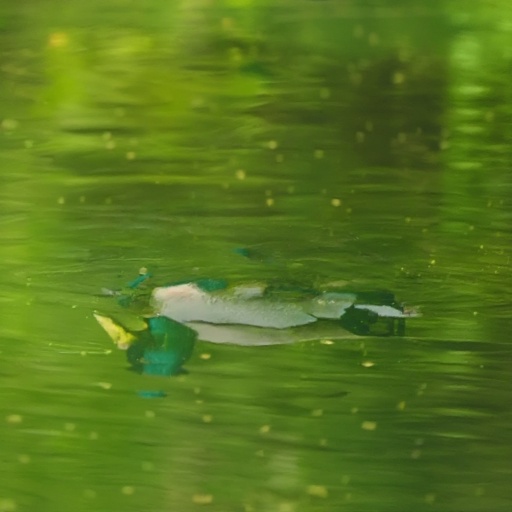} &
        \includegraphics[width=0.16\linewidth]{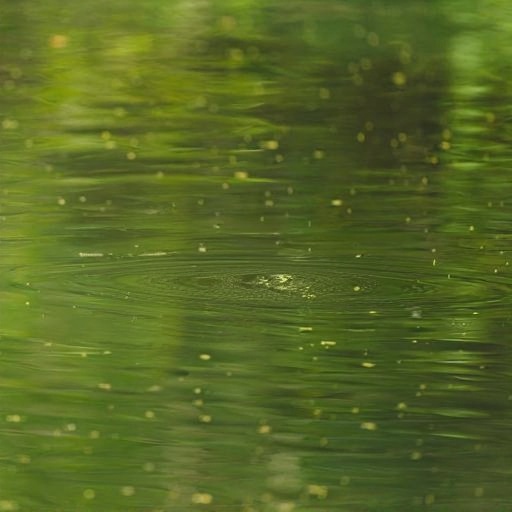} &
        \includegraphics[width=0.16\linewidth]{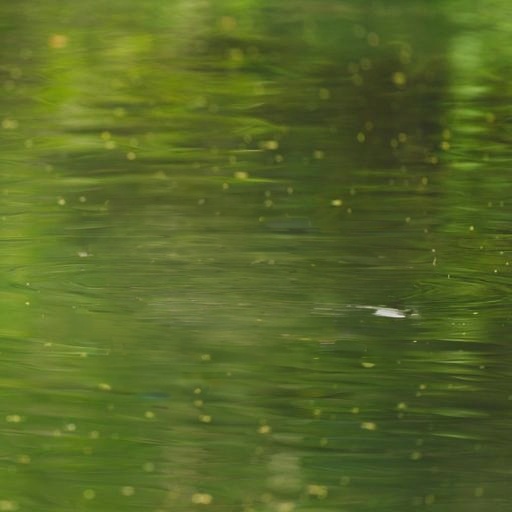} \\[-2pt]

        \includegraphics[width=0.16\linewidth]{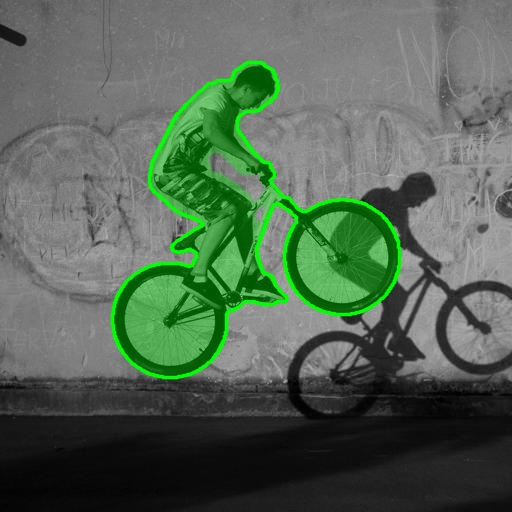} &
        \includegraphics[width=0.16\linewidth]{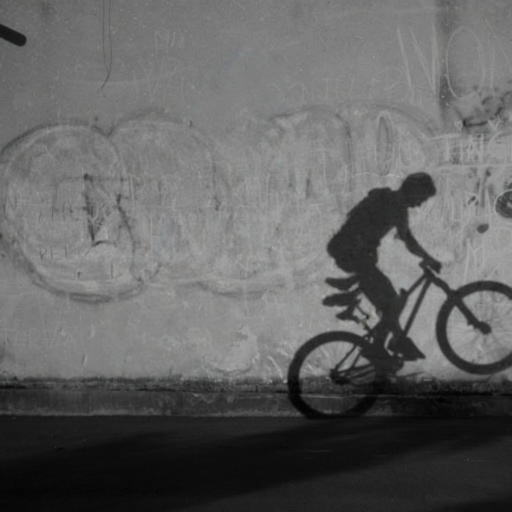} &
        \includegraphics[width=0.16\linewidth]{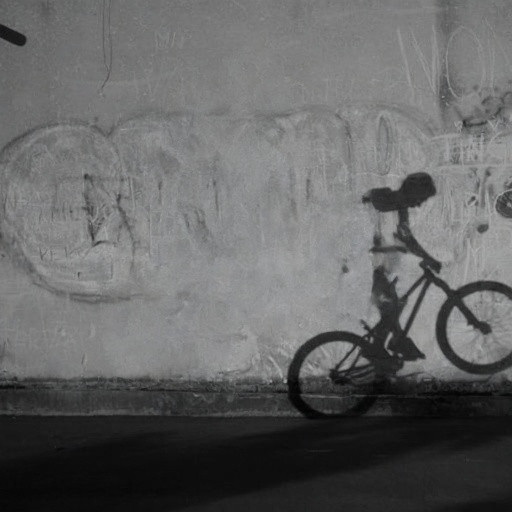} &
        \includegraphics[width=0.16\linewidth]{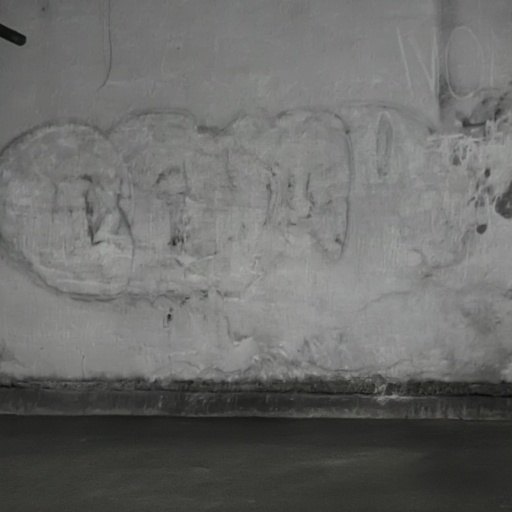} &
        \includegraphics[width=0.16\linewidth]{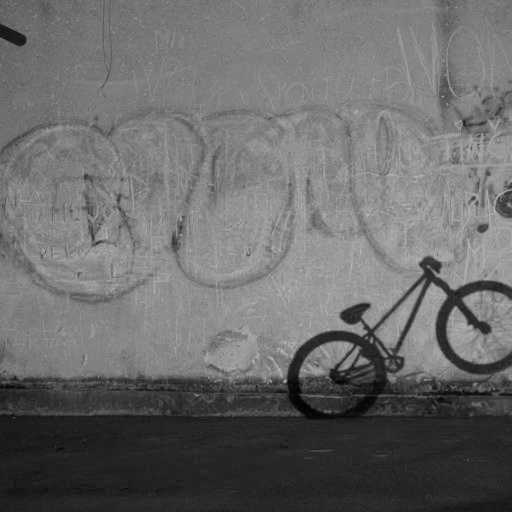} &
        \includegraphics[width=0.16\linewidth]{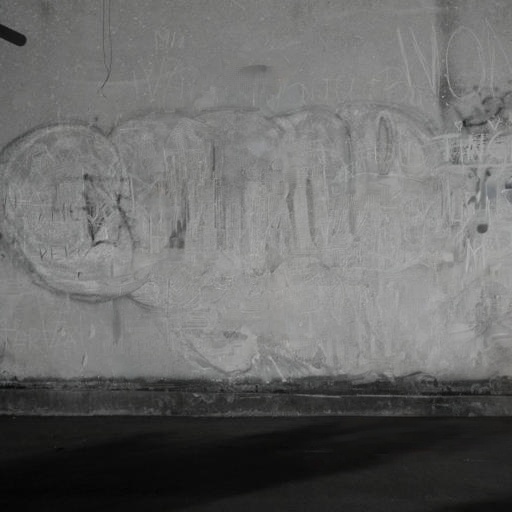} \\[-2pt]

        \includegraphics[width=0.16\linewidth]{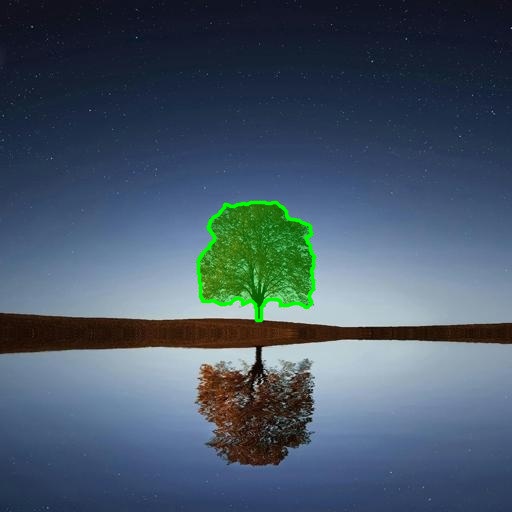} &
        \includegraphics[width=0.16\linewidth]{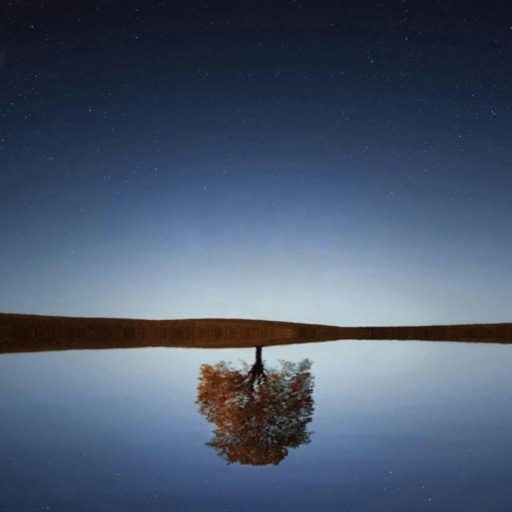} &
        \includegraphics[width=0.16\linewidth]{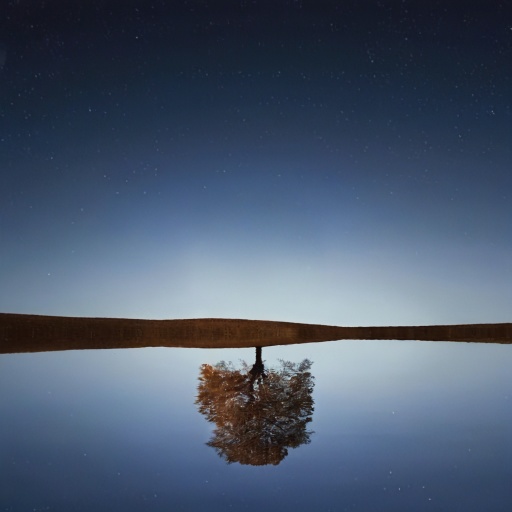} &
        \includegraphics[width=0.16\linewidth]{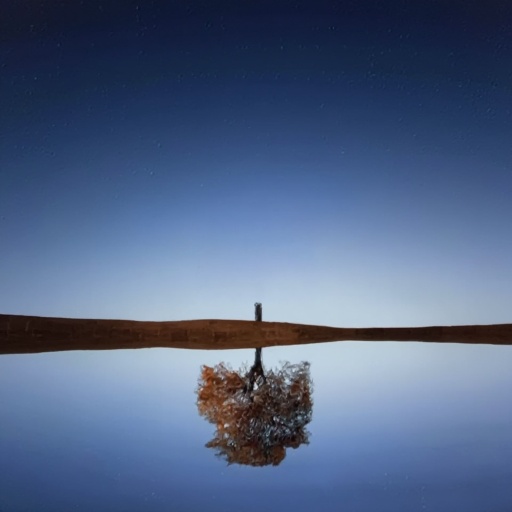} &
        \includegraphics[width=0.16\linewidth]{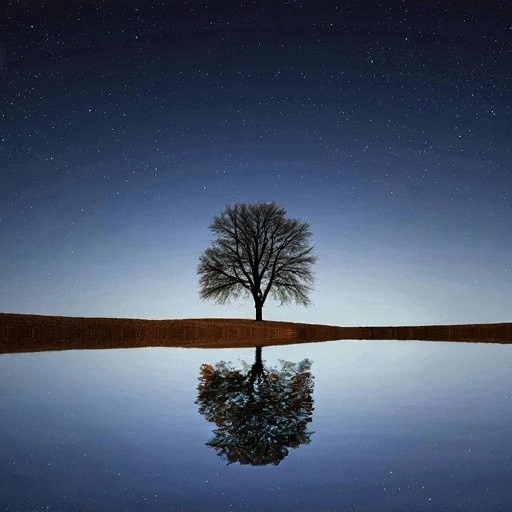} &
        \includegraphics[width=0.16\linewidth]{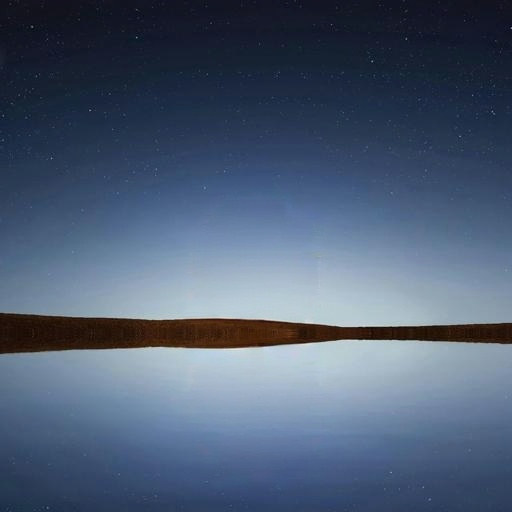} \\[-2pt]

        \includegraphics[width=0.16\linewidth]{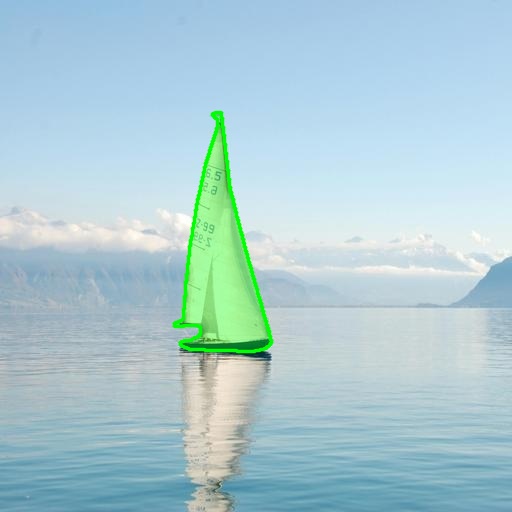} &
        \includegraphics[width=0.16\linewidth]{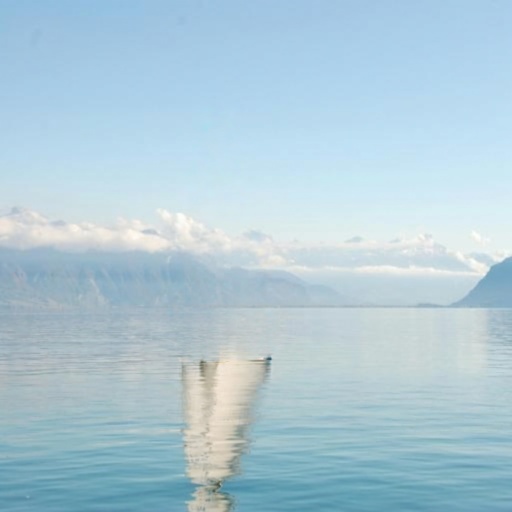} &
        \includegraphics[width=0.16\linewidth]{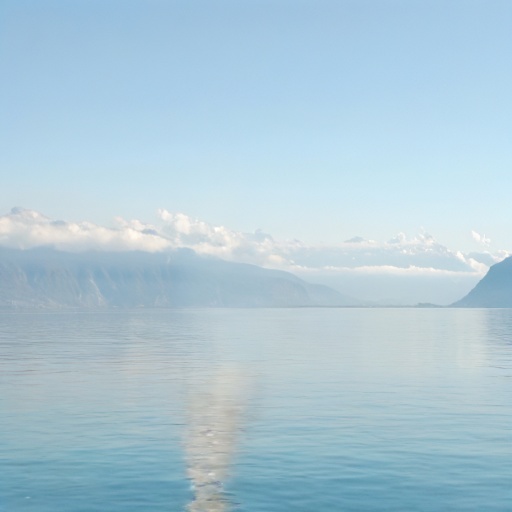} &
        \includegraphics[width=0.16\linewidth]{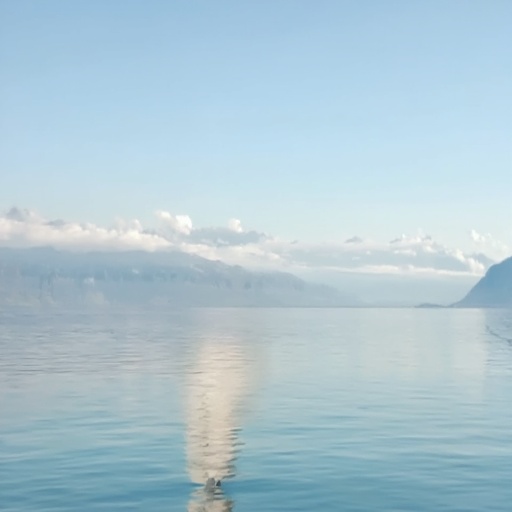} &
        \includegraphics[width=0.16\linewidth]{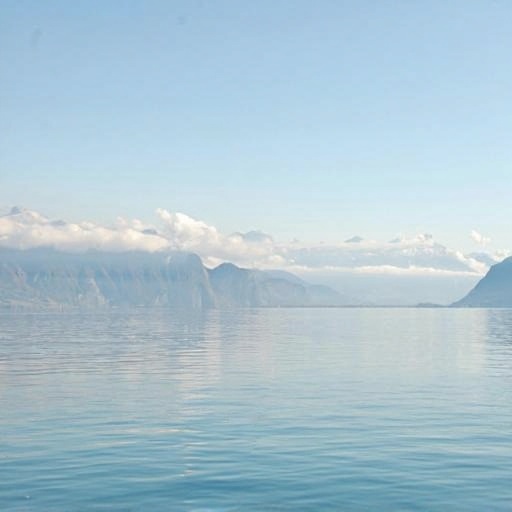} &
        \includegraphics[width=0.16\linewidth]{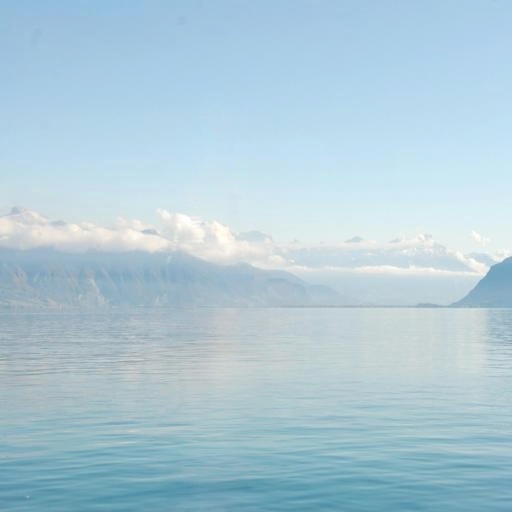} \\[-2pt]

        \includegraphics[width=0.16\linewidth]{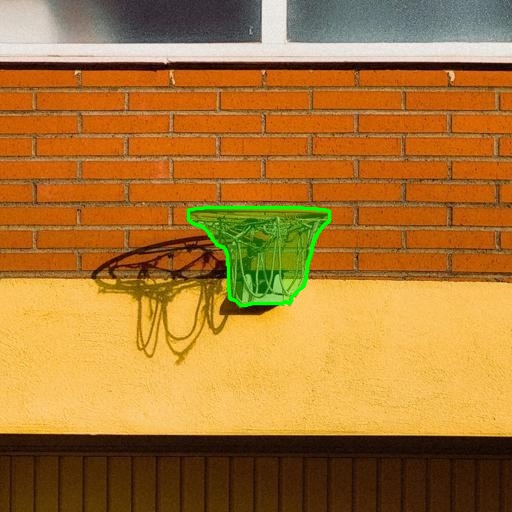} &
        \includegraphics[width=0.16\linewidth]{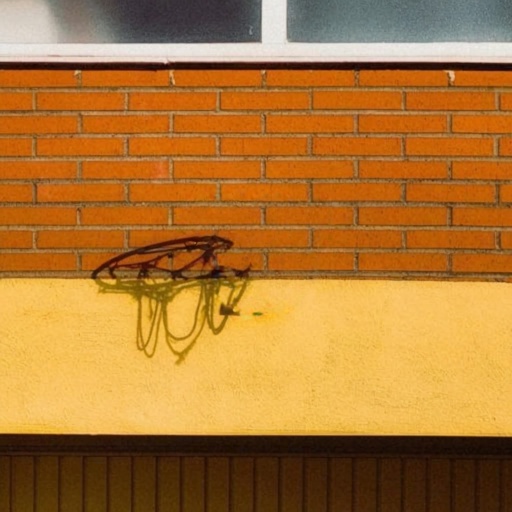} &
        \includegraphics[width=0.16\linewidth]{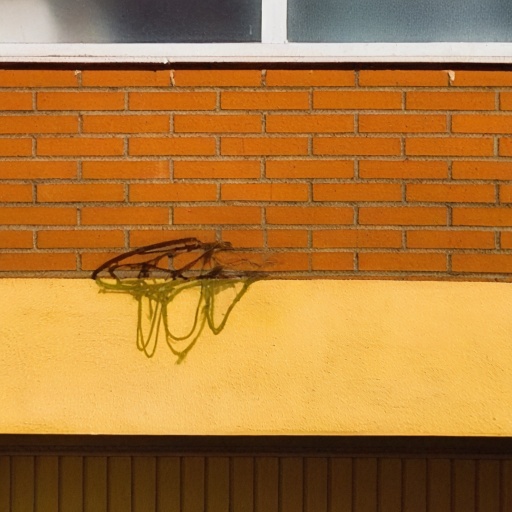} &
        \includegraphics[width=0.16\linewidth]{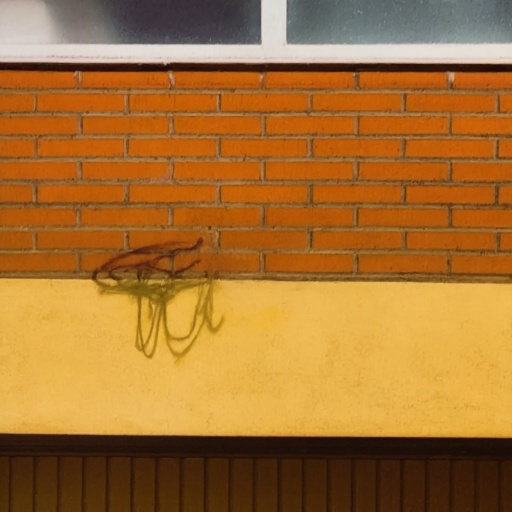} &
        \includegraphics[width=0.16\linewidth]{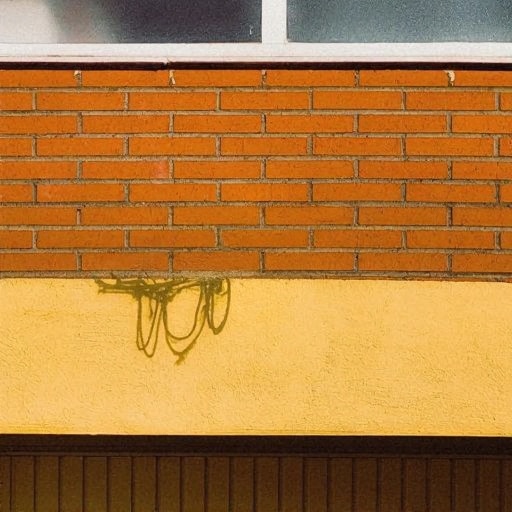} &
        \includegraphics[width=0.16\linewidth]{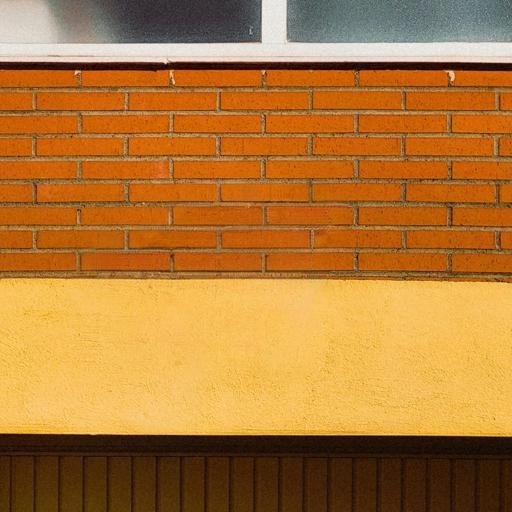} \\[-2pt]

        \footnotesize Input w/ Mask & 
        \footnotesize AttentiveEraser~\cite{sun2025attentive} & 
        \footnotesize RORem~\cite{li2025rorem} & 
        \footnotesize OmniEraser~\cite{wei2025omnieraser} & 
        \footnotesize OmniPaint~\cite{yu2025omnipaint} & 
        \footnotesize FlashClear (ours) \\
        
    \end{tabular}
    
\vspace{-2mm} 
\caption{More visual comparison of ours and other object removal methods on \textit{OBER-Wild} dataset.}
\label{fig:more visual 1}
\vspace{-5mm}
\end{figure}

\begin{figure}[h] 
    \centering

    \setlength{\tabcolsep}{0.5pt}     
    \renewcommand{\arraystretch}{1.2}

    \begin{tabular}{@{}cccccc@{}}

        \includegraphics[width=0.16\linewidth]{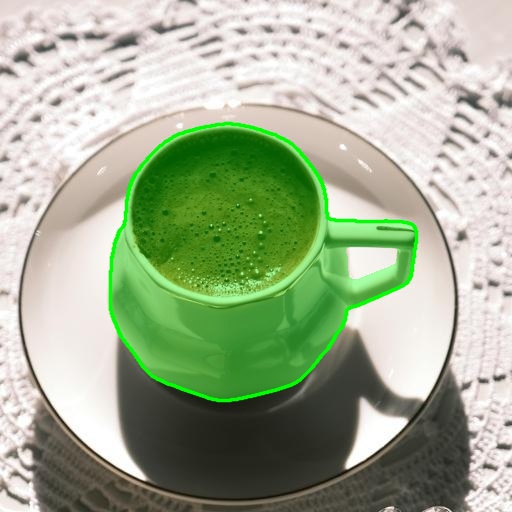} &
        \includegraphics[width=0.16\linewidth]{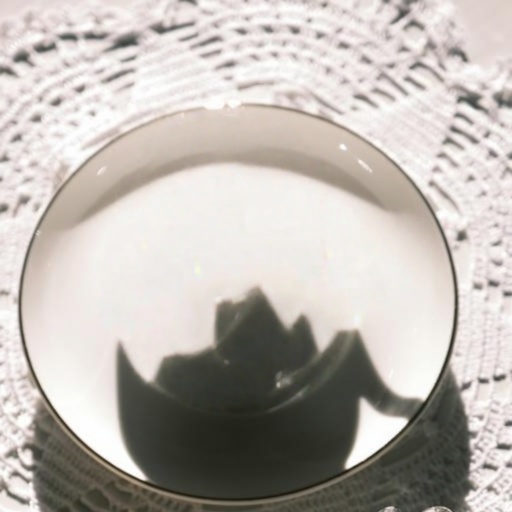} &
        \includegraphics[width=0.16\linewidth]{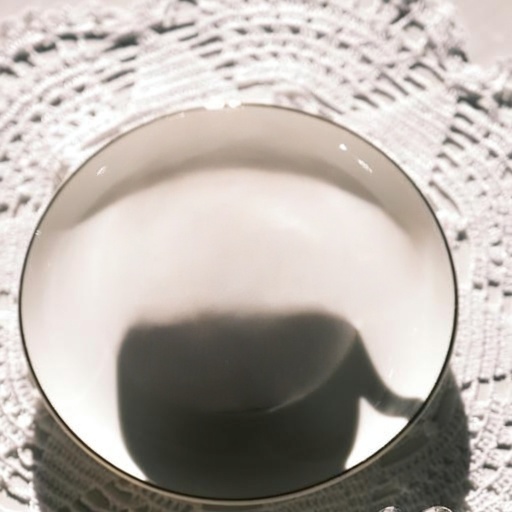} &
        \includegraphics[width=0.16\linewidth]{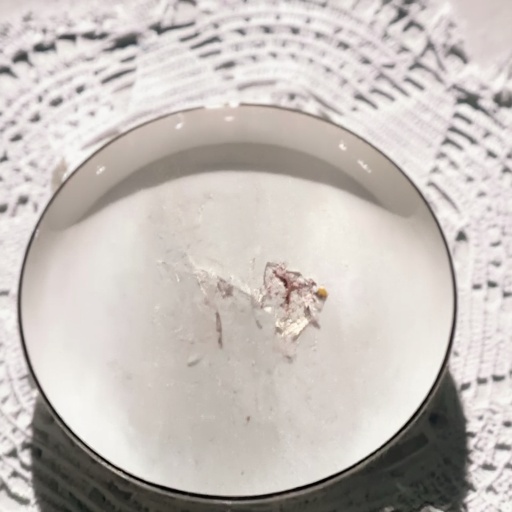} &
        \includegraphics[width=0.16\linewidth]{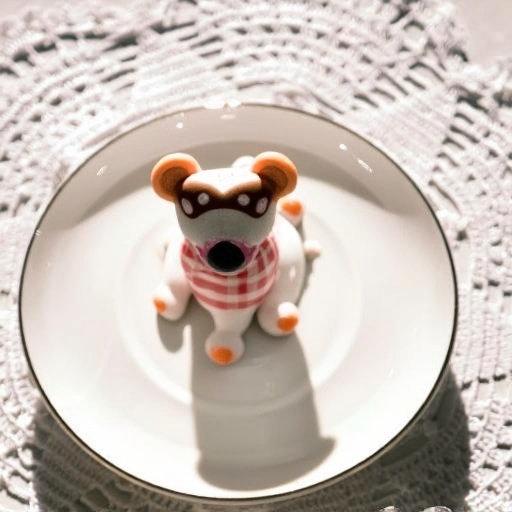} &
        \includegraphics[width=0.16\linewidth]{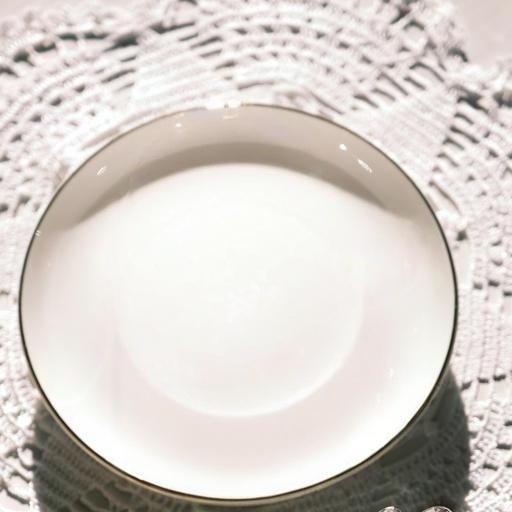} \\[-2pt]

        \includegraphics[width=0.16\linewidth]{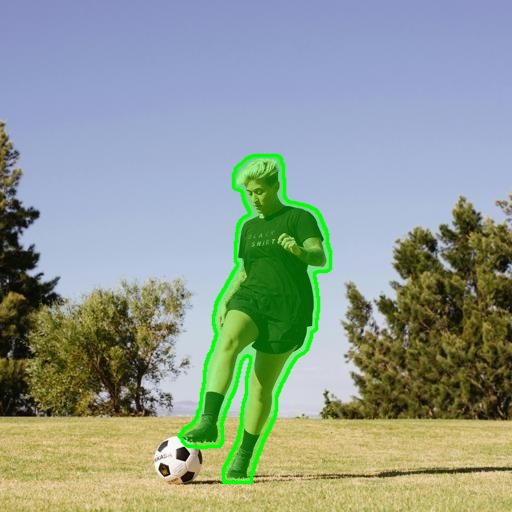} &
        \includegraphics[width=0.16\linewidth]{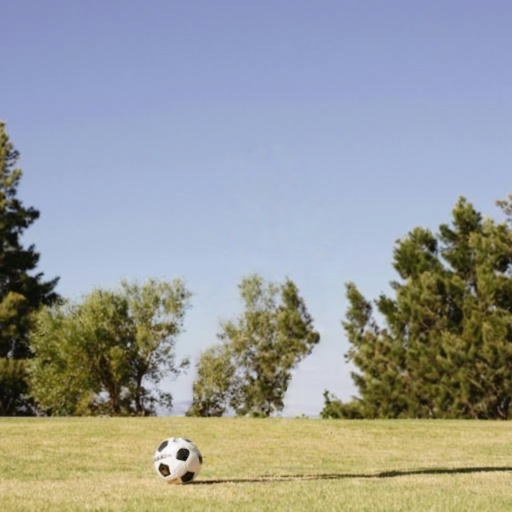} &
        \includegraphics[width=0.16\linewidth]{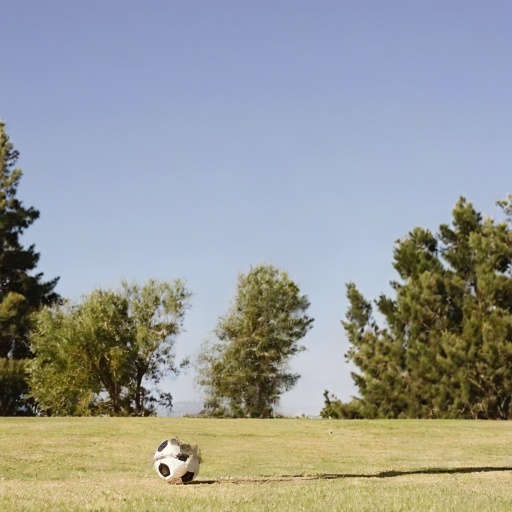} &
        \includegraphics[width=0.16\linewidth]{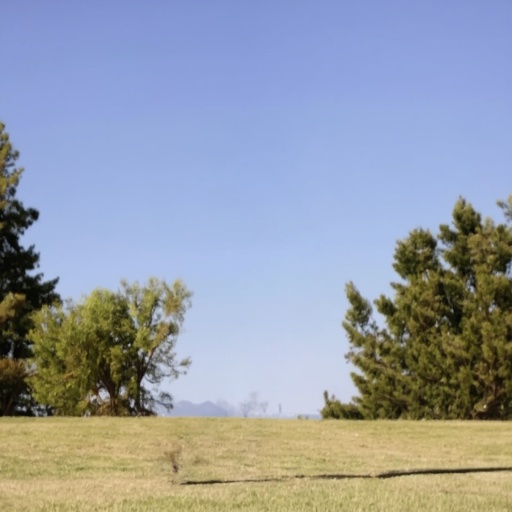} &
        \includegraphics[width=0.16\linewidth]{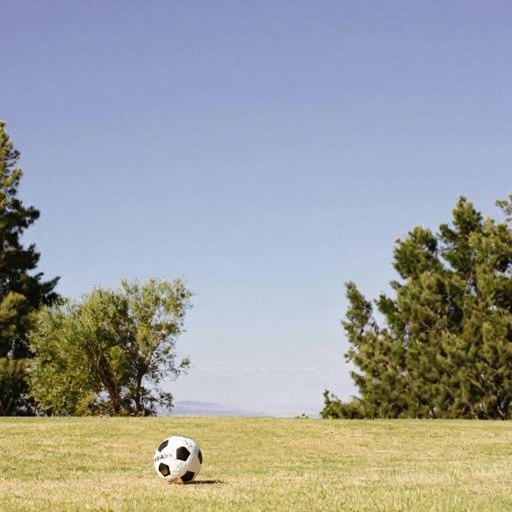} &
        \includegraphics[width=0.16\linewidth]{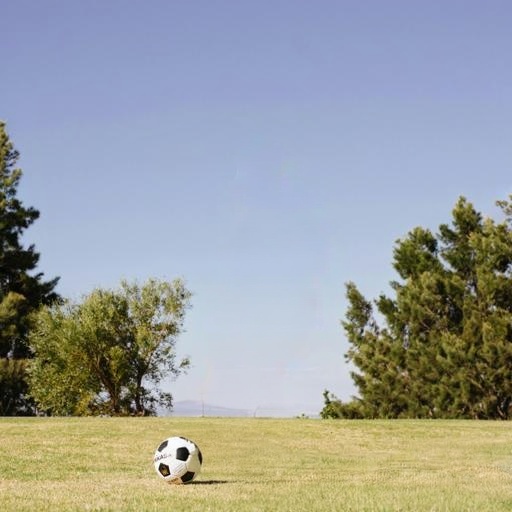} \\[-2pt]

        \includegraphics[width=0.16\linewidth]{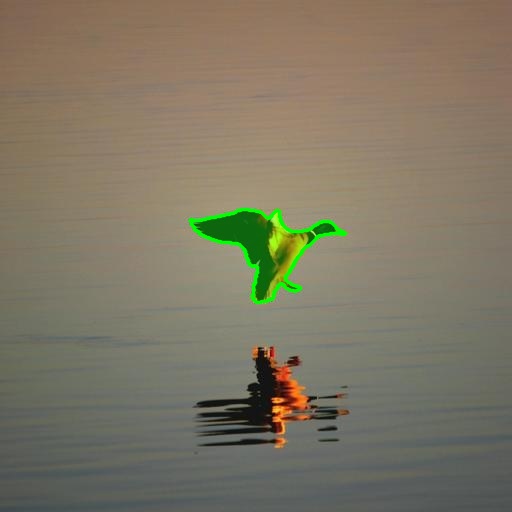} &
        \includegraphics[width=0.16\linewidth]{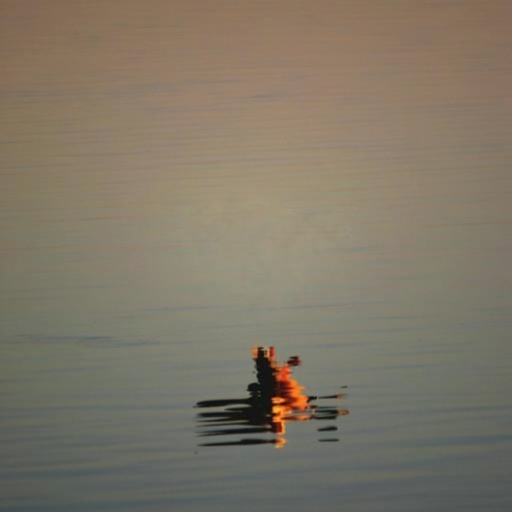} &
        \includegraphics[width=0.16\linewidth]{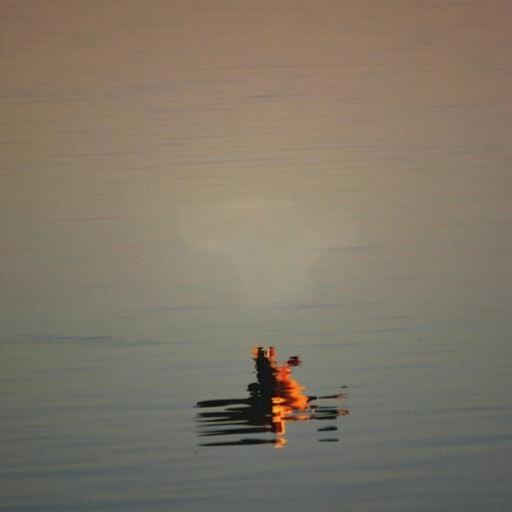} &
        \includegraphics[width=0.16\linewidth]{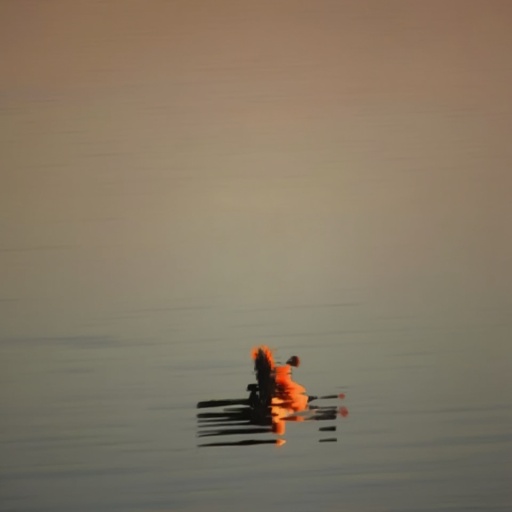} &
        \includegraphics[width=0.16\linewidth]{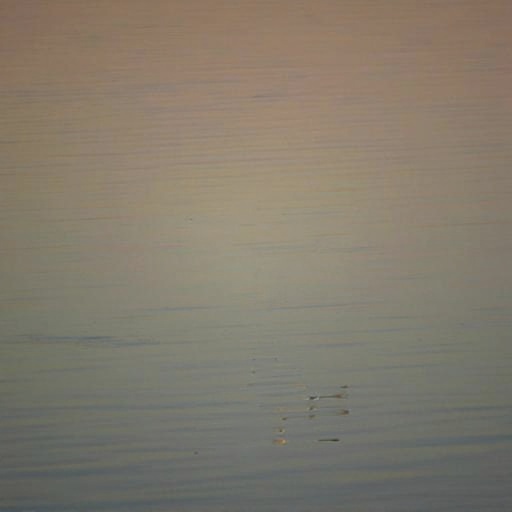} &
        \includegraphics[width=0.16\linewidth]{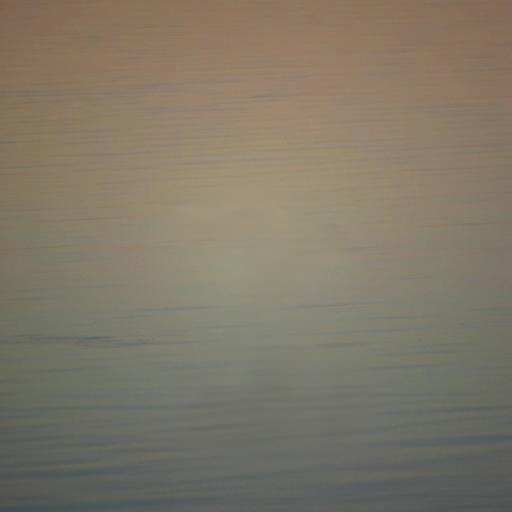} \\[-2pt]

        \includegraphics[width=0.16\linewidth]{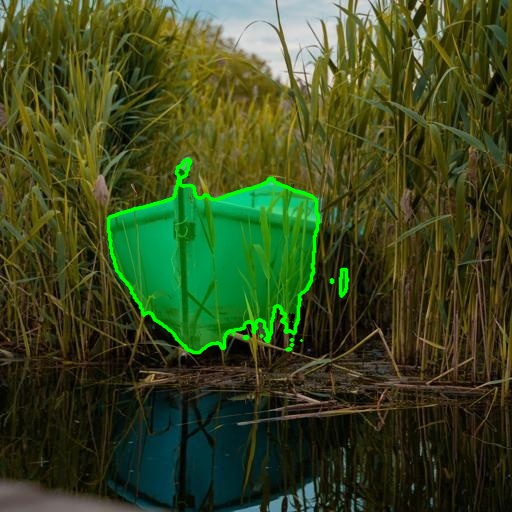} &
        \includegraphics[width=0.16\linewidth]{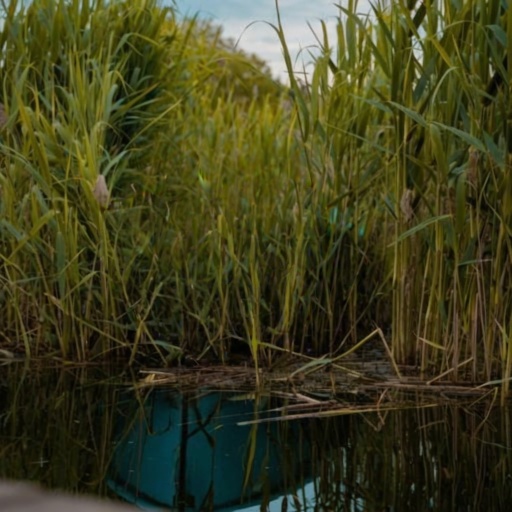} &
        \includegraphics[width=0.16\linewidth]{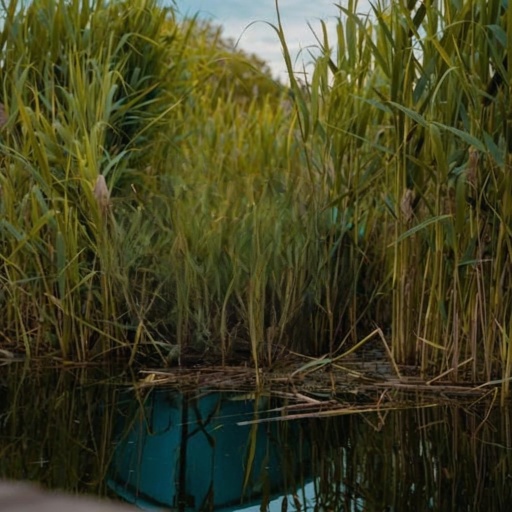} &
        \includegraphics[width=0.16\linewidth]{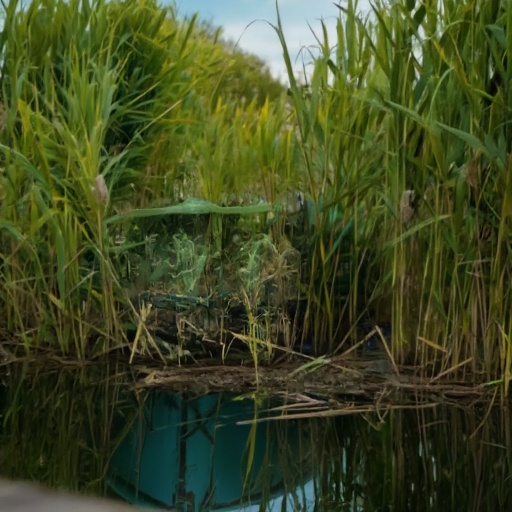} &
        \includegraphics[width=0.16\linewidth]{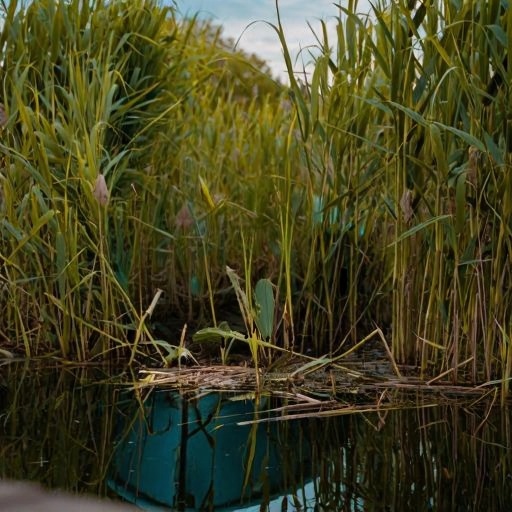} &
        \includegraphics[width=0.16\linewidth]{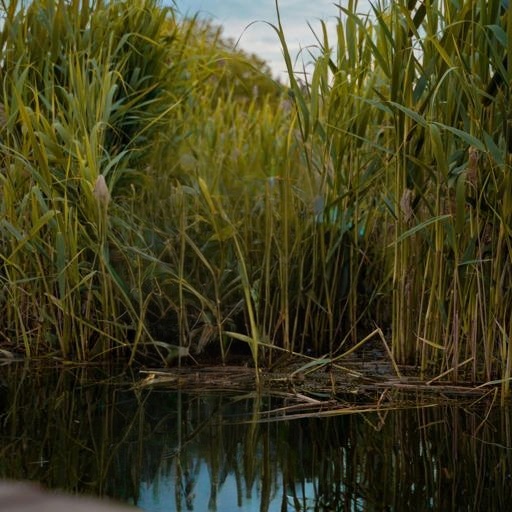} \\[-2pt]

        \includegraphics[width=0.16\linewidth]{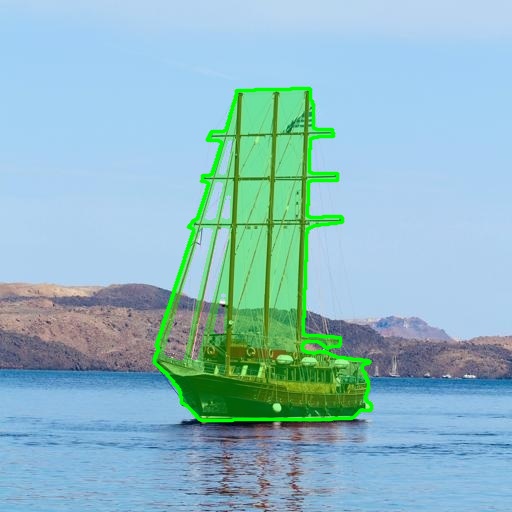} &
        \includegraphics[width=0.16\linewidth]{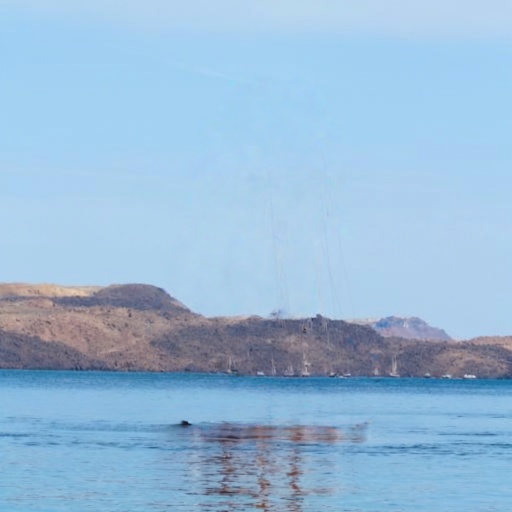} &
        \includegraphics[width=0.16\linewidth]{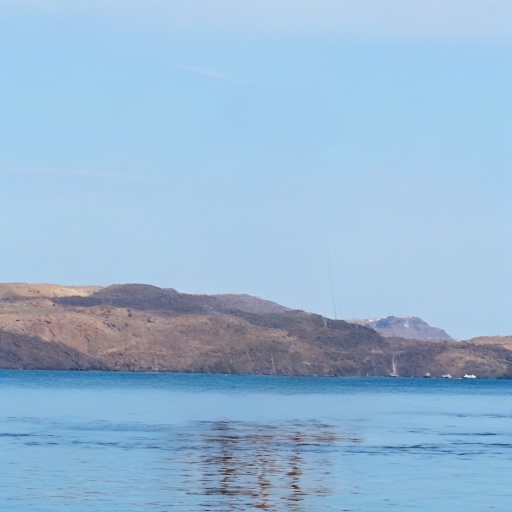} &
        \includegraphics[width=0.16\linewidth]{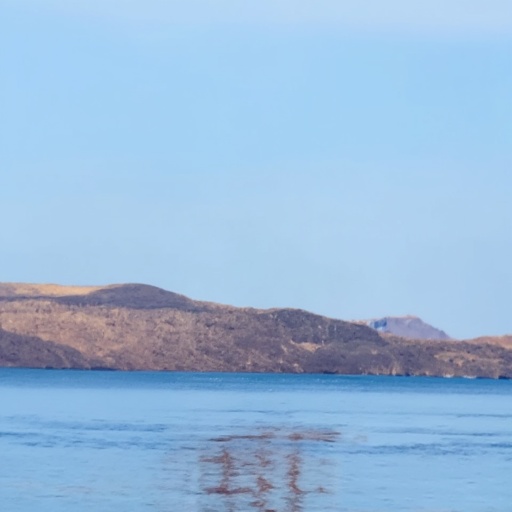} &
        \includegraphics[width=0.16\linewidth]{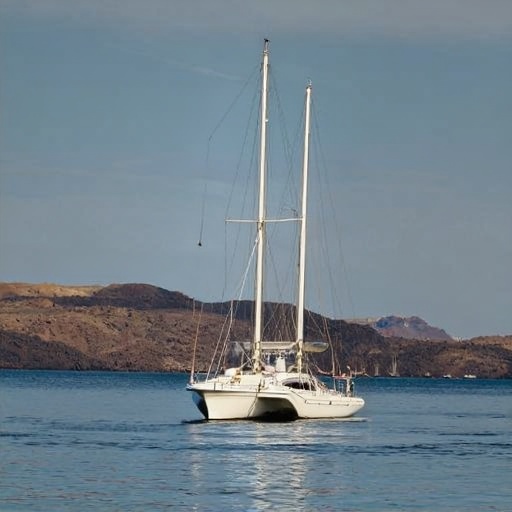} &
        \includegraphics[width=0.16\linewidth]{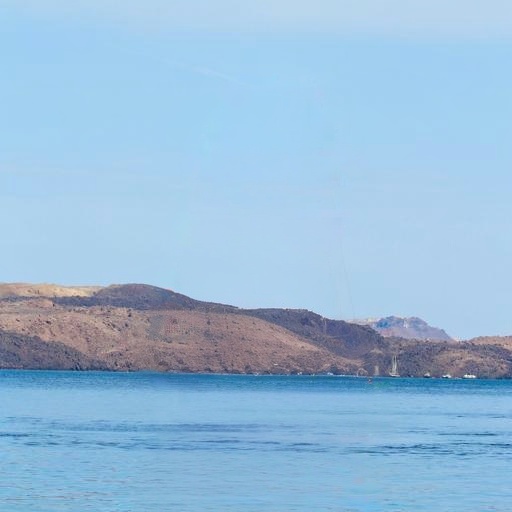} \\[-2pt]

        \includegraphics[width=0.16\linewidth]{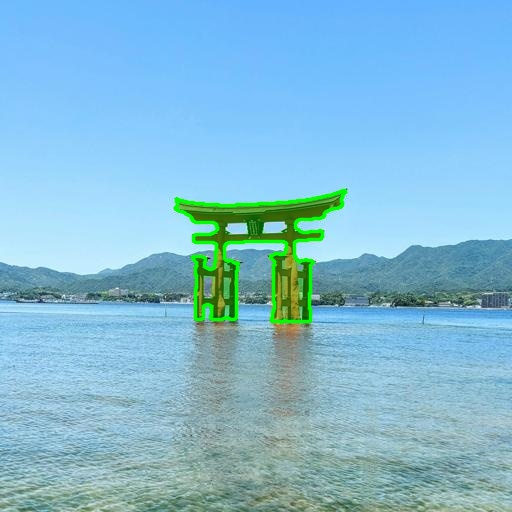} &
        \includegraphics[width=0.16\linewidth]{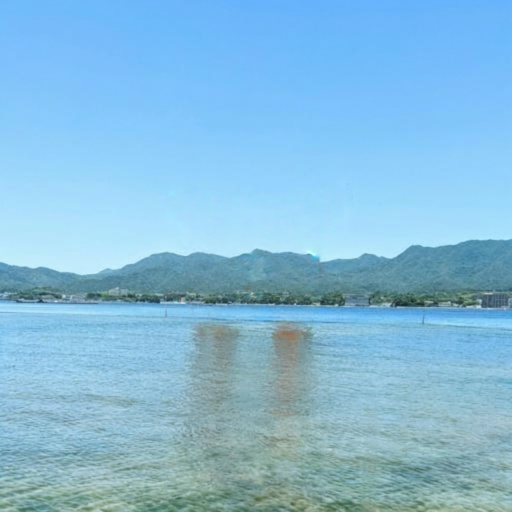} &
        \includegraphics[width=0.16\linewidth]{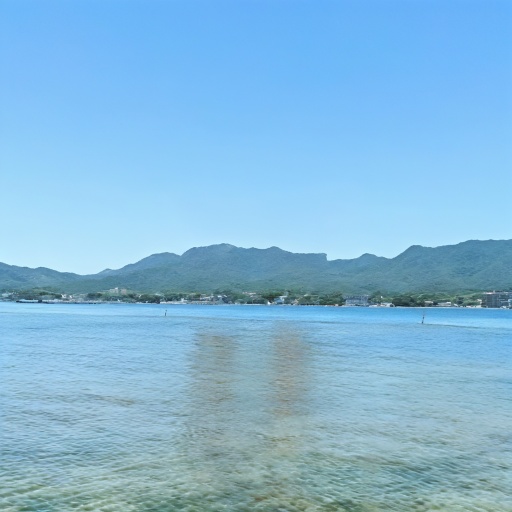} &
        \includegraphics[width=0.16\linewidth]{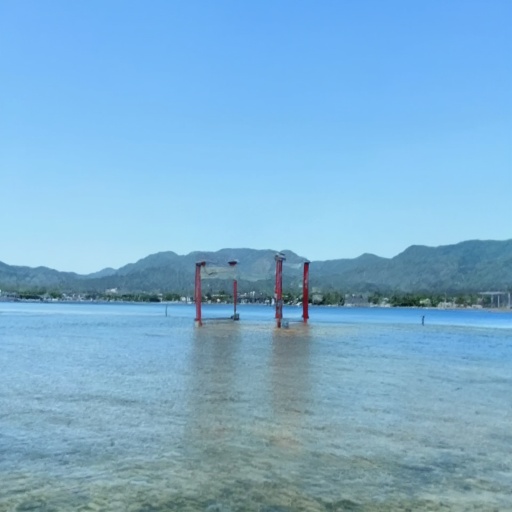} &
        \includegraphics[width=0.16\linewidth]{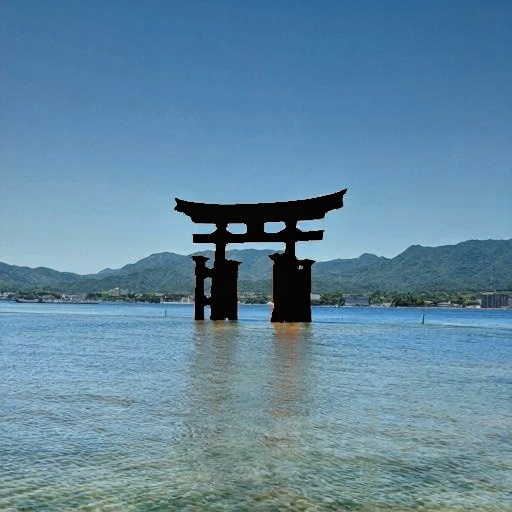} &
        \includegraphics[width=0.16\linewidth]{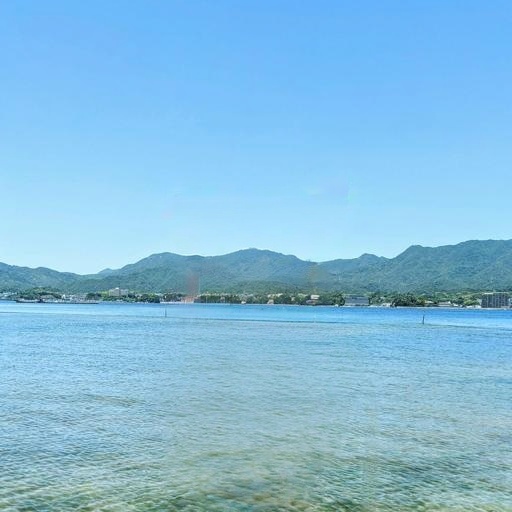} \\[-2pt]

        \includegraphics[width=0.16\linewidth]{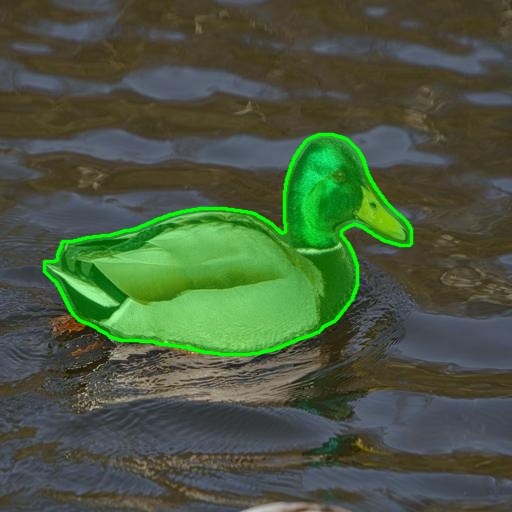} &
        \includegraphics[width=0.16\linewidth]{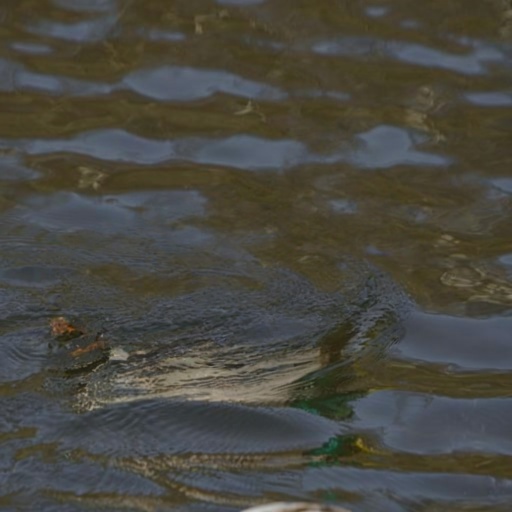} &
        \includegraphics[width=0.16\linewidth]{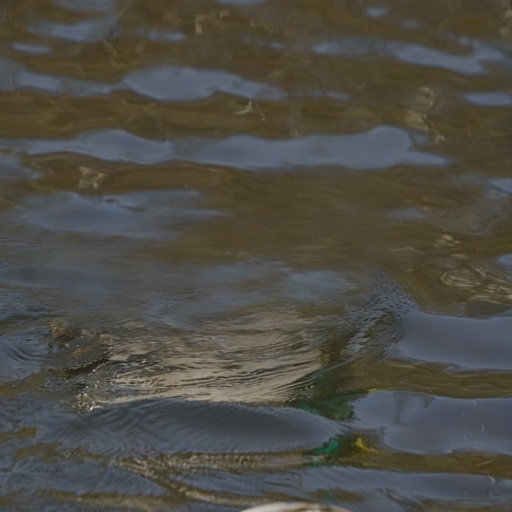} &
        \includegraphics[width=0.16\linewidth]{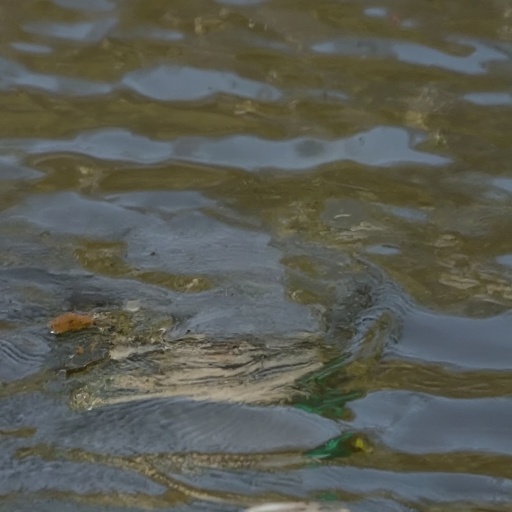} &
        \includegraphics[width=0.16\linewidth]{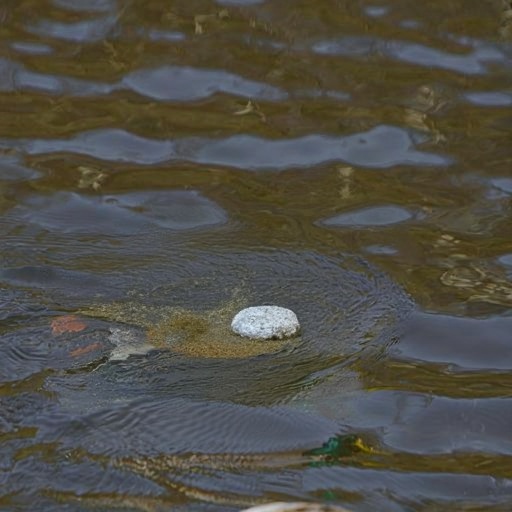} &
        \includegraphics[width=0.16\linewidth]{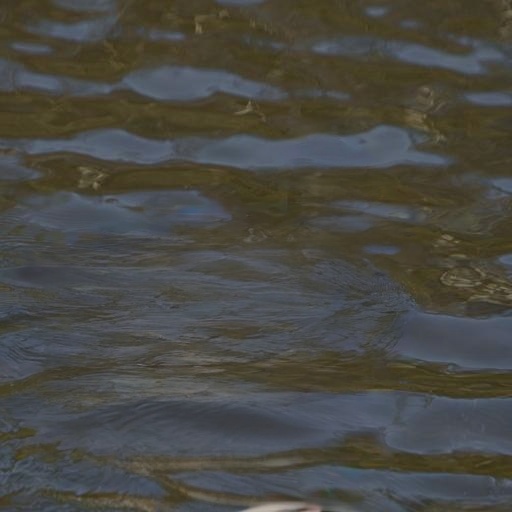} \\[-2pt]

        \includegraphics[width=0.16\linewidth]{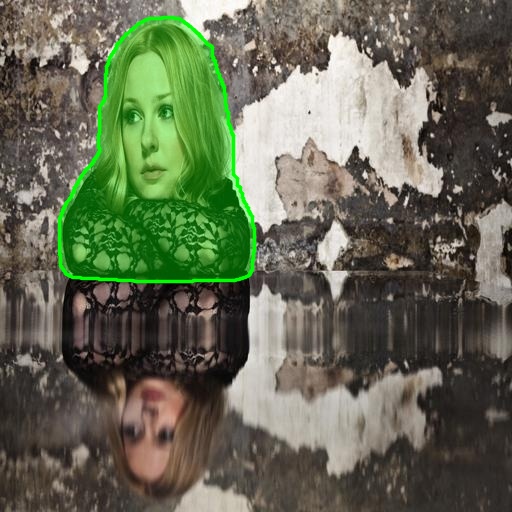} &
        \includegraphics[width=0.16\linewidth]{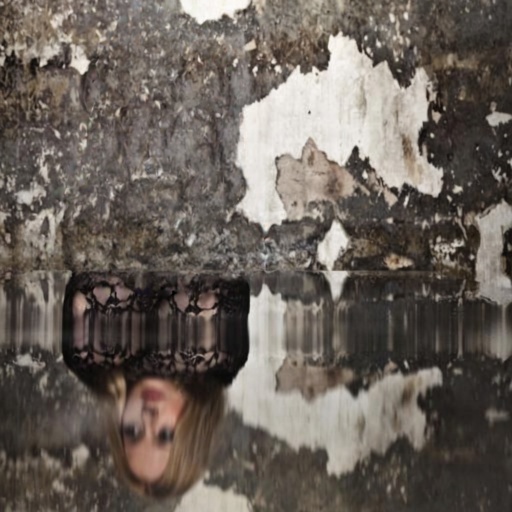} &
        \includegraphics[width=0.16\linewidth]{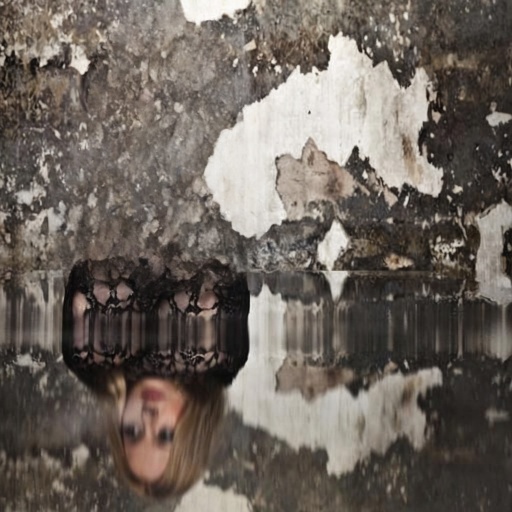} &
        \includegraphics[width=0.16\linewidth]{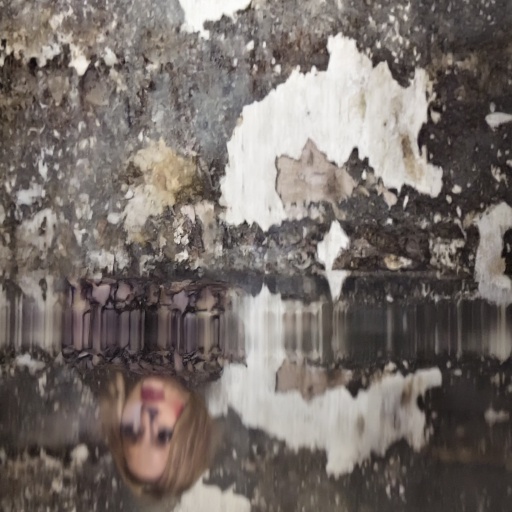} &
        \includegraphics[width=0.16\linewidth]{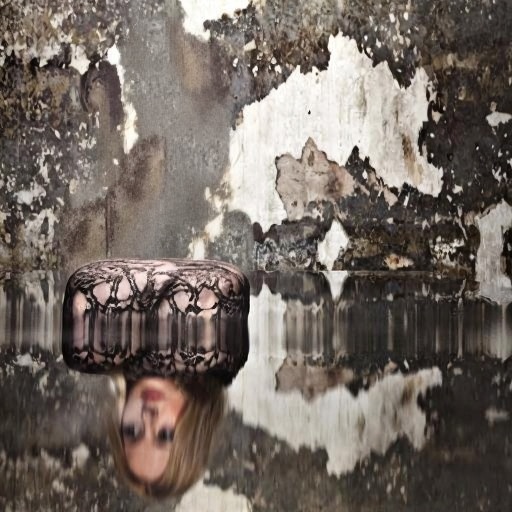} &
        \includegraphics[width=0.16\linewidth]{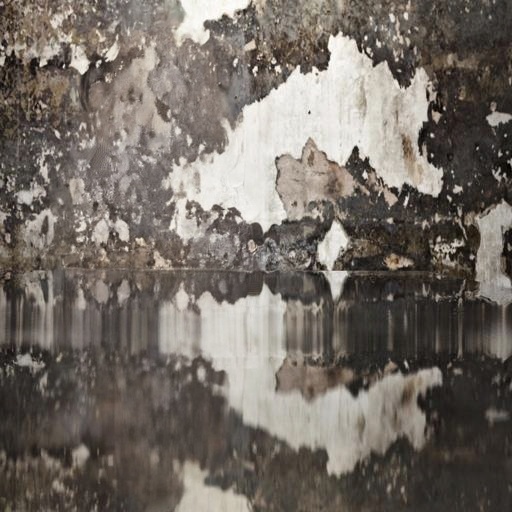} \\[-2pt]

        \includegraphics[width=0.16\linewidth]{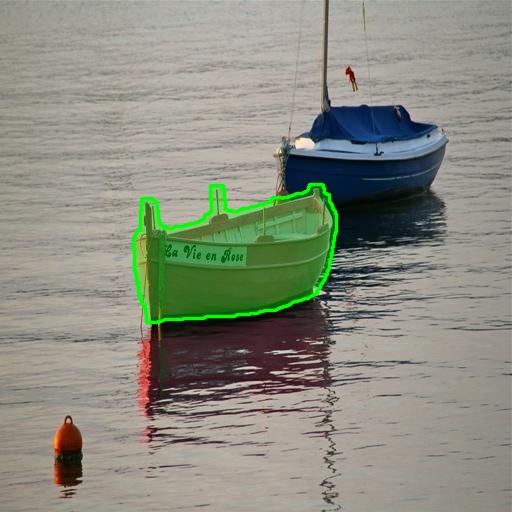} &
        \includegraphics[width=0.16\linewidth]{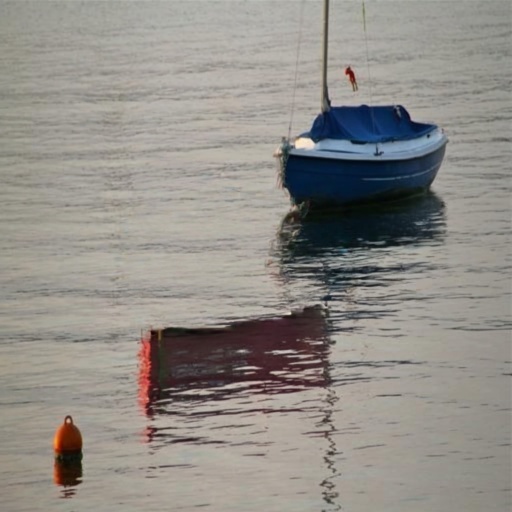} &
        \includegraphics[width=0.16\linewidth]{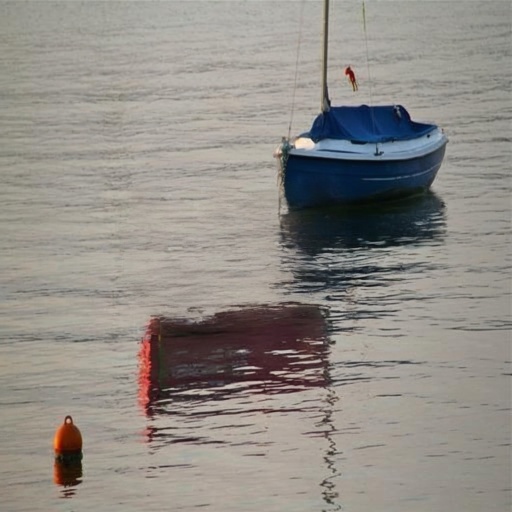} &
        \includegraphics[width=0.16\linewidth]{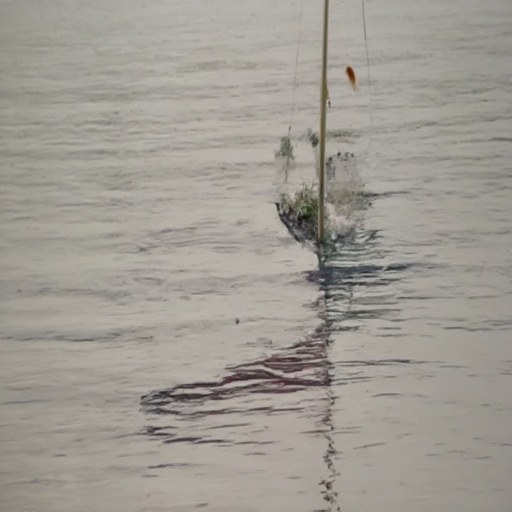} &
        \includegraphics[width=0.16\linewidth]{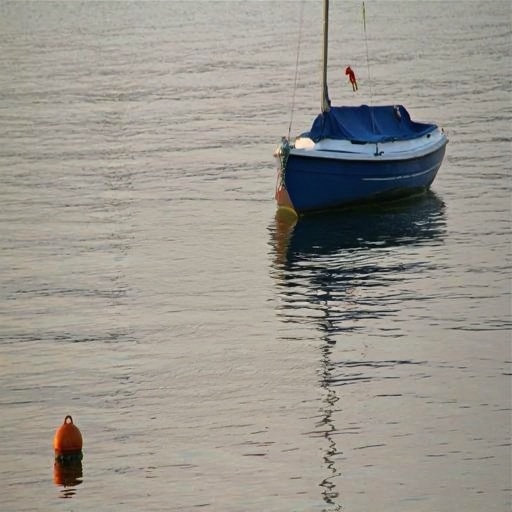} &
        \includegraphics[width=0.16\linewidth]{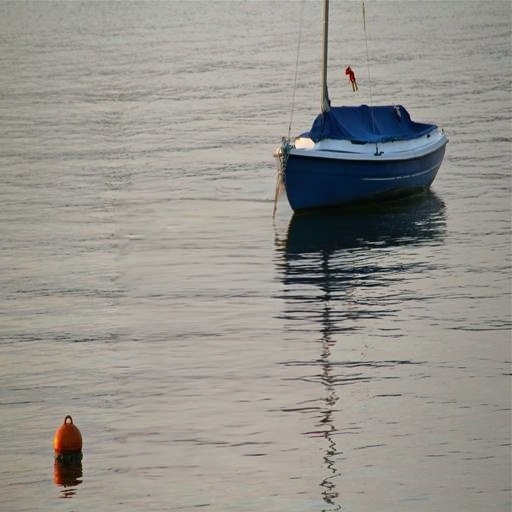} \\[-2pt]

        \footnotesize Input w/ Mask & 
        \footnotesize AttentiveEraser~\cite{sun2025attentive} & 
        \footnotesize RORem~\cite{li2025rorem} & 
        \footnotesize OmniEraser~\cite{wei2025omnieraser} & 
        \footnotesize OmniPaint~\cite{yu2025omnipaint} & 
        \footnotesize FlashClear (ours) \\
        
    \end{tabular}
    
\vspace{-2mm} 
\caption{More visual comparison of ours and other object removal methods on \textit{OBER-Wild} dataset.}
\label{fig:more visual 2}
\vspace{-5mm}
\end{figure}

\begin{figure}[t] 
    \centering

    \setlength{\tabcolsep}{0.5pt}      
    \renewcommand{\arraystretch}{1.2}

    \begin{tabular}{@{}ccccc@{}}

        \includegraphics[width=0.19\linewidth]{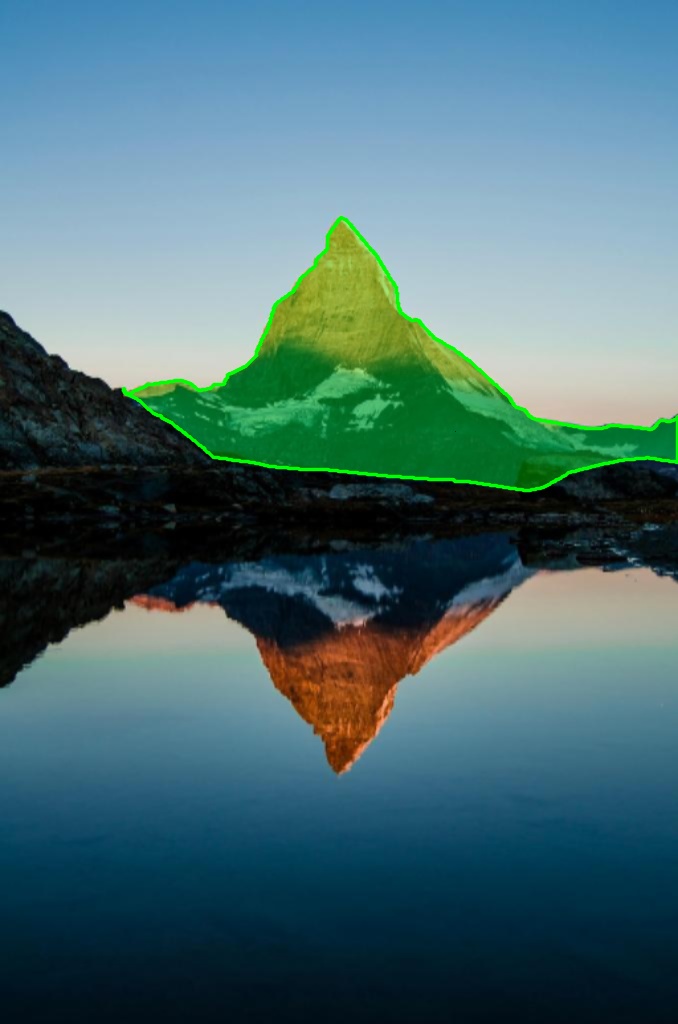} &
        \includegraphics[width=0.19\linewidth]{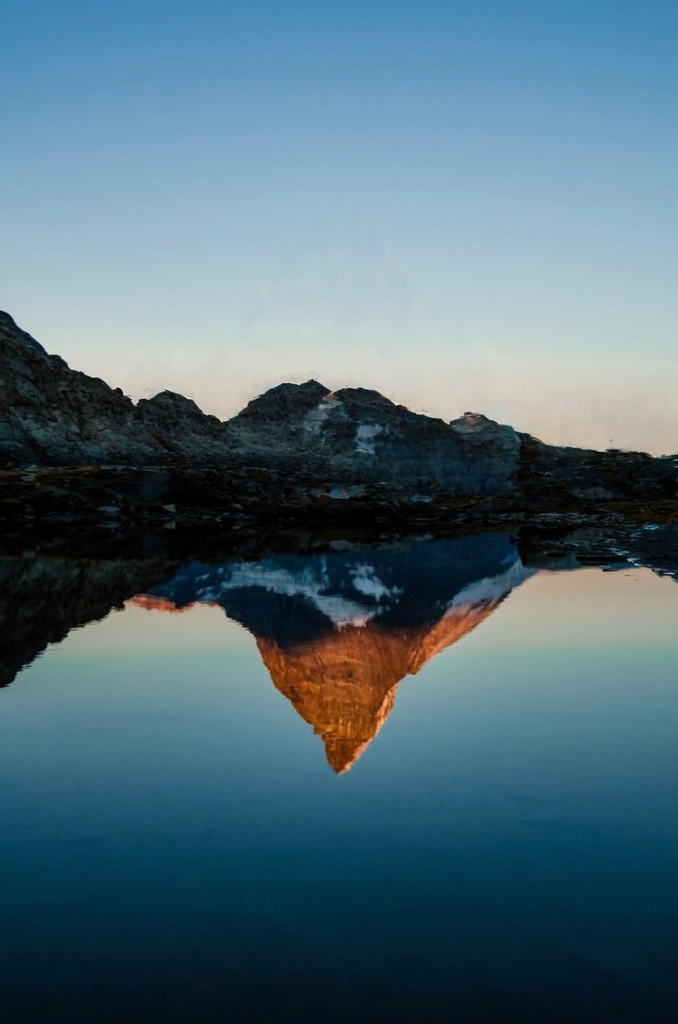} &
        \includegraphics[width=0.19\linewidth]{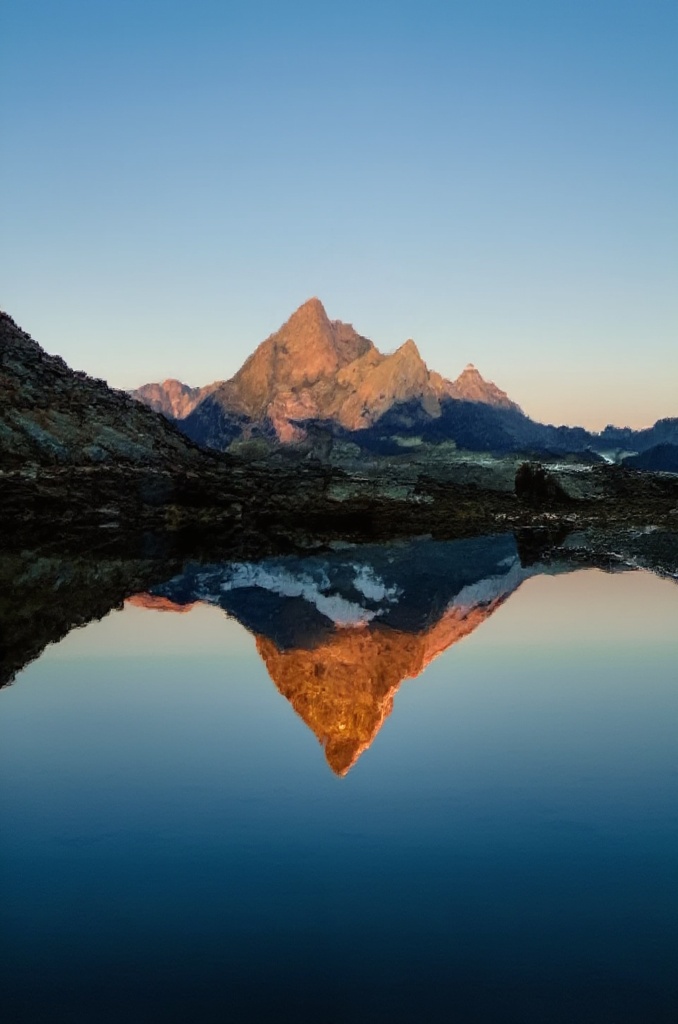} &
        \includegraphics[width=0.19\linewidth]{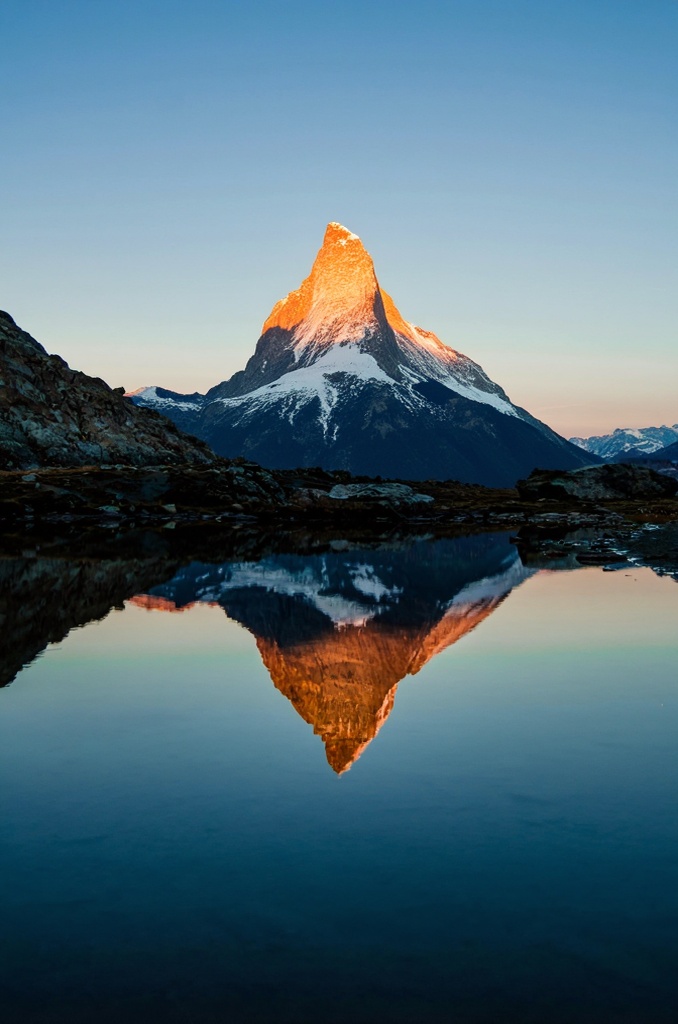} &
        \includegraphics[width=0.19\linewidth]{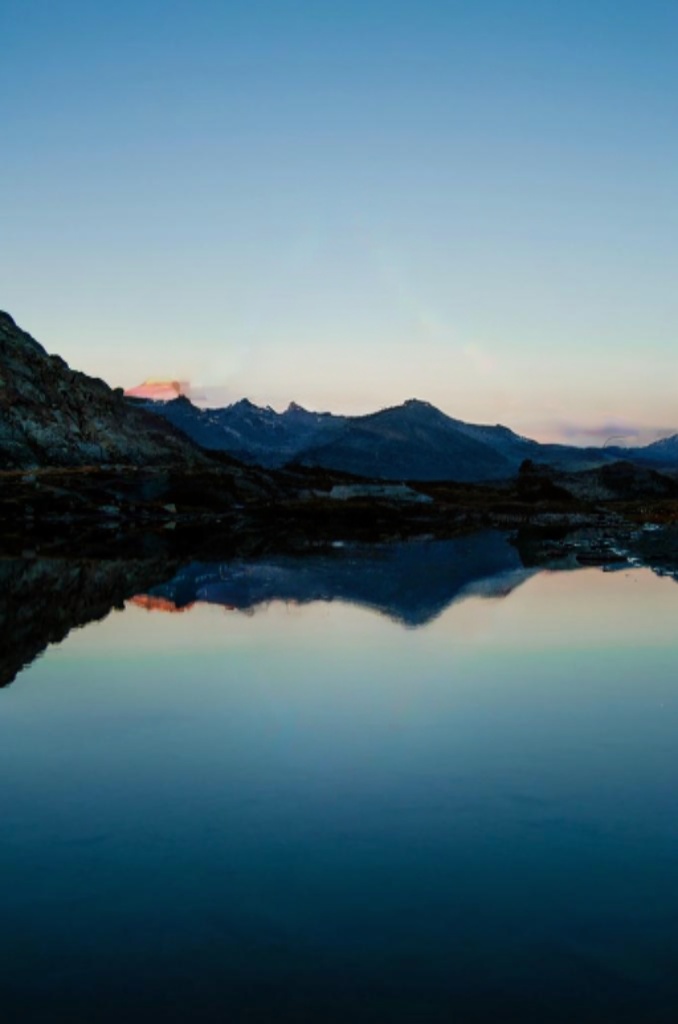} \\[-2pt]
        
       \includegraphics[width=0.19\linewidth]{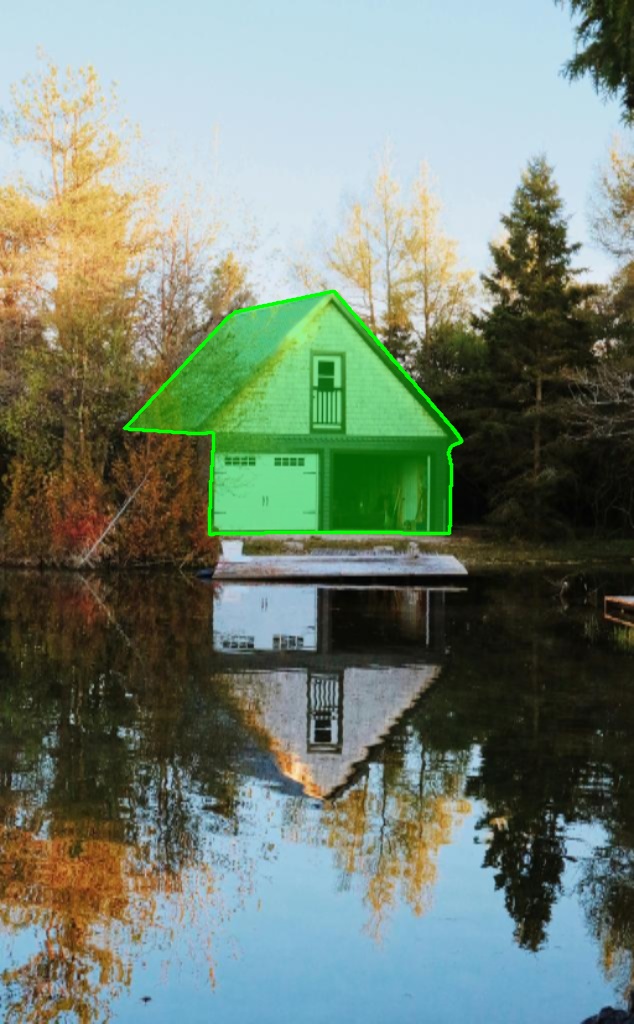} &
        \includegraphics[width=0.19\linewidth]{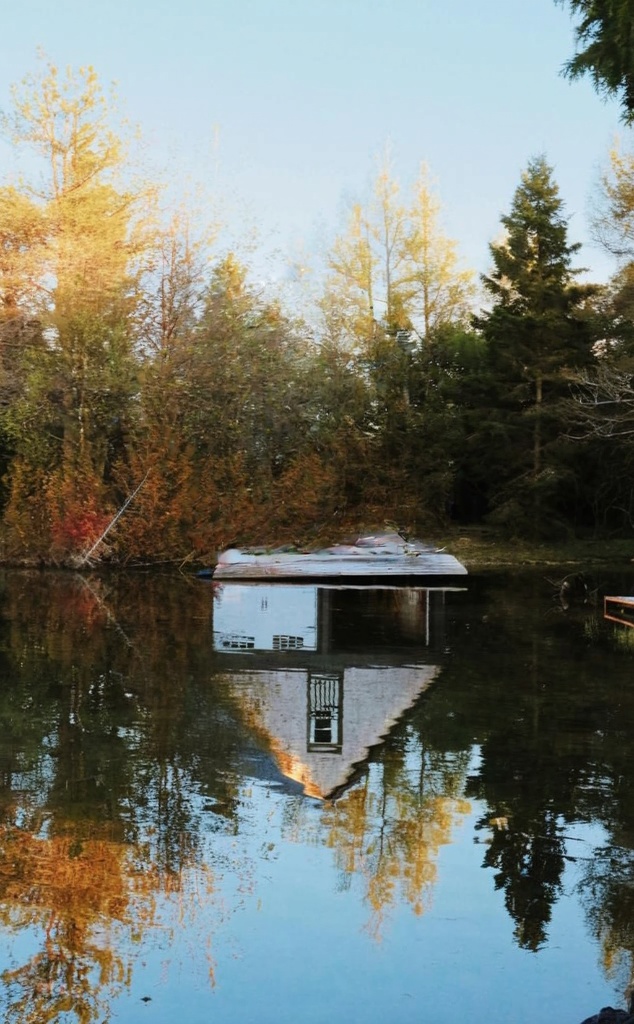} &
        \includegraphics[width=0.19\linewidth]{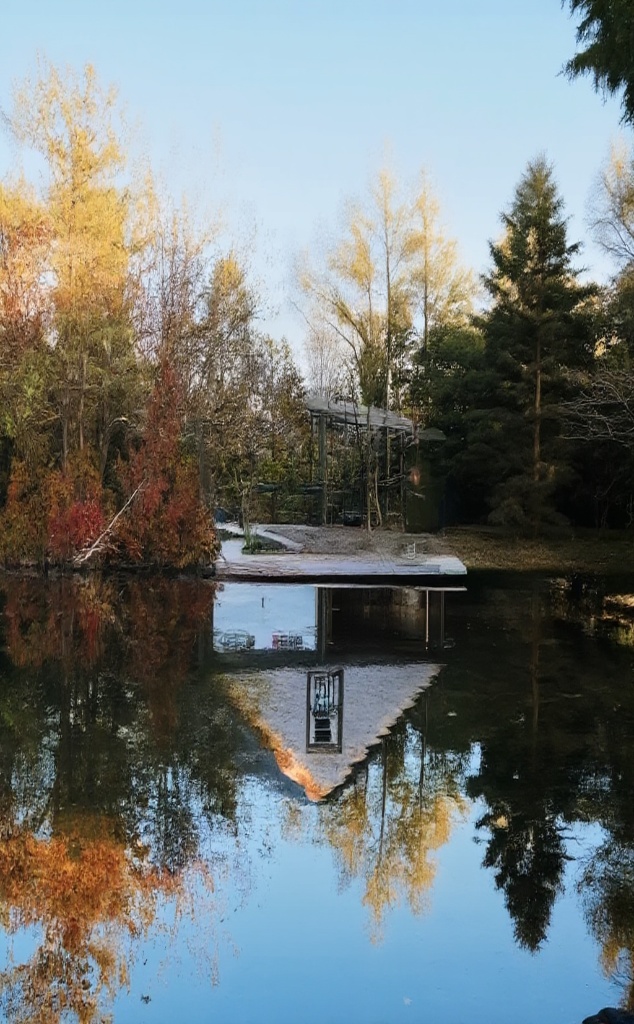} &
        \includegraphics[width=0.19\linewidth]{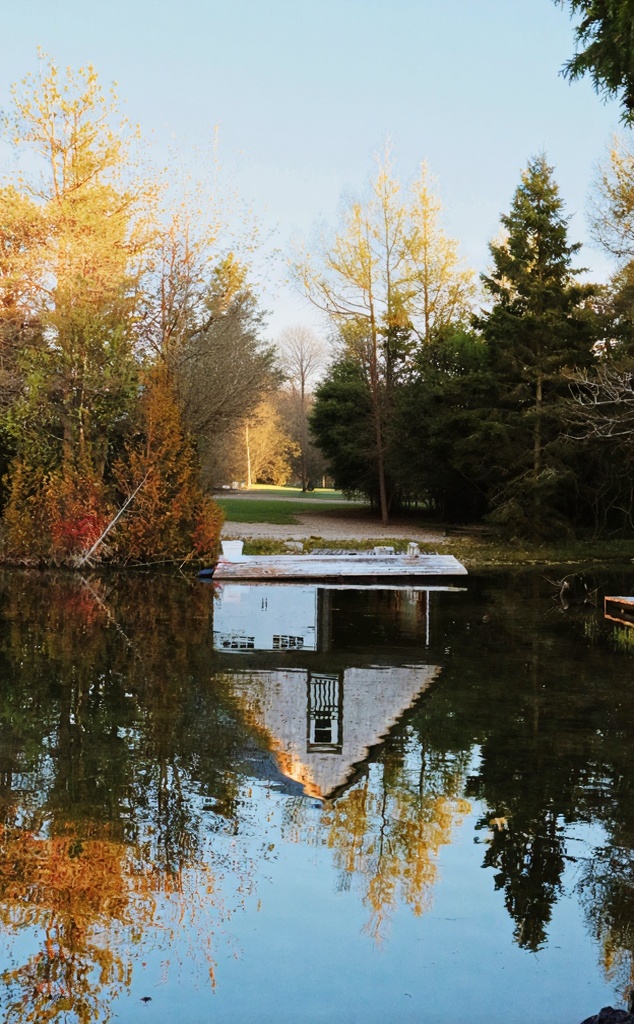} &
        \includegraphics[width=0.19\linewidth]{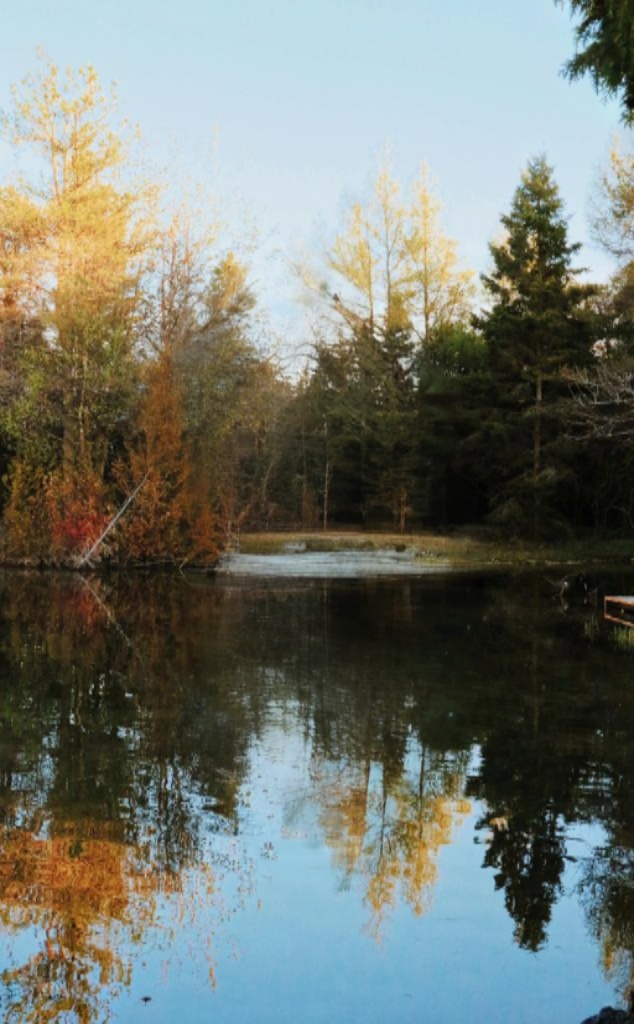} \\[-2pt]

        \includegraphics[width=0.19\linewidth]{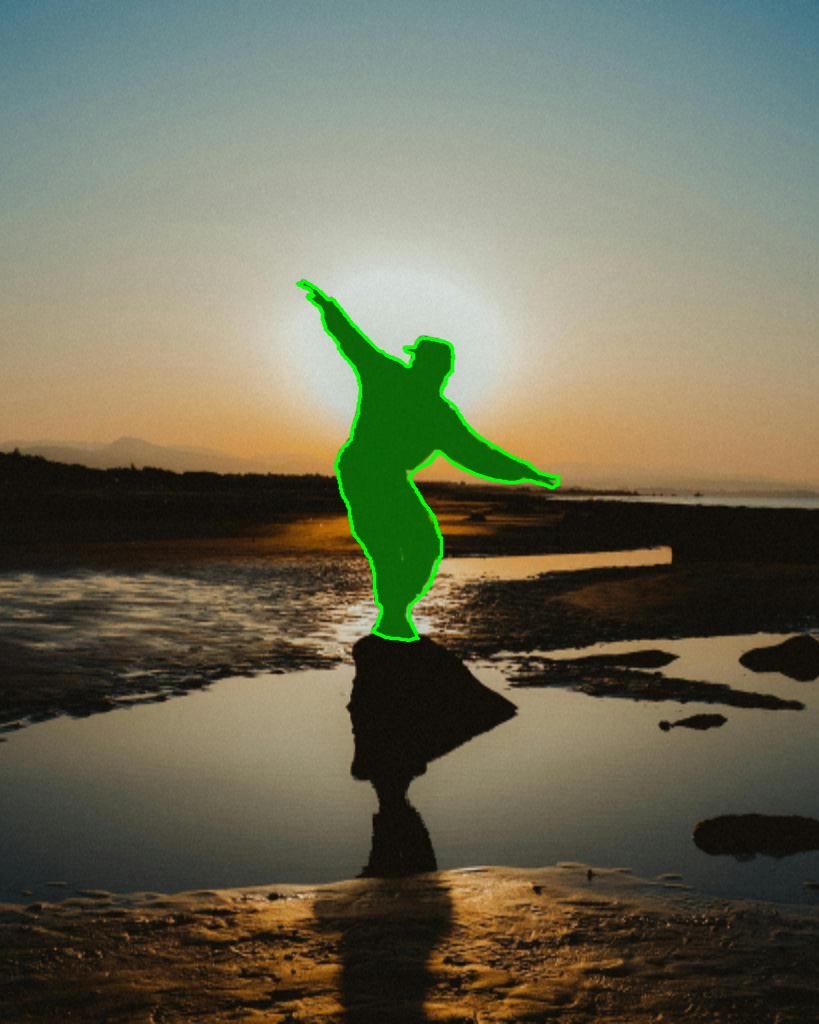} &
        \includegraphics[width=0.19\linewidth]{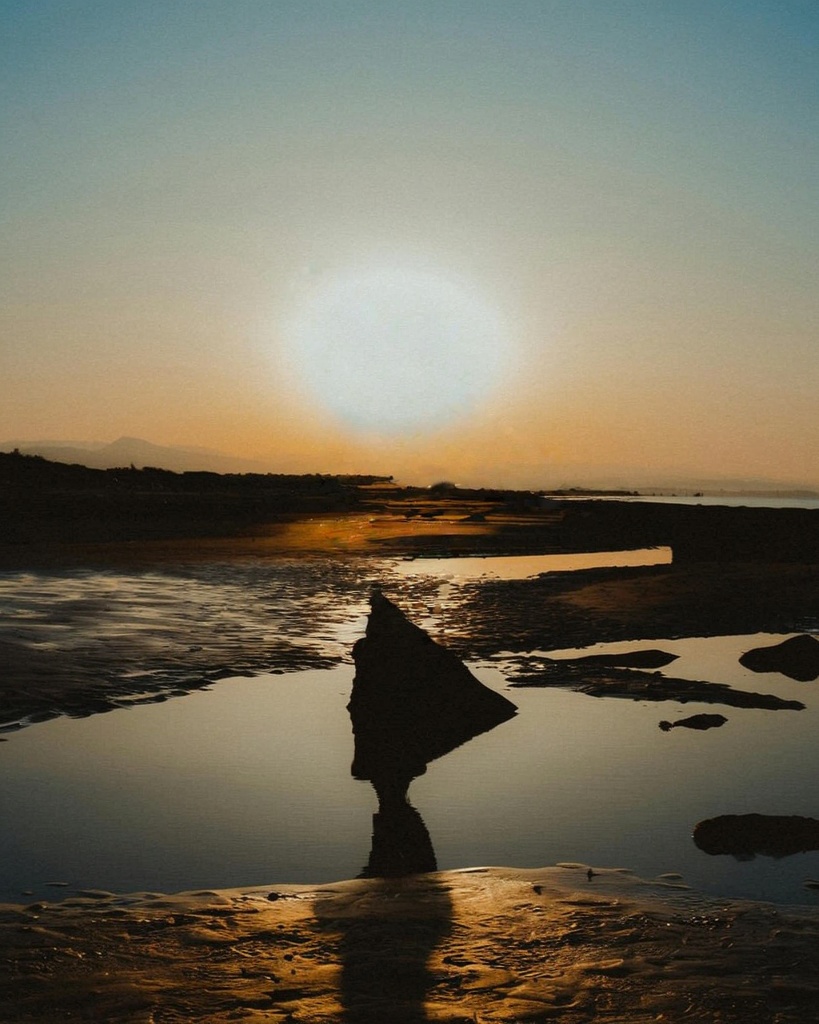} &
        \includegraphics[width=0.19\linewidth]{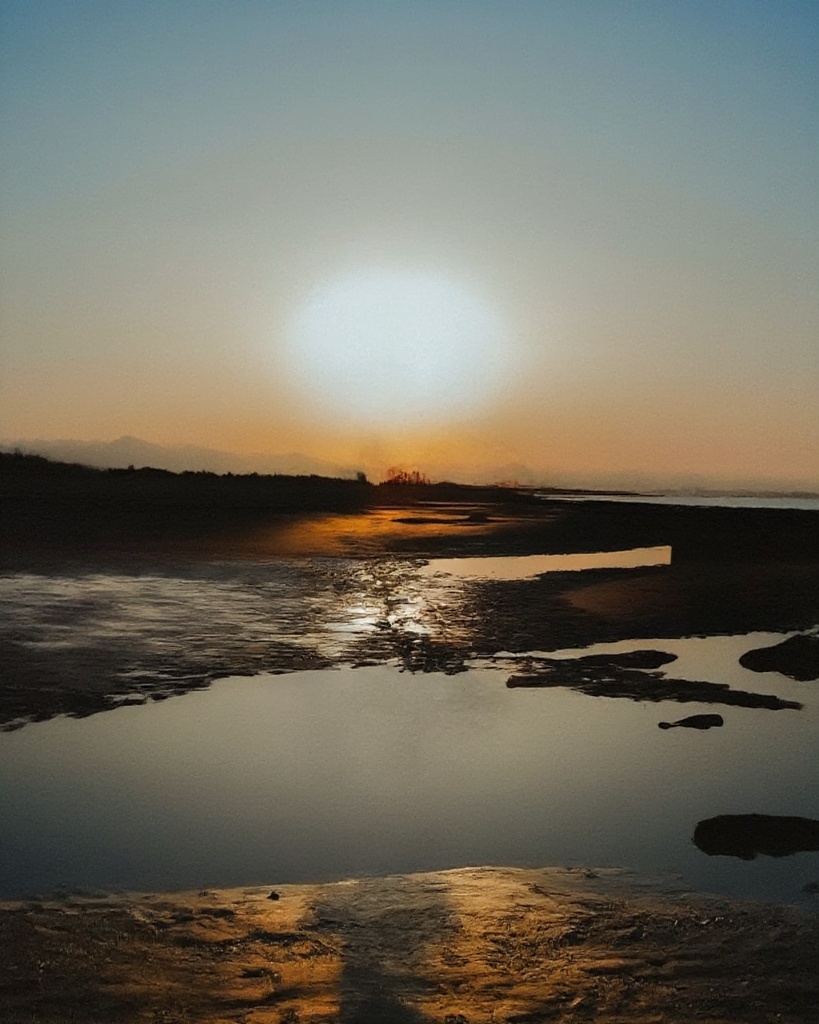} &
        \includegraphics[width=0.19\linewidth]{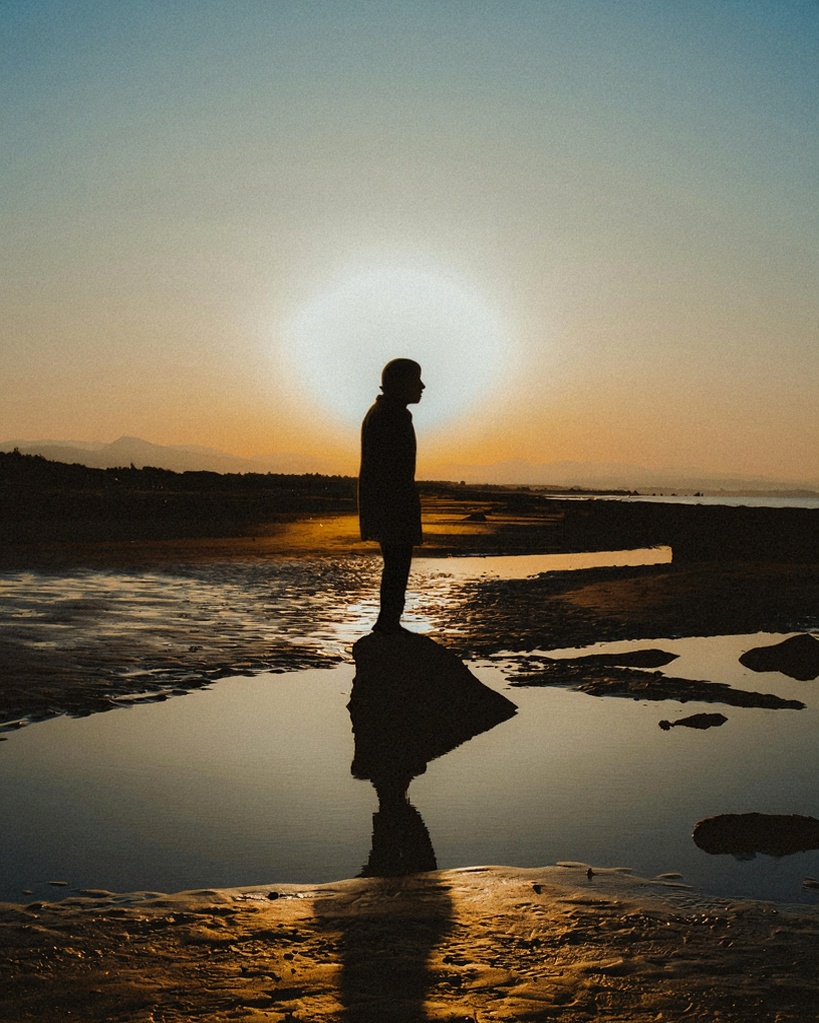} &
        \includegraphics[width=0.19\linewidth]{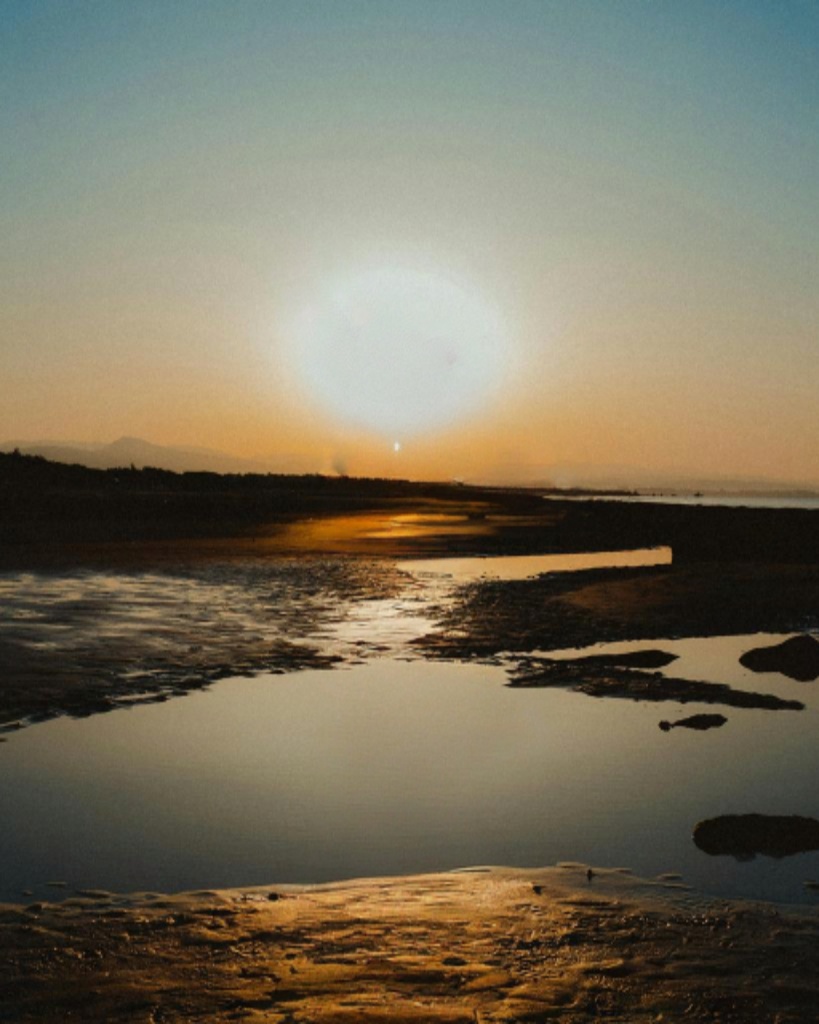} \\[-2pt]
        
        \includegraphics[width=0.19\linewidth]{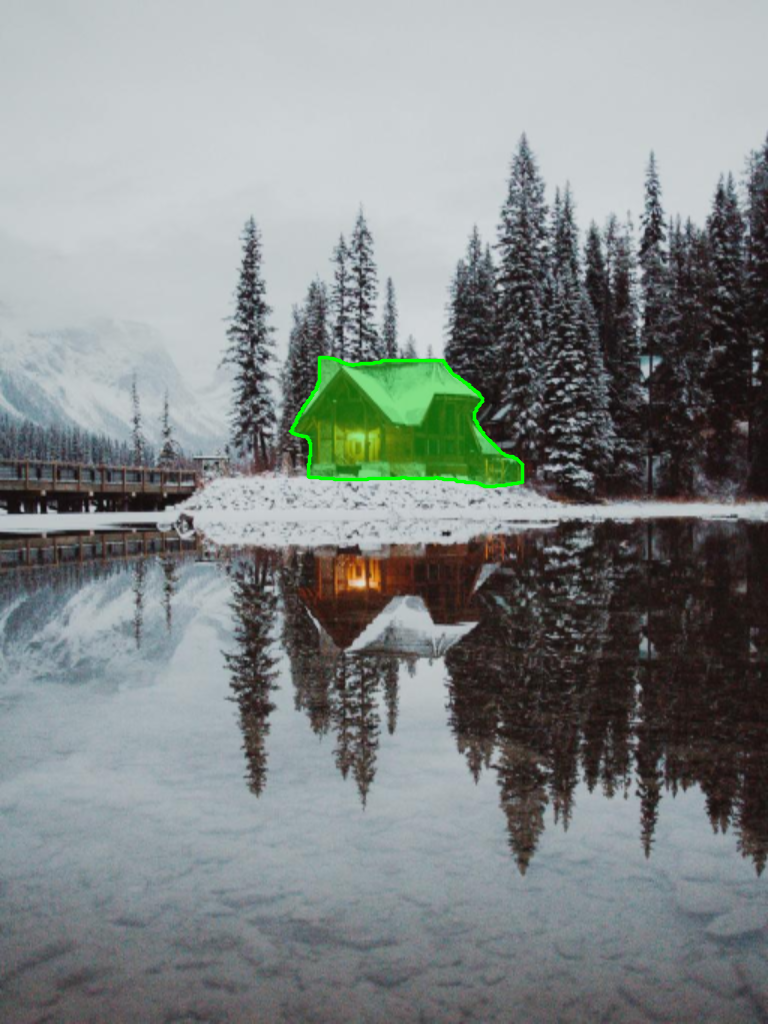} &
        \includegraphics[width=0.19\linewidth]{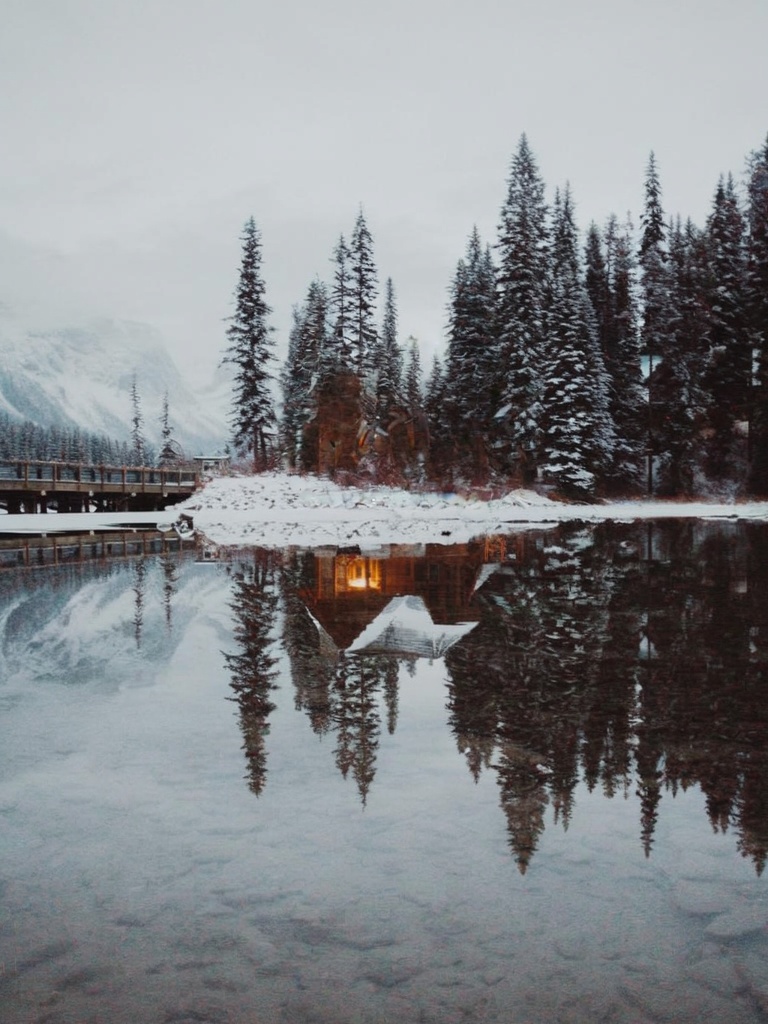} &
        \includegraphics[width=0.19\linewidth]{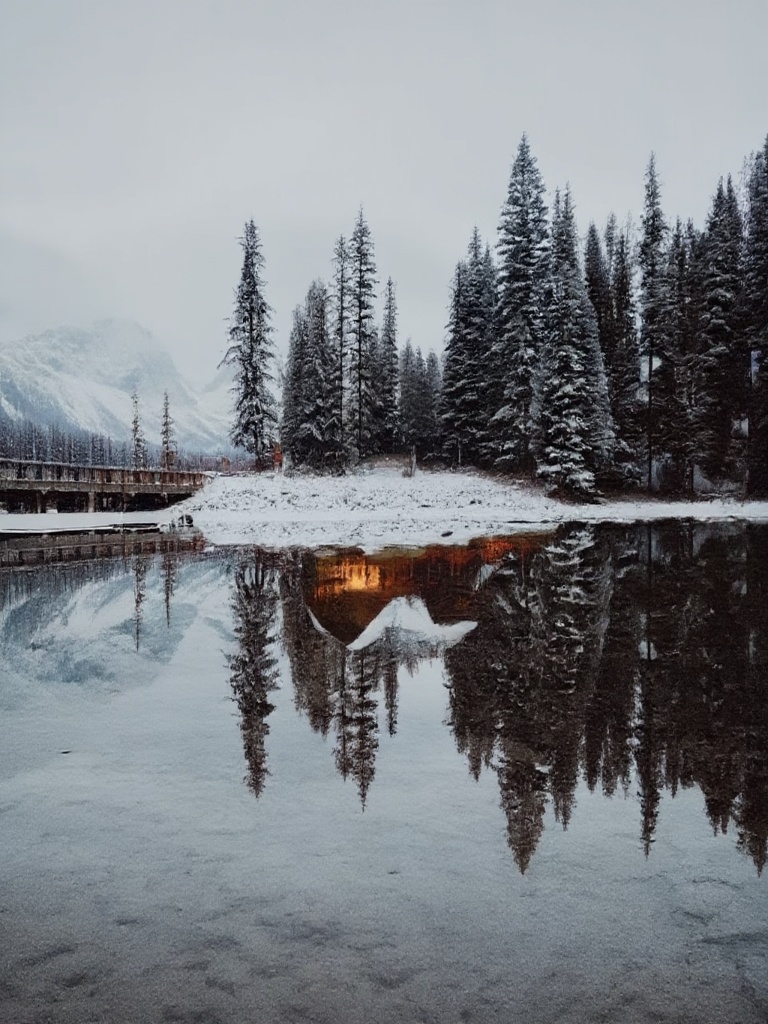} &
        \includegraphics[width=0.19\linewidth]{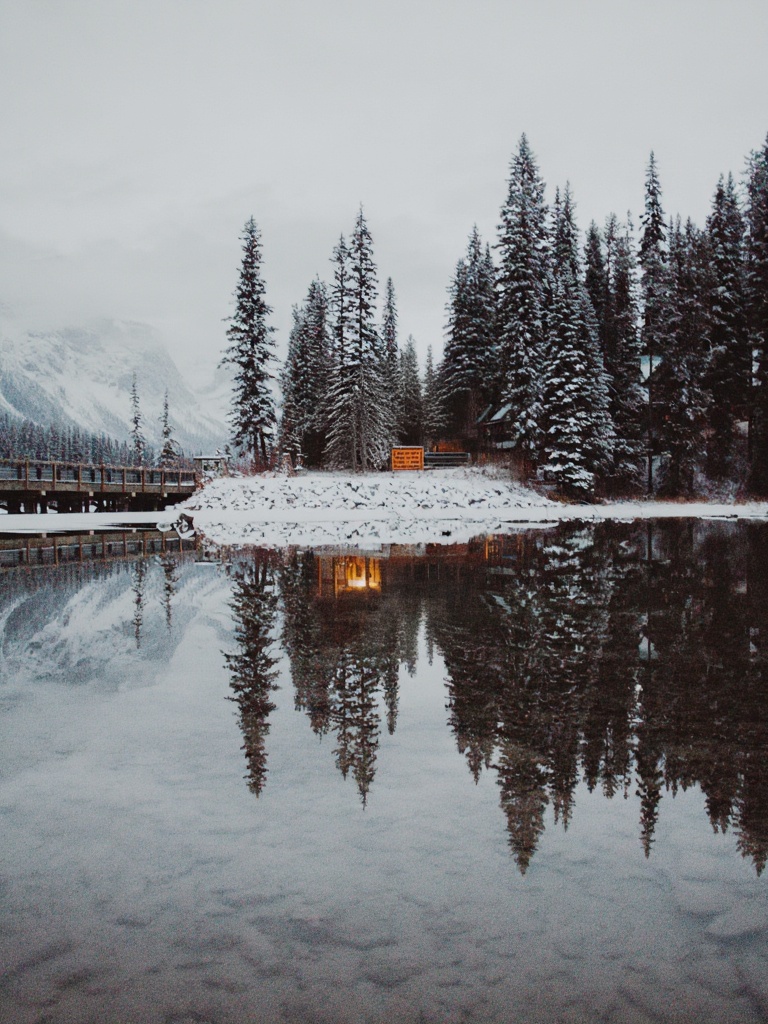} &
        \includegraphics[width=0.19\linewidth]{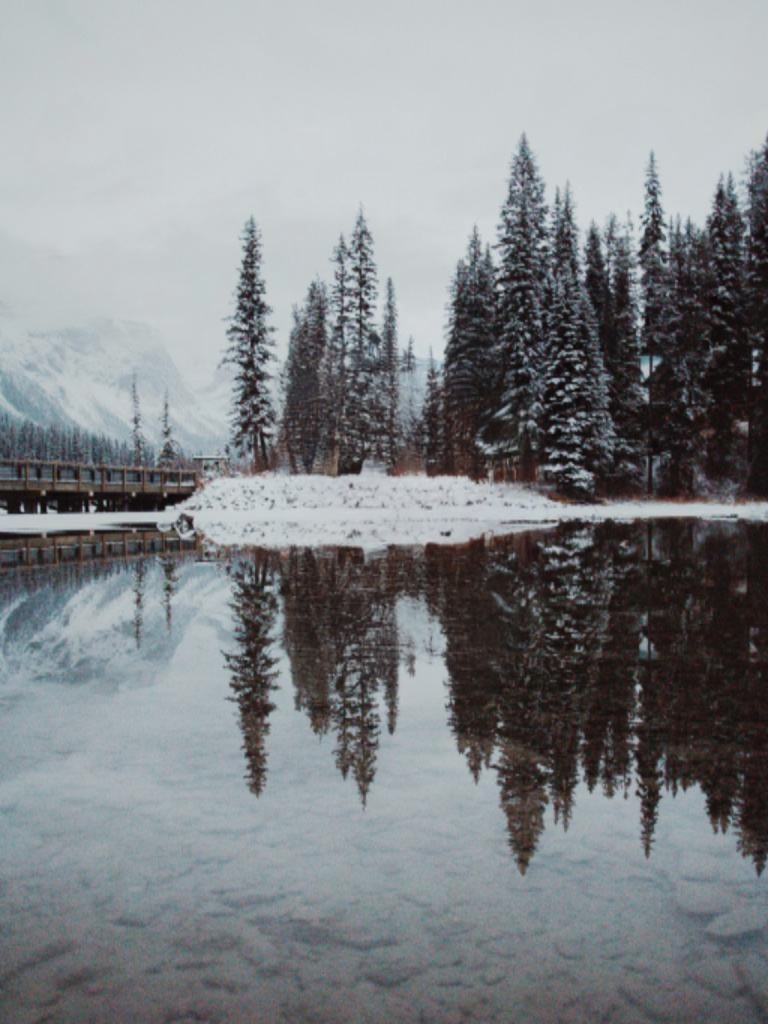} \\[-2pt]
        
        \includegraphics[width=0.19\linewidth]{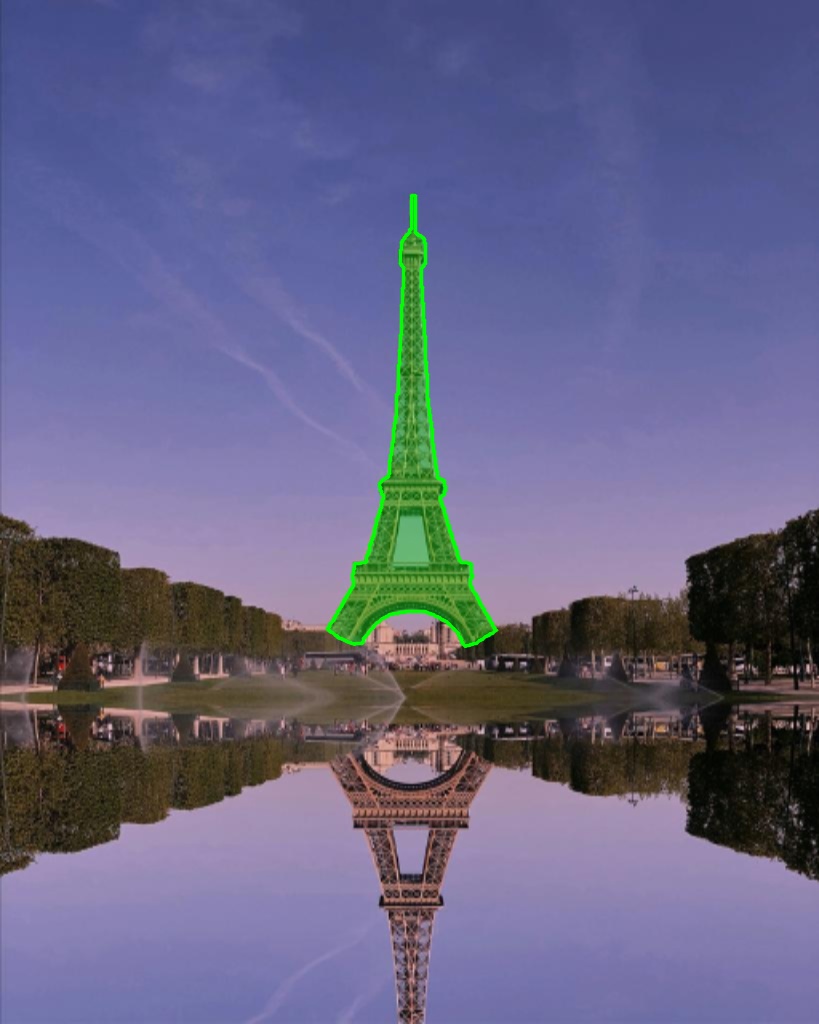} &
        \includegraphics[width=0.19\linewidth]{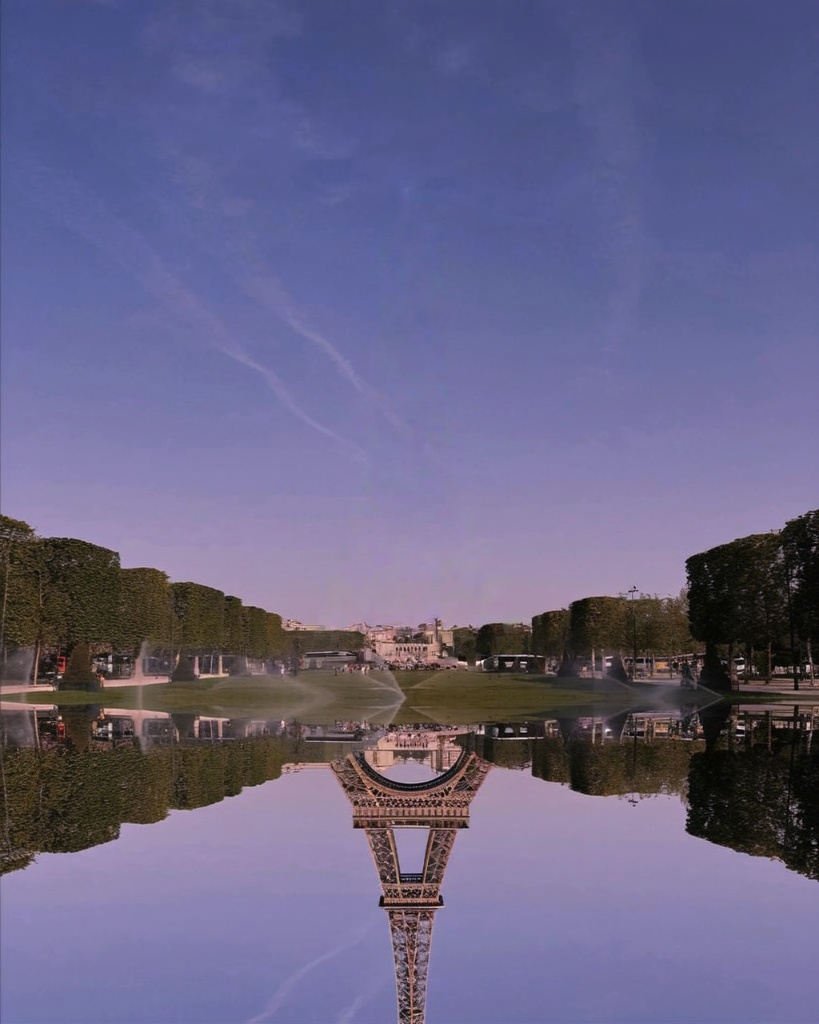} &
        \includegraphics[width=0.19\linewidth]{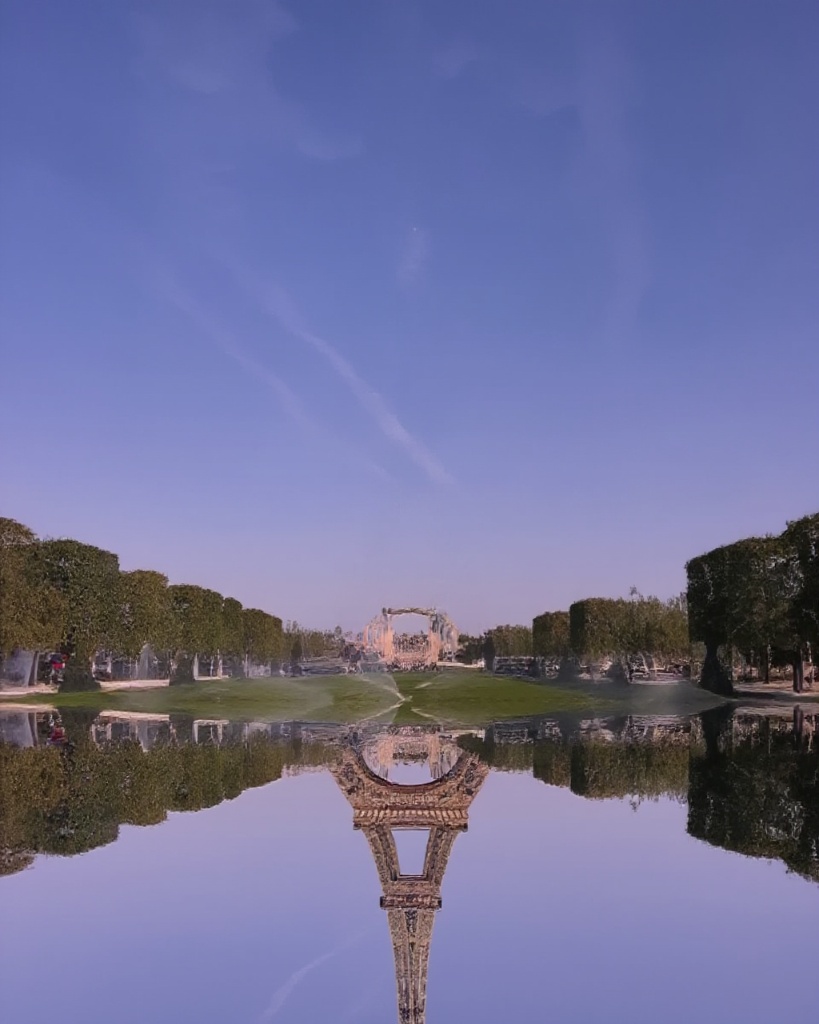} &
        \includegraphics[width=0.19\linewidth]{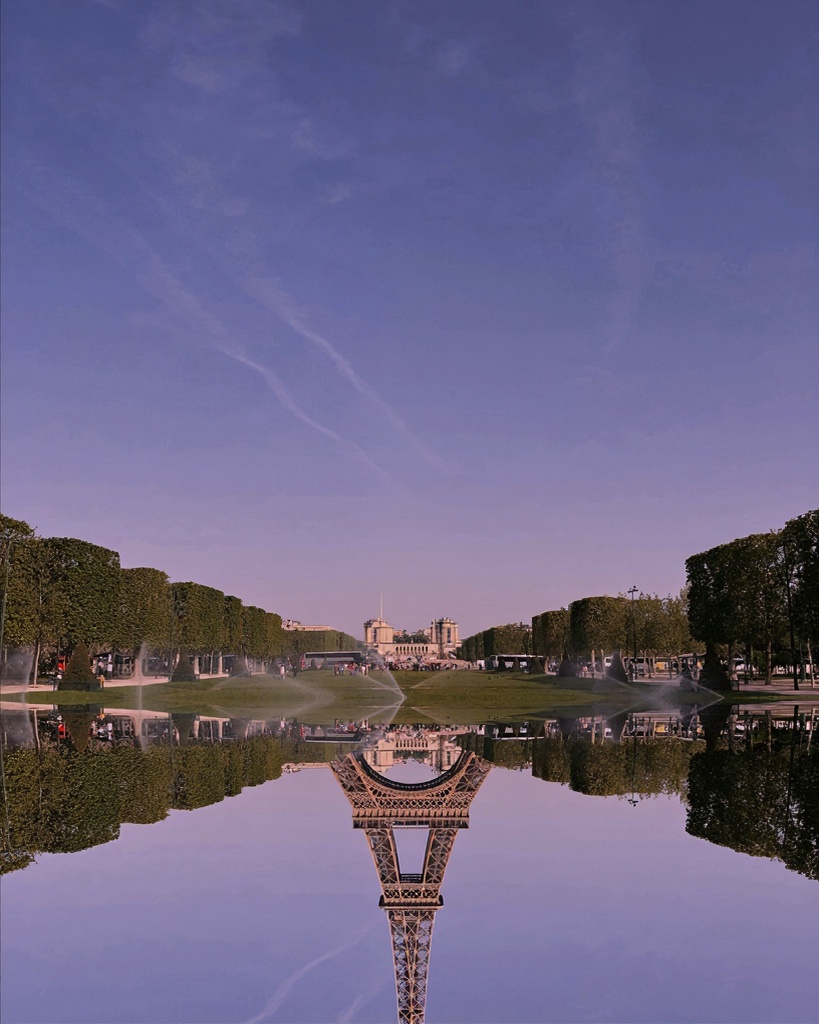} &
        \includegraphics[width=0.19\linewidth]{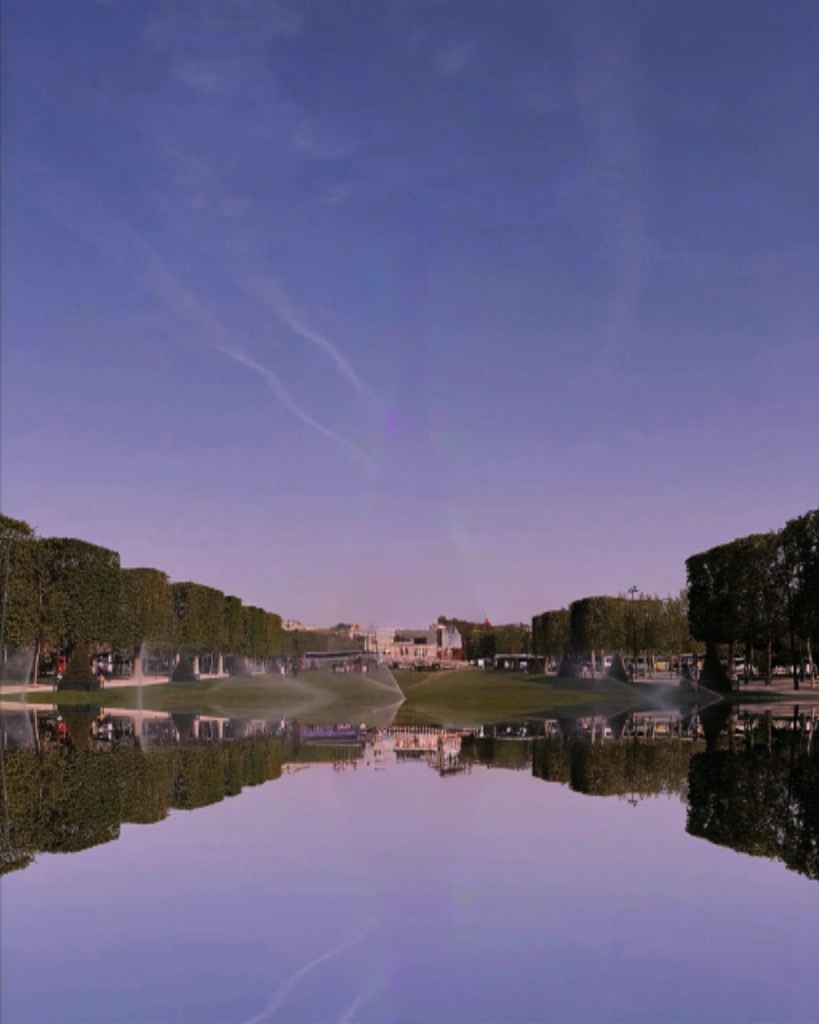} \\[-2pt]

        \footnotesize Input w/ Mask & 
        \footnotesize AttentiveEraser~\cite{sun2025attentive} & 
        \footnotesize OmniEraser~\cite{wei2025omnieraser} & 
        \footnotesize OmniPaint~\cite{yu2025omnipaint} & 
        \footnotesize FlashClear (ours) \\
        
    \end{tabular}
    
\vspace{-2mm}
\caption{More qualitative comparison of our method and others on \textit{CausRem}\cite{zhu2025georemover} reflection dataset.}
\label{fig:causrem_1}
\vspace{-5mm}
\end{figure}

\begin{figure}[t] 
    \centering

    \setlength{\tabcolsep}{0.5pt}      
    \renewcommand{\arraystretch}{1.2}

    \begin{tabular}{@{}ccccc@{}}

        \includegraphics[width=0.19\linewidth]{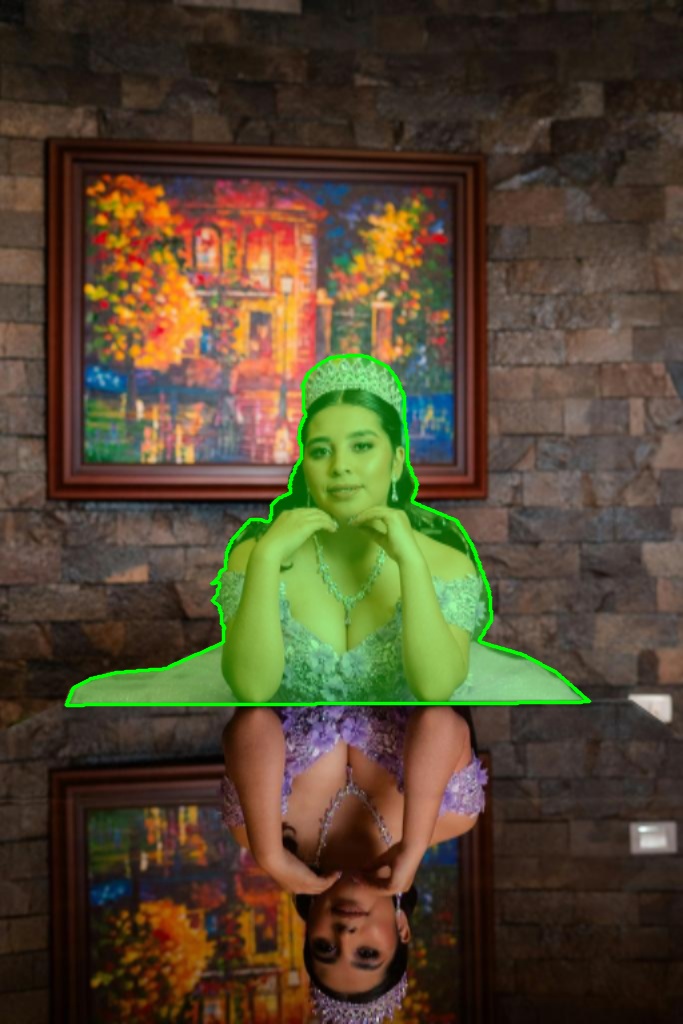} &
        \includegraphics[width=0.19\linewidth]{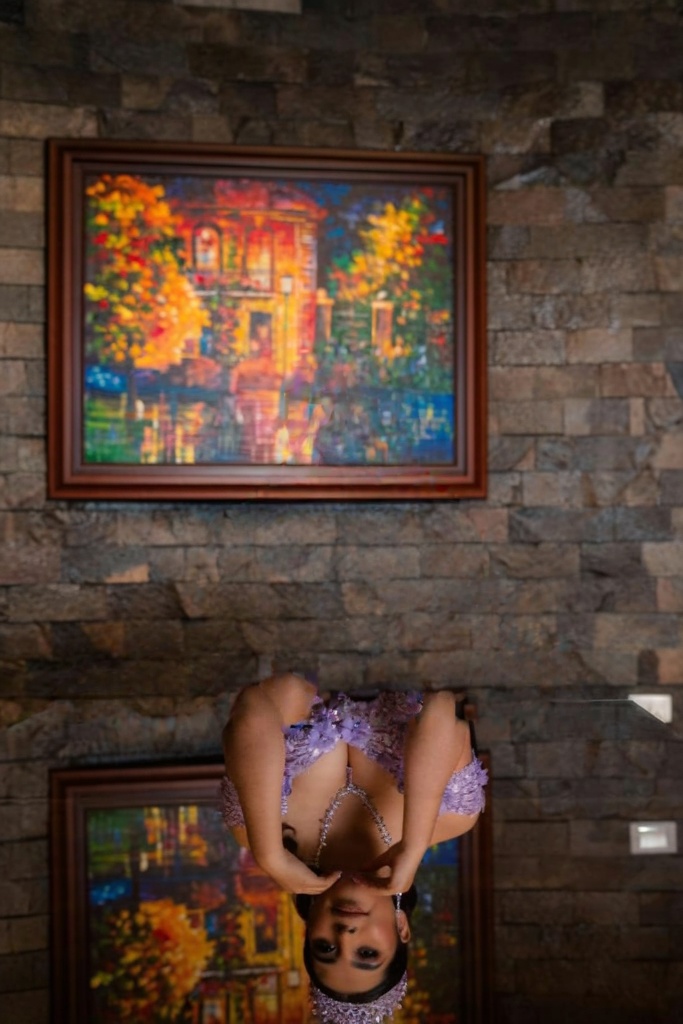} &
        \includegraphics[width=0.19\linewidth]{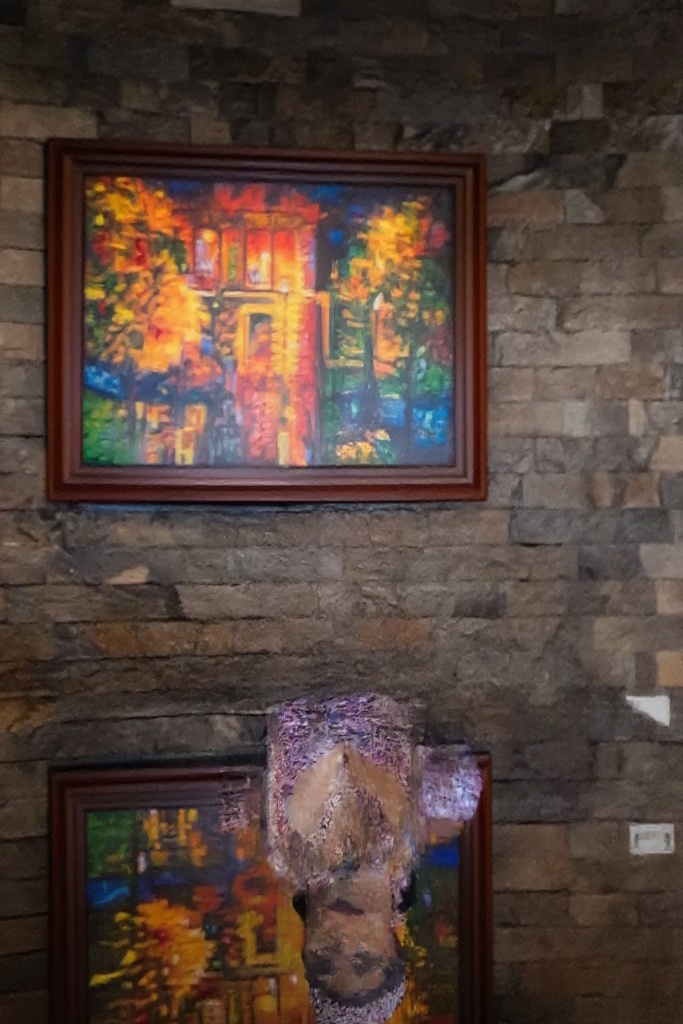} &
        \includegraphics[width=0.19\linewidth]{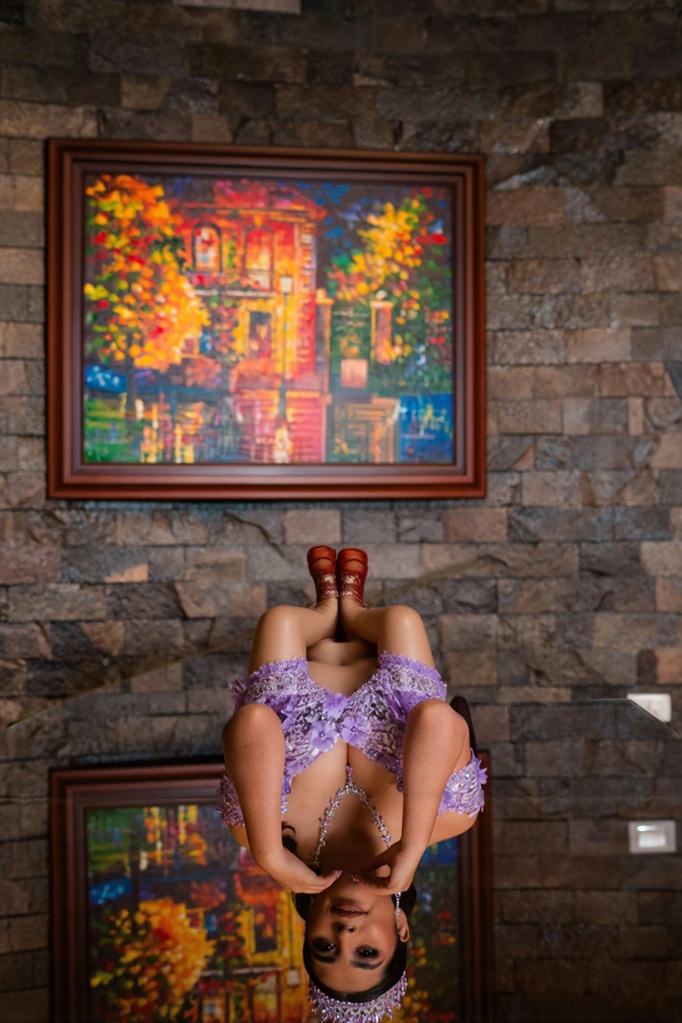} &
        \includegraphics[width=0.19\linewidth]{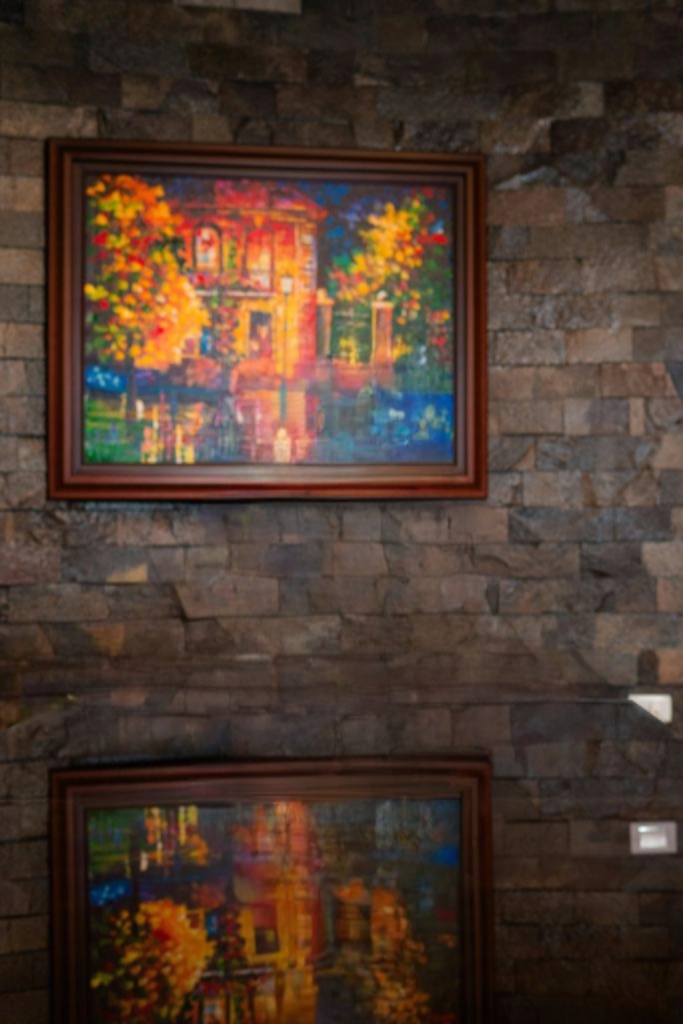} \\[-2pt]

       \includegraphics[width=0.19\linewidth]{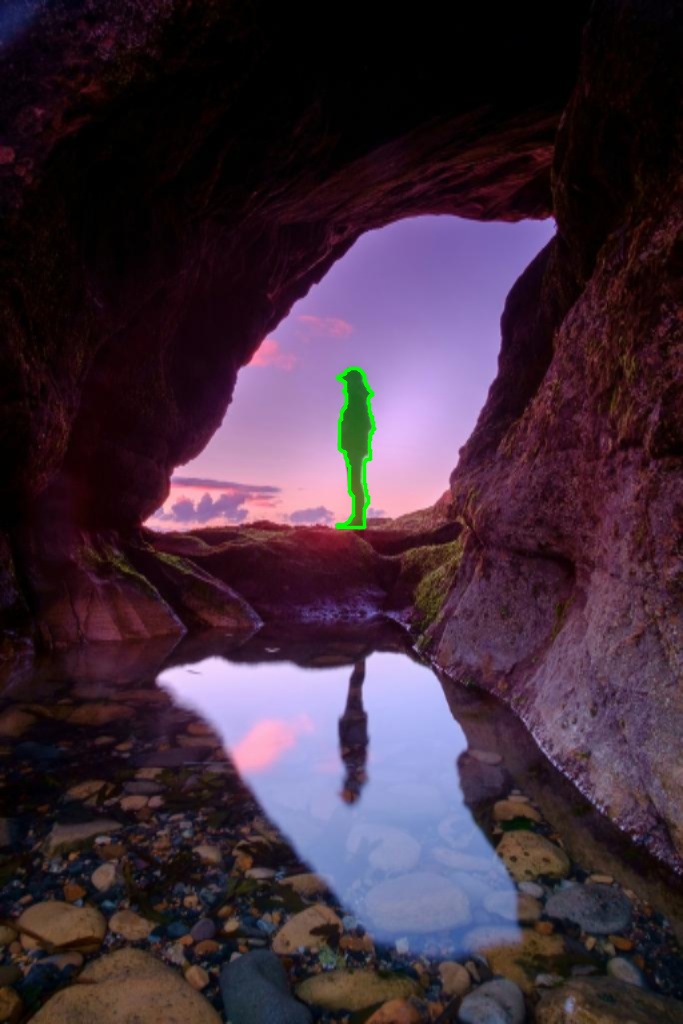} &
        \includegraphics[width=0.19\linewidth]{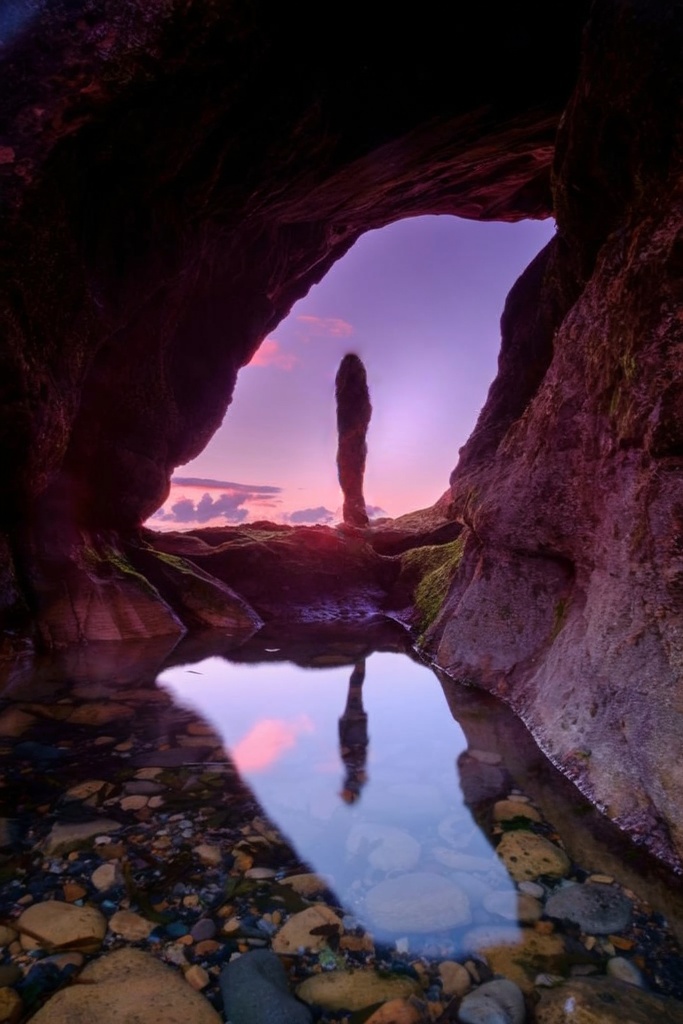} &
        \includegraphics[width=0.19\linewidth]{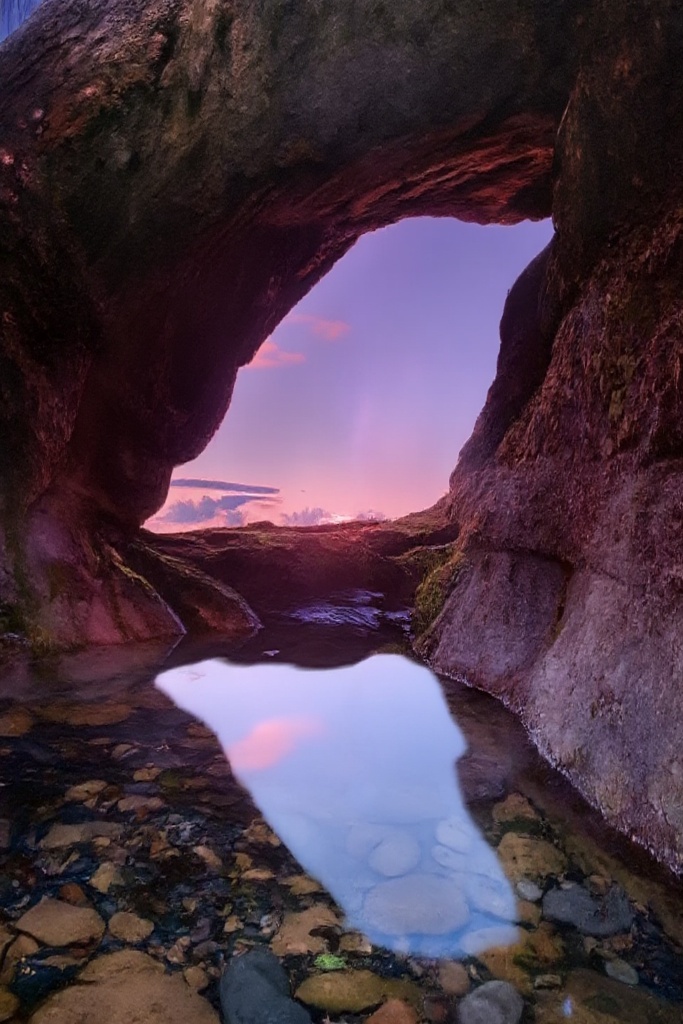} &
        \includegraphics[width=0.19\linewidth]{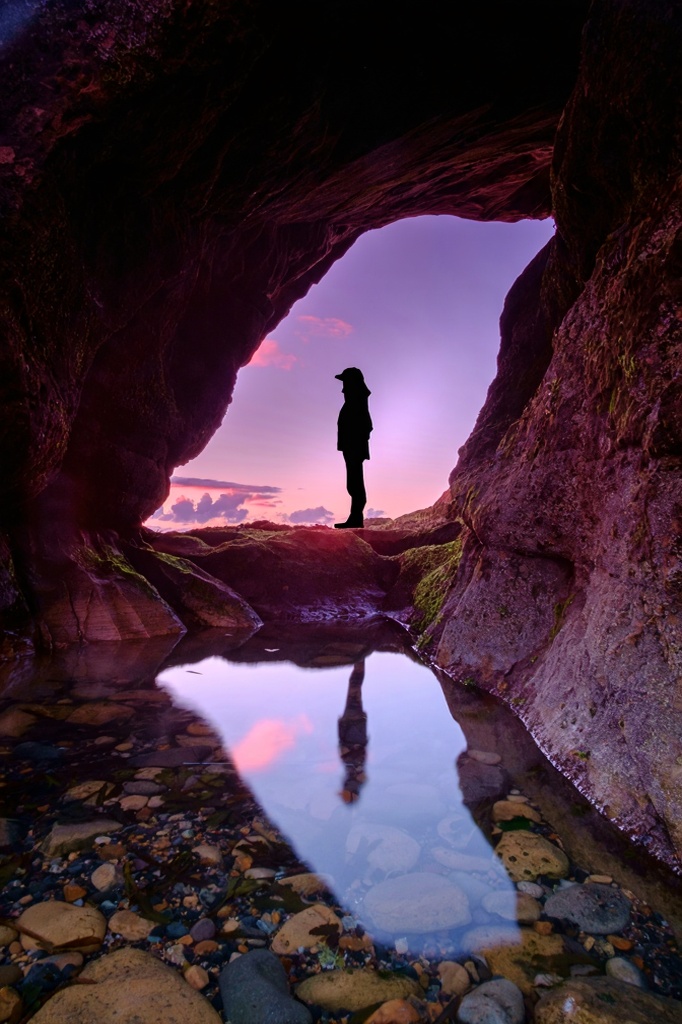} &
        \includegraphics[width=0.19\linewidth]{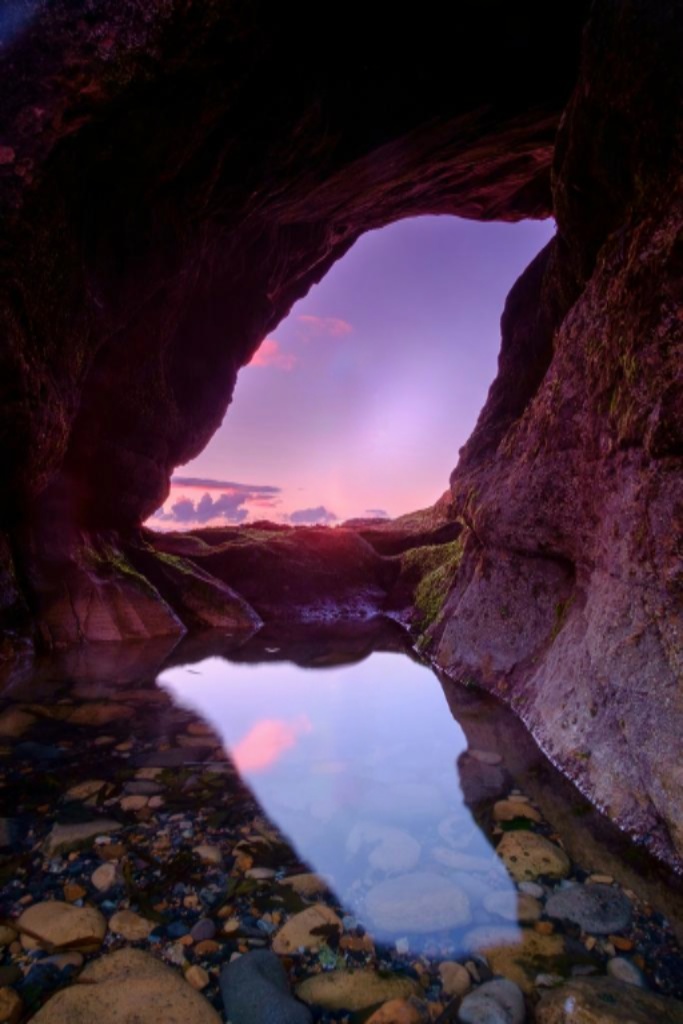} \\[-2pt]

        \includegraphics[width=0.19\linewidth]{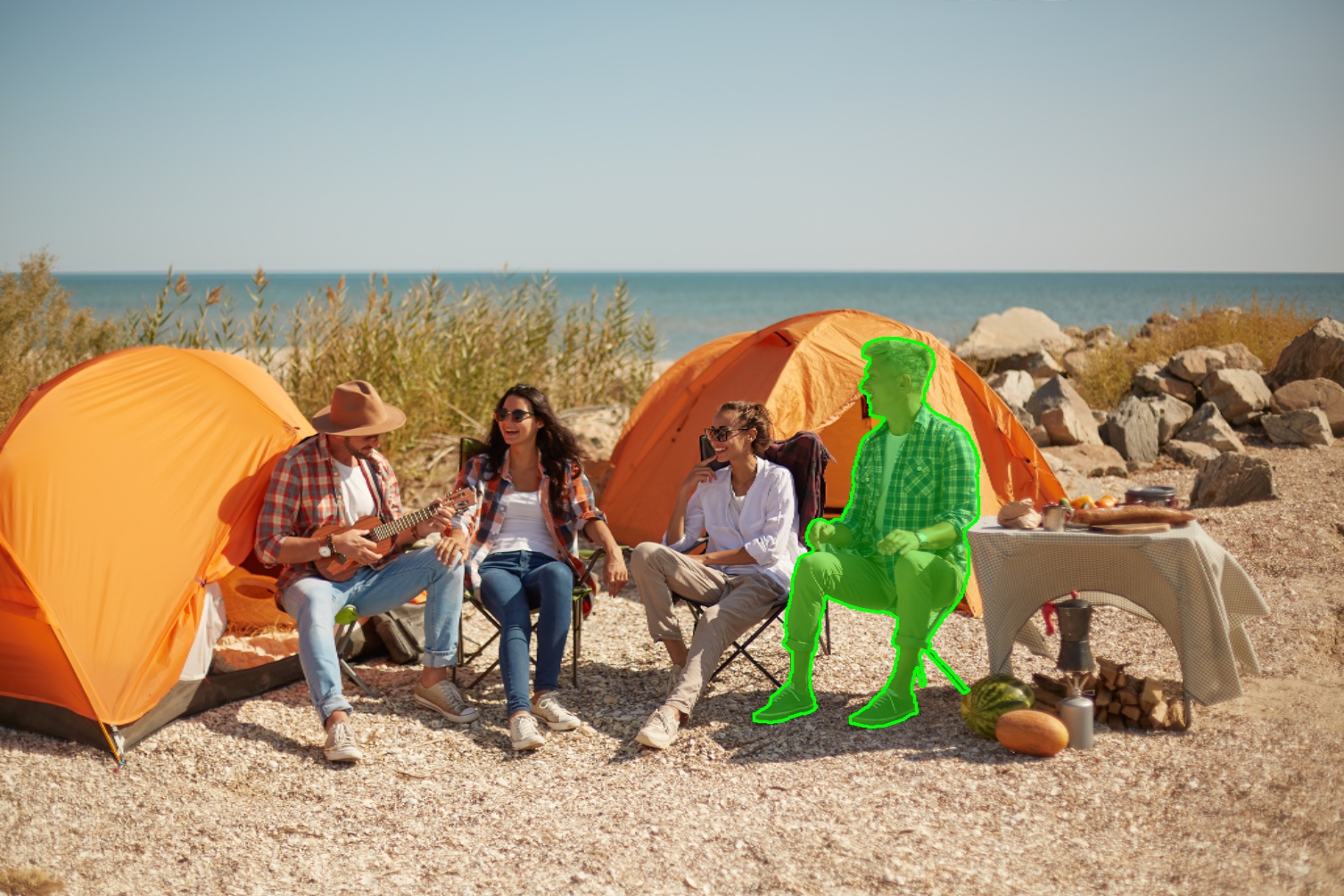} &
        \includegraphics[width=0.19\linewidth]{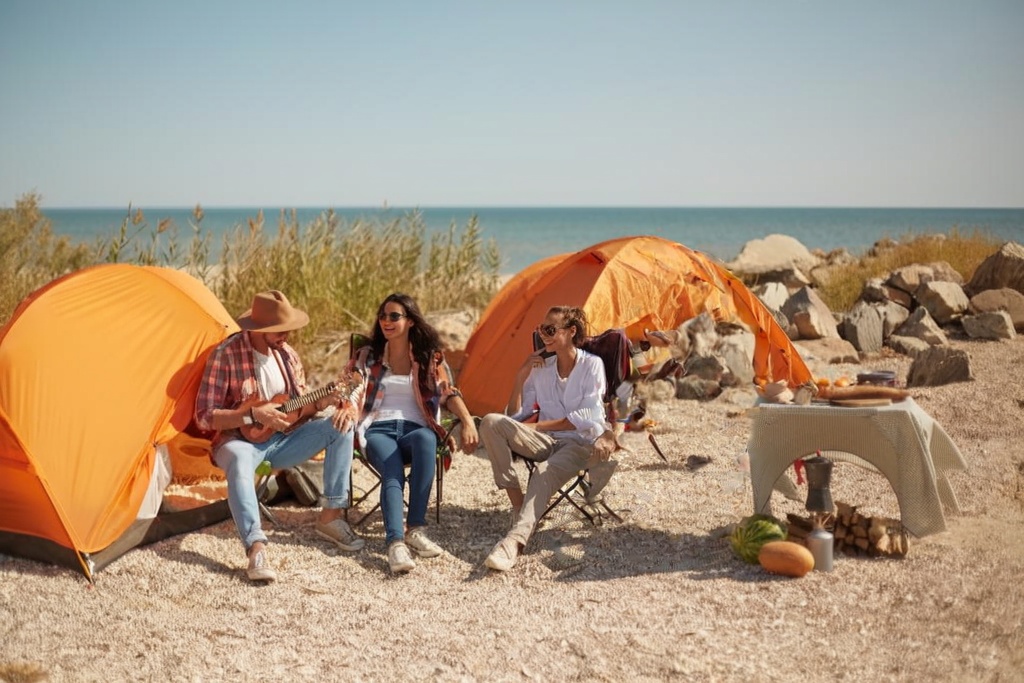} &
        \includegraphics[width=0.19\linewidth]{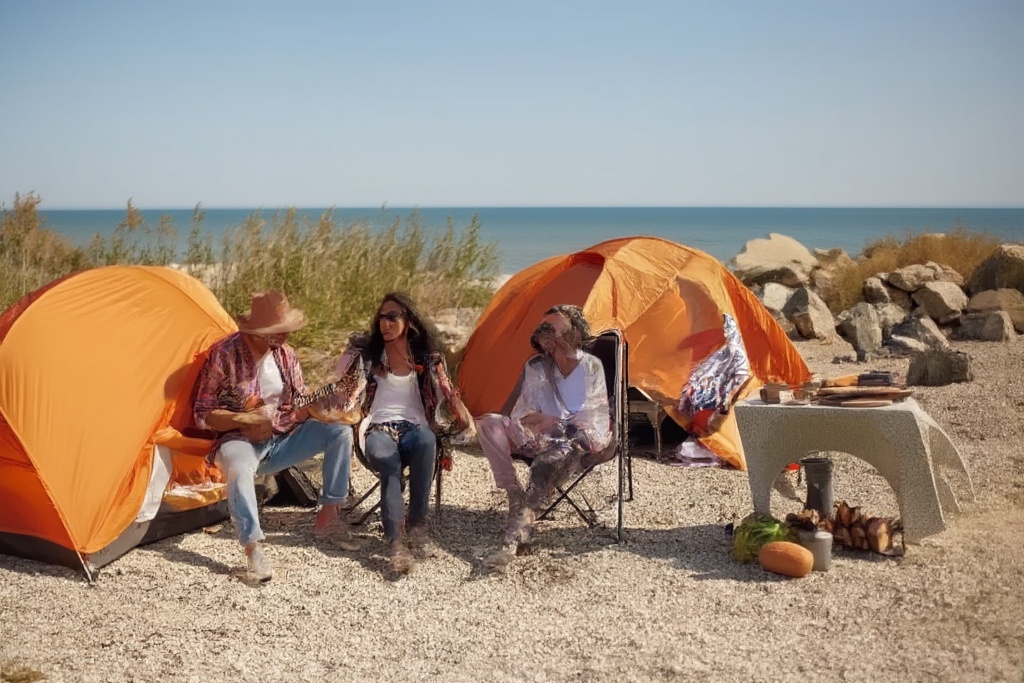} &
        \includegraphics[width=0.19\linewidth]{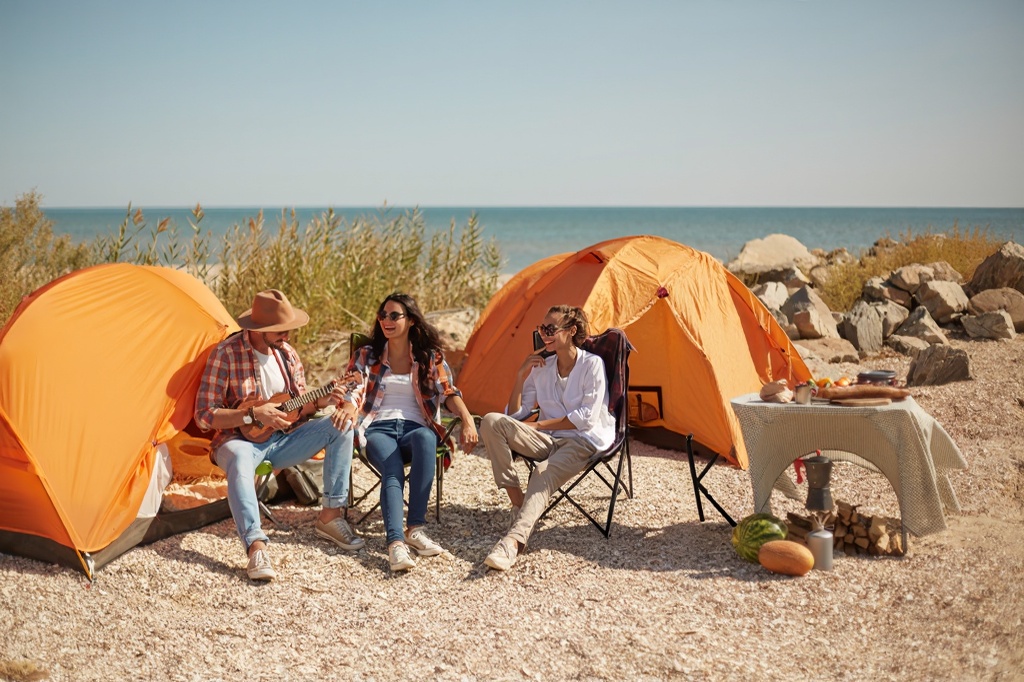} &
        \includegraphics[width=0.19\linewidth]{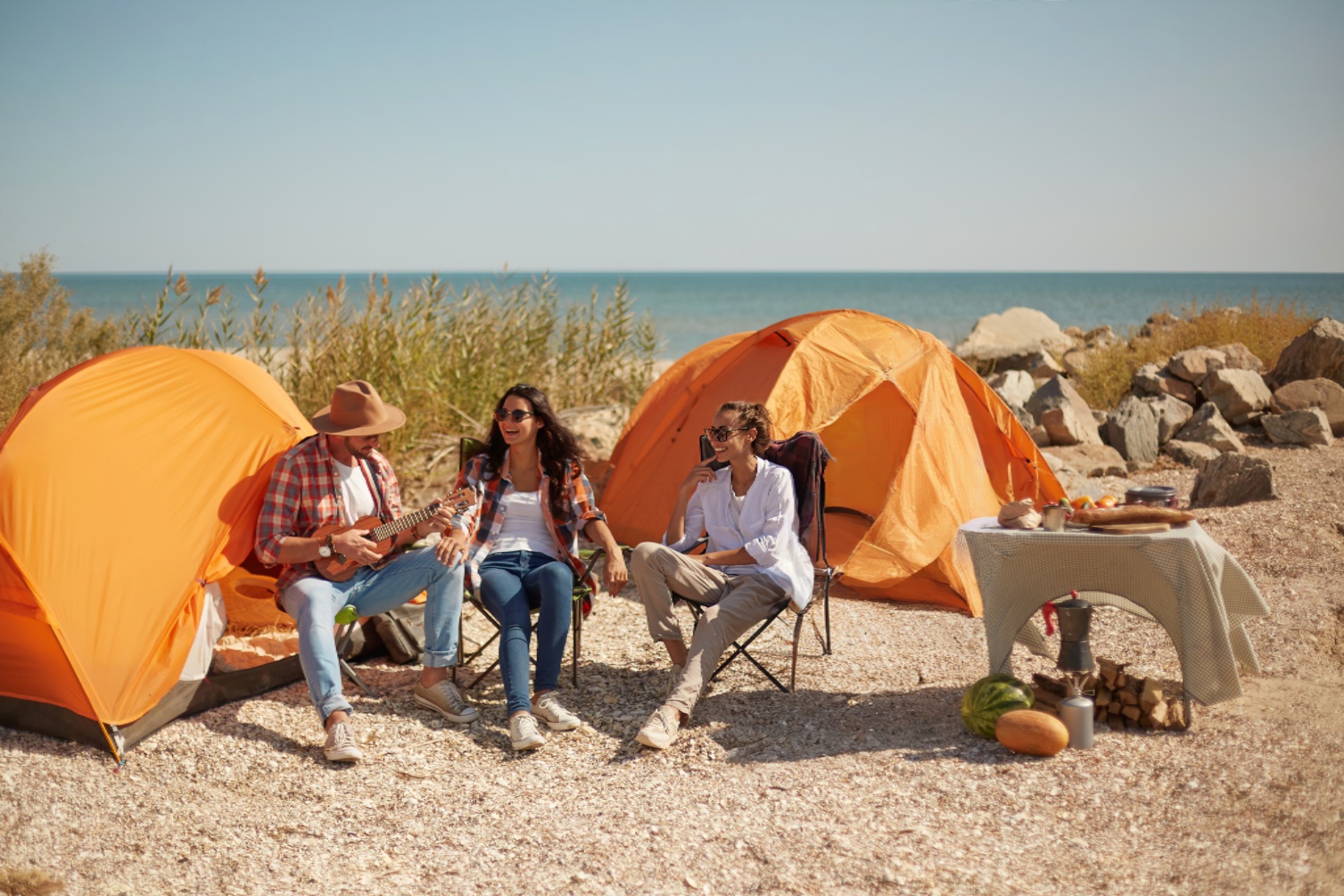} \\[-2pt]
        
        \includegraphics[width=0.19\linewidth]{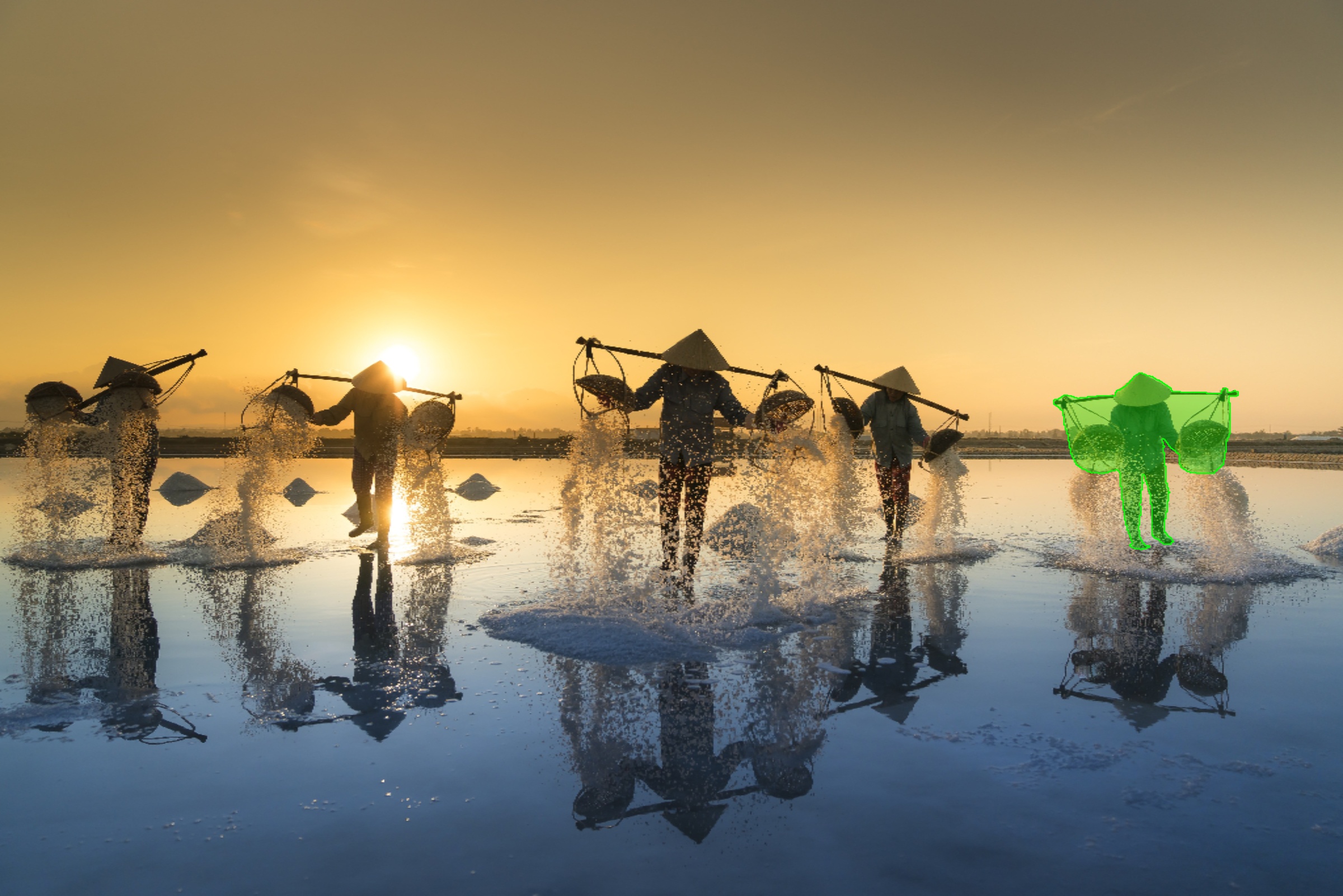} &
        \includegraphics[width=0.19\linewidth]{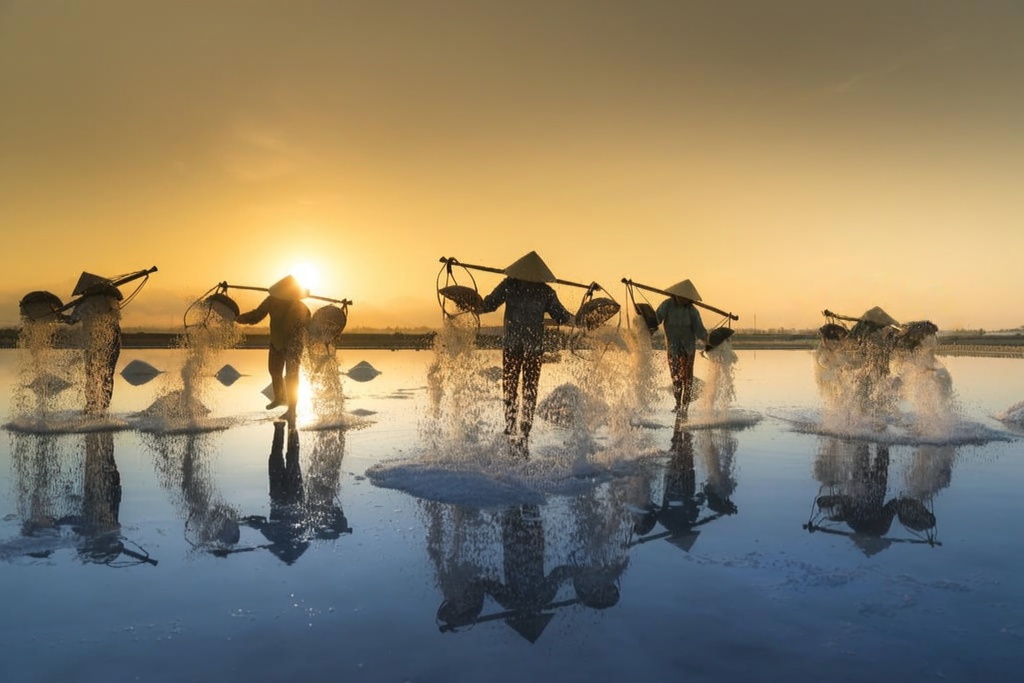} &
        \includegraphics[width=0.19\linewidth]{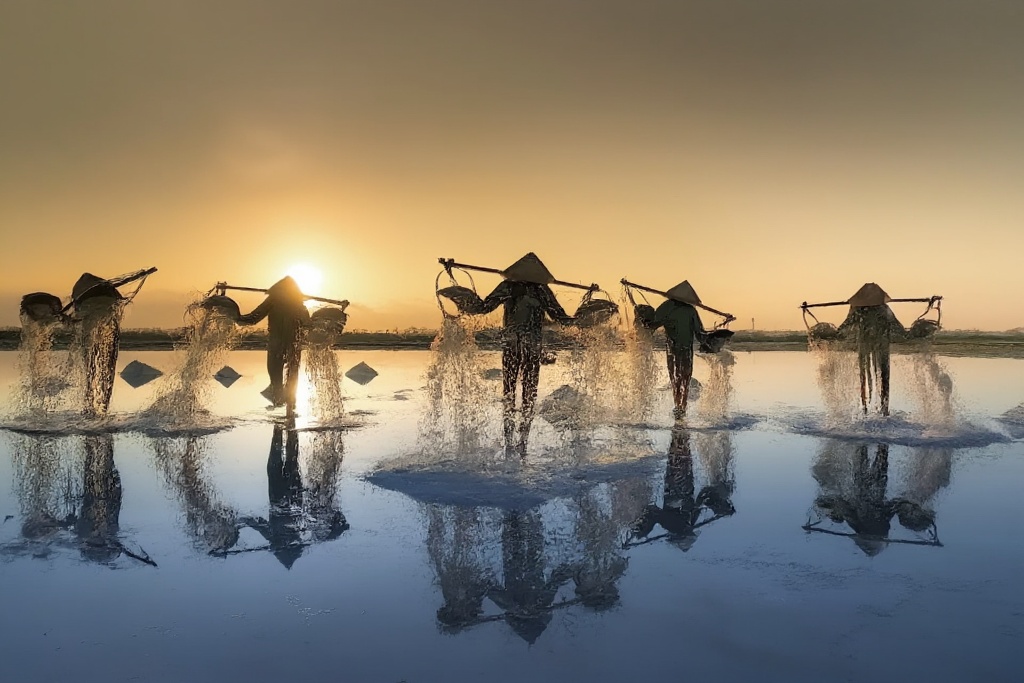} &
        \includegraphics[width=0.19\linewidth]{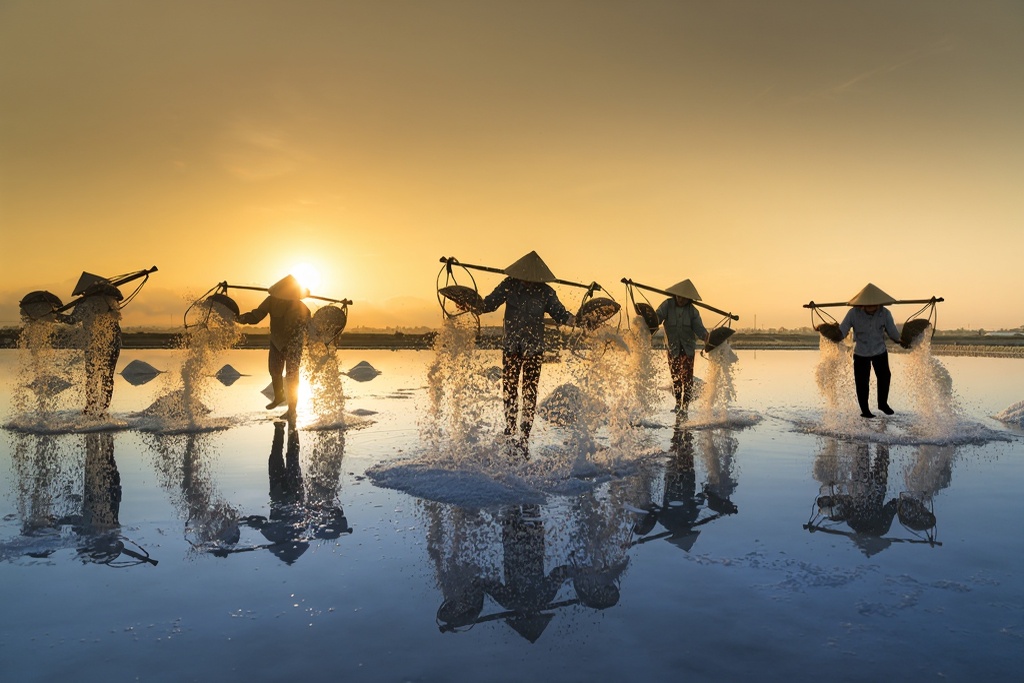} &
        \includegraphics[width=0.19\linewidth]{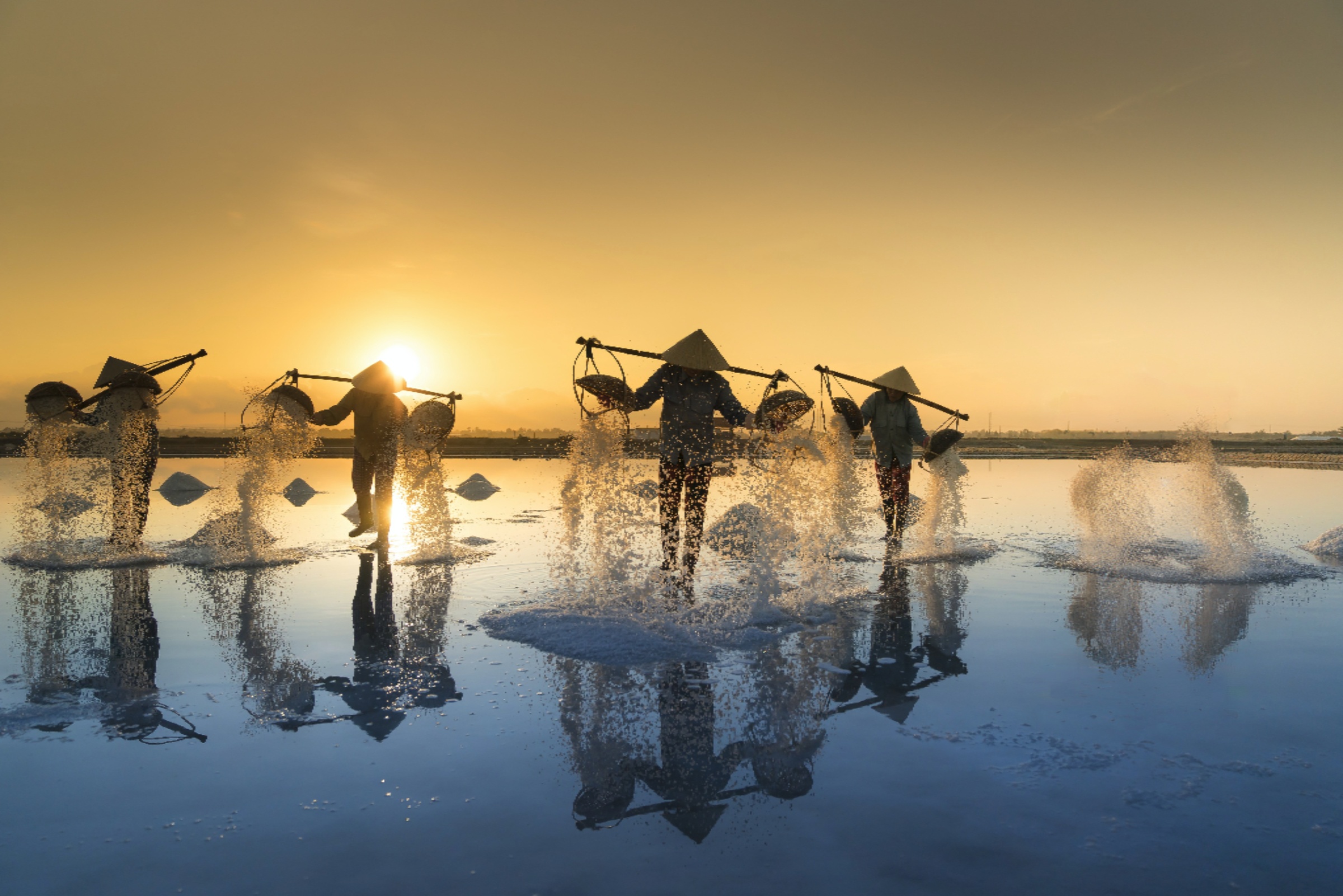} \\[-2pt]
        
        \includegraphics[width=0.19\linewidth]{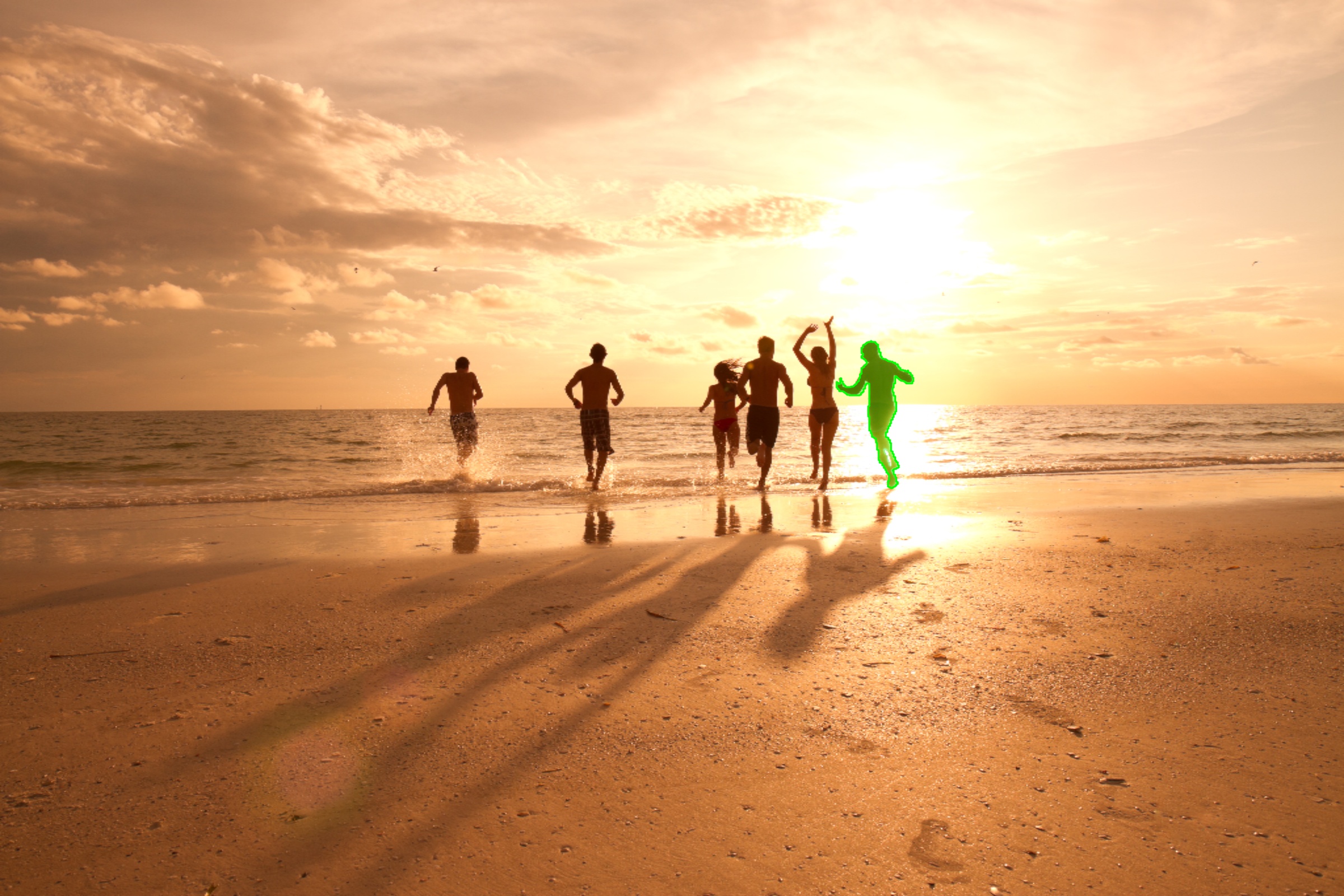} &
        \includegraphics[width=0.19\linewidth]{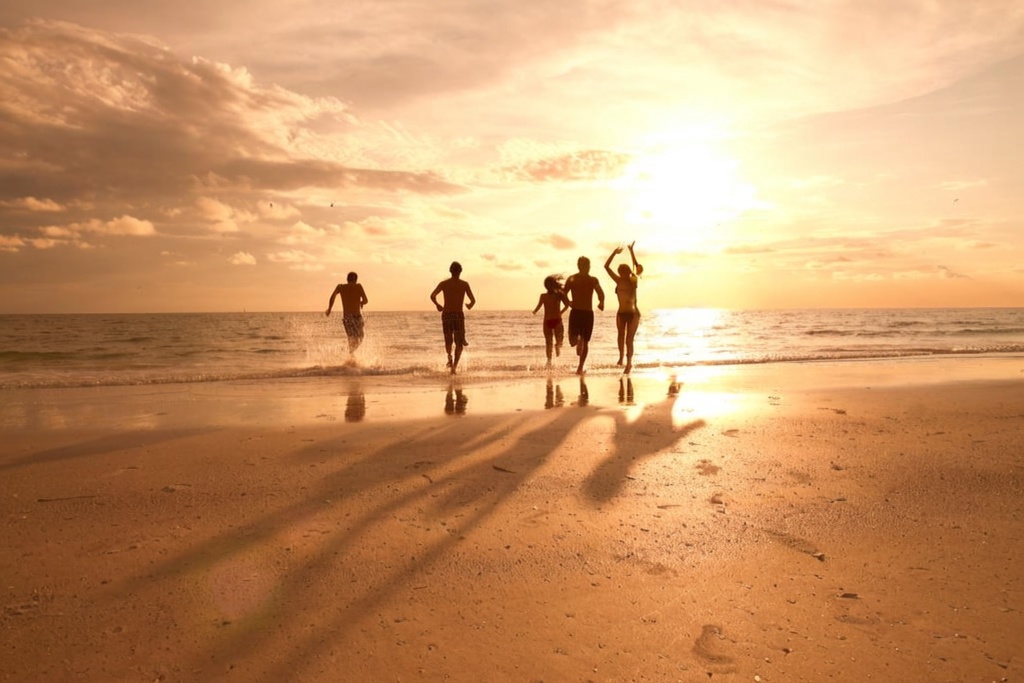} &
        \includegraphics[width=0.19\linewidth]{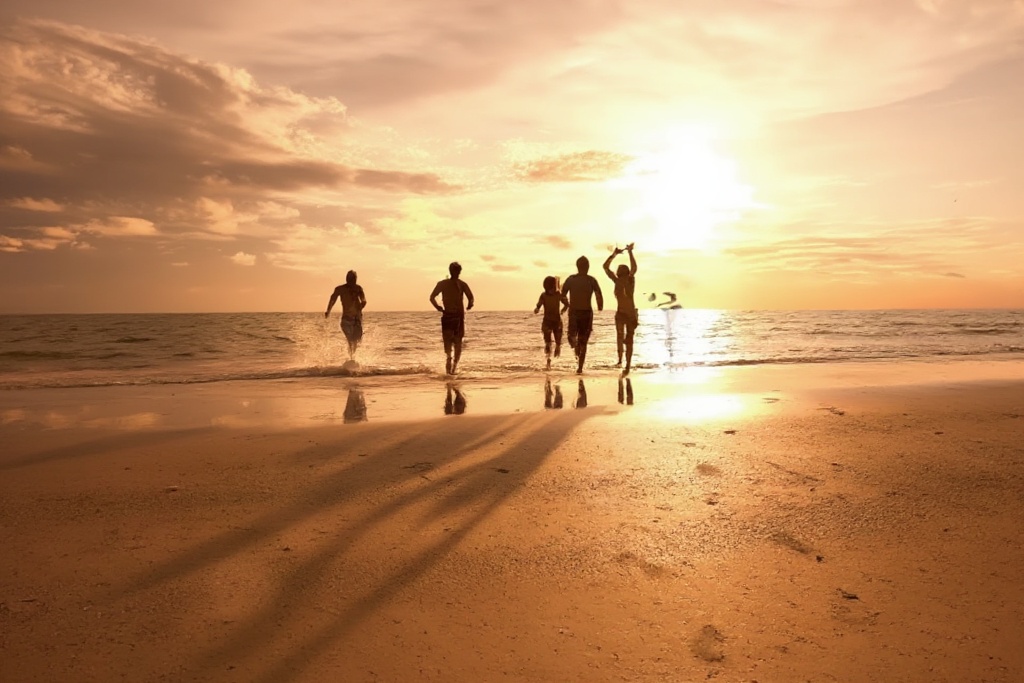} &
        \includegraphics[width=0.19\linewidth]{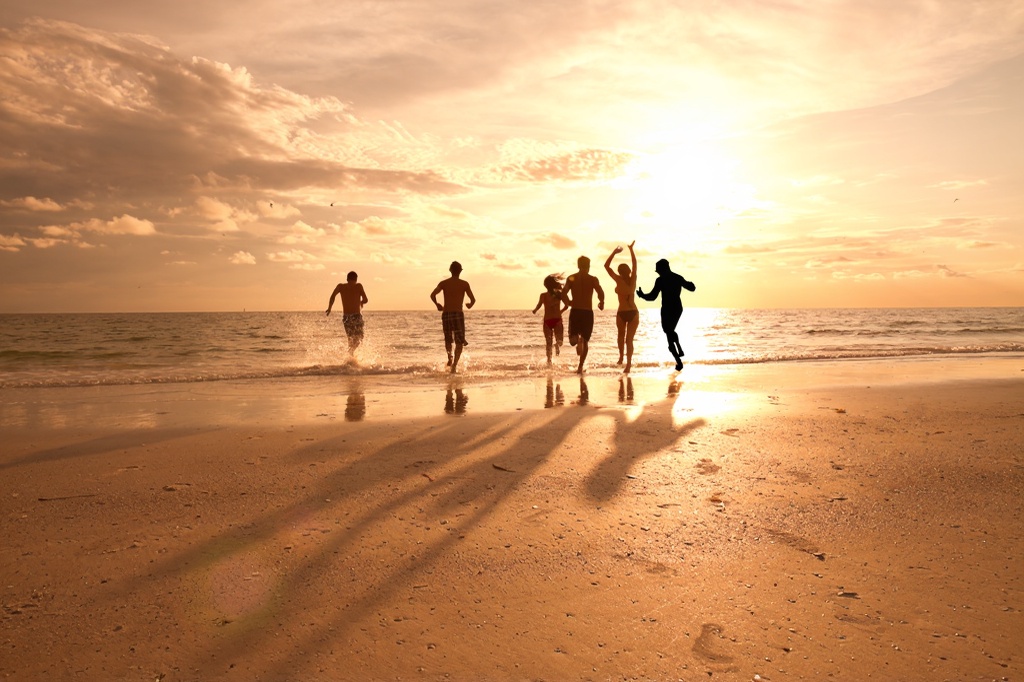} &
        \includegraphics[width=0.19\linewidth]{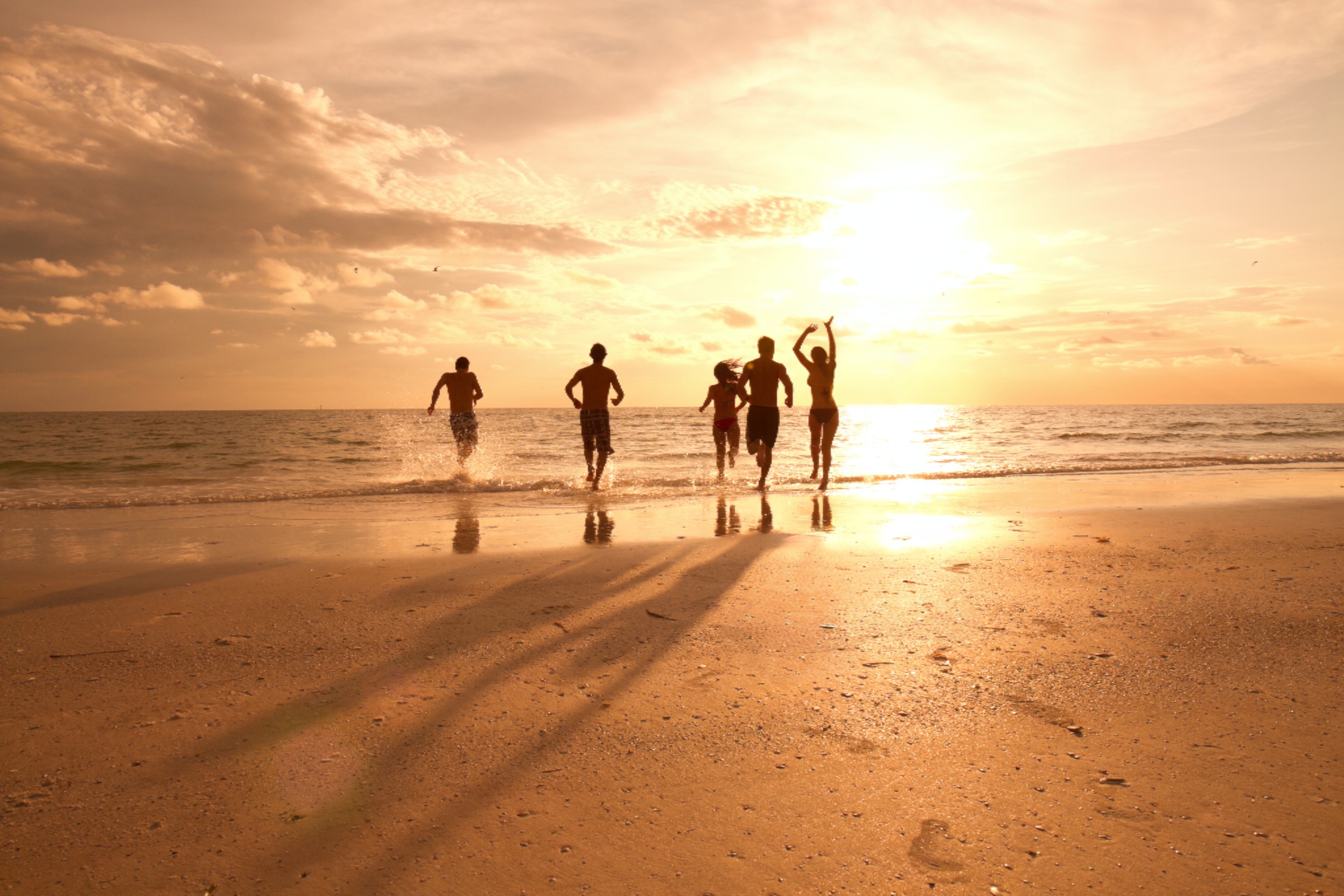} \\[-2pt]

        \includegraphics[width=0.19\linewidth]{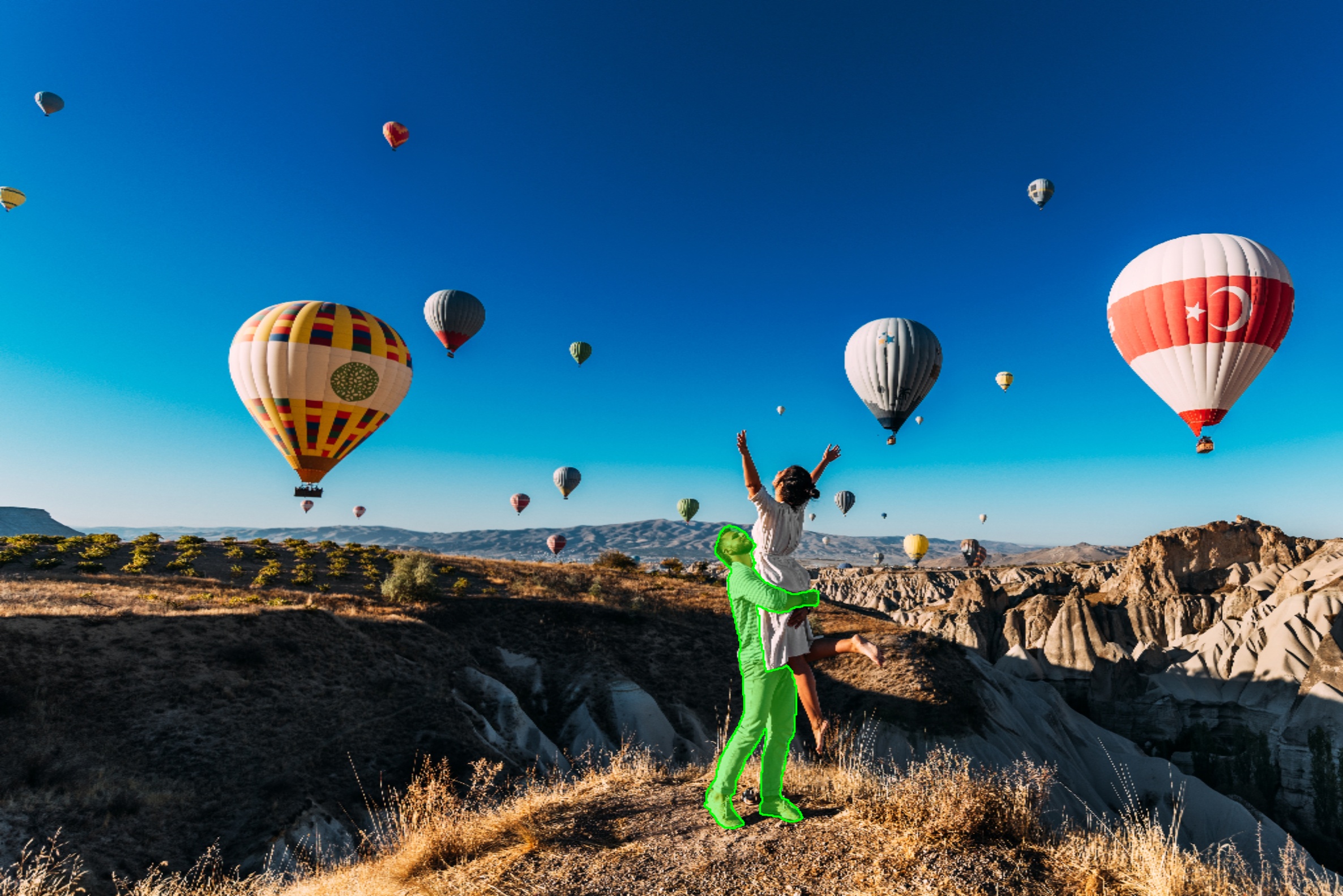} &
        \includegraphics[width=0.19\linewidth]{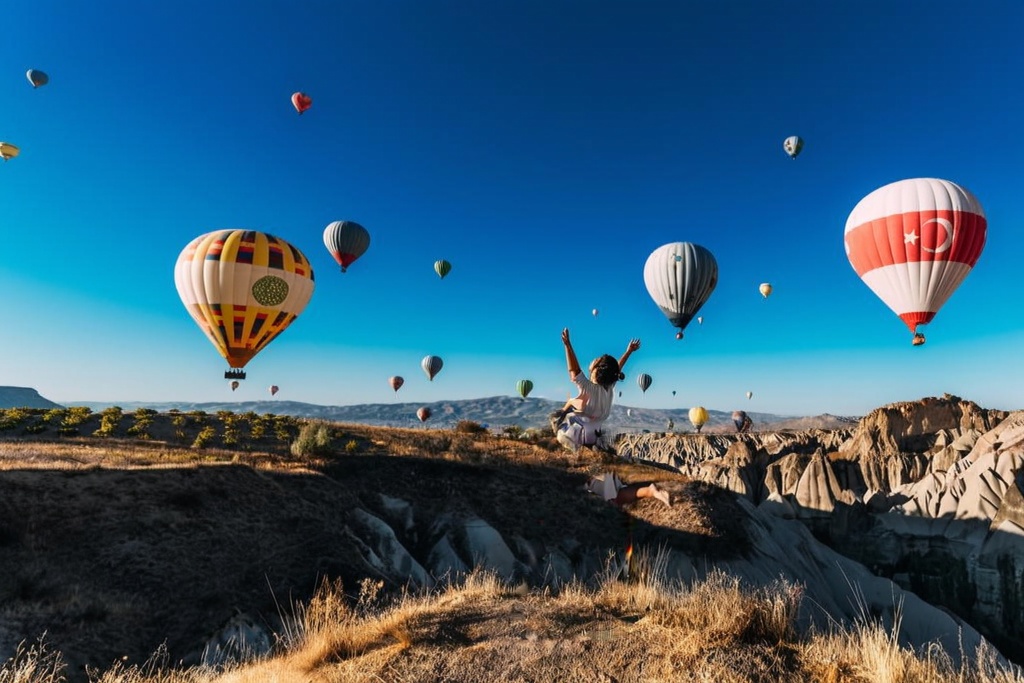} &
        \includegraphics[width=0.19\linewidth]{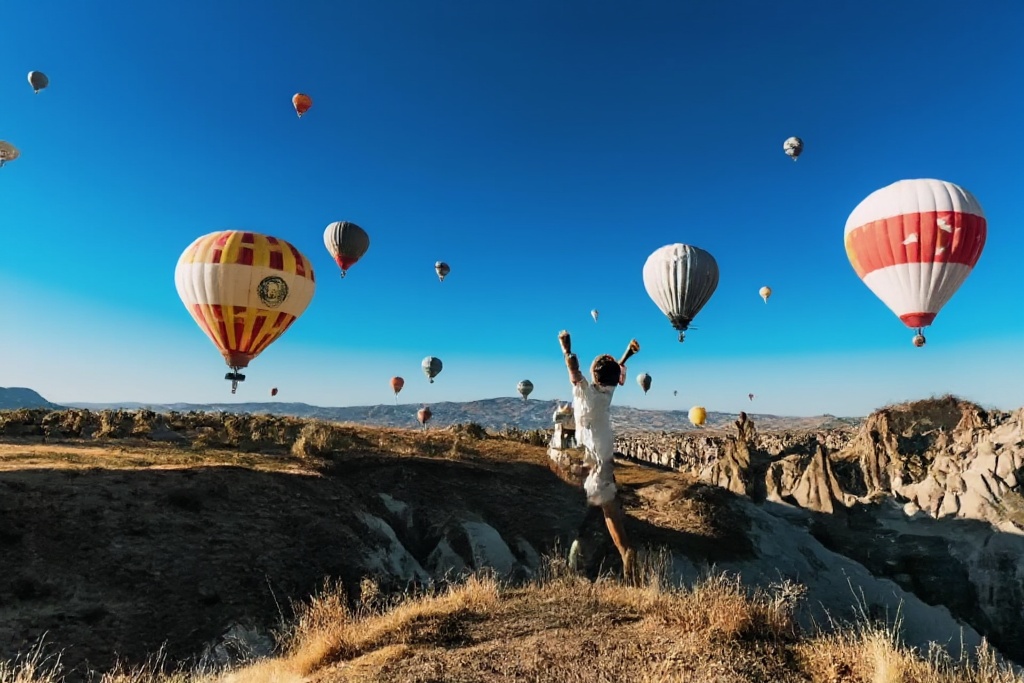} &
        \includegraphics[width=0.19\linewidth]{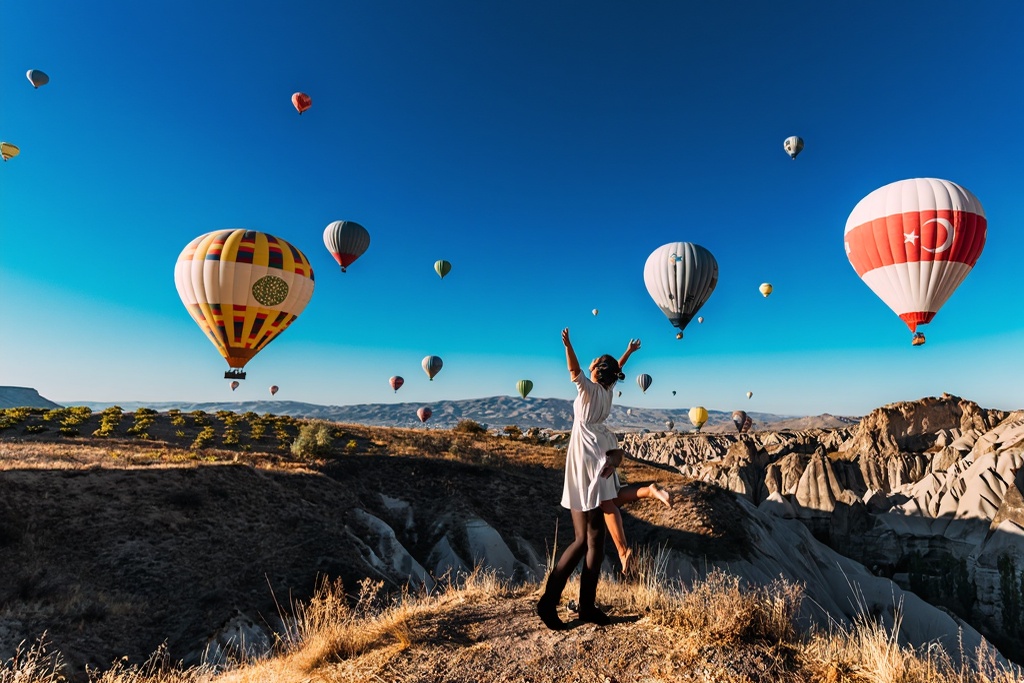} &
        \includegraphics[width=0.19\linewidth]{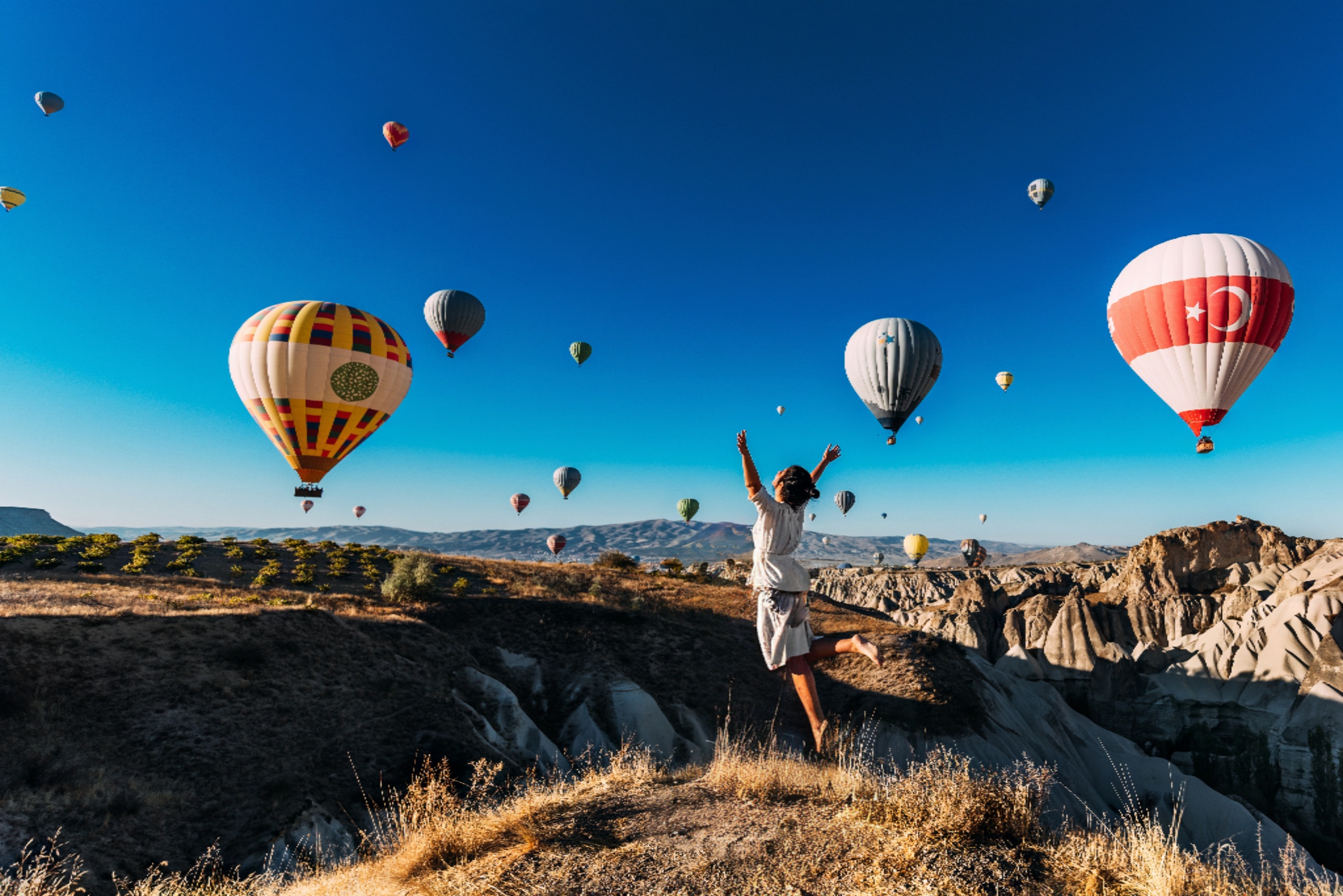} \\[-2pt]

        \includegraphics[width=0.19\linewidth]{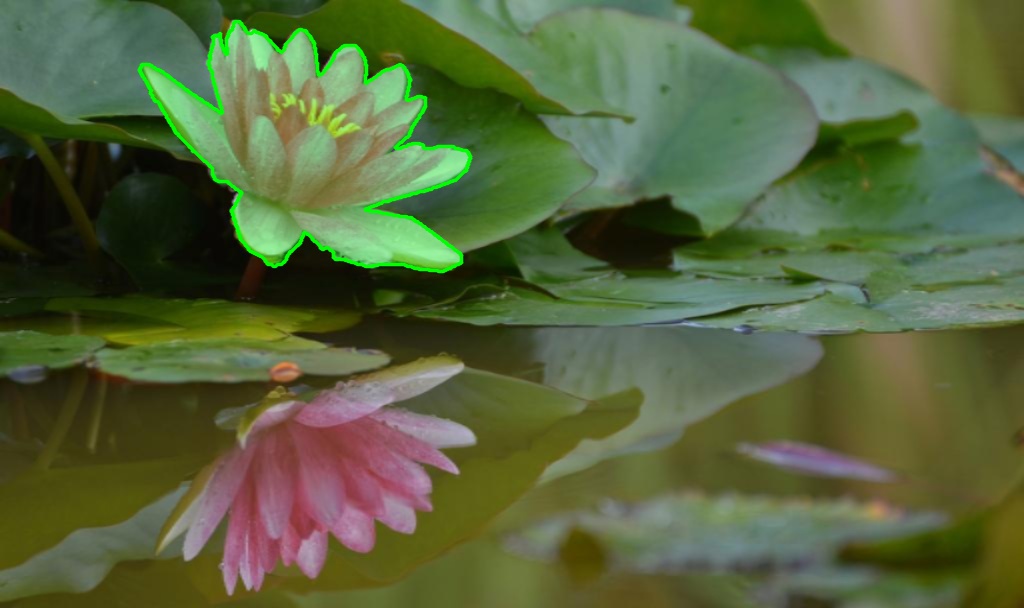} &
        \includegraphics[width=0.19\linewidth]{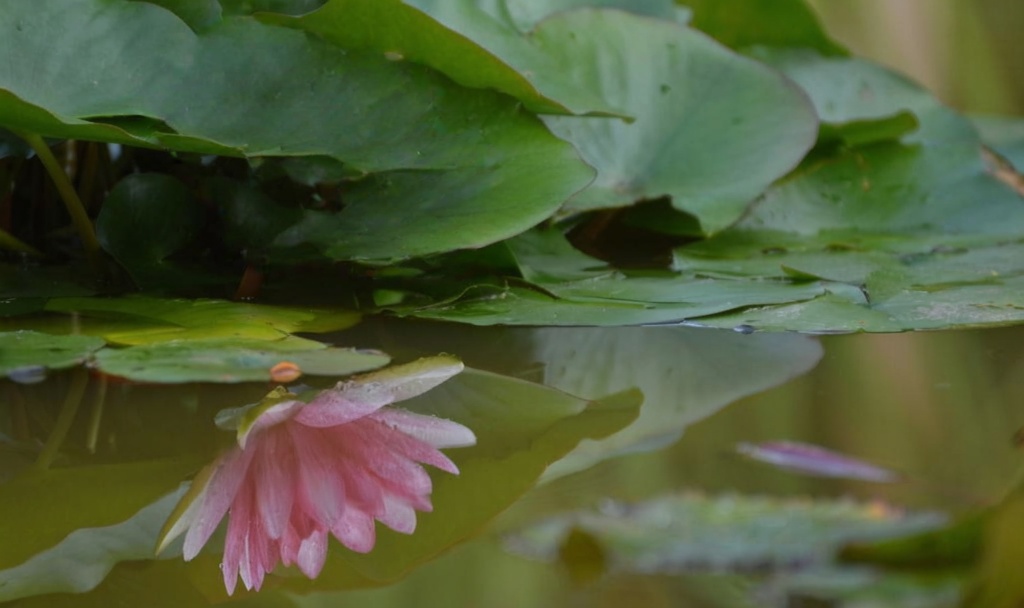} &
        \includegraphics[width=0.19\linewidth]{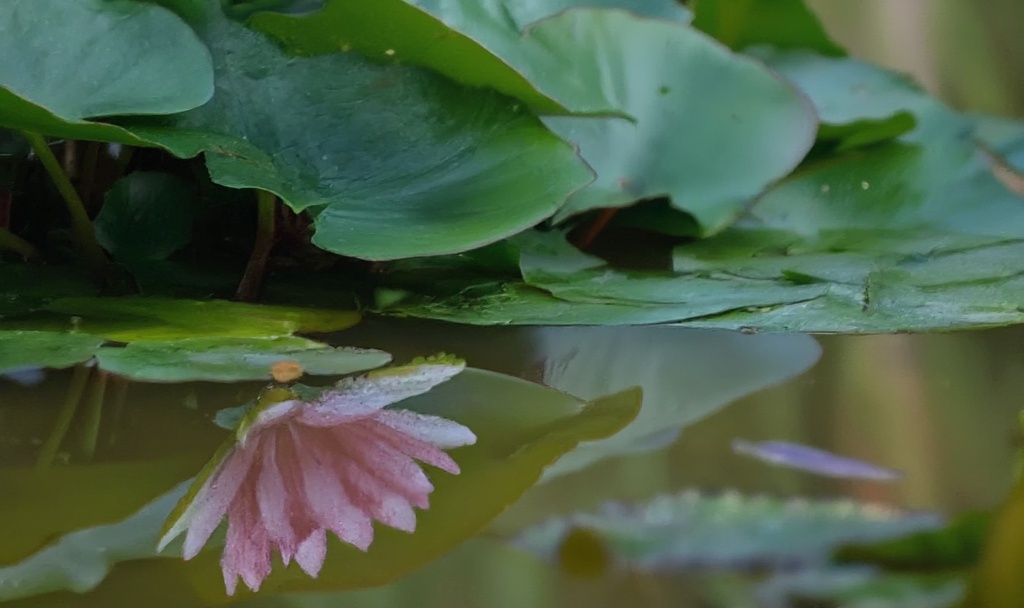} &
        \includegraphics[width=0.19\linewidth]{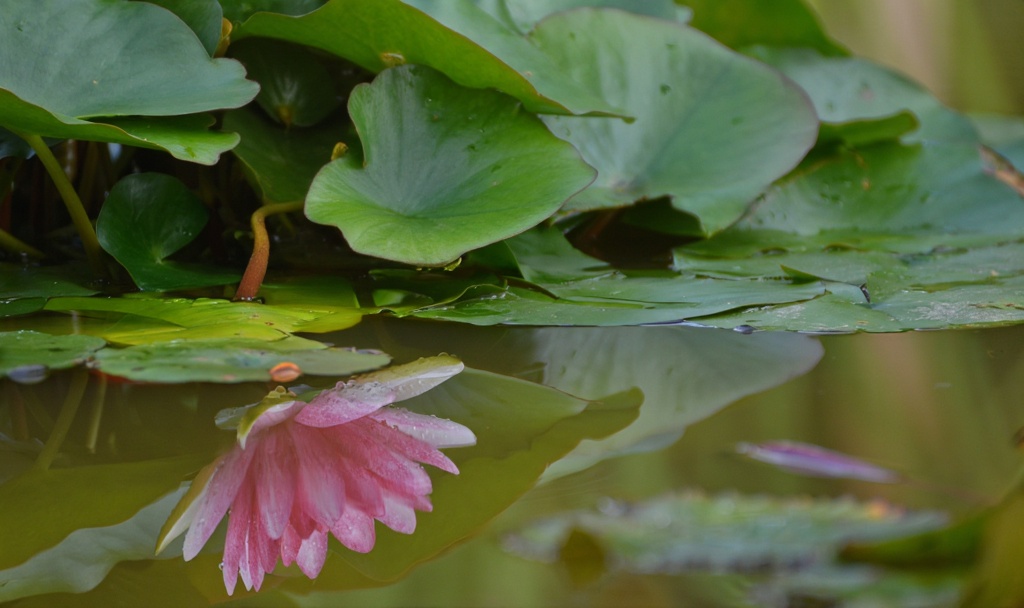} &
        \includegraphics[width=0.19\linewidth]{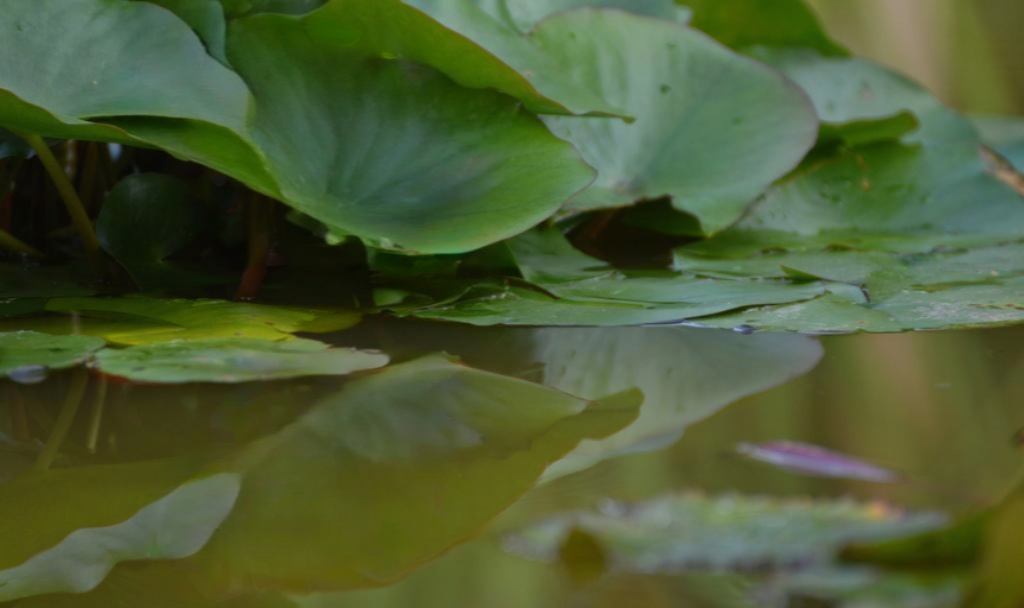} \\[-2pt]

        \includegraphics[width=0.19\linewidth]{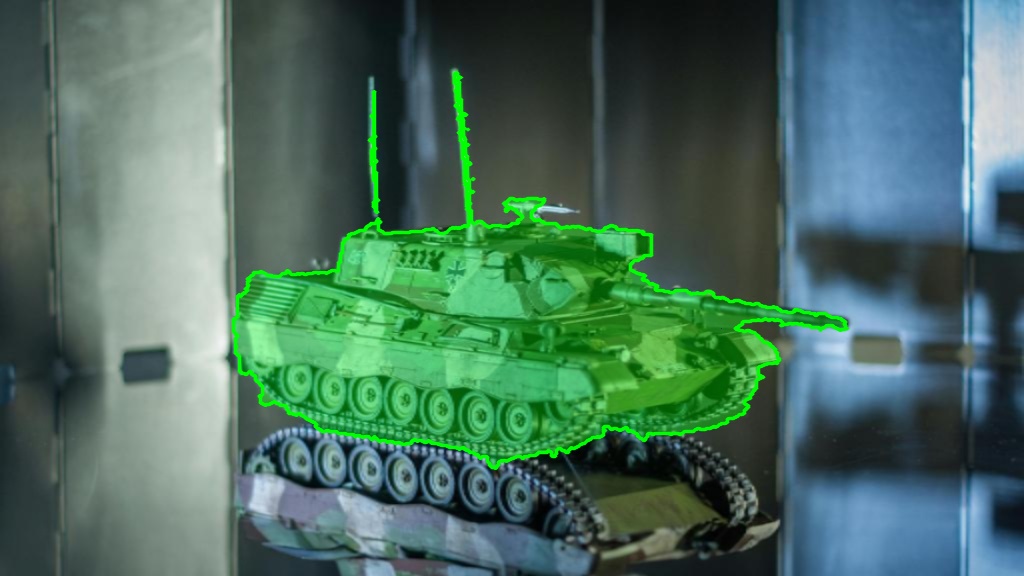} &
        \includegraphics[width=0.19\linewidth]{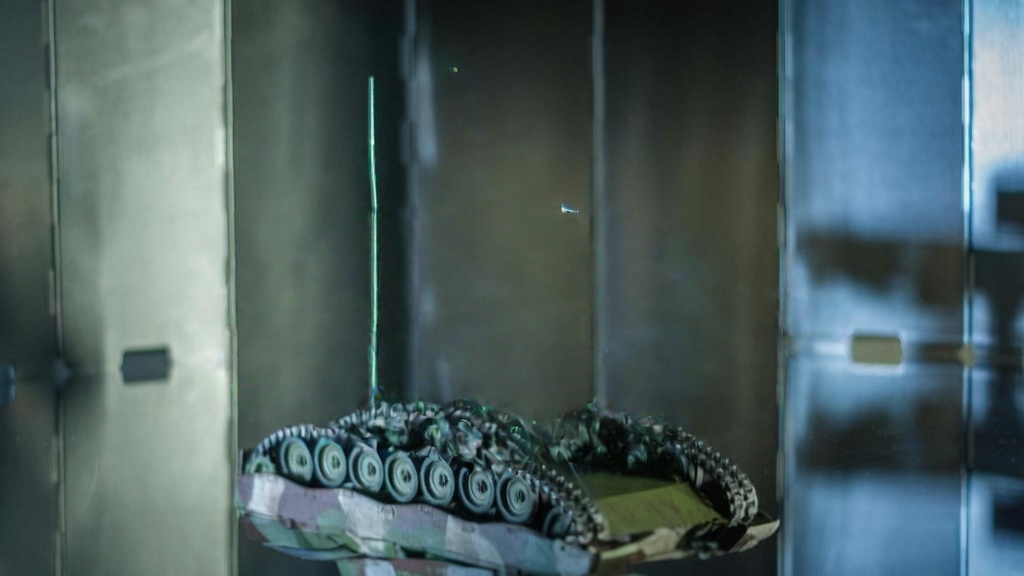} &
        \includegraphics[width=0.19\linewidth]{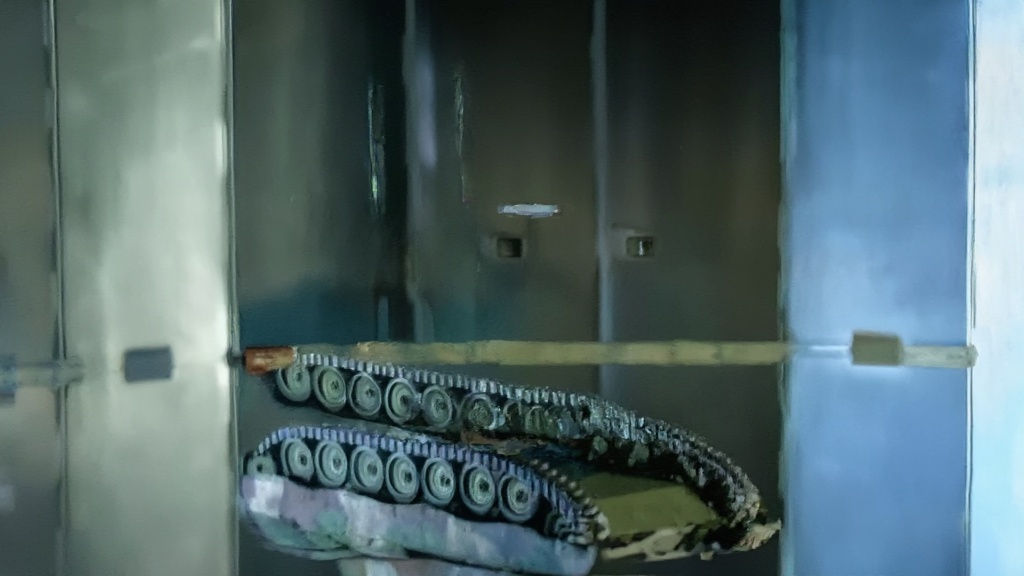} &
        \includegraphics[width=0.19\linewidth]{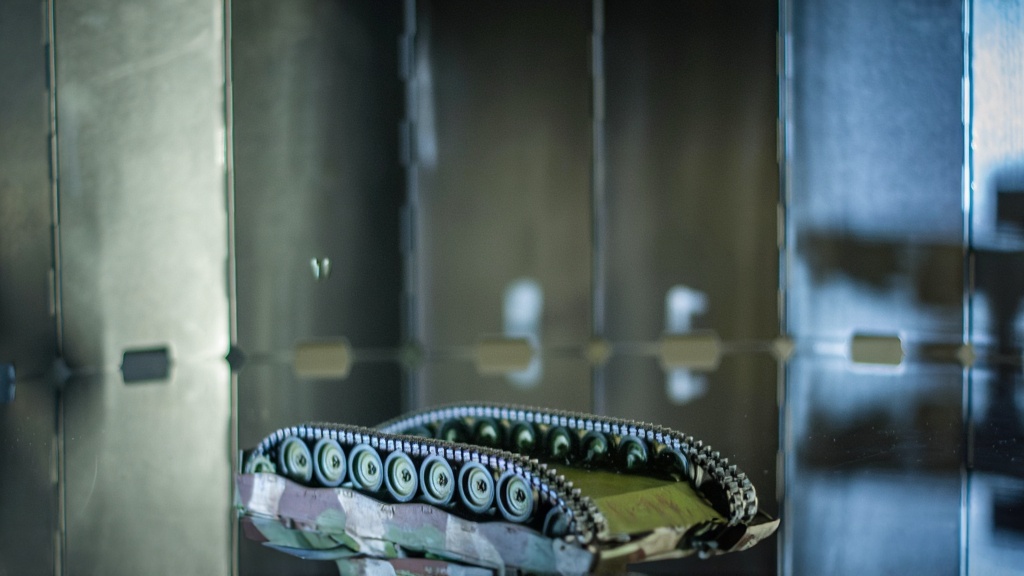} &
        \includegraphics[width=0.19\linewidth]{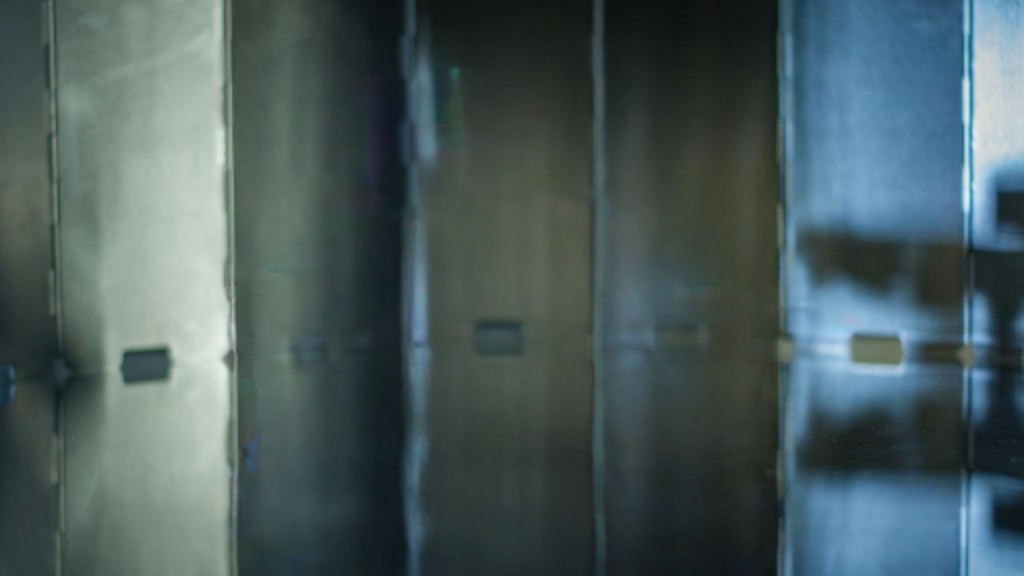} \\[-2pt]

        \footnotesize Input w/ Mask & 
        \footnotesize AttentiveEraser~\cite{sun2025attentive} & 
        \footnotesize OmniEraser~\cite{wei2025omnieraser} & 
        \footnotesize OmniPaint~\cite{yu2025omnipaint} & 
        \footnotesize FlashClear (ours) \\
        
    \end{tabular}
    
\vspace{-2mm} 
\caption{More qualitative comparison of our method and others on \textit{CausRem}\cite{zhu2025georemover} reflection and shadow dataset. Please zoom in on the image for a better viewing experience.}
\label{fig:causrem_2}
\vspace{-5mm}
\end{figure}

As shown in Figures~\ref{fig:acc_qualitative_comparison_more},~\ref{fig:more visual 1},~\ref{fig:more visual 2},~\ref{fig:causrem_1}, and~\ref{fig:causrem_2}, we provide more visual results as supplements to Section~\ref{sec:removal} and Section~\ref{sec:Acc}. More evidence demonstrates our methods' superiority.
\vspace{-2mm}
\section{Limitations and Possible Impact}\label{sec:limitations}
\vspace{-2mm}
While FlashClear achieves strong object removal quality and efficiency, it still has limitations and potential societal risks. First, it is tailored to object removal and relies on task-specific priors such as localized editing regions and background redundancy. Second, although FlashClear substantially reduces theoretical computation, practical end-to-end speedup can be affected by system-level overhead, since I/O operations and memory communication may occupy a larger fraction of the total runtime when model inference is already lightweight. Beyond technical limitations, FlashClear can benefit interactive editing, photo restoration, privacy-preserving content editing, and on-device applications by making high-quality object removal more efficient and accessible. However, object removal may also be misused to manipulate visual evidence or alter image context in misleading ways. We therefore encourage responsible use of this technology in practical deployment.

\newpage

\clearpage

\clearpage

\clearpage

\end{document}